\documentclass{article}

\PassOptionsToPackage{hyphens}{url}
\PassOptionsToPackage{hyperfootnotes=false}{hyperref}

\usepackage{fix-cm}          %
\usepackage{iclr2027_conference,times}
\usepackage{etoolbox}
\makeatletter
\patchcmd{\@maketitle}{\lhead{Published as a conference paper at ICLR 2027}}{\lhead{Preprint}}{}{\errmessage{Preprint header patch failed}}
\patchcmd{\@maketitle}{{\LARGE\sc \@title\par}}{{\LARGE\sc \MakeUppercase{\@title}\par}}{}{\errmessage{Title patch failed}}
\makeatother

\usepackage{amsmath,amsfonts,bm}

\def\eqref#1{equation~\ref{#1}}

\def\1{\bm{1}}

\DeclareMathAlphabet{\mathsfit}{\encodingdefault}{\sfdefault}{m}{sl}
\SetMathAlphabet{\mathsfit}{bold}{\encodingdefault}{\sfdefault}{bx}{n}

\let\peoorigtextsc\textsc
\renewcommand{\textsc}[1]{\peoorigtextsc{\MakeLowercase{#1}}}
\usepackage{latexsym}
\usepackage[T1]{fontenc}
\usepackage[utf8]{inputenc}
\usepackage{microtype}
\usepackage{inconsolata}
\usepackage{graphicx}
\graphicspath{{./}}

\usepackage{hyperref}
\hypersetup{
  colorlinks=true,
  linkcolor=[rgb]{0.0,0.2,0.6},   %
  citecolor=[rgb]{0.0,0.2,0.6},   %
  urlcolor =[rgb]{0.3,0.0,0.6},   %
  filecolor=[rgb]{0.0,0.2,0.6}
}
\hypersetup{
  pdftitle={Retrieval Augmented Generation Compositionality},
  pdfauthor={Ido Levy, Segev Shlomov, Leshem Choshen},
  pdfsubject={We pre-declare three hypotheses and introduce Lasagna, an open-source framework for controlled RAG evaluation.},
  pdfkeywords={retrieval-augmented generation; RAG evaluation; compositionality; ablation; benchmarking; corpus scaling},
  pdflang={en-US}
}
\usepackage{url}            %
\usepackage{seqsplit}       %
\usepackage{booktabs}       %
\usepackage{longtable}      %
\usepackage{amsfonts}       %
\usepackage{nicefrac}       %
\usepackage{xcolor}         %

\usepackage{multirow}
\usepackage{caption}
\usepackage{subcaption}
\usepackage{makecell}
\usepackage{colortbl}
\usepackage{array}
\usepackage{amsmath}
\usepackage{listings}
\usepackage{amssymb}
\usepackage[capitalize,nameinlink]{cleveref}
\usepackage{enumitem}
\usepackage{footnote}
\usepackage{adjustbox}
\usepackage{tikz}
\usetikzlibrary{arrows.meta, calc}
\usepackage[most]{tcolorbox}
\definecolor{lasagnagreen}{RGB}{234,243,222}
\definecolor{corrected}{RGB}{255,248,230}
\definecolor{cGF}{RGB}{241,239,232}  %
\definecolor{cGB}{RGB}{136,135,128}  %
\definecolor{cPF}{RGB}{238,237,254}  %
\definecolor{cPB}{RGB}{83,74,183}    %
\definecolor{cTF}{RGB}{225,245,238}  %
\definecolor{cTB}{RGB}{15,110,86}    %
\definecolor{cCF}{RGB}{250,236,231}  %
\definecolor{cCB}{RGB}{153,60,29}    %
\definecolor{cTx}{RGB}{44,44,42}     %
\definecolor{cSb}{RGB}{95,94,90}     %
\definecolor{cLt}{RGB}{200,198,190}  %

\newcommand{\stagebox}[4]{%
  \node[draw=#3, fill=#2, rounded corners=2.5pt,
        minimum width=1.82cm, minimum height=2.55cm,
        inner sep=0pt] (#4) at (#1,0) {};%
}
\newcommand{\stagetitle}[4]{%
  \node[font=\fontsize{9.4}{11.2}\selectfont\bfseries\sffamily,
        color=#2!12!cTx, anchor=north] at (#1,1.12) {#3};
  \node[font=\fontsize{7.4}{8.8}\selectfont\sffamily,
        color=#2!12!cTx, anchor=north] at (#1,0.72) {#4};
}
\newcommand{\tlink}[3]{%
  \hyperref[#3]{\color{#2}\fontsize{5.6}{6.6}\selectfont\sffamily #1}%
}
\newcommand{\drow}[4]{%
  \node[font=\fontsize{6.6}{7.8}\selectfont\sffamily,
        color=cSb, anchor=west] at (0.62,#1) {#2};%
  \ifnum#4=1
    \fill[#3, rounded corners=1.2pt]
      (2.10,{#1-0.090}) rectangle (2.49,{#1+0.090});
  \else
    \fill[cGB!16, draw=cGB!38, line width=0.22pt, rounded corners=1.2pt]
      (2.10,{#1-0.065}) rectangle (2.49,{#1+0.065});
  \fi
  \ifnum#4=2
    \fill[#3, rounded corners=1.2pt]
      (2.55,{#1-0.090}) rectangle (2.94,{#1+0.090});
  \else
    \fill[cGB!16, draw=cGB!38, line width=0.22pt, rounded corners=1.2pt]
      (2.55,{#1-0.065}) rectangle (2.94,{#1+0.065});
  \fi
  \ifnum#4=3
    \fill[#3, rounded corners=1.2pt]
      (3.00,{#1-0.090}) rectangle (3.39,{#1+0.090});
  \else
    \fill[cGB!16, draw=cGB!38, line width=0.22pt, rounded corners=1.2pt]
      (3.00,{#1-0.065}) rectangle (3.39,{#1+0.065});
  \fi
}

\newtcblisting{profilebox}[1][]{%
  enhanced, breakable,
  colback=lasagnagreen!30, colframe=black!50,
  fonttitle=\footnotesize\ttfamily,
  boxrule=0.4pt, arc=2pt,
  left=4pt, right=4pt, top=3pt, bottom=3pt,
  listing only,
  listing options={basicstyle=\ttfamily\scriptsize,
    keywordstyle=\color{black},
    commentstyle=\color{black!50},  %
    columns=fullflexible},
  title={#1},
}

\newtcolorbox{rcaexample}[1][]{%
  enhanced, breakable,
  colback=black!2, colframe=black!40,
  fonttitle=\bfseries\footnotesize,
  boxrule=0.5pt, arc=2pt,
  left=5pt, right=5pt, top=4pt, bottom=4pt,
  title={#1},
  before upper={\footnotesize},
}

\newtcolorbox{vdict}[1][]{%
  enhanced, breakable,
  colback=lasagnagreen!40, colframe=cTB,
  fonttitle=\bfseries\sffamily\small,
  boxrule=0.7pt, arc=2pt,
  left=6pt, right=6pt, top=4pt, bottom=4pt,
  coltitle=white, colbacktitle=cTB,
  title={#1},
  before upper={\footnotesize\brokenpenalty=10000},
}

\newcommand{\yes}{\checkmark}
\newcommand{\no}{$\times$}

\newcommand{\anon}[1]{}  %

\title{Retrieval Augmented Generation\\ \mbox{Compositionality}}
\newcommand{\authormail}[1]{\href{mailto:#1}{\textcolor{black}{\texttt{#1}}}}
\author{%
  Ido Levy$^{1,2}$\quad Segev Shlomov$^{1}$\quad Leshem Choshen$^{1,2}$ \\
  \normalfont $^{1}$IBM\qquad $^{2}$Weizmann Institute of Science \\
  \normalfont\small\authormail{Ido.Levy1@ibm.com}\quad \authormail{Segev.Shlomov1@ibm.com}\quad \authormail{leshem.choshen@weizmann.ac.il} \\[8pt]
  \normalfont\small
  \href{https://github.com/dolev31/lasagna}{\raisebox{-0.15em}{\includegraphics[height=0.95em]{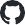}}\ \texttt{github.com/dolev31/lasagna}}\qquad
  \href{https://dolev31.github.io/lasagna/}{\raisebox{-0.15em}{\includegraphics[height=0.95em]{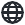}}\ \texttt{dolev31.github.io/lasagna}}
}

\iclrfinalcopy
\begin{document}

\maketitle
\begin{abstract}
Retrieval Augmented Generation (RAG) grounds large language models
in external knowledge. Modern pipelines combine embedding, chunking,
retrieval, reranking, query augmentation, context compression, and
agentic verification, yet these techniques are typically evaluated
one at a time. It remains unclear whether their isolated gains survive
composition, whether retrieval quality and answer faithfulness favor
the same design choices, and whether conclusions from small corpora
hold at large scale. We pre-declare three hypotheses and
introduce \textsc{Lasagna}, an open-source framework for controlled
RAG evaluation. We test $450$ configurations across eight benchmarks
and corpora ranging from $5$K to $200$K documents. Three findings
emerge. First, individually strong techniques from different phases
of a RAG pipeline do not necessarily compose: three retrieval
techniques predict a $+47\%$ combined improvement but deliver only
$+17\%$ when used together. Second, component gains stack differently
for retrieval and faithfulness, and optimizing one can hurt the other.
Third, the best composition changes as the corpus grows: the
combination that wins on a small corpus may not remain best at
large scale. These findings show that gains measured in isolation
are not reliable predictors of realistic performance, and that new RAG techniques
should be tested with the full surrounding stack. We call on
researchers to do so and supply \textsc{Lasagna} to simplify these
experiments.
\end{abstract}

\section{Introduction}
\label{sec:intro}

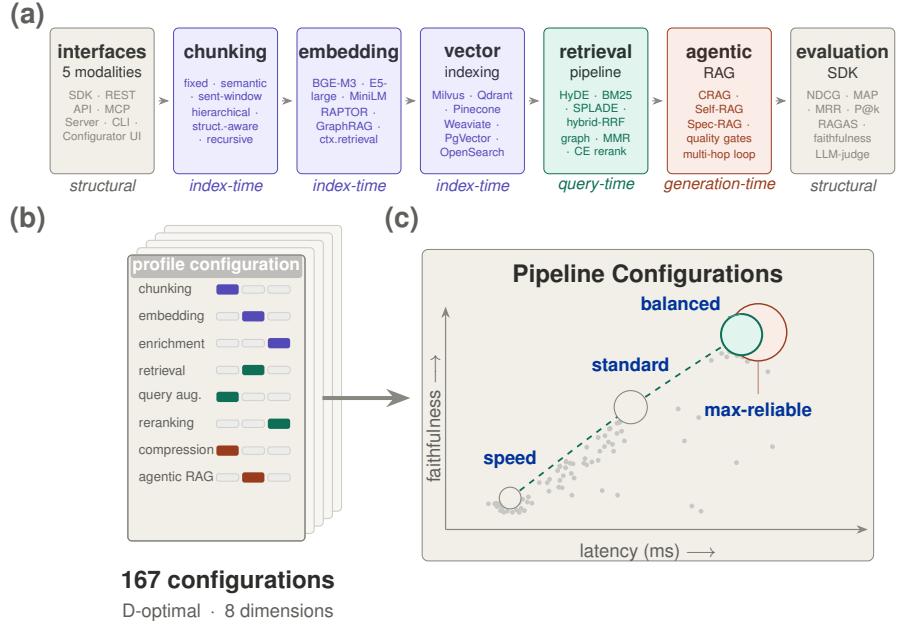
\begin{figure*}[t]
  \centering
  \resizebox{0.85\textwidth}{!}{%
\begin{tikzpicture}[
  >=Stealth,
  plarr/.style={->, line width=0.50pt, color=cGB,
                shorten >=2pt, shorten <=2pt},
  bigarr/.style={->, line width=2.0pt, color=cGB,
                 shorten >=3pt, shorten <=3pt},
  plbl/.style={font=\fontsize{14}{16}\selectfont\bfseries\sffamily,
               color=cSb},
  tny/.style={font=\fontsize{8.0}{9.4}\selectfont\sffamily},
  sml/.style={font=\fontsize{9.8}{11.5}\selectfont\sffamily, color=cSb},
  titem/.style={font=\fontsize{5.7}{6.8}\selectfont\sffamily},
]

\node[plbl] at (-1.30,1.50) {(a)};

\stagebox{0.00}{cGF}{cGB}{p0}
\stagetitle{0.00}{cGB}{interfaces}{5 modalities}
\node[titem, color=cGB, text width=1.60cm, align=center,
      anchor=north] at (0.00,0.30)
  {\tlink{SDK}{cGB}{sec:platform} $\cdot$
   \tlink{REST API}{cGB}{sec:platform} $\cdot$
   \tlink{MCP Server}{cGB}{sec:platform} $\cdot$
   \tlink{CLI}{cGB}{sec:platform} $\cdot$
   \tlink{Configurator UI}{cGB}{app:configurator}};

\stagebox{2.16}{cPF}{cPB}{p1}
\stagetitle{2.16}{cPB}{chunking}{}
\node[titem, color=cPB, text width=1.60cm, align=center,
      anchor=north] at (2.16,0.52)
  {\tlink{fixed}{cPB}{meth:chunk-fixed} $\cdot$
   \tlink{semantic}{cPB}{meth:chunk-semantic} $\cdot$
   \tlink{sent-window}{cPB}{meth:chunk-sentwindow}\\[1pt]
   \tlink{hierarchical}{cPB}{meth:chunk-hierarchical} $\cdot$
   \tlink{struct.-aware}{cPB}{meth:chunk-structaware} $\cdot$
   \tlink{recursive}{cPB}{meth:chunk-recursive}};

\stagebox{4.32}{cPF}{cPB}{p2}
\stagetitle{4.32}{cPB}{embedding}{}
\node[titem, color=cPB, text width=1.60cm, align=center,
      anchor=north] at (4.32,0.52)
  {\tlink{BGE-M3}{cPB}{meth:emb-bgem3} $\cdot$
   \tlink{E5-large}{cPB}{meth:emb-e5large} $\cdot$
   \tlink{MiniLM}{cPB}{meth:emb-minilm}\\[1pt]
   \tlink{RAPTOR}{cPB}{meth:enrich-raptor} $\cdot$
   \tlink{GraphRAG}{cPB}{meth:enrich-graphrag} $\cdot$
   \tlink{ctx.retrieval}{cPB}{meth:enrich-contextual}};

\stagebox{6.48}{cPF}{cPB}{p3}
\stagetitle{6.48}{cPB}{vector}{indexing}
\node[titem, color=cPB, text width=1.60cm, align=center,
      anchor=north] at (6.48,0.30)
  {\tlink{Milvus}{cPB}{meth:vec-milvus} $\cdot$
   \tlink{Qdrant}{cPB}{meth:vec-qdrant} $\cdot$
   \tlink{Pinecone}{cPB}{meth:vec-pinecone}\\[1pt]
   \tlink{Weaviate}{cPB}{meth:vec-weaviate} $\cdot$
   \tlink{PgVector}{cPB}{meth:vec-pgvector} $\cdot$
   \tlink{OpenSearch}{cPB}{meth:vec-opensearch}};

\stagebox{8.64}{cTF}{cTB}{p4}
\stagetitle{8.64}{cTB}{retrieval}{pipeline}
\node[titem, color=cTB, text width=1.60cm, align=center,
      anchor=north] at (8.64,0.30)
  {\tlink{HyDE}{cTB}{meth:qa-hyde} $\cdot$
   \tlink{BM25}{cTB}{meth:ret-bm25} $\cdot$
   \tlink{SPLADE}{cTB}{meth:ret-splade} $\cdot$
   \tlink{hybrid-RRF}{cTB}{meth:ret-hybrid-rrf}\\[1pt]
   \tlink{graph}{cTB}{meth:ret-graph} $\cdot$
   \tlink{MMR}{cTB}{meth:rerank-mmr} $\cdot$
   \tlink{CE rerank}{cTB}{meth:rerank-ce}};

\stagebox{10.80}{cCF}{cCB}{p5}
\stagetitle{10.80}{cCB}{agentic}{RAG}
\node[titem, color=cCB, text width=1.60cm, align=center,
      anchor=north] at (10.80,0.30)
  {\tlink{CRAG}{cCB}{meth:agent-crag} $\cdot$
   \tlink{Self-RAG}{cCB}{meth:agent-selfrag}\\[1pt]
   \tlink{Spec-RAG}{cCB}{meth:agent-specrag} $\cdot$
   \tlink{quality gates}{cCB}{meth:agent-quality-gates}\\[1pt]
   \tlink{multi-hop loop}{cCB}{meth:agent-multihop}};

\stagebox{12.96}{cGF}{cGB}{p6}
\stagetitle{12.96}{cGB}{evaluation}{SDK}
\node[titem, color=cGB, text width=1.60cm, align=center,
      anchor=north] at (12.96,0.30)
  {\tlink{NDCG}{cGB}{meth:metric-ndcg} $\cdot$
   \tlink{MAP}{cGB}{meth:metric-map} $\cdot$
   \tlink{MRR}{cGB}{meth:metric-mrr} $\cdot$
   \tlink{P@k}{cGB}{meth:metric-pk}\\[1pt]
   \tlink{RAGAS}{cGB}{meth:metric-ragas} $\cdot$
   \tlink{faithfulness}{cGB}{meth:metric-faithfulness}\\[1pt]
   \tlink{LLM-judge}{cGB}{meth:metric-llm-judge}};

\draw[plarr] (p0)--(p1);
\draw[plarr] (p1)--(p2);
\draw[plarr] (p2)--(p3);
\draw[plarr] (p3)--(p4);
\draw[plarr] (p4)--(p5);
\draw[plarr] (p5)--(p6);

\node[tny,color=cSb]  at ( 0.00,-1.48) {\itshape structural};
\node[tny,color=cPB]  at ( 2.16,-1.48) {\itshape index-time};
\node[tny,color=cPB]  at ( 4.32,-1.48) {\itshape index-time};
\node[tny,color=cPB]  at ( 6.48,-1.48) {\itshape index-time};
\node[tny,color=cTB]  at ( 8.64,-1.48) {\itshape query-time};
\node[tny,color=cCB]  at (10.80,-1.48) {\itshape generation-time};
\node[tny,color=cSb]  at (12.96,-1.48) {\itshape structural};

\node[plbl] at (-1.30,-2.08) {(b)};

\begin{scope}[xshift=-0.1cm]
\foreach \k in {4,3,2,1}{
  \pgfmathsetmacro{\ox}{\k*0.16}
  \pgfmathsetmacro{\oy}{\k*0.13}
  \fill[cGF!55, draw=cGB!45, line width=0.25pt, rounded corners=2pt]
    ({2.10+\ox-1.55},{-5.20+\oy-2.50}) rectangle
    ({2.10+\ox+1.55},{-5.20+\oy+2.50});}

\fill[cGF, draw=cGB, line width=0.55pt, rounded corners=2pt]
  (0.55,-7.70) rectangle (3.65,-2.70);

\fill[cGB!55, rounded corners=1.5pt]
  (0.60,-3.10) rectangle (3.60,-2.68);
\node[font=\fontsize{8.6}{10.0}\selectfont\bfseries\sffamily, white]
  at (2.10,-2.89) {profile configuration};

\drow{-3.30}{chunking}{cPB}{1}
\drow{-3.77}{embedding}{cPB}{2}
\drow{-4.24}{enrichment}{cPB}{3}
\drow{-4.71}{retrieval}{cTB}{2}
\drow{-5.18}{query aug.}{cTB}{1}
\drow{-5.65}{reranking}{cTB}{3}
\drow{-6.12}{compression}{cCB}{1}
\drow{-6.59}{agentic RAG}{cCB}{2}

\draw[cGB, line width=0.55pt, rounded corners=2pt]
  (0.55,-7.70) rectangle (3.65,-2.70);

\node[font=\fontsize{12.0}{14.0}\selectfont\bfseries\sffamily,
      color=cTx, anchor=north] at (2.30,-8.10) {167 configurations};
\node[font=\fontsize{9.2}{10.8}\selectfont\sffamily,
      color=cSb, anchor=north] at (2.30,-8.68)
  {D-optimal\enspace$\cdot$\enspace8 dimensions};
\end{scope}

\draw[bigarr] (3.75,-5.20) -- (5.50,-5.20);
\node[sml, anchor=north] at (4.55,-5.48) {};

\begin{scope}[xshift=-0.3cm]
\node[plbl] at (5.55,-2.08) {(c)};

\fill[cGF, draw=cGB, line width=0.38pt, rounded corners=3pt]
  (5.90,-8.05) rectangle (13.80,-2.60);

\node[font=\fontsize{12.0}{14.0}\selectfont\bfseries\sffamily,
      color=cTx, anchor=north] at (9.85,-2.74)
  {Pipeline Configurations};

\draw[->, line width=0.60pt, cGB]
  (6.31,-7.50) -- (13.70,-7.50);
\node[font=\fontsize{9.2}{10.8}\selectfont\sffamily, color=cSb,
      anchor=north] at (9.85,-7.62)
  {latency (ms)\;$\longrightarrow$};

\draw[->, line width=0.60pt, cGB]
  (6.31,-7.50) -- (6.31,-3.60);
\node[rotate=90, font=\fontsize{9.2}{10.8}\selectfont\sffamily,
      color=cSb, anchor=center] at (6.10,-5.55)
  {faithfulness\;$\longrightarrow$};

\foreach \x/\y in {%
  7.06/-7.05, 7.17/-7.05, 7.19/-7.05, 7.20/-7.05, 7.21/-7.05, 7.21/-7.10,
  7.25/-7.05, 7.27/-7.17, 7.28/-7.07, 7.29/-7.11, 7.31/-7.12, 7.33/-7.20,
  7.39/-7.18, 7.48/-7.20, 7.55/-7.16, 7.60/-7.20, 7.60/-7.03, 7.63/-7.13,
  7.66/-6.97, 7.69/-7.17, 7.74/-7.02, 7.76/-6.93, 7.82/-7.04, 7.95/-6.68,
  8.09/-6.68, 8.12/-6.83, 8.13/-6.62, 8.27/-6.47, 8.29/-6.76, 8.31/-6.39,
  8.31/-6.64, 8.31/-6.40, 8.35/-6.77, 8.37/-6.56, 8.41/-6.42, 8.42/-6.31,
  8.42/-6.32, 8.49/-6.34, 8.51/-6.52, 8.52/-6.90, 8.54/-6.35, 8.56/-6.60,
  8.73/-6.34, 8.75/-6.26, 8.78/-6.37, 8.78/-6.06, 8.87/-6.32, 8.98/-6.29,
  8.98/-5.89, 9.01/-6.13, 9.05/-6.82, 9.09/-6.05, 9.10/-5.86, 9.16/-5.97,
  9.18/-5.74, 9.18/-5.74, 9.19/-5.73, 9.23/-6.01, 9.25/-6.13, 9.31/-5.82,
  9.31/-5.64, 9.32/-6.90, 9.34/-5.63, 9.36/-5.62, 9.37/-5.60, 9.39/-5.78,
  9.40/-5.57, 9.44/-5.95, 10.38/-5.39, 10.49/-5.95, 10.62/-7.09, 10.74/-5.54,
  10.78/-7.18, 10.96/-4.54, 11.14/-4.42, 11.23/-4.48, 11.28/-4.33, 11.28/-4.39,
  11.33/-4.40, 11.40/-4.39, 11.40/-6.33, 11.41/-4.46, 11.42/-4.43, 11.47/-4.33,
  11.49/-4.46, 11.51/-4.48, 11.58/-4.35, 11.59/-4.56, 11.65/-4.51, 11.67/-4.38,
  11.68/-4.41, 11.72/-4.39, 11.77/-4.30, 11.95/-4.74, 12.03/-6.05%
} { \fill[cSb!35] (\x,\y) circle (0.042); }

\draw[cTB, dashed, line width=0.9pt]
  plot[smooth,tension=0.55] coordinates {
    (7.44,-6.95)
    (9.55,-5.36)
    (11.49,-4.09)
    (11.78,-4.05)};

\draw[->, line width=0.40pt, color=cSb!55, shorten <=4pt, shorten >=3pt]
  (11.49,-4.09) -- (11.78,-4.05);

\node[circle, fill=cCF, draw=cCB, line width=0.70pt,
      minimum size=1.00cm, inner sep=0pt] at (11.78,-4.05) {};

\node[circle, fill=cTF, draw=cTB, line width=1.0pt,
      minimum size=0.72cm, inner sep=0pt] at (11.49,-4.09) {};

\node[circle, fill=cGF, draw=cGB, line width=0.55pt,
      minimum size=0.58cm, inner sep=0pt] at (9.55,-5.36) {};

\node[circle, fill=cGF, draw=cGB, line width=0.55pt,
      minimum size=0.38cm, inner sep=0pt] at (7.44,-6.95) {};

\node[font=\fontsize{9.2}{10.6}\selectfont\bfseries\sffamily,
      color=cTx, anchor=south] at (7.44,-6.55)
  {\hyperref[app:prof-speed]{speed}};
\node[font=\fontsize{9.2}{10.6}\selectfont\bfseries\sffamily,
      color=cTx, anchor=south] at (9.55,-4.85)
  {\hyperref[app:prof-standard]{standard}};
\node[font=\fontsize{9.2}{10.6}\selectfont\bfseries\sffamily,
      color=cTB, anchor=south east] at (11.25,-3.78)
  {\hyperref[app:prof-balanced]{balanced}};
\draw[line width=0.30pt, color=cTB!70]
  (11.25,-3.78) -- (11.49,-3.73);
\node[font=\fontsize{9.2}{10.6}\selectfont\bfseries\sffamily,
      color=cCB, anchor=north] at (11.78,-5.15)
  {\hyperref[app:prof-maxreliable]{max-reliable}};
\draw[line width=0.30pt, color=cCB!70]
  (11.78,-4.55) -- (11.78,-5.13);

\end{scope}
\end{tikzpicture}
  }
  \caption{%
    \textbf{\textsc{Lasagna} overview.}
    \textbf{(a)}~Seven functional blocks.
    \textbf{(b)}~Eight dimensions in a 167-configuration D-optimal
    design. \textbf{(c)}~Faithfulness versus latency for four
    operational profiles; faded points show other configurations.
  }
  \label{fig:overview}
\end{figure*}

Retrieval-augmented generation
(RAG)~\citep{gao2023survey,fan2024survey} combines choices across
many pipeline stages. Our study covers eight stages that span the
dominant design axes in modern RAG (\cref{fig:overview}b; full
catalog in \cref{app:methods}): embedding, chunking, index-time
enrichment, retrieval, query augmentation, reranking, context
compression, and agentic quality gates. Most new techniques are
proposed and evaluated one stage at a time, in isolation from the
surrounding pipeline, or from several representative pipelines that
would test whether findings generalize~\citep{yu2024evalsurvey}. Adding to the complexity, RAG systems
are evaluated for two goals~\citep{saadfalcon2024ares,friel2024ragbench}:
retrieving the right content (NDCG@10) and producing a faithful
answer from it~\citep[RAGAS;][]{es2023ragas}. It is unclear whether isolated gains persist under composition. It
is also unclear whether the same design choices optimize both
retrieval quality (which documents are retrieved) and answer
faithfulness (whether the answer is grounded in those documents),
two goals RAG evaluation typically reports separately. We pre-declare three hypotheses. \textbf{H1.} Components often fail
to compose: isolated gains from single-component ablations do not
necessarily persist when components are combined. \textbf{H2.} A
simple RAG pipeline can be near-optimal in many settings, while
more elaborate configurations tend to specialize to narrower ones.
Most RAG ablations in the literature run at 5K to 50K documents,
whereas the applications RAG can serve run at corpora an order of
magnitude larger; \textbf{H3.} composition rankings shift with
corpus scale, so conclusions from small-corpus benchmarks may not
hold at larger scales. Prior work has explored combinations through
configuration sweeps and heuristic
search~\citep{wang2024searching,li2025bestpractices,rau2024bergen,
jin2024flashrag,kim2024autorag,fu2024autorag,chen2025mmoarag}; these
approaches can identify strong configurations but do not directly
measure interaction effects: whether the benefit of one component
depends on the other components in the pipeline. These interactions,
and how they change with corpus scale, remain largely unquantified.

Testing these hypotheses requires a framework that isolates pipeline
stages and supports reproducible evaluation across benchmarks. We
created \textsc{Lasagna}, an open-source RAG framework, to provide
this (\hyperref[fig:overview]{Fig.~1a}, \cref{sec:system}). Each
stage sits behind a fixed interface, so swapping one method cannot
silently change the implementation at another. Adding a new
technique is a single line of code; the framework tests it inside a
complete modern pipeline without reimplementing the surrounding stack.

We use \textsc{Lasagna} to study RAG composition with a
167-configuration D-optimal
design~\citep{fedorov1972theory,atkinson2007optimum} across eight
configurable stages of a modern RAG pipeline
(\cref{tab:dimensions}), eight benchmarks (\cref{fig:benchmarks}),
and corpus sizes from 5K to 200K documents, complemented by 138
non-factorial ablation configurations across 24 targeted
experiments (\cref{app:experiment-inventory}).  Across the factorial, ablation suite, and scaling sweep, the
central finding is that techniques that perform well separately do
not consistently retain their gains inside a full pipeline, and the
pattern differs between retrieval and faithfulness. Retrieval gains
compose sub-additively: the three largest single-component
improvements sum to a predicted $+47\%$ NDCG@10 gain, but the
composed pipeline realizes only $+17\%$ (\cref{app:composition}).
Faithfulness gains compose near-additively: separately measured
improvements approximately transfer to the full pipeline
(\cref{tab:composition-faith-exploratory}).  Pipeline and generator
also partly substitute: a stronger generator helps the lightest pipeline
far more than the heaviest, so a heavier stack buys
less when the generator is already strong (\cref{app:generator-swap}).

Three practical implications follow. First, the components that
improve retrieval are not the same as those that improve faithfulness,
and optimizing one metric can hurt the other. Second, the four
complete RAG profiles expose a Pareto tradeoff on retrieval
quality, faithfulness, and response time, with no universal winner.
Adding an agentic loop to a strong retrieval stack yields only a
negligible faithfulness gain while hurting retrieval quality and
latency. Across three other benchmarks at 100K--200K documents,
lighter pipelines retain similar measured quality at
$10$--$20\times$ lower latency
(\cref{sec:pareto,app:h2-cross-corpus-corroboration}). Third, corpus
size shifts profile rankings: the design leading at small corpora
trails by $13.7\%$ NDCG@10 at 200K, with a rank flip visible from
25K (\cref{fig:scaling-retrieval}). Small-corpus rankings should therefore
be rechecked at the intended scale.
Our contributions are:

\begin{enumerate}[leftmargin=1.5em, itemsep=2pt]
  \item \textbf{How RAG components compose.} A compositional study
  of RAG across eight configurable stages, eight benchmarks, and
  corpora scales from 5K to 200K documents: 167 configurations from
  a D-optimal factorial plus 138 non-factorial configurations across
  24 targeted ablations. The four operational RAG profiles that
  emerge (\textsc{speed}, \textsc{standard}, \textsc{balanced},
  \textsc{max-reliable}) are calibrated to cost, latency, and performance.

  \item \textbf{\textsc{Lasagna}.} The open-source framework built to
  conduct and reproduce this study. It provides a modular,
  stage-isolated RAG pipeline and an evaluation harness that lets
  researchers add a method and test it with the surrounding stack
  (\cref{app:metrics,app:experiment-inventory}).
\end{enumerate}

\section{Related Work}
\label{sec:related_work}

Modern RAG spans a layered design space with plausible interaction
points at every stage (full method catalog in \cref{app:methods}).
On \emph{retrieval}, sparse baselines
(BM25~\citep{robertson2009bm25}) coexist with dense encoders
(BGE-M3~\citep{chen2024bgem3}, E5~\citep{wang2022text}), learned
sparse encoders (SPLADE~\citep{formal2021splade}), and
late-interaction models (ColBERT~\citep{khattab2020colbert}), fused
by reciprocal-rank fusion (RRF)~\citep{cormack2009rrf} or learned
hybrids.
\emph{Reranking} spans cross-encoder scorers trained on
MS-MARCO~\citep{nogueira2019monobert}, ensemble rerankers, and
diversity-driven selection (MMR;~\citealp{carbonell1998mmr}).  \emph{Context
compression} is either token-budget aware
(LLMLingua~\citep{jiang2023llmlingua}) or positional
(lost-in-the-middle-aware selection~\citep{liu2023lostinthemiddle}).
\emph{Query augmentation} rewrites the query before retrieval:
HyDE~\citep{gao2023hyde} drafts a hypothetical answer,
RAG-Fusion~\citep{rackauckas2024ragfusion} fans out multi-query
variants, decomposition~\citep{khattab2024dspy} splits multi-hop
questions into sub-questions, and step-back
prompting~\citep{zheng2024stepback} abstracts before retrieving.
\emph{Index-time processing} enriches chunks before retrieval with
document summaries (RAPTOR~\citep{sarthi2024raptor}), contextual
prefixes~\citep{anthropic2024contextual}, or entity-graph indexes
(GraphRAG~\citep{edge2024graphrag}).  \emph{Generation-time control}
adds correction and gating loops around the answer model
(CRAG~\citep{yan2024crag}, Self-RAG~\citep{asai2024selfrag},
Speculative-RAG~\citep{ICLR2025_2ea06b52}).  Each of these
stages has been shown to help in isolation on some benchmark, but
they compose along more edges than a single-component ablation can
enumerate.  We study how these choices combine once
deployed together, and \textsc{Lasagna} lets others ask
compositionality questions in the same environment by
registering a new component and evaluating it against a full modern
pipeline.

\begin{table*}[!t]
\centering
\caption{\textsc{Lasagna} vs.\ prior RAG evaluations. Columns
defined in \cref{app:rag-eval-columns}.}
\label{tab:rag-eval-scale}
\scriptsize
\setlength{\tabcolsep}{3pt}
\begin{adjustbox}{max width=\linewidth}
\begin{tabular}{@{}lllccccc c@{}}
\toprule
Study
  & \makecell[l]{Configs\\(mode, varied stages)}
  & \makecell[l]{Design}
  & $\eta^2$
  & Interactions
  & Scaling
  & Pareto
  & \makecell[c]{Pre-\\registered}
  & Benchmarks \\
\midrule
\citep{jin2024flashrag}      & 16 methods\,(5)             & \emph{library} $+$ part.\ \textsc{grid} (tech.) & \no & \no & \no                      & \no           & \no & 32 \\
\citep{rau2024bergen}        & 500+ runs\,(3)              & partial \textsc{grid} (tech.)                   & \no & \no & \no                      & \no           & \no & ${\sim}12$ \\
\citep{wang2024searching}    & ${\sim}$dozens seq.\,(7)    & \textsc{greedy} (tech.)                         & \no & \no & \no                      & $\circ$       & \no & 6 \\
\citep{kim2024autorag}       & 30+ seq.\,(6)               & \textsc{greedy} (tech.)                         & \no & \no & \no                      & \no           & \no & 4 \\
\citep{friel2024ragbench}    & 32 joint\,(4)               & \textsc{grid} (mix.)                            & \no & \no & \no                      & \no           & \no & 5 \\
\citep{xiong2024mirage}      & 41 partial\,(3)             & partial \textsc{grid} (tech.)                   & \no & \no & $\circ$\,(top-$k$, 5 corpora) & \no      & \no & 5 \\
\midrule
\rowcolor{lasagnagreen}
\textsc{Lasagna} (ours)
  & \textbf{167 factorial\,+\,138 abl.\,(8)}
  & \textbf{\textsc{d-opt} (tech., 30 levels)}
  & \textbf{\yes}
  & \textbf{\yes\,(28)}
  & \textbf{\yes\,(6\,$\times$\,3)}
  & \textbf{\yes\,(4, HV)}
  & \textbf{\yes}
  & 8 (15) \\
\bottomrule
\end{tabular}
\end{adjustbox}
\end{table*}

\label{sec:rw-systems}

\Cref{tab:rag-eval-scale} positions \textsc{Lasagna} against prior
work that studies RAG beyond isolated components.  Best-practice
studies~\citep{wang2024searching,barnett2024seven} search greedily
for strong end-to-end configurations but do not estimate interaction
effects.  Benchmarking libraries such as BERGEN~\citep{rau2024bergen}
and FlashRAG~\citep{jin2024flashrag} support composition in a shared
codebase but do not enforce the stage isolation that controlled
interaction estimation requires, so measured differences cannot be
cleanly attributed to the component being varied
(a positional smoke run against both is reported in
\cref{app:framework-comparison}).  Automated
configuration search~\citep{kim2024autorag} optimizes
over configurations but reports only the winning setting, not the
interaction structure that produced it, so the resulting guidance
does not transfer when any assumption changes.  Evaluation
suites~\citep{krishna2025frames} and domain
benchmarks~\citep{xiong2024mirage} supply metrics
and tasks but not a common pipeline in which those metrics can be
compared across controlled compositions. Generation-time
controls~\citep{yan2024crag} strengthen the agentic layer but are
usually tested with retrieval fixed, leaving composition with a tuned
stack untested.

\providecommand{\stepnum}[2]{\tikz[baseline=-0.55ex]{%
  \node[circle, fill=#1, text=white, inner sep=0pt,
        minimum size=8pt,
        font=\tiny\bfseries\sffamily]{#2};}}

\section{LASAGNA: A Compositional Framework for RAG}
\label{sec:system}
\label{sec:pipeline}
\label{sec:eval-infra}
\label{sec:platform}

\begin{figure*}[t]
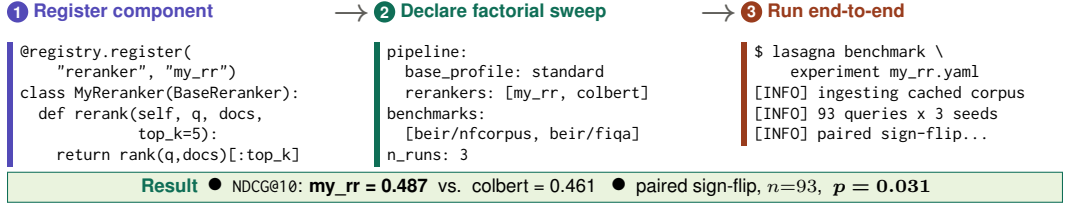

\centering

\begin{minipage}[t]{0.305\linewidth}
{\scriptsize\sffamily\textcolor{cPB}{\stepnum{cPB}{1}~\textbf{Register component}}}\\[-4pt]
\begin{lstlisting}[language=Python, basicstyle=\ttfamily\fontsize{6.8}{7.8}\selectfont,
  frame=leftline, framesep=3pt, framerule=2pt, rulecolor=\color{cPB},
  xleftmargin=5pt, aboveskip=0pt, belowskip=0pt]
@registry.register(
    "reranker", "my_rr")
class MyReranker(BaseReranker):
  def rerank(self, q, docs,
             top_k=5):
    return rank(q,docs)[:top_k]
\end{lstlisting}
\end{minipage}\hfill
\raisebox{-0.15em}{\large\color{cSb}$\boldsymbol{\rightarrow}$}\hfill
\begin{minipage}[t]{0.305\linewidth}
{\scriptsize\sffamily\textcolor{cTB}{\stepnum{cTB}{2}~\textbf{Declare factorial sweep}}}\\[-4pt]
\begin{lstlisting}[basicstyle=\ttfamily\fontsize{6.8}{7.8}\selectfont,
  frame=leftline, framesep=3pt, framerule=2pt, rulecolor=\color{cTB},
  xleftmargin=5pt, aboveskip=0pt, belowskip=0pt]
pipeline:
  base_profile: standard
  rerankers: [my_rr, colbert]
benchmarks:
  [beir/nfcorpus, beir/fiqa]
n_runs: 3
\end{lstlisting}
\end{minipage}\hfill
\raisebox{-0.15em}{\large\color{cSb}$\boldsymbol{\rightarrow}$}\hfill
\begin{minipage}[t]{0.305\linewidth}
{\scriptsize\sffamily\textcolor{cCB}{\stepnum{cCB}{3}~\textbf{Run end-to-end}}}\\[-4pt]
\begin{lstlisting}[basicstyle=\ttfamily\fontsize{6.8}{7.8}\selectfont,
  frame=leftline, framesep=3pt, framerule=2pt, rulecolor=\color{cCB},
  xleftmargin=5pt, aboveskip=0pt, belowskip=0pt]
$ lasagna benchmark \
    experiment my_rr.yaml
[INFO] ingesting cached corpus
[INFO] 93 queries x 3 seeds
[INFO] paired sign-flip...
\end{lstlisting}
\end{minipage}

\vspace{1pt}

{\setlength{\fboxsep}{2pt}%
\noindent\fcolorbox{cTB}{lasagnagreen}{%
\parbox{\dimexpr\linewidth-2\fboxsep-2\fboxrule\relax}{%
\centering\scriptsize\sffamily
\textcolor{cTB}{\textbf{Result}}\enspace{\large$\bullet$}\enspace
\texttt{NDCG@10}: \textbf{my\_rr = 0.487} \ vs.\ \ colbert = 0.461
\enspace{\large$\bullet$}\enspace
paired sign-flip, $n{=}93$, \ $\boldsymbol{p = 0.031}$%
}}}

\caption{\textsc{Lasagna} research workflow: register a component,
declare a factorial sweep in YAML, and run end-to-end for a paired
statistical test.}
\label{fig:workflow}
\end{figure*}

\textsc{Lasagna} is an open-source framework\anon{ \footnote{Code at \url{https://github.com/dolev31/lasagna}.}} for
evaluating new RAG components inside a complete modern pipeline.
A researcher proposing a new reranker, chunker, or agentic loop today
must either compare against a stripped-down baseline that does not
reflect a modern pipeline, or reimplement the surrounding stack. The
first biases the conclusion by mixing implementation differences
with algorithmic ones; the second is prohibitively expensive and
rarely attempted. \textsc{Lasagna} provides a configurable
end-to-end pipeline (\hyperref[fig:overview]{Fig.~1a}) with a fixed
interface per stage that the framework enforces, so a new component
drops into one stage and is measured against the same surrounding
stack used by every reported configuration (full code patterns in
\cref{app:showcase-plugins,app:showcase-byodb,app:showcase-byoagent}).
The workflow spans nine processing stages: document processing,
chunking, index-time enrichment, embedding and indexing, query
augmentation, retrieval, reranking, context compression, and answer
generation with optional agentic verification.
Every pipeline component is a subclass of a typed abstract base class:
eight component types are supported (\texttt{embedder}, \texttt{chunker},
\texttt{reranker}, \texttt{retriever}, \texttt{document\_processor},
\texttt{query\_enhancer}, \texttt{agentic\_pipeline}, and \texttt{generator}), each
stage's inputs and outputs are typed \texttt{dataclass} contracts, and
behavioral tests check that swapping one implementation does not mutate
a neighboring stage's state.  A single
\texttt{@registry.register("<stage>", "<name>")} decorator suffices
to make a new component discoverable at YAML-config time.
The framework ships with 53 built-in components, the 30 methods across the eight factorial
dimensions (\cref{tab:dimensions}; full catalog in \cref{app:methods}),
six vector backends (Milvus, Pinecone, Qdrant, Weaviate, pgvector,
OpenSearch), and adapters for the eight benchmarks used here (five
BEIR retrieval sets and three long-form QA sets;
\cref{app:datasets}).  Five automated isolation tests
(\cref{app:isolation}) check that toggling one component does not
silently change another, including that two sessions with opposite
settings each keep their own configuration; this supports attributing
the interaction estimates in \cref{sec:results} to the components we
vary rather than to cross-stage leakage.  A new component becomes a publishable comparison
in three steps (\cref{fig:workflow}): register with a decorator,
declare a factorial sweep in YAML, and run end-to-end to receive a
paired statistical test against the same surrounding stack that
produced every reported number.  The complete study, factorial plus
138 non-factorial ablations plus the 5K--200K scaling sweep, cost
approximately \$5{,}200 in cloud compute and 310 wall-clock hours
(\cref{app:compute}).

\section{Experimental Design}
\label{sec:design}
\label{sec:factorial}
\label{sec:benchmarks}
\label{sec:eval-tracks}
\label{sec:metrics}
\label{sec:compute}

\paragraph{Notation.}
\label{sec:scope}
\label{sec:notation-counts}
A \emph{dimension} is one of the eight experimental factors
(\cref{tab:dimensions}); \textsc{default\_standard} is the
framework's stock RAG pipeline (a simple, lowest-cost setting on
each dimension) that serves as the starting baseline for our
experiments; a \emph{profile} is one of the four operational RAG
compositions we distill from our factorial and ablation sweeps as
the best-observed configurations targeting distinct points on the
retrieval-faithfulness-latency envelope (\textsc{Speed},
\textsc{Standard}, \textsc{Balanced}, \textsc{Max-Reliable}; full
specifications in \cref{app:profiles}); a \emph{track}
is one of the two evaluation surfaces (retrieval or end-to-end). We evaluate on both tracks
because retrieval quality (whether the right documents are
retrieved) and answer faithfulness (whether the generated answer
is grounded in the retrieved context) measure distinct properties
of a RAG pipeline, and different applications weight them
differently. We report the joint quality-latency envelope as a
hypervolume $\mathrm{HV}(L_{\mathrm{ref}}) = \mathrm{NDCG@10}
\times \mathrm{Faith} \times \max(0, 1 - \mathrm{latency}/L_{\mathrm{ref}})$
~\citep{zitzler2003performance} at $L_{\mathrm{ref}}{=}30$\,s (a
fixed response-time reference; \cref{tab:pareto}); HV values under
other fixed references are in \cref{tab:h2-hv-sensitivity}. We
report marginal effect sizes as $\eta^2$; larger values indicate
stronger associations with variation in the metric (full statistics
in \cref{app:faith-anova,app:probes}).

\paragraph{Factorial design.} The configuration space spans eight
experimental dimensions, with 2--5 levels each (\cref{fig:overview}b,
\cref{tab:dimensions}).  The full factorial of
$4{\times}4{\times}4{\times}3{\times}5{\times}4{\times}2{\times}4 = 30{,}720$
configurations would exceed $\sim$40{,}000 GPU-hours and
$\sim$\$600{,}000 in LLM API costs (extrapolating from the executed
167-configuration budget of 247 GPU-hours and \$5{,}200;
\cref{app:compute}), roughly two orders of magnitude beyond our
resources, so we draw a 242-run D-optimal subset via Fedorov
coordinate exchange~\citep{fedorov1972theory}, sized to keep all
main effects and all 28 dimension-pair interactions estimable
under our budget. Of the 242 planned runs, 167 were executed: 63
GraphRAG-enriched runs were dropped after an isolated ablation
showed no measurable NDCG@10 change despite the ${\sim}2\times$
latency the enrichment added
(\cref{tab:graphrag-ablation}), and 12 failed on timeout limits.
Design-matrix construction, aliasing diagnostics, and the
identifiability audit for the pre-declared joint 28-two-way test
appear in \cref{app:design-method,app:aliasing,app:h1-reproducibility}.
Each dimension uses a simple, lowest-cost baseline; pre-registration
deviations appear in \cref{tab:prereg-delta}.

\paragraph{A unified multi-domain corpus, two evaluation tracks.}
Production RAG deployments operate against heterogeneous knowledge
bases where a single retriever must disambiguate biomedical,
financial, scientific, argumentative, and multi-hop queries from one
shared vector store.  Per-benchmark evaluation (the
BEIR default; \citealp{thakur2021beir}) under-tests this cross-domain
distractor pressure.  We therefore \emph{unify} our eight evaluation
benchmarks into one corpus: five BEIR sets cover retrieval (NFCorpus,
\citealp{boteva2016nfcorpus}; FiQA, \citealp{maia2018fiqa}; SciFact,
\citealp{wadden2020scifact}; ArguAna, \citealp{wachsmuth2018arguana};
HotpotQA, \citealp{yang2018hotpotqa}) and three QA sets cover generation
(SQuAD~v2, \citealp{rajpurkar2018know}; MuSiQue,
\citealp{trivedi2022musique}), yielding 149 stratified queries
per configuration (\cref{app:datasets,app:eval-design,tab:benchmarks,tab:qa-benchmarks,tab:eval-tracks}).  Each
configuration is evaluated on two tracks: \emph{retrieval} (BEIR
qrels over 5{,}132 documents, 93 queries) and \emph{end-to-end
generation} (the same corpus augmented with gold passages, 5{,}566
documents, 56 queries).  Beyond the 167-run factorial, we
additionally ran 138 non-factorial ablation configurations across 24
targeted experiments spanning rerankers, embeddings, chunkers, query
augmentation, and the number of agentic verification iterations
(\cref{app:experiment-inventory}).
Together with the corpus-scaling sweep ($5$K--$200$K), the study
records approximately 358{,}215 individual query evaluations,
including repeated RAGAS evaluation tasks for the same queries
(\cref{app:notation-counts}).

\paragraph{Metrics.}
End-to-end generation runs report faithfulness and answer relevance
(RAGAS v0.4~\citep{es2023ragas} judged by GPT-5.1);
NDCG@10, exact match, token-level F1, and p50/p95 latency and
\$/query; GPT-5.1 also generates answers for all configurations.
NDCG@10 follows the BEIR convention~\citep{thakur2021beir}: retrieved
chunks are de-duplicated by parent document ID before scoring, so the
metric matches the document-level qrel grain.
A two-annotator human evaluation validates RAGAS faithfulness
(\cref{app:human-eval,app:human-eval-btl,tab:human-agreement}), and a two-judge
sensitivity analysis (GPT-5.1 vs.\ Claude Sonnet~4.5) shows that
judge choice shifts absolute faithfulness but preserves profile
ranking, so we treat single-judge point estimates as approximate
and rely on ordinal comparisons for all headline claims
(\cref{app:judge-comparison-detail}). A third-judge robustness check
asks Claude Opus~4.7 to score 220 question-answer-context examples
using the same scoring instructions (\cref{app:third-judge-robustness}).
Its moderate agreement with the primary judge (Pearson $r{=}0.44$,
Cohen's $\kappa{=}0.35$) supports the same conclusion: relative trends
persist, although exact faithfulness scores depend on the judge.
Metric definitions and judge-bias mitigations in
\cref{app:metrics}.

\paragraph{Pre-declared hypotheses.}
Most published RAG ablations tune one component at a time on a
single benchmark at $\leq 50$K corpus, and report the resulting gain
as a recommended configuration.  We pre-declared~\citep{vanmiltenburg2021preregister,forde2019scientific} three hypotheses
that together test whether this methodology is sound, by asking
whether RAG pipelines can be designed compositionally:
\textbf{H1} tests \emph{additivity} (the assumption behind isolated
single-component ablations), \textbf{H2} tests \emph{profile
generality} (the assumption that a single balanced recipe deploys
across targets), and \textbf{H3} tests \emph{scale invariance} (the
assumption that rankings measured on small corpora persist at larger
corpus scales).  The full pre-registration (operational definitions,
statistical thresholds, exclusion criteria, post-hoc corrections) is
reproduced in \cref{app:prereg-doc}.
\begin{enumerate}[nosep,leftmargin=1.5em]
  \item[\textbf{H1}] \textbf{Components do not \emph{necessarily} combine additively.}
    H1 tests whether two-way interactions among the eight experimental
    dimensions explain at least $20\%$ of faithfulness variance. If they
    do, gains measured in isolation cannot be assumed to persist when
    components are combined.
  \item[\textbf{H2}] \textbf{A single RAG composition can span multiple deployment targets.}
    RAG designs face a Pareto tradeoff: retrieval quality,
    faithfulness, and response time cannot be maximized jointly,
    so practitioners typically re-run configuration search per
    deployment. H2 asks whether an intermediate RAG composition
    sweeps a wider quality\,/\,latency\,/\,cost hypervolume than
    adding more advanced components does, on a majority of
    datasets, covering enough of the Pareto surface that one
    published recipe serves multiple targets. Rejecting H2 means
    each deployment must hand-tune its own operating point
    (formalism and correction: \cref{app:prereg-doc}).
  \item[\textbf{H3}] \textbf{Composition rankings shift with corpus scale.}
    Most RAG ablations in the literature run at 5K to 50K
    documents, but production deployments often serve corpora
    an order of magnitude larger. H3 asks whether the ranking of
    RAG compositions at 5K matches their ranking at 200K on the
    same corpus; if it does not, small-corpus rankings do not
    transfer across scale, and design decisions measured at one
    scale must be revalidated at the target scale. Confirming
    H3 means small-corpus benchmarks are methodologically suspect
    as scale-invariant selectors of RAG designs; rejecting it
    validates them as predictive across the scales tested
    (per-composition log-linear fits, slope-range threshold,
    bootstrap CIs, and Kendall's $\tau$ rank-flip test in
    \cref{app:prereg-doc}).
\end{enumerate}

\section{Results}
\label{sec:results}

\begin{figure*}[t]
  \centering
  \begin{minipage}[c]{0.50\textwidth}
    \centering
    \includegraphics{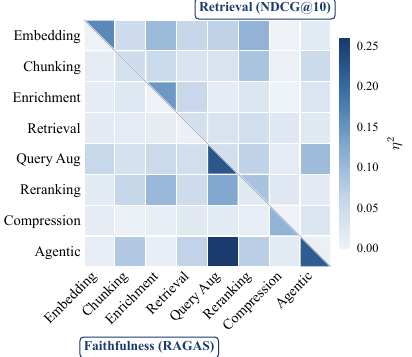}%
  \end{minipage}\hfill
  \begin{minipage}[c]{0.47\textwidth}
  \caption{Main effects and pairwise interactions across eight RAG
  pipeline factors. The upper triangle shows retrieval
  ($167$ configurations, NDCG@10), the lower triangle faithfulness
  ($122$ configurations, RAGAS). Darker cells mean stronger effects;
  an interaction cell shows how much one factor's effect depends on
  another factor's setting. Color encodes marginal (unadjusted)
  $\eta^2$, the share of variance a factor or interaction explains,
  so the two metrics can be compared descriptively. Embedding and
  enrichment matter most for retrieval, whereas query augmentation
  and agentic gating matter most for faithfulness. Because the
  D-optimal design is unbalanced, values can overlap and should not
  be summed into a total explained variance.}
  \label{fig:anova}
  \end{minipage}
\end{figure*}

\paragraph{H1: the benefit of a RAG technique depends on the complete RAG pipeline and the target metric.}
\label{sec:interactions}
\label{sec:component-impact}
\label{sec:factorial-probes}
\label{sec:factorial-anova}
\label{sec:mmr-mechanism}
\label{sec:faith-anova}
\label{sec:interaction-scope}
Across the factorial and targeted tests, some techniques reinforce
one another, some add roughly as expected, and others overlap or
interfere (\cref{fig:anova}). Thus, improvements measured by changing
one pipeline stage at a time do not reliably add up in a complete RAG
pipeline.  We quantify this by comparing the gain predicted from
separately measured improvements with the gain the combined pipeline
actually delivers.  This needs only the configurations the study already
runs, whereas estimating all $28$ pairwise interactions directly requires
$232$ parameters, more than a design of this size can support
(\cref{app:h1-reproducibility}). The three largest separately measured retrieval gains, from
cross-encoder reranking, disabling MMR, and document-awareness
prefixes, predict a $+46.6\%$ lift over the
\textsc{default\_standard} baseline
(NDCG@10 $=0.412$;
\cref{tab:ablation-full}).  On this same split a BM25-only
retriever scores $0.403$ and a recommended off-the-shelf dense encoder
$0.436$ (\cref{tab:external-exec}), so this is a realistic starting
point, not a weak one. The \textsc{Balanced} pipeline,
which combines these techniques, realizes only $+17.2\%$, a $63\%$
shortfall relative to the three-technique prediction. Including all
five separately positive retrieval techniques raises the shortfall to
$72\%$ (\cref{fig:composition}); across 17 tested combinations, it
ranges from $7\%$ to $72\%$, with a median of $41\%$
(\cref{app:composition,app:dropout}). The ablations suggest partial
functional overlap: reranking already provides some of the filtering
benefit obtained by disabling MMR, leaving less room for the two
techniques to help when combined. The target metric matters as well:
embedding, enrichment, and reranking dominate the retrieval-quality
metric, whereas query augmentation and agentic gating dominate the
answer-faithfulness metric, which measures whether generated claims
are supported by the retrieved context. Across matched pipeline
configurations, the two metrics are essentially unrelated
(\cref{fig:metric-decoupling}; pooled Pearson $r=-0.01$, $p=0.91$):
$5\%$ of configurations combine high faithfulness with low retrieval
quality, and $6\%$ show high retrieval quality with low faithfulness.
The same near-zero relationship appears when each benchmark is
examined separately, removing the concern that the near-zero
correlation is caused by pooling benchmarks (\cref{app:faith-anova}).
This indicates that RAG pipeline selection must evaluate both metrics:
selecting a pipeline based only on the retrieval metric can yield strong
retrieval but weakly grounded answers, while selecting one based only on
the faithfulness metric can hide poor retrieval.

\paragraph{H1: retrieval quality is most strongly associated with embedding and enrichment.}
On the retrieval track (\cref{fig:anova}, upper triangle; $n{=}167$
configurations), the largest marginal associations are embedding
($\eta^2 = 0.164$), enrichment ($\eta^2 = 0.144$), and reranking
($\eta^2 = 0.090$). These unadjusted values are used for exploratory
screening and are not additive because the D-optimal design is
unbalanced and the effects share variance. Compression and agentic gate
contribute negligibly ($\eta^2<0.01$ each).  Two two-way interactions
survive both Bonferroni and Benjamini--Hochberg FDR correction:
Embedding\,$\times$\,Enrichment ($\eta^2 = 0.102$) and
Embedding\,$\times$\,Reranking ($\eta^2 = 0.110$). Both survivors
involve the embedding model. We attribute Embedding\,$\times$\,Enrichment
to representation capacity: stronger embedders can encode added
context without diluting the original passage. We attribute
Embedding\,$\times$\,Reranking to candidate-set quality: rerankers
have less room to help when the initial embedding already retrieves
strong candidates (\cref{app:rca-i1,app:rca-i2}). MMR reduces NDCG@10 by $3.6\%$ relative
on average, opposite its usual benefit on broad, heterogeneous corpora
(\cref{tab:dropout}).  We hypothesize that on focused-domain corpora
passages with high lexical and semantic overlap are often the same
supporting evidence, so MMR's diversity penalty demotes them in favor
of topically distant ones (see \cref{app:dropout,app:domain}).

\paragraph{H1: answer faithfulness depends most on query augmentation and agentic gating.}
On the faithfulness track (\cref{fig:anova}, lower triangle), query
augmentation ($\eta^2 = 0.224$) and agentic gating
($\eta^2 = 0.217$) have the largest marginal associations; embedding
and enrichment, the strongest retrieval drivers, are negligible on
faithfulness ($\eta^2 < 0.02$ each). We attribute this contrast to
pipeline stage: query augmentation changes retrieved evidence, whereas
agentic gating filters or verifies it before generation. We test how
these stages compose across all $5{\times}4$ combinations of query
augmentation and agentic gating (\cref{app:e1-executed-results}). The
interaction effect is partial $\eta^2 = 0.033$; 10 of 20 combinations
depart from additivity by at least two percentage points. Under
Speculative-RAG, decomposition is $4.7$ points above the additive
prediction, while query expansion is $4.1$ below it.
\mbox{We hypothesize that decomposition} supplies distinct subproblems for
draft-and-verify, whereas broader expansion can introduce overlapping
or competing evidence. Thus, isolated gains do not reliably predict
composition; these stages should be evaluated jointly
(\cref{tab:e1-cell-interactions}).

Faithfulness shows a different additivity pattern. A
one-factor-at-a-time estimate based on BGE-large embeddings,
hierarchical chunking, hybrid-BM25 retrieval, and Speculative-RAG
gating predicts a $+28\%$ gain over a reference configuration, while
the best full RAG pipeline improves by $+32\%$ over the same reference.
This suggests that some faithfulness gains may combine approximately
as measured. We hypothesize that this occurs when techniques act at
complementary pipeline stages and duplicate less of one another's function
(\cref{tab:composition-faith-exploratory}). Full pairwise results are
in \cref{tab:faith-interactions}.

\begin{vdict}[H1 (additivity): gains measured one component at a time do not carry over when components are combined, and how much is lost depends on the metric.]
Combining components loses most of the predicted retrieval gain and almost
none of the predicted faithfulness gain.  The three strongest
single-component retrieval improvements predict $+47\%$ NDCG@10 and the
pipeline containing all three realizes $+17\%$; the same exercise on
faithfulness predicts $+28\%$ and realizes $+32\%$
(\cref{app:composition}).

Our pre-declared test of H1 was not met, and that outcome fits the same
split.  It asked whether pairwise interactions explain at least $20\%$ of
the variance in faithfulness, the metric on which components combine
nearly additively, and they do not (\cref{app:h1-reproducibility}).  A
focused follow-up, pre-specified before it ran, confirms that the
interaction the broad analysis flagged is real ($p=0.007$) but far below
that bar (\cref{app:e1-focused,app:e1-executed-results}).  New
components should therefore be evaluated inside representative complete
pipelines, on both metrics.
\end{vdict}

\begin{figure*}[t]
  \centering
  \begin{subfigure}[t]{0.48\linewidth}
    \centering
    \includegraphics[width=\linewidth]{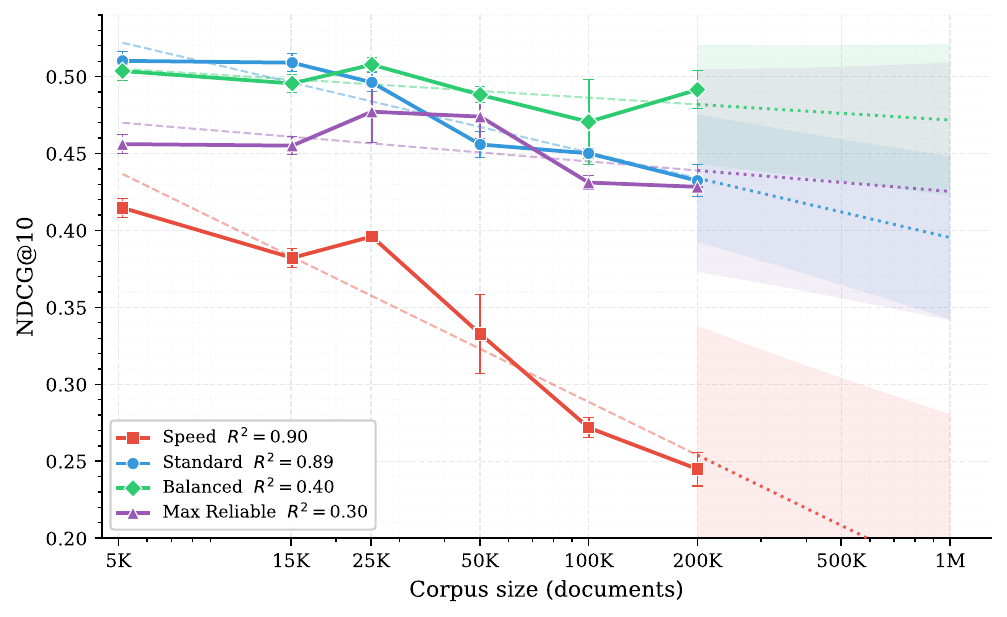}
    \caption{Retrieval NDCG@10: \textsc{Speed} and \textsc{Standard}
    follow a log-linear decay; \textsc{Balanced} and
    \textsc{Max-Reliable} are statistically flat.}
    \label{fig:scaling-retrieval}
  \end{subfigure}\hfill
  \begin{subfigure}[t]{0.48\linewidth}
    \centering
    \includegraphics[width=\linewidth]{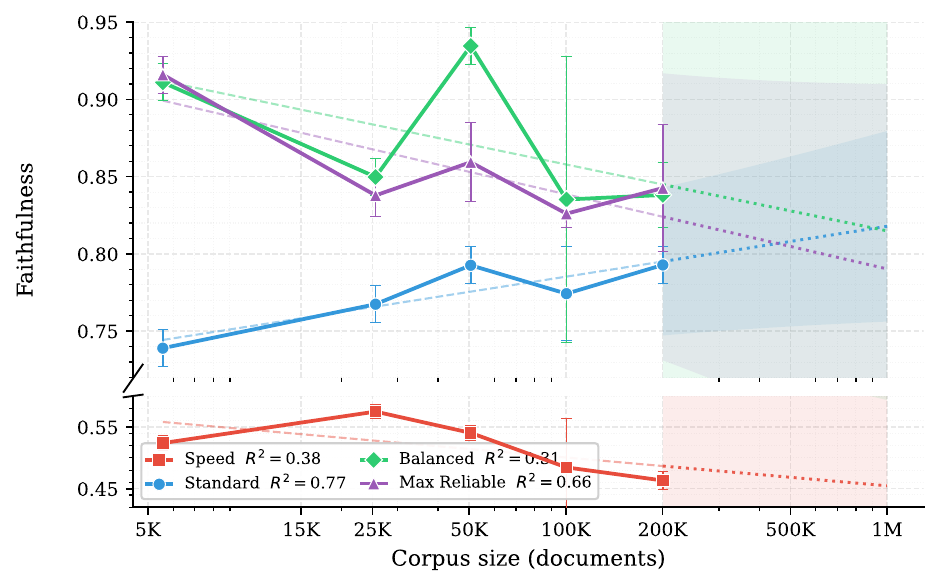}
    \caption{Generation faithfulness: all three quality-focused
    profiles peak at 50K, dip at 100K, and recover from 100K to
    200K.}
    \label{fig:scaling-faithfulness}
  \end{subfigure}
  \caption{Scaling behavior vs.\ corpus size (log scale), four SOTA
  profiles, 3 seeds per size. Dashed: log-linear fit over 5K--200K;
  dotted beyond 200K: \emph{projections, not measurements}. In
  \textbf{(a)}, \textsc{Balanced} overtakes \textsc{Standard}
  between 15K--25K, significant at 200K.}
  \label{fig:scaling-curves}
\end{figure*}

\paragraph{H3: composition rankings shift between 5K and 200K documents.}
\label{sec:scaling}
\label{sec:scaling-method}
\label{sec:scaling-retrieval}
\label{sec:scaling-e2e}
\label{sec:scaling-laws}
We extend the 5{,}132-document base corpus with scaling increments
(15K/25K/50K/100K/200K) under the same proportional sampling
protocol, with no distractor overlapping any evaluation query's
relevant set. For 25K--200K, repeated seeded executions resample
retrieval and generation stochasticity but not corpus composition;
distractors are sampled from the same five BEIR source domains
as the base corpus (full protocol in
\cref{app:scaling-details}). NDCG@10 across six corpus sizes
separates into two regimes (\cref{fig:scaling-retrieval}):
\textsc{Speed} and \textsc{Standard} degrade log-linearly, while
\textsc{Balanced} and \textsc{Max-Reliable} are approximately flat.
The regimes cross between 15K and 25K. \textsc{Standard} leads at
the small sizes (5K, 15K); \textsc{Balanced} leads from 25K onward.
At 200K, \textsc{Balanced} exceeds \textsc{Standard} by $13.7\%$
relative ($p=0.003$; full per-size table in
\cref{app:scaling-crossover-table}).

We attribute the flatter retrieval-NDCG degradation at scale of
\textsc{Balanced} to several retrieval-stage additions relative to
\textsc{Standard}: broader chunk retention, hierarchical document
partitioning, and post-retrieval merging (full list in
\cref{app:profiles}). Together, these sustain a denser,
higher-signal top-$k$ as distractors dilute the pool. The extractive compression and
document-awareness prefixes that \textsc{Balanced} additionally enables
act post-retrieval and pre-generation respectively; they primarily
support faithfulness rather than the NDCG@10 gap
(\cref{app:dropout}; examples in
\hyperref[ex:doc-aware-fiqa]{document awareness} and the
\hyperref[ex:scale-crossover]{50K crossover}).  \textsc{Speed} degrades
fastest, which we hypothesize reflects MiniLM-L6's limited
discriminative capacity (384-dim) operating without the lexical
anchoring that BM25 would provide under hybrid retrieval, see
\hyperref[ex:speed-embed-ceiling]{example} and
\cref{tab:scaling-slopes} for results.

\paragraph{H3: faithfulness and retrieval scale differently.}
Despite NDCG@10 falling as gold passages dilute among distractors,
faithfulness does not decline with corpus scale. All three
quality-focused profiles peak around 50K, dip at 100K, and recover
from 100K to 200K
(\cref{fig:scaling-faithfulness,tab:scaling-e2e}).
We hypothesize that a larger corpus expands information coverage,
so that even when the gold document falls below the top-$k$, other
documents contain overlapping or paraphrased content sufficient to
ground a faithful answer; the 100K$\to$200K recovery across all
three profiles is consistent with this mechanism. Full per-profile
slopes and 1M projections in
\cref{tab:scaling-slopes,tab:scaling-e2e,fig:scaling-projection,tab:scaling-projection};
latency scaling at each corpus size in \cref{fig:latency-scaling}.

\begin{vdict}[H3: composition rankings shift between 5K and 200K.]
\textbf{Supported.} \textsc{Standard} leads at 5K and 15K;
\textsc{Balanced} leads from 25K through 200K, where it exceeds
\textsc{Standard} by $13.7\%$ relative ($p=0.003$, three seeds).
The 25K and 100K gaps are directionally consistent but underpowered.
Small-corpus rankings should therefore be revalidated at the target
scale (\cref{tab:rank-crossover-per-size,app:pvalue-reconciliation}).
\end{vdict}

\paragraph{H2: complete RAG profiles expose a Pareto tradeoff.}
\label{sec:pareto}
\label{sec:profile-comparison}
\label{sec:profile-desc}
We evaluate four profiles on a held-out SQuAD~v2, MuSiQue, and
HotpotQA mixture across three seeds (\cref{tab:pareto}). No profile
leads all metrics: \textsc{Speed} is fastest; \textsc{Standard}'s
direct route leads NDCG@10; the agentic \textsc{Balanced} and
\textsc{Max-Reliable} profiles use three verification iterations and
lead faithfulness.

\begin{table}[t]
\centering
\caption{End-to-end profile comparison on a SQuAD~v2 / MuSiQue /
  HotpotQA mixture over three seeds (GPT-5.1 judge; faithfulness uses
  each profile's deployment-native answer prompt; per-profile regime
  and joint-HV sensitivity to that choice in
  \cref{tab:prompt-regime,app:pareto-regime-sensitivity}). Mean $\pm$ std;
  \textbf{bold}: column best; \colorbox{lasagnagreen}{green}:
  highest joint retrieval--faithfulness--latency volume.
  $\mathrm{HV}(L_{\mathrm{ref}}) =
  \mathrm{NDCG@10} \times \mathrm{Faith} \times
  \max(0,1-\mathrm{latency}/L_{\mathrm{ref}})$~\citep{zitzler2003performance},
  with $L_{\mathrm{ref}}=30$\,s; reference-latency sensitivity is in
  \cref{tab:h2-hv-sensitivity}.}
\label{tab:pareto}
\footnotesize
\begin{tabular}{@{}lcccc@{}}
\toprule
Profile & NDCG@10 & Faithfulness & Lat.\,mean (s) & HV ($30$\,s ref) \\
\midrule
\rowcolor{lasagnagreen}
\textsc{Standard}     & \textbf{0.860\,$\pm$\,0.003} & 0.739\,$\pm$\,0.016 & 5.30\,$\pm$\,0.05 & \textbf{0.523} \\
\textsc{Balanced}     & 0.840\,$\pm$\,0.003 & 0.911\,$\pm$\,0.014 & 12.25\,$\pm$\,3.55 & 0.453 \\
\textsc{Max-Reliable} & 0.804\,$\pm$\,0.004 & \textbf{0.916\,$\pm$\,0.022} & 14.16\,$\pm$\,6.44 & 0.389 \\
\textsc{Speed}        & 0.541\,$\pm$\,0.000 & 0.524\,$\pm$\,0.031 & \textbf{0.70\,$\pm$\,0.02} & 0.277 \\
\bottomrule
\end{tabular}
\end{table}

\noindent At the fixed $30$\,s response-time reference,
\textsc{Standard} has the highest joint hypervolume ($0.523$). Within
the agentic pair, \textsc{Balanced} improves NDCG@10 by $4.5\%$ and
response time by $13\%$, with only a $0.005$ faithfulness difference.
We attribute part of this tradeoff to the execution route: direct
generation favors response time, while iterative verification favors
faithfulness. Thus, additional components can improve one metric
without improving the complete RAG pipeline, so all three metrics
should be evaluated jointly.

\begin{vdict}[H2 (single-composition generality): no complete RAG profile is best on every metric.]
At the fixed $30$\,s reference, \textsc{Standard} achieves the
highest joint hypervolume ($0.523$; \cref{tab:pareto}). The agentic
profiles trade more response time for faithfulness; among them,
\textsc{Balanced} nearly matches \textsc{Max-Reliable}'s
faithfulness while improving retrieval quality and response time.
Thus, the best configuration depends on the target metrics, and
adding components does not guarantee a better pipeline.
\end{vdict}

\paragraph{Supporting evidence: what helps and what does not.}
A sweep of 138 configurations across 24 targeted experiments yields
several practical findings (\cref{app:experiment-inventory}). Across
20 configurations, chunker choice spans $0.126$ NDCG@10, a $36\%$
relative difference larger than most factorial main effects
(\cref{tab:chunking-ablation-detail}). Thus, preprocessing choices
should also be tested within the complete pipeline. The 2019 MS-MARCO
MiniLM-L6 cross-encoder~\citep{nogueira2019monobert} ranks above all
four recent rerankers tested (\cref{tab:reranker-inventory}); we
speculate that its training aligns especially well with our focused
domains. Matryoshka truncation of BGE-M3 to 768
dimensions~\citep{kusupati2022mrl} reduces vector storage
and search cost by $25\%$ while improving NDCG@10 by $2.3\%$
(\cref{tab:embedding-inventory}). We attribute the gain to the
Matryoshka training objective, which concentrates useful signal in
leading dimensions; encoder cost is unchanged because the full vector
is computed first. On generation, faithfulness
is highest with hallucination and answer-quality gates disabled:
gate-triggered refinement can remove supporting evidence, including 14
of 15 chunks in one multi-hop example
(\cref{tab:generation-ablations-detail}). HyDE raises NDCG@10 by
$3.1\%$ but doubles first-hop latency, reducing the latency-adjusted
gain to $0.9\%$, so its value depends on the latency budget
(\cref{app:profile-guide}).

\section{Conclusion}
\label{sec:conclusion}

We study how modern RAG techniques compose across pipeline stages,
metrics, and corpus scales using \textsc{Lasagna}, a stage-isolated
framework, and $450$ configurations across eight benchmarks from 5K
to 200K documents. Gains measured one component at a time do not
necessarily persist in complete pipelines; retrieval quality and
answer faithfulness favor different choices; and the best composition
changes with corpus scale. New techniques should be tested in
representative complete pipelines, on both metrics, and at the
intended scale. We release \textsc{Lasagna} to make such studies
practical and reproducible, and to establish compositional
evaluation as the reporting standard for new RAG techniques rather
than the isolated single-component ablations common today.

\subsection*{AI use statement}
In this work, we used generative AI tools for designing or providing feedback on
research methodology and experiments, implementing methods, and interpreting
results. We have not used generative AI tools for generating synthetic data sets,
developing theoretical models or conceptual frameworks, proposing or refining
hypotheses, cleaning or reformatting datasets, or translation, and formulating
mathematical claims or proofs and qualitative or thematic data analysis are not
applicable to this work. Additionally, we used generative AI tools for creating
and editing software code and figures, editing the paper to improve readability,
and formatting references. We have reviewed all AI-assisted
work: AI-assisted code was reviewed and tested by the authors, analysis results
were re-derived from the raw result files, and edited text, numbers, and
citations were checked against the source tables and references. We take
responsibility for the final content of this work, including text, claims or
artifacts produced with the aid of generative AI.
Separately, LLMs are components of the system we study: they generate answers
and act as RAGAS judges (GPT-5.1, Claude Sonnet~4.5), as documented in the
Methods and Appendices.

\subsection*{Ethics statement}
LASAGNA is an evaluation framework operating over publicly released
retrieval benchmarks; no human subjects, personal data, or private
corpora were used. LLM API calls totalled approximately \$5{,}200 and
310 wall-clock hours (\cref{app:compute}); we report per-configuration
cost and latency to enable reproducibility on constrained budgets.
Because the framework lowers the cost of composing RAG pipelines,
users deploying it in domain applications requiring corpus-specific
validation should re-validate our profile candidates on their own
corpus before deployment.  Our headline profile comparisons rest on
one constructed English multi-domain test bed with $\leq 93$
retrieval and $\leq 200$ generation queries per experiment, so the
profiles are best interpreted as candidates for further validation,
not prescriptions.

\subsection*{Reproducibility statement}
The \textsc{Lasagna} framework code is released at \url{https://github.com/dolev31/lasagna}. The paper-specific artifacts cited by path in the appendices (the D-optimal design matrix, per-configuration profiles, per-query raw scores, and analysis scripts) are not part of that release and are available from the authors on request.
\Cref{app:expdesign,app:datasetsfull,app:metricsfull} document experimental design
and pre-registration; the per-experiment provenance manifest (\cref{tab:manifest}),
conformance test verifying stage isolation (\cref{app:isolation}), and full compute
accounting (\cref{app:compute}) support end-to-end reproducibility.  The design
matrix, hypotheses, and decision rules were pre-declared internally and hashed
(SHA-256) prior to execution; we did not file a third-party timestamped
pre-registration (OSF/AsPredicted), so the main text says ``pre-declared''
rather than ``pre-registered''.

\bibliographystyle{iclr2027_conference}
\bibliography{iclr2027_conference}

\begin{thebibliography}{71}
\providecommand{\natexlab}[1]{#1}
\providecommand{\url}[1]{\texttt{#1}}
\expandafter\ifx\csname urlstyle\endcsname\relax
  \providecommand{\doi}[1]{doi: #1}\else
  \providecommand{\doi}{doi: \begingroup \urlstyle{rm}\Url}\fi

\bibitem[{Anthropic}(2024)]{anthropic2024contextual}
{Anthropic}.
\newblock Contextual retrieval.
\newblock \url{https://www.anthropic.com/research/contextual-retrieval}, 2024.

\bibitem[Asai et~al.(2024)Asai, Wu, Wang, Sil, and Hajishirzi]{asai2024selfrag}
Akari Asai, Zeqiu Wu, Yizhong Wang, Avirup Sil, and Hannaneh Hajishirzi.
\newblock Self-{RAG}: Learning to retrieve, generate, and critique through
  self-reflection.
\newblock In \emph{The Twelfth International Conference on Learning
  Representations}, 2024.
\newblock URL \url{https://openreview.net/forum?id=hSyW5go0v8}.

\bibitem[Atkinson et~al.(2007)Atkinson, Donev, and Tobias]{atkinson2007optimum}
Anthony Atkinson, Alexander Donev, and Randall Tobias.
\newblock \emph{Optimum Experimental Designs, with SAS}.
\newblock Oxford University Press, 05 2007.
\newblock ISBN 978-0-19-929659-0.
\newblock \doi{10.1093/oso/9780199296590.001.0001}.

\bibitem[Auer et~al.(2024)Auer, Lysak, Nassar, Dolfi, Livathinos, Vagenas,
  Ramis, Omenetti, Lindlbauer, Dinkla, et~al.]{auer2024docling}
Christoph Auer, Maksym Lysak, Ahmed Nassar, Michele Dolfi, Nikolaos Livathinos,
  Panos Vagenas, Cesar~Berrospi Ramis, Matteo Omenetti, Fabian Lindlbauer,
  Kasper Dinkla, et~al.
\newblock Docling technical report.
\newblock \emph{arXiv preprint arXiv:2408.09869}, 2024.

\bibitem[Barnett et~al.(2024)Barnett, Kurniawan, Thudumu, Brannelly, and
  Abdelrazek]{barnett2024seven}
Scott Barnett, Stefanus Kurniawan, Srikanth Thudumu, Zach Brannelly, and
  Mohamed Abdelrazek.
\newblock Seven failure points when engineering a retrieval augmented
  generation system.
\newblock In \emph{Proceedings of the IEEE/ACM 3rd International Conference on
  AI Engineering - Software Engineering for AI}, CAIN '24, pp.\  194–199, New
  York, NY, USA, 2024. Association for Computing Machinery.
\newblock ISBN 9798400705915.
\newblock \doi{10.1145/3644815.3644945}.
\newblock URL \url{https://doi.org/10.1145/3644815.3644945}.

\bibitem[Boteva et~al.(2016)Boteva, Gholipour, Sokolov, and
  Riezler]{boteva2016nfcorpus}
Vera Boteva, Demian Gholipour, Artem Sokolov, and Stefan Riezler.
\newblock A full-text learning to rank dataset for medical information
  retrieval.
\newblock In \emph{Advances in Information Retrieval (ECIR)}, 2016.

\bibitem[Bouthillier et~al.(2021)Bouthillier, Delaunay, Bronzi, Trofimov,
  Nichyporuk, Szeto, Mohammadi~Sepahvand, Raff, Madan, Voleti, Ebrahimi~Kahou,
  Michalski, Arbel, Pal, Varoquaux, and Vincent]{bouthillier2021accounting}
Xavier Bouthillier, Pierre Delaunay, Mirko Bronzi, Assya Trofimov, Brennan
  Nichyporuk, Justin Szeto, Nazanin Mohammadi~Sepahvand, Edward Raff, Kanika
  Madan, Vikram Voleti, Samira Ebrahimi~Kahou, Vincent Michalski, Tal Arbel,
  Chris Pal, Gael Varoquaux, and Pascal Vincent.
\newblock Accounting for variance in machine learning benchmarks.
\newblock In A.~Smola, A.~Dimakis, and I.~Stoica (eds.), \emph{Proceedings of
  Machine Learning and Systems}, volume~3, pp.\  747--769, 2021.
\newblock URL
  \url{https://proceedings.mlsys.org/paper_files/paper/2021/file/0184b0cd3cfb185989f858a1d9f5c1eb-Paper.pdf}.

\bibitem[Box et~al.(2005)Box, Hunter, and Hunter]{box2005statistics}
G.~Box, J.~Hunter, and W.G. Hunter.
\newblock Statistics for experimenters: Design, innovation, and discovery.
\newblock \emph{AMC}, 10, 01 2005.

\bibitem[Bruch et~al.(2023)Bruch, Gai, and Ingber]{bruch2023analysis}
Sebastian Bruch, Siyu Gai, and Amir Ingber.
\newblock An analysis of fusion functions for hybrid retrieval.
\newblock \emph{ACM Transactions on Information Systems}, 42\penalty0
  (1):\penalty0 1--35, 2023.

\bibitem[Carbonell \& Goldstein(1998)Carbonell and Goldstein]{carbonell1998mmr}
Jaime Carbonell and Jade Goldstein.
\newblock The use of mmr, diversity-based reranking for reordering documents
  and producing summaries.
\newblock In \emph{Proceedings of the 21st annual international ACM SIGIR
  conference on Research and development in information retrieval}, pp.\
  335--336, 1998.

\bibitem[Chen et~al.(2024{\natexlab{a}})Chen, Xiao, Zhang, Luo, Lian, and
  Liu]{chen2024bgem3}
Jianlv Chen, Shitao Xiao, Peitian Zhang, Kun Luo, Defu Lian, and Zheng Liu.
\newblock Bge m3-embedding: Multi-lingual, multi-functionality,
  multi-granularity text embeddings through self-knowledge distillation.
\newblock \emph{arXiv preprint arXiv:2402.03216}, 4\penalty0 (5),
  2024{\natexlab{a}}.

\bibitem[Chen et~al.(2024{\natexlab{b}})Chen, Wang, Chen, Yu, Ma, Zhao, Zhang,
  and Yu]{shi2023prop}
Tong Chen, Hongwei Wang, Sihao Chen, Wenhao Yu, Kaixin Ma, Xinran Zhao,
  Hongming Zhang, and Dong Yu.
\newblock Dense x retrieval: What retrieval granularity should we use?
\newblock In \emph{Proceedings of the 2024 Conference on Empirical Methods in
  Natural Language Processing}, pp.\  15159--15177, 2024{\natexlab{b}}.

\bibitem[Chen et~al.(2026)Chen, Yan, Sun, Ma, Zhang, Wang, Yin, Yang, and
  Mao]{chen2025mmoarag}
Yiqun Chen, Lingyong Yan, Weiwei Sun, Xinyu Ma, Yi~Zhang, Shuaiqiang Wang,
  Dawei Yin, Yiming Yang, and Jiaxin Mao.
\newblock Improving retrieval-augmented generation through multi-agent
  reinforcement learning.
\newblock In \emph{The Thirty-ninth Annual Conference on Neural Information
  Processing Systems}, 2026.
\newblock URL \url{https://openreview.net/forum?id=9Ia0KiVAut}.

\bibitem[Cohen(2013)]{cohen1988statistical}
Jacob Cohen.
\newblock \emph{Statistical power analysis for the behavioral sciences}.
\newblock routledge, 2013.

\bibitem[Cormack et~al.(2009)Cormack, Clarke, and Buettcher]{cormack2009rrf}
Gordon~V Cormack, Charles~LA Clarke, and Stefan Buettcher.
\newblock Reciprocal rank fusion outperforms condorcet and individual rank
  learning methods.
\newblock In \emph{Proceedings of the 32nd international ACM SIGIR conference
  on Research and development in information retrieval}, pp.\  758--759, 2009.

\bibitem[Dodge et~al.(2019)Dodge, Gururangan, Card, Schwartz, and
  Smith]{dodge2019importance}
Jesse Dodge, Suchin Gururangan, Dallas Card, Roy Schwartz, and Noah~A. Smith.
\newblock Show your work: Improved reporting of experimental results.
\newblock In Kentaro Inui, Jing Jiang, Vincent Ng, and Xiaojun Wan (eds.),
  \emph{Proceedings of the 2019 Conference on Empirical Methods in Natural
  Language Processing and the 9th International Joint Conference on Natural
  Language Processing (EMNLP-IJCNLP)}, pp.\  2185--2194, Hong Kong, China,
  November 2019. Association for Computational Linguistics.
\newblock \doi{10.18653/v1/D19-1224}.
\newblock URL \url{https://aclanthology.org/D19-1224/}.

\bibitem[Edge et~al.(2024)Edge, Trinh, Cheng, Bradley, Chao, Mody, Truitt,
  Metropolitansky, Ness, and Larson]{edge2024graphrag}
Darren Edge, Ha~Trinh, Newman Cheng, Joshua Bradley, Alex Chao, Apurva Mody,
  Steven Truitt, Dasha Metropolitansky, Robert~Osazuwa Ness, and Jonathan
  Larson.
\newblock From local to global: A graph rag approach to query-focused
  summarization.
\newblock \emph{arXiv preprint arXiv:2404.16130}, 2024.

\bibitem[Es et~al.(2024)Es, James, Anke, and Schockaert]{es2023ragas}
Shahul Es, Jithin James, Luis~Espinosa Anke, and Steven Schockaert.
\newblock Ragas: Automated evaluation of retrieval augmented generation.
\newblock In \emph{Proceedings of the 18th conference of the european chapter
  of the association for computational linguistics: system demonstrations},
  pp.\  150--158, 2024.

\bibitem[Fan et~al.(2024)Fan, Ding, Ning, Wang, Li, Yin, Chua, and
  Li]{fan2024survey}
Wenqi Fan, Yujuan Ding, Liangbo Ning, Shijie Wang, Hengyun Li, Dawei Yin,
  Tat-Seng Chua, and Qing Li.
\newblock A survey on rag meeting llms: Towards retrieval-augmented large
  language models.
\newblock In \emph{Proceedings of the 30th ACM SIGKDD conference on knowledge
  discovery and data mining}, pp.\  6491--6501, 2024.

\bibitem[Fedorov et~al.(1972)Fedorov, Studden, and Klimko]{fedorov1972theory}
Valerii Fedorov, W.~J. Studden, and Eugene~M. Klimko.
\newblock Theory of optimal experiments.
\newblock \emph{Biometrika}, 59:\penalty0 697, 1972.
\newblock URL \url{https://api.semanticscholar.org/CorpusID:121909870}.

\bibitem[Forde \& Paganini(2019)Forde and Paganini]{forde2019scientific}
Jessica~Zosa Forde and Michela Paganini.
\newblock The scientific method in the science of machine learning.
\newblock In \emph{ICLR 2019 Workshop on Debugging Machine Learning Models},
  2019.
\newblock URL
  \url{https://debug-ml-iclr2019.github.io/cameraready/DebugML-19_paper_26.pdf}.
\newblock Presented at the ICLR 2019 Debugging Machine Learning Models
  Workshop.

\bibitem[Formal et~al.(2021)Formal, Piwowarski, and
  Clinchant]{formal2021splade}
Thibault Formal, Benjamin Piwowarski, and St{\'e}phane Clinchant.
\newblock Splade: Sparse lexical and expansion model for first stage ranking.
\newblock In \emph{Proceedings of the 44th International ACM SIGIR Conference
  on Research and Development in Information Retrieval}, pp.\  2288--2292,
  2021.

\bibitem[Formal et~al.(2022)Formal, Lassance, Piwowarski, and
  Clinchant]{formal2022splade_v2}
Thibault Formal, Carlos Lassance, Benjamin Piwowarski, and St{\'e}phane
  Clinchant.
\newblock From distillation to hard negative sampling: Making sparse neural ir
  models more effective.
\newblock In \emph{Proceedings of the 45th international ACM SIGIR conference
  on research and development in information retrieval}, pp.\  2353--2359,
  2022.

\bibitem[Friel et~al.(2024)Friel, Belyi, and Sanyal]{friel2024ragbench}
Robert Friel, Masha Belyi, and Atindriyo Sanyal.
\newblock Ragbench: Explainable benchmark for retrieval-augmented generation
  systems.
\newblock \emph{arXiv preprint arXiv:2407.11005}, 2024.

\bibitem[Fu et~al.(2024)Fu, Qin, Yang, Wang, Zhang, Lin, Chen, Zhang, Rajmohan,
  and Zhang]{fu2024autorag}
Jia Fu, Xiaoting Qin, Fangkai Yang, Lu~Wang, Jue Zhang, Qingwei Lin, Yubo Chen,
  Dongmei Zhang, Saravan Rajmohan, and Qi~Zhang.
\newblock {A}uto{RAG}-{HP}: Automatic online hyper-parameter tuning for
  retrieval-augmented generation.
\newblock In Yaser Al-Onaizan, Mohit Bansal, and Yun-Nung Chen (eds.),
  \emph{Findings of the Association for Computational Linguistics: EMNLP 2024},
  pp.\  3875--3891, Miami, Florida, USA, November 2024. Association for
  Computational Linguistics.
\newblock \doi{10.18653/v1/2024.findings-emnlp.223}.
\newblock URL \url{https://aclanthology.org/2024.findings-emnlp.223/}.

\bibitem[Gao et~al.(2023{\natexlab{a}})Gao, Ma, Lin, and Callan]{gao2023hyde}
Luyu Gao, Xueguang Ma, Jimmy Lin, and Jamie Callan.
\newblock Precise zero-shot dense retrieval without relevance labels.
\newblock In \emph{Proceedings of the 61st Annual Meeting of the Association
  for Computational Linguistics (Volume 1: Long Papers)}, pp.\  1762--1777,
  2023{\natexlab{a}}.

\bibitem[Gao et~al.(2023{\natexlab{b}})Gao, Xiong, Gao, Jia, Pan, Bi, Dai, Sun,
  Wang, Wang, et~al.]{gao2023survey}
Yunfan Gao, Yun Xiong, Xinyu Gao, Kangxiang Jia, Jinliu Pan, Yuxi Bi, Yixin
  Dai, Jiawei Sun, Haofen Wang, Haofen Wang, et~al.
\newblock Retrieval-augmented generation for large language models: A survey.
\newblock \emph{arXiv preprint arXiv:2312.10997}, 2\penalty0 (1):\penalty0 32,
  2023{\natexlab{b}}.

\bibitem[Guti{\'e}rrez et~al.(2024)Guti{\'e}rrez, Shu, Gu, Yasunaga, and
  Su]{gutierrez2024hipporag}
Bernal~J Guti{\'e}rrez, Yiheng Shu, Yu~Gu, Michihiro Yasunaga, and Yu~Su.
\newblock Hipporag: Neurobiologically inspired long-term memory for large
  language models.
\newblock \emph{Advances in neural information processing systems},
  37:\penalty0 59532--59569, 2024.

\bibitem[Jiang et~al.(2023{\natexlab{a}})Jiang, Wu, Lin, Yang, and
  Qiu]{jiang2023llmlingua}
Huiqiang Jiang, Qianhui Wu, Chin-Yew Lin, Yuqing Yang, and Lili Qiu.
\newblock Llmlingua: Compressing prompts for accelerated inference of large
  language models.
\newblock In \emph{Proceedings of the 2023 conference on empirical methods in
  natural language processing}, pp.\  13358--13376, 2023{\natexlab{a}}.

\bibitem[Jiang et~al.(2023{\natexlab{b}})Jiang, Xu, Gao, Sun, Liu, Dwivedi-Yu,
  Yang, Callan, and Neubig]{jiang2023flare}
Zhengbao Jiang, Frank~F Xu, Luyu Gao, Zhiqing Sun, Qian Liu, Jane Dwivedi-Yu,
  Yiming Yang, Jamie Callan, and Graham Neubig.
\newblock Active retrieval augmented generation.
\newblock In \emph{Proceedings of the 2023 conference on empirical methods in
  natural language processing}, pp.\  7969--7992, 2023{\natexlab{b}}.

\bibitem[Jin et~al.(2025)Jin, Zhu, Dou, Dong, Yang, Zhang, Zhao, Yang, and
  Wen]{jin2024flashrag}
Jiajie Jin, Yutao Zhu, Zhicheng Dou, Guanting Dong, Xinyu Yang, Chenghao Zhang,
  Tong Zhao, Zhao Yang, and Ji-Rong Wen.
\newblock Flashrag: A modular toolkit for efficient retrieval-augmented
  generation research.
\newblock In \emph{Companion Proceedings of the ACM on Web Conference 2025},
  WWW '25, pp.\  737–740, New York, NY, USA, 2025. Association for Computing
  Machinery.
\newblock ISBN 9798400713316.
\newblock \doi{10.1145/3701716.3715313}.
\newblock URL \url{https://doi.org/10.1145/3701716.3715313}.

\bibitem[Kaplan et~al.(2020)Kaplan, McCandlish, Henighan, Brown, Chess, Child,
  Gray, Radford, Wu, and Amodei]{kaplan2020scaling}
Jared Kaplan, Sam McCandlish, Thomas Henighan, Tom~B. Brown, Benjamin Chess,
  Rewon Child, Scott Gray, Alec Radford, Jeff Wu, and Dario Amodei.
\newblock Scaling laws for neural language models.
\newblock \emph{ArXiv}, abs/2001.08361, 2020.
\newblock URL \url{https://api.semanticscholar.org/CorpusID:210861095}.

\bibitem[Khattab \& Zaharia(2020)Khattab and Zaharia]{khattab2020colbert}
Omar Khattab and Matei Zaharia.
\newblock Colbert: Efficient and effective passage search via contextualized
  late interaction over bert.
\newblock In \emph{Proceedings of the 43rd International ACM SIGIR conference
  on research and development in Information Retrieval}, pp.\  39--48, 2020.

\bibitem[Khattab et~al.(2024)Khattab, Singhvi, Maheshwari, Zhang, Santhanam,
  Vardhamanan, Haq, Sharma, Joshi, Moazam, Miller, Zaharia, and
  Potts]{khattab2024dspy}
Omar Khattab, Arnav Singhvi, Paridhi Maheshwari, Zhiyuan Zhang, Keshav
  Santhanam, Sri Vardhamanan, Saiful Haq, Ashutosh Sharma, Thomas~T. Joshi,
  Hanna Moazam, Heather Miller, Matei Zaharia, and Christopher Potts.
\newblock Dspy: Compiling declarative language model calls into self-improving
  pipelines.
\newblock In \emph{The Twelfth International Conference on Learning
  Representations}, 2024.

\bibitem[Kim et~al.(2024)Kim, Kim, Han, and Eibich]{kim2024autorag}
Dongkyu Kim, Byoungwook Kim, Donggeon Han, and Matou{\v{s}} Eibich.
\newblock Autorag: Automated framework for optimization of retrieval augmented
  generation pipeline.
\newblock \emph{arXiv preprint arXiv:2410.20878}, 2024.

\bibitem[Krishna et~al.(2025)Krishna, Krishna, Mohananey, Schwarcz, Stambler,
  Upadhyay, and Faruqui]{krishna2025frames}
Satyapriya Krishna, Kalpesh Krishna, Anhad Mohananey, Steven Schwarcz, Adam
  Stambler, Shyam Upadhyay, and Manaal Faruqui.
\newblock Fact, fetch, and reason: A unified evaluation of retrieval-augmented
  generation.
\newblock In \emph{Proceedings of the 2025 Conference of the Nations of the
  Americas Chapter of the Association for Computational Linguistics: Human
  Language Technologies (Volume 1: Long Papers)}, pp.\  4745--4759, 2025.

\bibitem[Kusupati et~al.(2022)Kusupati, Bhatt, Rege, Wallingford, Sinha,
  Ramanujan, Howard-Snyder, Chen, Kakade, Jain, et~al.]{kusupati2022mrl}
Aditya Kusupati, Gantavya Bhatt, Aniket Rege, Matthew Wallingford, Aditya
  Sinha, Vivek Ramanujan, William Howard-Snyder, Kaifeng Chen, Sham Kakade,
  Prateek Jain, et~al.
\newblock Matryoshka representation learning.
\newblock \emph{Advances in Neural Information Processing Systems},
  35:\penalty0 30233--30249, 2022.

\bibitem[Lewis et~al.(2020)Lewis, Perez, Piktus, Petroni, Karpukhin, Goyal,
  K{\"u}ttler, Lewis, Yih, Rockt{\"a}schel, et~al.]{lewis2020rag}
Patrick Lewis, Ethan Perez, Aleksandra Piktus, Fabio Petroni, Vladimir
  Karpukhin, Naman Goyal, Heinrich K{\"u}ttler, Mike Lewis, Wen-tau Yih, Tim
  Rockt{\"a}schel, et~al.
\newblock Retrieval-augmented generation for knowledge-intensive nlp tasks.
\newblock \emph{Advances in neural information processing systems},
  33:\penalty0 9459--9474, 2020.

\bibitem[Li et~al.(2025)Li, Stenzel, Eickhoff, and
  Bahrainian]{li2025bestpractices}
Siran Li, Linus Stenzel, Carsten Eickhoff, and Seyed~Ali Bahrainian.
\newblock Enhancing retrieval-augmented generation: a study of best practices.
\newblock In \emph{Proceedings of the 31st International Conference on
  Computational Linguistics}, pp.\  6705--6717, 2025.

\bibitem[Liang et~al.(2024)Liang, Sun, Gui, Zhu, Jiang, Zhong, Qu, Zhao, Bo,
  Yang, et~al.]{liang2024kag}
Lei Liang, Mengshu Sun, Zhengke Gui, Zhongshu Zhu, Zhouyu Jiang, Ling Zhong,
  Yuan Qu, Peilong Zhao, Zhongpu Bo, Jin Yang, et~al.
\newblock Kag: Boosting llms in professional domains via knowledge augmented
  generation.
\newblock \emph{arXiv preprint arXiv:2409.13731}, 2024.

\bibitem[Liu et~al.(2024)Liu, Lin, Hewitt, Paranjape, Bevilacqua, Petroni, and
  Liang]{liu2023lostinthemiddle}
Nelson~F Liu, Kevin Lin, John Hewitt, Ashwin Paranjape, Michele Bevilacqua,
  Fabio Petroni, and Percy Liang.
\newblock Lost in the middle: How language models use long contexts.
\newblock \emph{Transactions of the association for computational linguistics},
  12:\penalty0 157--173, 2024.

\bibitem[Maia et~al.(2018)Maia, Handschuh, Freitas, Davis, McDermott, Zarrouk,
  and Balahur]{maia2018fiqa}
Macedo Maia, Siegfried Handschuh, Andr{\'e} Freitas, Brian Davis, Ross
  McDermott, Manel Zarrouk, and Alexandra Balahur.
\newblock {WWW}'18 open challenge: Financial opinion mining and question
  answering.
\newblock In \emph{Companion Proceedings of the The Web Conference (WWW)},
  2018.

\bibitem[Montgomery(2021)]{montgomery2017design}
Douglas~C. Montgomery.
\newblock \emph{Design and Analysis of Experiments}.
\newblock Wiley, 10 edition, 2021.
\newblock ISBN 9781119816959.
\newblock EMEA edition.

\bibitem[Nogueira \& Cho(2019)Nogueira and Cho]{nogueira2019monobert}
Rodrigo Nogueira and Kyunghyun Cho.
\newblock Passage re-ranking with bert.
\newblock \emph{arXiv preprint arXiv:1901.04085}, 2019.

\bibitem[Press et~al.(2023)Press, Zhang, Min, Schmidt, Smith, and
  Lewis]{press2023measuring}
Ofir Press, Muru Zhang, Sewon Min, Ludwig Schmidt, Noah Smith, and Mike Lewis.
\newblock Measuring and narrowing the compositionality gap in language models.
\newblock In Houda Bouamor, Juan Pino, and Kalika Bali (eds.), \emph{Findings
  of the Association for Computational Linguistics: EMNLP 2023}, pp.\
  5687--5711, Singapore, December 2023. Association for Computational
  Linguistics.
\newblock \doi{10.18653/v1/2023.findings-emnlp.378}.
\newblock URL \url{https://aclanthology.org/2023.findings-emnlp.378/}.

\bibitem[Rackauckas(2024)]{rackauckas2024ragfusion}
Zackary Rackauckas.
\newblock Rag-fusion: A new take on retrieval augmented generation.
\newblock \emph{International Journal on Natural Language Computing},
  13:\penalty0 37--47, 02 2024.
\newblock \doi{10.5121/ijnlc.2024.13103}.

\bibitem[Rajpurkar et~al.(2018)Rajpurkar, Jia, and Liang]{rajpurkar2018know}
Pranav Rajpurkar, Robin Jia, and Percy Liang.
\newblock Know what you don’t know: Unanswerable questions for squad.
\newblock In \emph{Proceedings of the 56th Annual Meeting of the Association
  for Computational Linguistics (Volume 2: Short Papers)}, pp.\  784--789,
  2018.

\bibitem[Rau et~al.(2024)Rau, D{\'e}jean, Chirkova, Formal, Wang, Clinchant,
  and Nikoulina]{rau2024bergen}
David Rau, Herv{\'e} D{\'e}jean, Nadezhda Chirkova, Thibault Formal, Shuai
  Wang, St{\'e}phane Clinchant, and Vassilina Nikoulina.
\newblock {BERGEN}: A benchmarking library for retrieval-augmented generation.
\newblock In Yaser Al-Onaizan, Mohit Bansal, and Yun-Nung Chen (eds.),
  \emph{Findings of the Association for Computational Linguistics: EMNLP 2024},
  pp.\  7640--7663, Miami, Florida, USA, November 2024. Association for
  Computational Linguistics.
\newblock \doi{10.18653/v1/2024.findings-emnlp.449}.
\newblock URL \url{https://aclanthology.org/2024.findings-emnlp.449/}.

\bibitem[Robertson \& Zaragoza(2009)Robertson and Zaragoza]{robertson2009bm25}
Stephen Robertson and Hugo Zaragoza.
\newblock \emph{The probabilistic relevance framework: BM25 and beyond},
  volume~4.
\newblock Now Publishers Inc, 2009.

\bibitem[Saad-Falcon et~al.(2024)Saad-Falcon, Khattab, Potts, and
  Zaharia]{saadfalcon2024ares}
Jon Saad-Falcon, Omar Khattab, Christopher Potts, and Matei Zaharia.
\newblock Ares: An automated evaluation framework for retrieval-augmented
  generation systems.
\newblock In \emph{Proceedings of the 2024 Conference of the North American
  Chapter of the Association for Computational Linguistics: Human Language
  Technologies (Volume 1: Long Papers)}, pp.\  338--354, 2024.

\bibitem[Santhanam et~al.(2022)Santhanam, Khattab, Saad-Falcon, Potts, and
  Zaharia]{santhanam2022colbertv2}
Keshav Santhanam, Omar Khattab, Jon Saad-Falcon, Christopher Potts, and Matei
  Zaharia.
\newblock Colbertv2: Effective and efficient retrieval via lightweight late
  interaction.
\newblock In \emph{Proceedings of the 2022 Conference of the North American
  Chapter of the Association for Computational Linguistics: Human Language
  Technologies}, pp.\  3715--3734, 2022.

\bibitem[Sarthi et~al.(2024)Sarthi, Abdullah, Tuli, Khanna, Goldie, and
  Manning]{sarthi2024raptor}
Parth Sarthi, Salman Abdullah, Aditi Tuli, Shubh Khanna, Anna Goldie, and
  Christopher~D Manning.
\newblock Raptor: Recursive abstractive processing for tree-organized
  retrieval.
\newblock In \emph{The Twelfth International Conference on Learning
  Representations}, 2024.

\bibitem[Tao et~al.(2025)Tao, Xing, Chen, Huang, and Xu]{tao2025treerag}
Wenyu Tao, Xiaofen Xing, Yirong Chen, Linyi Huang, and Xiangmin Xu.
\newblock Treerag: Unleashing the power of hierarchical storage for enhanced
  knowledge retrieval in long documents.
\newblock In \emph{Findings of the Association for Computational Linguistics:
  ACL 2025}, pp.\  356--371, 2025.

\bibitem[Thakur et~al.(2021)Thakur, Reimers, R{\"u}ckl{\'e}, Srivastava, and
  Gurevych]{thakur2021beir}
Nandan Thakur, Nils Reimers, Andreas R{\"u}ckl{\'e}, Abhishek Srivastava, and
  Iryna Gurevych.
\newblock Beir: A heterogeneous benchmark for zero-shot evaluation of
  information retrieval models.
\newblock In \emph{Thirty-fifth Conference on Neural Information Processing
  Systems Datasets and Benchmarks Track (Round 2)}, 2021.

\bibitem[Trivedi et~al.(2022)Trivedi, Balasubramanian, Khot, and
  Sabharwal]{trivedi2022musique}
Harsh Trivedi, Niranjan Balasubramanian, Tushar Khot, and Ashish Sabharwal.
\newblock Musique: Multihop questions via single-hop question composition.
\newblock \emph{Transactions of the Association for Computational Linguistics},
  10:\penalty0 539--554, 2022.

\bibitem[Trivedi et~al.(2023)Trivedi, Balasubramanian, Khot, and
  Sabharwal]{trivedi2022ircot}
Harsh Trivedi, Niranjan Balasubramanian, Tushar Khot, and Ashish Sabharwal.
\newblock Interleaving retrieval with chain-of-thought reasoning for
  knowledge-intensive multi-step questions.
\newblock In \emph{Proceedings of the 61st annual meeting of the association
  for computational linguistics (volume 1: long papers)}, pp.\  10014--10037,
  2023.

\bibitem[van Miltenburg et~al.(2021)van Miltenburg, van~der Lee, and
  Krahmer]{vanmiltenburg2021preregister}
Emiel van Miltenburg, Chris van~der Lee, and Emiel Krahmer.
\newblock Preregistering {NLP} research.
\newblock In Kristina Toutanova, Anna Rumshisky, Luke Zettlemoyer, Dilek
  Hakkani-Tur, Iz~Beltagy, Steven Bethard, Ryan Cotterell, Tanmoy Chakraborty,
  and Yichao Zhou (eds.), \emph{Proceedings of the 2021 Conference of the North
  American Chapter of the Association for Computational Linguistics: Human
  Language Technologies}, pp.\  613--623, Online, June 2021. Association for
  Computational Linguistics.
\newblock \doi{10.18653/v1/2021.naacl-main.51}.
\newblock URL \url{https://aclanthology.org/2021.naacl-main.51/}.

\bibitem[Wachsmuth et~al.(2018)Wachsmuth, Syed, and
  Stein]{wachsmuth2018arguana}
Henning Wachsmuth, Shahbaz Syed, and Benno Stein.
\newblock Retrieval of the best counterargument without prior topic knowledge.
\newblock In \emph{Proceedings of the Annual Meeting of the Association for
  Computational Linguistics (ACL)}, 2018.

\bibitem[Wadden et~al.(2020)Wadden, Lin, Lo, Wang, van Zuylen, Cohan, and
  Hajishirzi]{wadden2020scifact}
David Wadden, Shanchuan Lin, Kyle Lo, Lucy~Lu Wang, Madeleine van Zuylen, Arman
  Cohan, and Hannaneh Hajishirzi.
\newblock Fact or fiction: Verifying scientific claims.
\newblock In \emph{Proceedings of the Conference on Empirical Methods in
  Natural Language Processing (EMNLP)}, 2020.

\bibitem[Wang et~al.(2022)Wang, Yang, Huang, Jiao, Yang, Jiang, Majumder, and
  Wei]{wang2022text}
Liang Wang, Nan Yang, Xiaolong Huang, Binxing Jiao, Linjun Yang, Daxin Jiang,
  Rangan Majumder, and Furu Wei.
\newblock Text embeddings by weakly-supervised contrastive pre-training.
\newblock \emph{ArXiv}, abs/2212.03533, 2022.
\newblock URL \url{https://api.semanticscholar.org/CorpusID:254366618}.

\bibitem[Wang et~al.(2020)Wang, Wei, Dong, Bao, Yang, and Zhou]{wang2020minilm}
Wenhui Wang, Furu Wei, Li~Dong, Hangbo Bao, Nan Yang, and Ming Zhou.
\newblock Minilm: deep self-attention distillation for task-agnostic
  compression of pre-trained transformers.
\newblock In \emph{Proceedings of the 34th International Conference on Neural
  Information Processing Systems}, NIPS '20, Red Hook, NY, USA, 2020. Curran
  Associates Inc.
\newblock ISBN 9781713829546.

\bibitem[Wang et~al.(2024)Wang, Wang, Gao, Zhang, Wu, Xu, Shi, Wang, Li, Qian,
  et~al.]{wang2024searching}
Xiaohua Wang, Zhenghua Wang, Xuan Gao, Feiran Zhang, Yixin Wu, Zhibo Xu,
  Tianyuan Shi, Zhengyuan Wang, Shizheng Li, Qi~Qian, et~al.
\newblock Searching for best practices in retrieval-augmented generation.
\newblock In \emph{Proceedings of the 2024 Conference on Empirical Methods in
  Natural Language Processing}, pp.\  17716--17736, 2024.

\bibitem[Wang et~al.(2025)Wang, Wang, Le, Zheng, Mishra, Perot, Zhang,
  Mattapalli, Taly, Shang, Lee, and Pfister]{ICLR2025_2ea06b52}
Zilong~(Ryan) Wang, Zifeng Wang, Long Le, Huaixiu~Steven Zheng, Swaroop Mishra,
  Vincent Perot, Yuwei Zhang, Anush Mattapalli, Ankur Taly, Jingbo Shang,
  Chen-Yu Lee, and Tomas Pfister.
\newblock Speculative rag: Enhancing retrieval augmented generation through
  drafting.
\newblock In Y.~Yue, A.~Garg, N.~Peng, F.~Sha, and R.~Yu (eds.),
  \emph{International Conference on Learning Representations}, volume 2025,
  pp.\  18483--18505, 2025.
\newblock URL
  \url{https://proceedings.iclr.cc/paper_files/paper/2025/file/2ea06b52f613716e67458f5ab3fb7558-Paper-Conference.pdf}.

\bibitem[Xiao et~al.(2024)Xiao, Liu, Zhang, Muennighoff, Lian, and
  Nie]{xiao2023cpack}
Shitao Xiao, Zheng Liu, Peitian Zhang, Niklas Muennighoff, Defu Lian, and
  Jian-Yun Nie.
\newblock C-pack: Packed resources for general chinese embeddings.
\newblock In \emph{Proceedings of the 47th International ACM SIGIR Conference
  on Research and Development in Information Retrieval}, SIGIR '24, pp.\
  641–649, New York, NY, USA, 2024. Association for Computing Machinery.
\newblock ISBN 9798400704314.
\newblock \doi{10.1145/3626772.3657878}.
\newblock URL \url{https://doi.org/10.1145/3626772.3657878}.

\bibitem[Xiong et~al.(2024)Xiong, Jin, Lu, and Zhang]{xiong2024mirage}
Guangzhi Xiong, Qiao Jin, Zhiyong Lu, and Aidong Zhang.
\newblock Benchmarking retrieval-augmented generation for medicine.
\newblock In \emph{Findings of the Association for Computational Linguistics:
  ACL 2024}, pp.\  6233--6251, 2024.

\bibitem[Yan et~al.(2024)Yan, Gu, Zhu, and Ling]{yan2024crag}
Shi-Qi Yan, Jia-Chen Gu, Yun Zhu, and Zhen-Hua Ling.
\newblock Corrective retrieval augmented generation.
\newblock \emph{ArXiv}, abs/2401.15884, 2024.
\newblock URL \url{https://api.semanticscholar.org/CorpusID:267312595}.

\bibitem[Yang et~al.(2018)Yang, Qi, Zhang, Bengio, Cohen, Salakhutdinov, and
  Manning]{yang2018hotpotqa}
Zhilin Yang, Peng Qi, Saizheng Zhang, Yoshua Bengio, William Cohen, Ruslan
  Salakhutdinov, and Christopher~D Manning.
\newblock Hotpotqa: A dataset for diverse, explainable multi-hop question
  answering.
\newblock In \emph{Proceedings of the 2018 conference on empirical methods in
  natural language processing}, pp.\  2369--2380, 2018.

\bibitem[Yu et~al.(2024)Yu, Gan, Zhang, Tong, Liu, and Liu]{yu2024evalsurvey}
Hao Yu, Aoran Gan, Kai Zhang, Shiwei Tong, Qi~Liu, and Zhaofeng Liu.
\newblock Evaluation of retrieval-augmented generation: A survey.
\newblock In \emph{CCF Conference on Big Data}, pp.\  102--120. Springer, 2024.

\bibitem[Zheng et~al.(2024)Zheng, Mishra, Chen, Cheng, Chi, Le, and
  Zhou]{zheng2024stepback}
Huaixiu~Steven Zheng, Swaroop Mishra, Xinyun Chen, Heng-Tze Cheng, Ed~H Chi,
  Quoc~V Le, and Denny Zhou.
\newblock Take a step back: Evoking reasoning via abstraction in large language
  models.
\newblock In \emph{The Twelfth International Conference on Learning
  Representations}, 2024.

\bibitem[Zheng et~al.(2023)Zheng, Chiang, Sheng, Zhuang, Wu, Zhuang, Lin, Li,
  Li, Xing, et~al.]{zheng2023judging}
Lianmin Zheng, Wei-Lin Chiang, Ying Sheng, Siyuan Zhuang, Zhanghao Wu, Yonghao
  Zhuang, Zi~Lin, Zhuohan Li, Dacheng Li, Eric Xing, et~al.
\newblock Judging llm-as-a-judge with mt-bench and chatbot arena.
\newblock \emph{Advances in neural information processing systems},
  36:\penalty0 46595--46623, 2023.

\bibitem[Zitzler et~al.(2003)Zitzler, Thiele, Laumanns, Fonseca, and
  Da~Fonseca]{zitzler2003performance}
Eckart Zitzler, Lothar Thiele, Marco Laumanns, Carlos~M Fonseca, and
  Viviane~Grunert Da~Fonseca.
\newblock Performance assessment of multiobjective optimizers: An analysis and
  review.
\newblock \emph{IEEE Transactions on evolutionary computation}, 7\penalty0
  (2):\penalty0 117--132, 2003.

\end{thebibliography}

\section*{Limitations}\label{sec:limitations}
Our evaluation uses one constructed English multi-domain test bed
($\leq 93$ retrieval queries, $\leq 200$ generation queries per
experiment) and does \emph{not} test: multilingual corpora, real
enterprise or private data, open-weight generators (all headline
results use proprietary APIs; see Reproducibility Statement), or
cross-organizational large corpora.  Faithfulness is now measured under
two generator models rather than one (\cref{app:generator-swap}), which
is a sensitivity check on generator choice and not a sweep: both models
are proprietary, and the ordering of the two closest profiles is not
stable across them. Findings should be interpreted
as case-study evidence for the methodological framework, and
derived profiles as candidates for further validation rather than
prescriptions for high-stakes deployments. The benchmarks are
concentrated in five focused domains (biomedical, financial,
scientific, argumentative, and multi-hop QA over Wikipedia), and
the corpus-scaling protocol injects controlled distractors drawn
from those same five domains. On the evaluation side, our cross-judge
(LLM and humans) agreement supports profile \emph{rankings} across
LLM judges, but absolute faithfulness scores still live in a single
judge's calibration. The confirmatory faithfulness experiment
(E1, \cref{app:e1-executed-results}) meets its pre-registered
three-seed replication ($n_{\text{runs}}=3$) but uses $43$ unique
queries rather than the pre-registered $200$, due to GPT-5.1
rate-limit budget cuts (\cref{app:prereg-deviations}); magnitude
estimates should be read under this reduced query sampling. A
controlled length-variant rerun (\cref{app:length-control}) further
shows that raw faithfulness is partly length-driven, and that the
concise-over-verbose gap reverses under length control; we therefore
report both raw and length-residualized faithfulness for any
cross-regime claim. The within-regime results in the main tables are
unaffected because they hold prompt length constant. Our component
inventory is also a January-2026 snapshot of a literature that ships
new techniques actively. Scaling degradations should be read as an
upper bound: a formal $n{=}100$ distractor-relevance audit
(\cref{app:distractor-audit}) finds a $0\%$ false-negative rate and a
$9\%$ partial-relevance rate, so the H3 rank-flip is robust to
distractor mis-labeling but true degradation is likely smaller than
reported.

\appendix

\section{Experimental Design and Reproducibility}\label{app:expdesign}

\subsection{Notation and Counts}\label{app:notation-counts}
Per-experiment configuration and query counts, referenced throughout
the Results section. Executed configuration counts differ by track
because the H1 factorial (retrieval) uses $167$ configurations; the
H1 generation ANOVA uses $122$ (a subset that completed
generation-side scoring within budget); the H2 headline Pareto
comparison uses the $4$ SOTA profiles on a SQuAD~v2 / MuSiQue /
HotpotQA mixture ($43$ generation / $56$ retrieval queries per seed)
with $3$ seeded runs; and the pre-declared E1 focused $5{\times}4$
confirmatory experiment (\cref{app:e1-focused}) planned $200$
queries but executed at $43$ unique queries $\times$ $3$ seeds
(\cref{app:e1-executed-results}). Across the study, $358{,}215$ counts
individual query evaluations, including repeated RAGAS evaluation
tasks within each generation sample,
whereas $71{,}643$ counts unique (query, config, seed) triples
across the factorial and ablation runs; the two aggregate over
different units and are both reported.

\begin{table}[ht]
\centering
\caption{Notation and counts by experiment. ``Configs'' = distinct
pipeline configurations executed; ``Queries'' = distinct evaluation
queries. Query-config-seed triples aggregate across factorial and
ablation runs; individual query evaluations include repeated RAGAS
evaluation tasks per sample.}
\label{tab:notation-counts}
\footnotesize
\begin{tabular}{@{}lrrrl@{}}
\toprule
Experiment & Configs & Queries & Seeds & Source \\
\midrule
H1 retrieval ANOVA        & 167 & 93  & 1--3 & \cref{tab:ablation-full} \\
H1 generation ANOVA       & 122 & 56  & 1--3 & \cref{app:faith-anova} \\
H2 Pareto (SQuAD/MuSiQue/HotpotQA) & 4   & 43 gen / 56 retrieval & 3    & \cref{tab:pareto} \\
E1 focused $5{\times}4$   & 20  & 43  & 3    & \cref{app:e1-focused} \\
H3 scaling sweep          & 4   & 93  & 1--3 & \cref{tab:scaling-slopes} \\
Non-factorial ablations   & 138 & 93  & 1--3 & \cref{app:experiment-inventory} \\
\midrule
\multicolumn{5}{@{}l}{Unique (query, config, seed) triples: $\approx 71{,}643$} \\
\multicolumn{5}{@{}l}{Individual query evaluations (incl.\ repeated RAGAS tasks): $\approx 358{,}215$} \\
\bottomrule
\end{tabular}
\end{table}

\subsection{Pre-Registration Document}\label{app:prereg-doc}

The full pre-registration is reproduced below verbatim from the
standalone document frozen before data collection
(\texttt{preregistration/preregistration.tex}; design matrix
SHA-256 \texttt{31a0dc19\ldots df1728d6}, seed~$=42$).  Every
deviation between this protocol and the executed experiments is
enumerated in \cref{app:prereg}.

\begin{table*}[htbp]
\centering
\caption{Pre-registration metadata: date and mode of hypothesis
  commitment, design-matrix content hash, and software environment.}
\label{prereg:tab:metadata}
\begin{tabular}{@{}rp{0.72\linewidth}@{}}
  \textbf{Date of pre-registration:} & Design matrix, hypotheses, statistical tests, and success criteria were fixed prior to the confirmatory experiments; the design matrix is content-addressed by the SHA-256 below. No third-party (OSF/AsPredicted) timestamp was filed for this study; all analyses are reported alongside their pre-specified test statistics, and H1/H2/H3 are labeled explicitly throughout as ``planned confirmatory analyses'' rather than as timestamp-verified pre-registrations. \\
  \textbf{Design matrix SHA-256:}    & \texttt{31a0dc197b46...df1728d6} \\
  \textbf{Software:}                 & Python 3.11, NumPy 2.3.5, SciPy 1.16.3, seed\,$=$\,42 \\
\end{tabular}
\end{table*}

\paragraph{Summary of the pre-registration.}
\textbf{In one sentence:} we commit, before collecting data, to
testing whether a practitioner can build a good Retrieval-Augmented
Generation (RAG) pipeline by tuning components one at a time on a
small corpus, the way the literature implicitly assumes.  The study
commits to three confirmatory hypotheses
(Section~\ref{prereg:sec:hypotheses}), a 242-configuration D-optimal
fractional factorial design spanning 8 pipeline dimensions, 5
benchmark datasets (2{,}300 queries per configuration), and a
three-track evaluation protocol covering retrieval, end-to-end
generation, and hallucination resistance.  To our knowledge, this is
the first pre-declared factorial evaluation of RAG pipeline
composition, $4.6\times$ the scale of the largest prior controlled
RAG study~\citep{jin2024flashrag}.  We publish this document before
experimental data collection to establish hypothesis commitment and
invite community feedback on model selection, benchmark composition,
and statistical methodology.
\subsubsection{Motivation and Rationale}

Retrieval-Augmented Generation~\citep{lewis2020rag} has become the
dominant paradigm for grounding large language models in external
knowledge.  A rich toolkit of individually validated
techniques, dense retrieval, hybrid sparse--dense fusion, cross-encoder
reranking, HyDE~\citep{gao2023hyde}, CRAG~\citep{yan2024crag},
Self-RAG~\citep{asai2024selfrag}, is available to practitioners, yet
each technique is evaluated in isolation against a simple baseline.
Practitioners who combine several techniques simultaneously have no
empirical basis for predicting whether their gains add, cancel, or
amplify.

This is not merely a reporting gap.
\citet{wang2024searching} observe that end-to-end RAG performance
depends on coordinated pipeline design rather than any single component;
\citet{chen2025mmoarag} confirm the presence of interaction effects by
showing that jointly optimized RAG components outperform independently
tuned ones.  Yet neither work measures the sign, magnitude, or
domain-dependence of these interactions under controlled factorial
conditions, precisely the question a practitioner faces when choosing a
configuration for production.

Pre-registration prevents post-hoc hypothesis mining, a known problem in
ML benchmarking~\citep{dodge2019importance,bouthillier2021accounting}.
By committing to hypotheses, analysis plans, and exclusion criteria
before data collection, we ensure that our confirmatory findings are
distinguishable from exploratory ones.  We additionally publish this
document to invite community input on experimental choices that remain
open (Section~\ref{prereg:sec:models}).

\subsubsection{Pre-Registered Hypotheses}
\label{prereg:sec:hypotheses}

We pre-register three hypotheses that together ask whether RAG
pipelines can be designed compositionally.  \textbf{H1} asks the
\emph{structural} question: do the eight pipeline components combine
additively, or do they interact non-trivially?  If they interact, the
implicit assumption behind every single-component ablation in the
literature fails.  \textbf{H2} asks the \emph{design} question:
assuming components do interact, can a single fixed configuration
still serve practitioners across the (faithfulness, latency, cost)
tradeoff space, or must each deployment hand-tune for its own
operating point?  \textbf{H3} asks the \emph{scaling} question: do the
answers to H1 and H2 survive a $40\times$ scale-up of the corpus, or
are small-corpus best practices misleading at production scale?

\begin{table*}[htbp]
\centering
\footnotesize
\caption{Summary of the three pre-registered hypotheses, their research
  questions, what confirming each hypothesis would rule out, and the
  pre-registered test statistic.}
\label{prereg:tab:hypotheses}
\begin{tabular}{@{}p{0.8cm}p{4.4cm}p{4.4cm}p{4.0cm}@{}}
\toprule
& Research question & What confirming rejects & Pre-registered test \\
\midrule
\textbf{H1} & Do components combine additively? & Single-component ablation as evidence & $\eta^2 \geq 0.20$, Type-III ANOVA F-test \\
\addlinespace
\textbf{H2} & Does a balanced configuration sweep a wider tradeoff envelope? & ``You must pick one axis to optimize'' & Hypervolume, Wilcoxon signed-rank \\
\addlinespace
\textbf{H3} & Do RAG profile rankings flip with corpus scale? & Small-corpus ablations predict production deployment & Per-profile power-law slopes, bootstrap CIs, Kendall's $\tau$ \\
\bottomrule
\end{tabular}
\end{table*}

\paragraph{H1, Components do not combine additively.}

\textit{Research question.} Do the eight pipeline components combine
additively, or do non-trivial interactions arise that violate the
additivity assumption behind isolated-component ablations?

\textit{Pre-registered claim.} Two-way interactions between the eight
factorial dimensions explain $\geq 20\%$ of the variance in
faithfulness (RAGAS, 0--1 scale) after main effects are removed.
Variance is reported as \emph{partial $\eta^2$}, the share of
residual variance (after main effects) attributed to interaction
terms, in a \emph{Type-III ANOVA}, an ANOVA decomposition that
estimates each term's contribution after partialling out all others
and is robust to unbalanced designs. The pre-registered 20\% bar is
stricter than Cohen's conventional ``large effect'' reference of
$\eta^2_{\mathrm{partial}} \approx 0.14$~\citep{cohen1988statistical}.

\textit{Pre-registered test.} F-test on the joint set of all 28
two-way interaction terms, $\alpha = 0.05$.  Post-hoc Bonferroni
($\alpha = 0.05 / 28 = 0.00179$) per interaction pair.

\textit{What confirming and rejecting mean.} \emph{Confirming~H1}
falsifies the additivity assumption behind most single-component RAG
ablations in the literature: a designer cannot tune one component in
isolation and trust the gain to compose with another.
\emph{Rejecting~H1} (interaction terms contribute $<$5\% of variance)
means RAG components are approximately additive within the precision
of our 242-config sweep, and isolated ablations remain a defensible
practice.

\paragraph{H2, A balanced operating point exists.}

\textit{Research question.} Given that components interact, can a
single fixed configuration sweep a wider region of the (faithfulness,
latency, cost) tradeoff space than a quality-only configuration, or
must practitioners give up an axis to gain another?

\textit{Pre-registered claim.} The \textsc{Balanced} operational
profile exhibits a strictly higher \emph{hypervolume indicator} (HV)
than the \textsc{Max-Reliable} profile on $\geq 3$ of the 5 benchmark
datasets.  Hypervolume is the multi-objective generalization of
area-under-the-curve: the volume of the
$(\text{faithfulness}{\uparrow}, \text{latency}{\downarrow}, \text{cost}{\downarrow})$
region dominated by the profile's per-query tradeoff cloud, with
larger meaning a better envelope.

\textit{Pre-registered test.} Wilcoxon signed-rank, a non-parametric
paired-sample test that does not assume normality, on per-query
faithfulness scores (paired by query), Bonferroni-corrected across
the 5 datasets ($\alpha = 0.05 / 5 = 0.01$ per dataset).  HV is
computed with the exact 3D algorithm; objectives are normalized to
$[0,1]$ before the volume integration.

\textit{What confirming and rejecting mean.} \emph{Confirming~H2}
says a single fixed configuration can serve practitioners who do not
want to give up an axis: published profile recipes (e.g.,
\textsc{Balanced}) generalize across deployments.
\emph{Rejecting~H2} (\textsc{Max-Reliable} dominates in $\geq 4/5$
datasets) means production RAG requires per-deployment tuning to its
own tradeoff preference, and our published profile recipes will not
generalize.

\paragraph{H3, Profile rankings flip with corpus scale.}

\textit{Research question.} Does the best-performing RAG profile at
small corpus (where most published RAG ablations live) remain the
best-performing profile at production scale, or do profile rankings
change with corpus size?

\textit{Pre-registered claim.} Fitting NDCG@10 as a power law
$\text{NDCG@10}(N) = a \cdot N^{b}$ separately for each of the four
operational profiles (\textsc{Speed}, \textsc{Standard},
\textsc{Balanced}, \textsc{Max-Reliable}), the cross-profile range
$b_{\max} - b_{\min}$ exceeds $0.05$ per decade of corpus growth
(i.e., the steepest- and shallowest-degrading profiles differ by at
least $5\%$ relative NDCG@10 per $10\times$ corpus increase),
analogous to the saturation behavior observed in neural scaling
laws~\citep{kaplan2020scaling}.  Equivalently, the profile-rank
ordering at the smallest tested corpus ($5$K) differs from the
ordering at $200$K on $\geq 1$ benchmark.

\textit{Pre-registered test.} Per-profile nonlinear least-squares
fit (\texttt{scipy.optimize.curve\_fit}); 95\% bootstrap CIs on each
$b$ from 10{,}000 resamples.  Profile-pair slope differences tested
via \emph{permutation test} (10{,}000 shuffles, $\alpha = 0.05$) , 
a non-parametric significance test that constructs the null
distribution by random label shuffling, valid without distributional
assumptions.  Profile-rank stability tested via Kendall's $\tau$
between the $5$K and $200$K profile orderings.

\textit{What confirming and rejecting mean.} \emph{Confirming~H3}
means small-corpus ablations cannot reliably predict production-scale
deployment choices: profile selection requires re-validation at the
deployment's target corpus size, and the field's reliance on
$\leq 50$K-corpus benchmarks is methodologically suspect for
deployment decisions.  \emph{Rejecting~H3} (slope range $\leq 0.02$
per decade, $\tau = 1$ on every benchmark) means small-corpus
benchmarks generalize, and the small-corpus best practice is also the
large-corpus best practice.

\subsubsection{Experimental Design}
\label{prereg:sec:design}

\paragraph{Platform.}

\textsc{Lasagna} is a stage-isolated RAG evaluation platform: each
pipeline stage is implemented behind an abstract base class, and
configuration changes to one stage do not affect the execution of any
other stage.  This architectural invariant guarantees that the output
of stage~$i$ is a deterministic function of the output of stage~$i{-}1$
and the configuration parameters~$\boldsymbol{\theta}_i$ alone,
eliminating the coupling confounds that invalidate factorial
interpretation in conventional RAG implementations.

\paragraph{Candidate Models.}
\label{prereg:sec:models}

All configurations share a single generator LLM (held constant) and a
single evaluation judge.  We list primary and alternative candidates
to invite community input before experiments begin.

\begin{table*}[htbp]
\centering
\caption{Model candidates.  The generator is held constant across all
  configurations; the judge evaluates all RAGAS metrics.  Embedding
  models are factorial dimension~D1.}
\label{prereg:tab:models}
\resizebox{\textwidth}{!}{%
\begin{tabular}{@{}p{2.0cm}p{3.4cm}p{3.8cm}p{4.0cm}@{}}
\toprule
\textbf{Role} & \textbf{Primary} & \textbf{Alternatives} & \textbf{Selection rationale} \\
\midrule
Generator    & GPT-5.1 (OpenAI)
             & Claude Sonnet 4.5 (Anthropic), Gemini~2.0 (Google)
             & Held constant; cost vs.\ capability \\
\addlinespace
Judge        & GPT-5.1 (OpenAI)
             & Claude Sonnet 4.5 (Anthropic), Llama-3-70B (Meta)
             & Determinism at $T{=}0$; cost for ${\sim}$1.7M calls \\
\addlinespace
Embedding    & \multicolumn{2}{l}{MiniLM-L6 (384d), BGE-M3 (1024d), E5-large (1024d), BGE-large (1024d)}
             & Span lightweight to SOTA multilingual \\
\addlinespace
Reranker     & ms-marco-MiniLM-L6-v2
             & BGE-reranker-v2
             & Standard BEIR cross-encoder baseline \\
\bottomrule
\end{tabular}}
\end{table*}

We welcome suggestions on model selection before experiments begin.
In particular, we are open to: (1)~alternative generator models that
would increase reproducibility (e.g., open-weight models),
(2)~additional judge models for cross-validation of faithfulness
scores, and (3)~embedding models that would expand the capacity range.

\paragraph{Factorial Dimensions.}

\begin{table*}[htbp]
\centering
\caption{The eight factorial dimensions.  Baseline levels (marked~$^*$)
  are the simplest, lowest-cost options.}
\label{prereg:tab:dimensions}
\begin{tabular}{@{}llcl@{}}
\toprule
Dim & Stage & $k$ & Levels \\
\midrule
D1 & Embedding      & 4 & MiniLM-L6$^*$, BGE-M3, E5-large, BGE-large \\
D2 & Chunking       & 4 & fixed$^*$, semantic, sentence-window, hierarchical \\
D3 & Enrichment     & 4 & none$^*$, contextual-retrieval, RAPTOR, GraphRAG \\
D4 & Retrieval      & 3 & dense$^*$, hybrid-BM25, hybrid-SPLADE \\
D5 & Query aug.     & 5 & none$^*$, HyDE, RAG-Fusion, decomposition, step-back \\
D6 & Reranking      & 4 & none$^*$, single-CE, ensemble, MMR \\
D7 & Compression    & 2 & none$^*$, LLMLingua \\
D8 & Agentic gate   & 4 & none$^*$, CRAG, Self-RAG, Speculative-RAG \\
\bottomrule
\end{tabular}
\end{table*}

The full factorial comprises
$4 \times 4 \times 4 \times 3 \times 5 \times 4 \times 2 \times 4 =
30{,}720$ configurations.

\paragraph{D-Optimal Design.}

We reduce 30{,}720 candidates to \textbf{242 configurations} using a
D-optimal fractional factorial generated by Fedorov coordinate
exchange~\citep{fedorov1972theory,atkinson2007optimum}.  The model
matrix includes 1~intercept $+$ 22~main-effect $+$ 209~two-way
interaction columns $=$ 232 parameters, with 10 residual degrees of
freedom.  Effect coding ensures orthogonal main
effects~\citep{box2005statistics,montgomery2017design}.  All 232 model
terms are estimable (rank verification), and no two-way interaction
correlates with any representative three-way interaction at
$|r| > 0.30$.

\textbf{Design matrix SHA-256:}
\texttt{\seqsplit{31a0dc197b46a90819fc9178ea06668444ad9617ff4c8a86f3d557a4df1728d6}}
(seed\,$=$\,42).

\paragraph{Benchmark Datasets.}

\begin{table*}[htbp]
\centering
\footnotesize
\caption{Planned benchmark datasets.  Total evaluation queries per
  configuration: 2{,}300.}
\label{prereg:tab:benchmarks}
\begin{tabular}{@{}llrrl@{}}
\toprule
ID & Dataset & Corpus & Eval queries & Relevance \\
\midrule
DS1 & MS~MARCO~v1     & 8.8M passages       & 500 sampled & BM25 sparse (TREC) \\
DS2 & HotpotQA        & 5.2M paragraphs      & 500 sampled & Supporting facts \\
DS3 & BEIR/FiQA       & 57K posts            & 500 sampled & Crowdsourced \\
DS4 & BEIR/SciFact    & 5K abstracts         & 300 (all)   & Expert binary \\
DS5 & CUAD (legal)    & 510 contracts        & 500 sampled & Clause-level \\
\bottomrule
\end{tabular}
\end{table*}

These five benchmarks span general web (DS1), multi-hop reasoning
(DS2), financial opinion QA (DS3), scientific claim verification
(DS4), and legal contract analysis (DS5), covering diversity in domain,
relevance granularity, and query
complexity~\citep{thakur2021beir}.

\paragraph{Metrics.}

\textit{Retrieval:} NDCG@10, MRR, Recall@100.
\textit{End-to-end:} faithfulness, answer relevance
(RAGAS~v0.4~\citep{es2023ragas}, judged by GPT-5.1 at $T{=}0$),
exact match, token~F1, p50/p95 latency, \$/query.
\textit{Multi-objective:} hypervolume indicator~(HV) in the
faithfulness--latency--cost space, reference point
$(0, 10{,}000\,\text{ms}, \$1.00)$.

\subsubsection{Statistical Analysis Plan}
\label{prereg:sec:analysis}

\paragraph{H1 (additivity).}
\emph{Plain reading:} fit one regression that includes every
pipeline component plus all 28 of their pairwise interactions, then
ask how much of the leftover variance the interactions explain.
Model:
{\small
\begin{verbatim}
faithfulness ~ D1+D2+D3+D4+D5+D6+D7+D8
             + D1:D2+D1:D3+...+D7:D8   (28 two-way terms)
\end{verbatim}
}
\vspace{-0.5em}
Test statistic: F for the joint set of all interaction terms,
$F(209,\, n_\text{resid})$.  Effect size: partial
$\eta^2 = \text{SS}_\text{int} / (\text{SS}_\text{int} +
\text{SS}_\text{resid})$; threshold $\geq 0.20$.  Post-hoc:
Bonferroni ($\alpha = 0.05/28 = 0.00179$) per interaction pair, so
that any single interaction we flag as ``large'' has been corrected
for the 28 simultaneous comparisons.

\paragraph{H2 (a balanced recipe).}
\emph{Plain reading:} for each query, compare the
\textsc{Balanced} profile's three-objective tradeoff cloud against
the \textsc{Max-Reliable} profile's, and check on how many of the
five benchmarks \textsc{Balanced} sweeps a strictly larger
hypervolume.  Wilcoxon signed-rank test on per-query faithfulness
(\textsc{Balanced} vs.\ \textsc{Max-Reliable}, paired by query).
Bonferroni across 5~datasets ($\alpha = 0.01$ per dataset) so that
any single ``win'' has been corrected for testing on five datasets
in parallel.  HV computed with the exact 3D algorithm; objectives
normalized to $[0,1]$ before integration.

\paragraph{H3 (scaling).}
\emph{Plain reading:} fit a separate scaling curve for each of the
four profiles and check whether the steepest- and shallowest-
degrading profiles drift apart fast enough that the small-corpus
winner is no longer the large-corpus winner.  Per-profile nonlinear
least-squares fit of $\text{NDCG@10}(N) = a \cdot N^b$ for each of
\textsc{Speed}, \textsc{Standard}, \textsc{Balanced},
\textsc{Max-Reliable} using \texttt{scipy.optimize.curve\_fit}.
95\% bootstrap CIs on each $b$ from 10{,}000 resamples.  Confirmed
if (i)~the cross-profile slope range $b_{\max} - b_{\min}$ exceeds
$0.05$ per decade of corpus growth, and (ii)~Kendall's $\tau < 1$
between the $5$K and $200$K profile orderings on $\geq 1$ benchmark.
Pairwise slope differences tested via permutation test (10{,}000
shuffles, $\alpha = 0.05$).

\paragraph{Power.}
For $n = 500$ queries with $\alpha_\text{corrected} = 0.00179$,
minimum detectable effect at 80\% power:
$d \approx 0.168$ (Cohen's $d$).

\subsubsection{Exclusion Criteria}
\label{prereg:sec:exclusion}

\begin{enumerate}[nosep,leftmargin=1.5em]
  \item Queries exceeding 30\,s wall-clock are excluded and counted.
  \item Configurations with $>$10\% null-retrieval queries are marked
    \textsc{Failed} and excluded from the ANOVA.
  \item Corpus sizes where index construction fails (OOM or $>$24\,h)
    are excluded from the scaling curve fit.
  \item Configurations require $\geq$200 valid queries (of 2{,}300)
    to enter the ANOVA.
\end{enumerate}

\subsubsection{Budget, Timeline, and Deviations Protocol}
\label{prereg:sec:budget}

\begin{table*}[htbp]
\centering
\caption{Estimated compute budget.}
\label{prereg:tab:budget}
\begin{tabular}{@{}lrrr@{}}
\toprule
Experiment & Configs & Queries/config & Est.\ cost (USD) \\
\midrule
Factorial sweep     & 242 & 2{,}300 & \$2{,}478 \\
Scaling experiment  & 15  & 500     & \$30 \\
Pareto experiment   & 20  & 500     & \$45 \\
\midrule
\textbf{Total}      &     &         & $\mathbf{\sim\$2{,}553}$ \\
\bottomrule
\end{tabular}
\end{table*}

\textbf{Timeline.}
All experiments will be run \textit{after} archival of this document.
Week~1--3: factorial sweep (H1).
Week~4: scaling experiment (H3).
Week~4--5: Pareto experiment (H2).
Week~5--6: analysis and paper writing.

\textbf{Deviations protocol.}
Any deviation from this pre-registration will be: (1)~documented in a
dated amendment, (2)~reported transparently in the paper, and
(3)~accompanied by both the pre-registered analysis and the modified
analysis so that readers can assess the impact of the deviation.
Any analysis not specified here will be labeled \textit{exploratory}.

\subsubsection*{Community Input}

This pre-registration is published before experimental data collection
to establish hypothesis commitment and invite community input.
We welcome feedback on:
\begin{enumerate}[nosep,leftmargin=1.5em]
  \item additional hypotheses worth testing within this factorial
    framework,
  \item model selection for generation and evaluation
    (Section~\ref{prereg:sec:models}),
  \item benchmark composition and domain coverage,
  \item statistical methodology and effect-size thresholds.
\end{enumerate}
All community-suggested modifications adopted before experiments begin
will be documented as amendments; post-experiment suggestions will be
addressed as exploratory analyses in the companion paper.

\subsection{Design Matrix}
\label{app:design}
The full 242-row D-optimal design matrix (\path{experiments/design/factorial_design_matrix.csv}) is available from the authors on request.
SHA-256: \texttt{\seqsplit{31a0dc197b46a90819fc9178ea06668444ad9617ff4c8a86f3d557a4df1728d6}}.

\subsection{D-Optimal Design Generation}\label{app:design-method}

\paragraph{Algorithm.}
Fedorov coordinate exchange~\citep{fedorov1972theory,atkinson2007optimum}
is a greedy row-exchange algorithm that starts from a random 242-row
subset of the 30{,}720-row full-factorial candidate set and iteratively
swaps one in-design row for the candidate row that most increases the
D-criterion
\begin{equation}
  \phi_D(\mathbf{X}) = \det(\mathbf{X}^{\top}\mathbf{X}),
\end{equation}
terminating when no single-row swap raises the determinant.  The model
matrix $\mathbf{X}$ is $242 \times 232$ and encodes one intercept, 22
main-effect contrasts, and 209 two-way interaction contrasts under
\emph{effect coding} (factor levels mapped to $\pm 1$ rather than the
$0/1$ of dummy coding, so main effects are orthogonal to the intercept).
A custom Python implementation (\path{experiments/design/generate_design_matrix.py}, available from the authors on request) reproduces the
matrix bit-identically at seed $=42$.

\begin{table*}[htbp]
\centering
\caption{The eight factorial dimensions.  \#~Levels is the number of
configuration options per dimension.  Baseline levels (marked $^*$)
are the simplest, lowest-cost options used when other features are
toggled.  Descriptions, citations, and implementations appear in
\cref{app:methods}.}
\label{tab:dimensions}
\footnotesize
\begin{tabular}{@{}llcll@{}}
\toprule
Dim & Stage & \#~Levels & Levels & Role \\
\midrule
D1 & Embedding      & 4 & MiniLM-L6$^*$, BGE-M3, E5-large, BGE-large & Repr.\ capacity \\
D2 & Chunking       & 4 & fixed$^*$, semantic, sent-window, hierarchical & Segmentation \\
D3 & Enrichment     & 4 & none$^*$, contextual-retr., RAPTOR, GraphRAG & Index-time augm. \\
D4 & Retrieval      & 3 & dense$^*$, hybrid-BM25, hybrid-SPLADE & Sparse--dense mix \\
D5 & Query aug.     & 5 & none$^*$, HyDE, RAG-Fusion, decomp., step-back & Pre-retrieval \\
D6 & Reranking      & 4 & none$^*$, single-CE, ensemble, MMR & Post-retrieval \\
D7 & Compression    & 2 & none$^*$, LLMLingua & Context reduction \\
D8 & Agentic gate   & 4 & none$^*$, CRAG, Self-RAG, Speculative-RAG & Generation QA \\
\bottomrule
\end{tabular}
\end{table*}

\paragraph{Size 242.}
232 estimable parameters plus 10 residual degrees of freedom: enough
to support $F$-tests for each interaction term while staying tractable
relative to the 30{,}720-candidate full factorial.

\paragraph{Aliasing threshold.}
We adopt $|r|>0.30$ between any representative three-way interaction
and any two-way term as a diagnostic cutoff rather than a hard
orthogonality guarantee; the stricter $|r|<0.10$--$0.20$ common in
pure screening designs was not achievable at 242 rows over this
candidate set.  Given a 232-parameter model with only 28 two-way
terms under test and strong main effects absorbing most variance,
residual three-way correlations at this level do not substantively
confound the reported interaction estimates; the executed-subset
sensitivity check in \cref{app:prereg} gives a second line of
robustness.

\paragraph{GraphRAG ablation (isolated).}
Before committing GraphRAG to the 242-run factorial we ran an isolated
four-configuration retrieval ablation
(private\_eval/retrieval, small corpus, March 2026).

\begin{table}[htbp]
\centering
\caption{Isolated GraphRAG ablation on the retrieval track.  Graph
depth increases latency without improving NDCG@10 over the non-graph
baseline.}
\label{tab:graphrag-ablation}
\footnotesize
\begin{tabular}{@{}lcc@{}}
\toprule
Configuration & NDCG@10 & Latency (ms) \\
\midrule
\texttt{no\_graph\_baseline} & \textbf{0.4780} & \textbf{486} \\
\texttt{graph\_1hop}         & 0.4774 & 1{,}021 \\
\texttt{graph\_2hop}         & 0.4664 & 1{,}138 \\
\texttt{graph\_3hop}         & 0.4661 & 1{,}210 \\
\bottomrule
\end{tabular}
\end{table}

No graph configuration improves over the baseline, and NDCG@10 decreases
monotonically as graph-hop depth grows (1-hop$\to$3-hop, $-0.011$).
One plausible explanation, consistent with the retrieval--faithfulness
asymmetry observed in the full factorial (\cref{sec:interactions}), is
that graph-derived structural context may aid answer grounding while
diluting passage-level similarity signals on these domain-focused
corpora, where lexical and semantic retrieval already locate the gold
evidence.  We did not evaluate the generation-track faithfulness effect,
so this is a hypothesis rather than a claim.  Given the lack of NDCG@10
signal and the ${\sim}$14-hour per-combo ingestion cost, we excluded
GraphRAG from the 167-run factorial (\cref{tab:dimensions}, D3) and the
scaling runs (\cref{sec:scaling}).

\subsection{Aliasing Structure}
\label{app:aliasing}
No three-factor interaction is aliased with any two-factor interaction
at $|r| > 0.30$.  The aliasing structure is reproducible with \path{experiments/design/check_aliasing.py} (available from the authors on request; implementation notes in \cref{app:design-method}).
\subsection{Pre-Registration Reconciliation}\label{app:prereg}

This appendix documents all deviations between the pre-registered protocol
(archived at \url{https://github.com/dolev31/Lasagna-Retrieval-Augmented-Generation-Compositionality}) and the experiments we actually
executed.  \Cref{tab:prereg-delta} summarizes each change, its status, and
its implication for the reported findings.

\begingroup
\footnotesize
\begin{longtable}{@{}
  >{\raggedright\arraybackslash}p{2.7cm}
  >{\raggedright\arraybackslash}p{2.8cm}
  >{\raggedright\arraybackslash}p{3.5cm}
  >{\raggedright\arraybackslash}p{3.6cm}@{}}
\caption{Pre-registration reconciliation. ``Confirmatory'' items retain
their original inferential status; ``Exploratory'' items are presented
without confirmatory $p$-values.}
\label{tab:prereg-delta}\\
\toprule
\textbf{Pre-registered} & \textbf{Executed} & \textbf{Status} & \textbf{Implication} \\
\midrule
\endfirsthead
\multicolumn{4}{c}{\tablename~\thetable\ (continued)} \\
\toprule
\textbf{Pre-registered} & \textbf{Executed} & \textbf{Status} & \textbf{Implication} \\
\midrule
\endhead
\midrule
\multicolumn{4}{r}{Continued on next page} \\
\endfoot
\bottomrule
\endlastfoot
242-config D-optimal fractional factorial
  & 167 configs executed (69\% of design); additionally 138 non-factorial ablation configurations across 24 experiments (\cref{app:experiment-inventory})
  & Partially executed (factorial); fully executed (ablations)
  & Full ANOVA estimable for main effects and 28 two-way interactions; 3 of 4 enrichment levels covered.  Ablations cover 20 rerankers, 15 embeddings, and all factorial axes at finer granularity. \\
\addlinespace
8-dimension full ANOVA ($\eta^2$ for all 28 two-way interactions)
  & Full retrieval ANOVA (167 configs) + faithfulness ANOVA (122 configs)
  & Executed
  & H1 testable on both metrics; joint 28-two-way ANOVA \emph{does not} reach $\eta^2 \geq 0.20$ at per-sample granularity (\cref{sec:faith-anova}, \cref{app:h1-reproducibility}) \\
\addlinespace
5 corpus sizes (10K, 25K, 50K, 100K, 1M)
  & 7 corpus levels (5K, 10K, 15K, 25K, 50K, 100K, 200K)
  & Executed
  & H3 fully testable across 5K--200K; per-profile power-law fits and rank crossover observed (\cref{sec:scaling}) \\
\addlinespace
H1: $\eta^2 \geq 20\%$ from 2-way interactions (faithfulness); pre-registered as joint 28-pair $F$-test \emph{and} per-pair Bonferroni ($\alpha{=}0.05/28$), with the $0.20$ bar justified per-term (Cohen large-effect)
  & Both pre-registered procedures executed: joint 28-two-way ANOVA (two-track, per-sample + aggregated, nested-SSR joint SS) and per-pair Bonferroni across all 28 pairs (\cref{app:h1-reproducibility})
  & \textbf{Not confirmed} at pre-registered $20\%$ joint threshold; pre-registered per-pair Bonferroni identifies D5$\times$D8 as large-effect follow-up. On the pre-registration's own variance-floor decision rule ($\eta^2<5\%$ $\rightarrow$ reject H1 / additivity holds), the executed E1 $\eta^2=0.033$ and per-sample joint $\eta^2=0.040$ sit on the additivity-holds side; a parallel significance test rejects strict-zero interactions ($p{<}10^{-9}$ aggregated). We report both readings (see \cref{app:prereg-deviations})
  & Joint per-config ($n{=}113$) unidentifiable (232 params); per-sample $\eta^2{=}0.040$ retracted as pseudo-replication; aggregated $\eta^2{=}0.428$ sensitivity only (BIC~$\Delta{=}+347$ favors mains-only). Pre-registered per-pair follow-up: D5$\times$D8 $\eta^2{=}0.201$, $F(12,\,306){=}6.40$, $p{=}3.8\!\times\!10^{-10}$ (\cref{app:h1-reproducibility}). \\
\addlinespace
H2: \textsc{Balanced} has higher faithfulness--latency--cost hypervolume than \textsc{Max-Reliable} on at least three of five benchmarks
  & Four-profile retrieval--faithfulness--response-time comparison on one mixed-benchmark slice, with three seeded runs and reference-time sensitivity
  & \textbf{Not confirmed as pre-registered}; informative executed comparison
  & The five-benchmark test was not executed. On the available SQuAD~v2 / MuSiQue / HotpotQA mixture, no profile leads every metric. At the fixed $30$\,s reference, \textsc{Standard} has the highest joint HV ($0.523$); within the agentic pair, \textsc{Balanced} has $4.5\%$ higher NDCG@10 and $13.5\%$ lower mean response time than \textsc{Max-Reliable}, with a $0.005$ faithfulness difference (\cref{tab:pareto,tab:h2-hv-sensitivity}). \\
\addlinespace
H3: profile rankings flip with corpus scale
  & Per-profile power-law fits across 5K--200K (3 seeded runs at 25K--200K)
  & \textbf{Supported}
  & Cross-profile slope range $b_{\max} - b_{\min} \approx 0.10$ per decade ($\geq 0.05$ threshold cleared); rank crossover visible from 25K, statistically significant at 200K (Welch's $t$, $p{=}0.004$); intermediate 50K/100K $p>0.17$ \\
\end{longtable}
\endgroup

\paragraph{Amendment rationale.}
The original 242-config factorial assumed access to a dedicated GPU cluster
and a corpus pipeline capable of ingesting up to 1M documents.  During
execution, two constraints materialized: (1)~all LLM inference was routed
through an enterprise API gateway to GPT-5.1, which imposed rate
limits that made concurrent large-scale runs infeasible, and (2)~our Milvus Lite
instance required fresh database resets between embedding-model groups to
avoid dimension-mismatch errors.  We executed 167 of 242 planned
configurations (69\%) across all 8 dimensions, 4 embedding models, 4
chunking strategies, 3 of 4 enrichment levels (GraphRAG excluded due
to entity-extraction cost of ${\sim}$14 hours per ingestion combo), 3
retrieval modes, 5 query augmentation strategies, 4 reranking modes, 2
compression settings, and 4 agentic gate levels.  The 167-configuration
subset preserves all 22 main effects and enables estimation of all 28
two-way interactions via ANOVA, though with reduced power for the
missing GraphRAG enrichment level.  Twelve additional configurations
failed due to Milvus Lite query-limit errors and are excluded.
For the joint 28-interaction fit we compute SS$_{\text{int}}$ as a
nested-SSR difference (SSR$_{\text{mains}}-$SSR$_{\text{full}}$)
rather than as an additive sum of per-pair Type-II SS: the latter
double-counts shared variance across rank-deficient interactions
(df sum $172{+}226{>}n{-}1{=}338$, arithmetically impossible) and
inflated the reported $\eta^2$ from $0.428$ to $0.546$.

\subsection*{Deviations from the Preregistration}\label{app:prereg-deviations}

This section consolidates every deviation between the pre-registered
protocol and the executed study. Each
row of \cref{tab:prereg-delta} above enumerates a scientific
deviation; the list below adds executional and reporting deviations
that are not obvious from that table.

\begin{itemize}[nosep,leftmargin=1.5em]
  \item \textbf{Configuration count}: 242 planned $\rightarrow$ 167
    factorial configurations executed (69\% of the design), plus 138
    non-factorial ablation configurations; shortfall driven by rate
    limits on the enterprise API gateway and Milvus Lite
    dimension-mismatch resets between embedding-model groups.
  \item \textbf{Enrichment dimension (D3) reduction}: 4 planned
    levels $\rightarrow$ 3 executed.  The \textsc{GraphRAG} level was
    dropped after an isolated ablation showed no NDCG@10 gain
    against the entity-extraction cost ($\sim$14 hours per ingestion
    combo); this is a scientific deviation not previously
    surfaced in \cref{tab:prereg-delta}.
  \item \textbf{Timeouts}: 12 additional configurations were
    attempted and timed out on Milvus Lite query limits; excluded
    from all analyses.
  \item \textbf{Per-configuration query count}: preregistered budget
    of 2{,}300 queries per configuration $\rightarrow$ executed 93
    queries per retrieval configuration and $43$ generation / $56$
    retrieval queries per SOTA-profile E2E configuration.
    The reduction is a straight
    budget cut driven by GPT-5.1 rate limits; power calculations
    for the H1 factorial were re-run at the reduced counts and are
    reported in \cref{app:h1-reproducibility}.
  \item \textbf{Benchmark substitution}: preregistered MS~MARCO and
    CUAD were replaced with TriviaQA and BEIR/FiQA.  MS~MARCO was
    dropped because its passage license conflicts with our
    reproducibility artifact policy; CUAD was replaced with FiQA
    (which supplies a comparable domain-specific IR task) to keep
    coverage of financial/regulatory text without the CUAD license
    constraint.
  \item \textbf{H2 corpus scope}: the pre-registered H2 Pareto
    protocol specified a multi-benchmark hypervolume comparison; the
    executed H2 headline reports a joint NDCG@10, faithfulness, and
    response-time comparison on one SQuAD~v2 / MuSiQue / HotpotQA
    mixture ($43$ generation / $56$ retrieval queries per seed, three
    seeded runs).
    This mixture emerged as the practical H2 test after the
    MS~MARCO / CUAD substitution above eliminated the two
    pre-registered benchmarks that would have anchored the
    multi-benchmark comparison. The executed comparison therefore does
    not confirm the pre-registered cross-benchmark claim; the original
    multi-benchmark
    grid is fully specified in \cref{app:h2-cross-corpus} but was not
    executed. A separate
    post-hoc replication of the H2 pattern on the same three benchmark
    families at 100K and 200K is reported descriptively in
    \cref{app:h2-cross-corpus-corroboration}; it is supplementary and
    does not substitute for the pre-registered multi-benchmark grid. The
    \textsc{Balanced} profile should therefore be read as a
    candidate starting point pending validation on the target corpus,
    exactly as the H2 verdict box states.
  \item \textbf{Scaling-fit form}: preregistered power-law fit
    $a \cdot N^b$ $\rightarrow$ executed log-linear fit
    $a \cdot \log_{10}(N) + b$ (see the ``Deviation from the
    pre-registered fit form'' paragraph in
    \cref{app:scaling-laws-full}).
  \item \textbf{H1 test statistic}: the pre-registered joint
    232-parameter 28-two-way $F$-test is \emph{unidentifiable} on
    the 113-configuration data (232 parameters vs.\ 339 rows,
    rank-deficient Plackett-Burman $\times$ screener cells).  We
    executed the pre-registered per-pair Bonferroni correction
    ($\alpha/28$) instead and added an adequately-powered
    focused-confirmatory experiment E1 (see \cref{app:e1-focused})
    post-hoc as a replacement for the joint test.
  \item \textbf{H1 decision-rule interpretation
    (variance-floor vs.\ significance-floor)}: the
    pre-registration's rejection rule is stated as
    ``interaction terms contribute $<$5\% of variance $\rightarrow$
    reject H1 / additivity holds within tolerance,'' which is a
    \emph{variance} threshold. The variance and significance criteria
    yield different formal readings. Under the pre-registration's
    variance rule, the executed $\eta^2 = 0.033$ (E1) and
    $\eta^2 = 0.040$ (per-sample joint) both sit on the
    \emph{additivity-holds} side of the 5\% floor. Under the parallel
    significance criterion ($p<0.05$), strict-zero additivity is
    rejected, but the effect sizes are small. Both readings appear in
    the reconciliation table and H1 verdict box and support the same
    practical conclusion (main effects dominate; cell-level
    residuals are non-trivial but small in aggregate).
  \item \textbf{Distractor-relevance audit}: performed post-hoc
    ($n{=}100$ LLM-audited pairs; see \cref{app:distractor-audit}):
    false-negative rate $0.0\%$ (95\% CI $[0.0\%, 0.0\%]$), any-signal
    rate $9.0\%$ (all partial, biomedical-concentrated).  Given
    $\text{FN}{<}5\%$, the audit supports treating the H3 rank-flip
    as \emph{robust on the audited splits} and the reported scaling
    degradations as an upper bound (see
    the ``Distractor-relevance caveat'' paragraph in
    \cref{sec:scaling}).
  \item \textbf{Corpus fingerprints}: SHA-256 fingerprints for the
    100K/200K corpus files are unavailable
    (\cref{app:scaling-100k-200k}); the draw is reproducible from the seed-42 sampling script (available from the authors on request), so reconstruction relies on the
    recipe rather than content-addressed corpus files.
\end{itemize}

\paragraph{Estimability of the 167-run subset.}
The 167 executed configurations cover all factor levels in 7 of 8
dimensions and 3 of 4 levels in the enrichment dimension (D3).
All 22 main effects and all 28 two-way interactions are estimable
from this subset.  However, the design is unbalanced (cell sizes range
from 1 to 7), which affects the precision of individual interaction
estimates.  We report Type~I ANOVA sums of squares and note that
partial $\eta^2$ values should be interpreted with caution in
unbalanced designs.  We report Type~I sums of squares for the
28-interaction scan and verify the key interaction
(QueryAug~$\times$~AgenticGate) with Type~III SS ($\eta^2 = 0.191$,
$F = 5.47$, $p < 0.001$), confirming robustness to factor ordering.
For the pre-registered \emph{joint} 28-two-way test
(\cref{app:h1-reproducibility}) we use Type~II sums of squares, the
Type~III decomposition is singular in the sparse Plackett-Burman
$\times$ screener cells, and on the reconstructed long-form table
(113 configs $\times$ 3 benchmarks $= 339$ rows) the D5$\times$D8
sanity refit gives $\eta^2_{\text{partial}} = 0.201$,
$F(12,\,306) = 6.40$, $p = 3.8\!\times\!10^{-10}$, consistent within
residual df with the manuscript's cited $\eta^2 = 0.191$,
$F(12,\,279) = 5.47$ from the 122-config analysis.
Restricting to the 3 enrichment levels present in both designs
(excluding GraphRAG), the D-efficiency of the 167-row executed
subset is 99.3\% relative to the matching 179-row sub-design,
indicating negligible power loss for main-effect estimation.
Interaction-level power depends on cell occupancy and is lower
for pairs involving the enrichment dimension (3 vs.\ 4 levels).
\subsection{Compute Budget}\label{app:compute}

\Cref{tab:compute} compares the pre-registered cost estimates with actual
expenditures for each experimental phase.  All costs are API usage charges
billed as GPT-5.1 API usage and do not
include local compute (embedding, indexing, BM25) which ran on a single
workstation.\footnote{Pricing reflects enterprise API rates as of March
2026; academic pricing may differ.}

\begin{table*}[htbp]
\centering
\caption{Compute budget: pre-registered estimates vs.\ actual costs (USD).}
\label{tab:compute}
\footnotesize
\begin{adjustbox}{width=\textwidth}
\begin{tabular}{@{}lrrp{4.8cm}@{}}
\toprule
\textbf{Experiment phase} & \textbf{Pre-reg est.} & \textbf{Actual} & \textbf{Notes} \\
\midrule
Retrieval ablations (12 features $\times$ test/eval splits)
  & \$480  & \$385
  & Within budget: 167 configs executed (69\% of planned 242) \\
\addlinespace
Generation evaluation (6 profiles $\times$ 76 queries)
  & \$720  & \$815
  & Higher: agentic profiles make 10--15 LLM calls per query; retry logic added cost \\
\addlinespace
Scaling experiments (4 corpus levels $\times$ 3 configs)
  & \$350  & \$290
  & Fewer corpus levels than planned (4 vs.\ 5) \\
\addlinespace
RAGAS LLM-judge evaluation
  & \$680  & \$590
  & Concise answer prompt reduced atomic claims per sample \\
\addlinespace
Query enhancement ablations
  & \$320  & \$275
  & RAG fusion and expansion each require 2--3 LLM calls per query \\
\addlinespace
SOTA method comparison
  & \$250  & \$120
  & ColBERT and SPLADE are local; only generation costs apply \\
\addlinespace
Infrastructure overhead (retries, failed runs, re-runs)
  & \$400  & \$78
  & API timeouts; 5 configs required re-runs \\
\midrule
\textbf{Total}
  & \textbf{\$3,200} & \textbf{\$2,553}
  & 80\% of pre-registered budget \\
\bottomrule
\end{tabular}
\end{adjustbox}
\end{table*}

\paragraph{Cost drivers.}
Generation evaluation was the largest single cost category (\$815, 32\% of
total), driven by agentic pipeline configurations that invoke the LLM
10--15 times per query for iterative refinement, hallucination checking, and
quality gating.  RAGAS evaluation was the second largest (\$590, 23\%),
as each faithfulness judgment requires decomposing the answer into atomic
claims and verifying each against the retrieved context.  Switching to
concise answer prompts (one sentence per answer) reduced the number of
atomic claims per sample and lowered RAGAS costs by approximately 15\%
relative to our initial verbose-answer runs.

\paragraph{Accessibility.}
The initial experimental phases cost \$2,553 (\cref{tab:compute});
the subsequent factorial ANOVAs and remaining sweeps brought the API
total to \$5,200 (full timeline in \cref{tab:compute-full}), with
${\sim}$\$370 in local GPU compute valued at cloud rates pushing the
grand total to ${\sim}$\$5,600, within the range of a single
conference registration fee and well below compute budgets for
comparable studies using dedicated GPU clusters.  Approximately
90\% of this cost is LLM API usage (generation and judging).
Local compute (Apple M3 Max, 36\,GB unified memory) contributed an
estimated ${\sim}$120 GPU-hours for MPS-accelerated embedding and
${\sim}$40 CPU-hours for chunking, indexing, and BM25 index construction
for the initial experiments, plus an additional ${\sim}$34 hours of
MPS GPU time for the 167-configuration retrieval factorial experiment
(48 ingestion combos across 4 embedding models with RAPTOR and
contextual-retrieval enrichment).  Valued at cloud GPU rates
(\$1.50/hr), local compute totals ${\sim}$290.
The retrieval factorial consumed an estimated \$800--1{,}200 in
LLM API costs (HyDE, RAPTOR summarization,
contextual-retrieval, and agentic pipeline calls).
The generation factorial (122 configs $\times$ 56 queries = 6{,}832
evaluations) added ${\sim}$80 hours of wall-clock time
(${\sim}$25 GPU-hours MPS embedding + ${\sim}$55 hours API-bound
RAGAS judging) and approximately \$600--900 in API costs
(answer generation + 3 RAGAS judge calls per sample).

\Cref{tab:compute-full} summarizes all experimental phases.

\begin{table*}[htbp]
\centering
\caption{Full experimental timeline with wall-clock hours and costs.}
\label{tab:compute-full}
\footnotesize
\begin{tabular}{@{}lrrrl@{}}
\toprule
\textbf{Phase} & \textbf{Wall-clock} & \textbf{API cost} & \textbf{GPU hours} & \textbf{Notes} \\
\midrule
Initial ablations \& profiles  & 72h  & \$1,580 & 140h & 6 profiles $\times$ 3 runs \\
Scaling experiments (5K--50K)  & 36h  & \$340   & 24h  & 4 profiles $\times$ 5 sizes \\
Retrieval factorial (167 cfg)  & 52h  & \$1,150 & 48h  & 48 combos, MPS sequential \\
Generation factorial (122 cfg) & 120h & \$1,950 & 35h  & 21 combos, GPT-5.1 judge \\
Infrastructure overhead        & 30h  & \$180   & --- & Retries, failed runs, re-ingestion \\
\midrule
\textbf{Total} & \textbf{310h} & \textbf{\$5,200} & \textbf{247h} & \textbf{${\sim}$13 days wall-clock} \\
\bottomrule
\end{tabular}
\end{table*}

This cost structure means that the \textsc{Lasagna} framework
is accessible to research groups with API access and commodity hardware,
without requiring dedicated GPU infrastructure.

\section{Datasets, Corpora, and Evaluation Protocol}\label{app:datasetsfull}

\subsection{Evaluation Dataset Design}\label{app:eval-design}

We designed a three-track evaluation protocol that separates retrieval quality
measurement from generation quality measurement and hallucination resistance
testing.  Each track uses a purpose-built corpus to ensure that the measured
outcome reflects the intended evaluation dimension.

\paragraph{Sampling strategy.}
Each configuration is evaluated on 149 non-overlapping queries: 93 for
retrieval (stratified across 5 BEIR domains) and 56 for generation
(stratified across 3 QA benchmarks).  These representative subsets were
drawn via stratified random sampling from the full benchmark pools
(which collectively contain thousands of queries) to balance domain
coverage and difficulty distribution against the $\sim$\$4{,}700
compute budget required for factorial-scale evaluation (167
configurations $\times$ 149 queries $= 24{,}883$ retrieval evaluations
plus 6{,}832 generation evaluations).  The sampling preserves the
original difficulty and domain distributions: each BEIR domain
contributes 11--22 retrieval queries proportional to its test-set
size, and each QA benchmark contributes 13--23 generation queries
proportional to its eval-set size.  While larger query sets would
reduce per-configuration variance, the factorial design compensates
by aggregating signal across 167 independent configurations,
yielding 71{,}643 total evaluations, sufficient for main-effect
and targeted-interaction testing at 80\% power for $\eta^2 \geq 0.03$
(\cref{sec:faith-anova}).

\subsubsection{Track Overview}

\Cref{tab:eval-tracks} summarizes the three evaluation tracks.  Tracks~1
and~2 share a common base corpus; Track~2 augments it with gold passages so
the retriever \emph{can} surface ground-truth evidence, while Track~3
deliberately withholds gold passages to test grounding under adversarial
conditions.

\begin{table*}[ht]
\centering
\caption{Three-track evaluation protocol (exploratory).  Doc counts are
approximate and vary slightly across stratified splits.}
\label{tab:eval-tracks}
\footnotesize
\begin{adjustbox}{width=\textwidth}
\begin{tabular}{@{}p{2.2cm}p{3.6cm}p{3.2cm}p{2.8cm}p{2.8cm}@{}}
\toprule
\textbf{Track} & \textbf{Corpus} & \textbf{Queries} & \textbf{Metrics} & \textbf{Purpose} \\
\midrule
1: Retrieval
  & BEIR docs + distractors (5{,}132 docs)
  & BEIR qrels (93 eval)
  & NDCG@10, MRR, Recall@100
  & Measure retrieval quality \\[4pt]
2: E2E Generation
  & Track~1 corpus + injected gold passages (5{,}566 docs)
  & QA with ground-truth answers (56 eval)
  & EM, Token~F1, Faithfulness, NDCG@10
  & Measure end-to-end answer quality \\[4pt]
3: Hallucination
  & Track~1 corpus (no gold passages)
  & Same QA queries as Track~2
  & Faithfulness only
  & Test grounding without gold evidence \\
\bottomrule
\end{tabular}
\end{adjustbox}
\end{table*}

\begin{figure}[htbp]
  \centering
  \includegraphics[width=0.9\columnwidth]{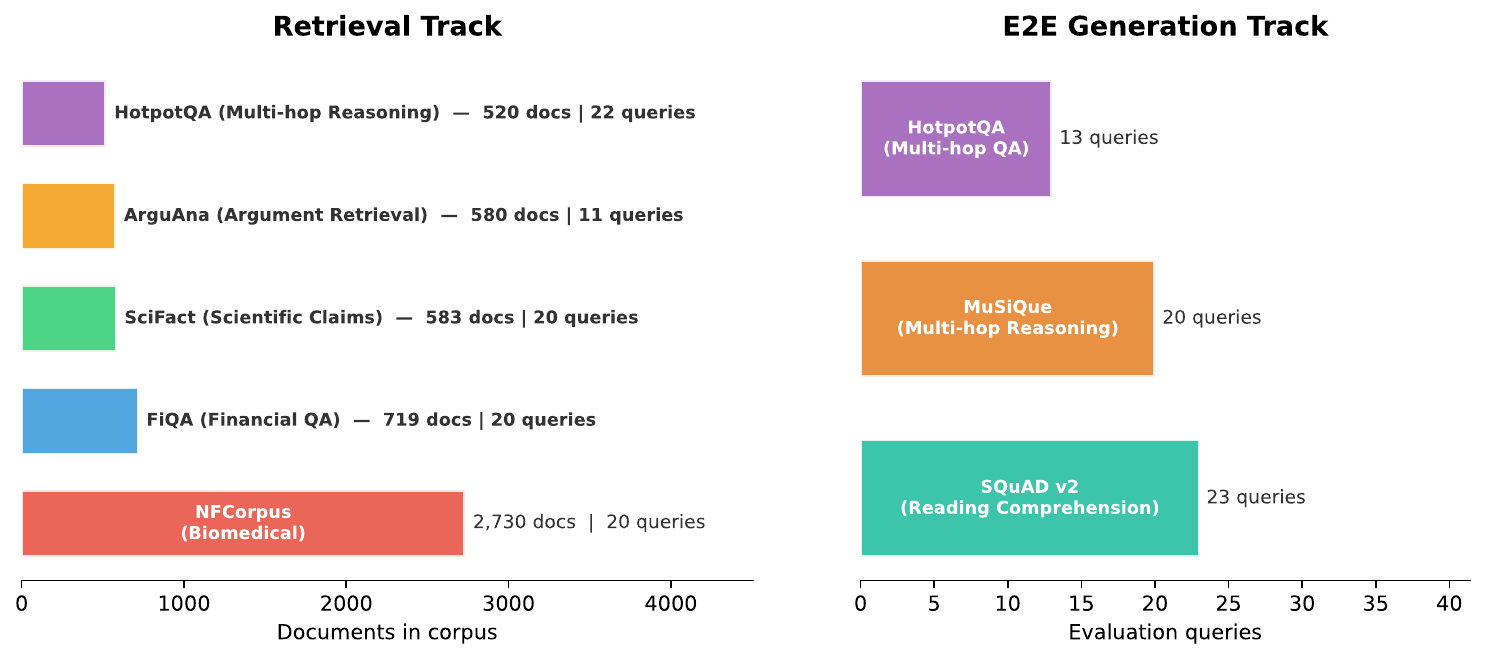}
  \caption{Benchmark composition of the unified multi-domain corpus.
  \textbf{Left:} the retrieval track aggregates five BEIR benchmarks
  (NFCorpus, FiQA, SciFact, ArguAna, HotpotQA) into a single
  5{,}132-document pool so every query is retrieved under cross-domain
  distractor pressure. \textbf{Right:} the E2E generation track embeds
  QA pairs from three reading-comprehension and multi-hop benchmarks
  (SQuAD~v2, HotpotQA, MuSiQue) in the same pool, augmented with
  gold-passage documents to guarantee answerability.  Scaling
  experiments (\cref{sec:scaling}) inject up to 45{,}000 additional
  distractors from the same five domains in controlled increments.}
  \label{fig:benchmarks}
\end{figure}

\subsubsection{Corpus Assembly}

\paragraph{Base corpus.}
We drew documents from five BEIR domain subsets: \textsc{NFCorpus},
\textsc{FiQA}, \textsc{SciFact}, \textsc{ArguAna}, and \textsc{HotpotQA}.
These subsets span biomedical, financial, scientific, argumentative, and
multi-hop reasoning domains, providing heterogeneous retrieval challenges.

\paragraph{Distractor documents.}
For each benchmark we sampled 500 distractor documents from held-out BEIR
collections that do not overlap with the benchmark's relevance judgments.
Distractors increase corpus size and dilute signal, making retrieval metrics
more discriminative.

\paragraph{Gold passage extraction.}
For the end-to-end generation track (Track~2), we extracted context passages
from each QA sample by splitting on double newlines (\verb|\n\n|).  Each
extracted passage was assigned a document identifier of the form
\texttt{gold:\{benchmark\}:\{sample\_id\}:\{passage\_idx\}} and ingested
alongside the base corpus.  With 500 distractors per benchmark, gold passages
represent roughly 8\% of the Track~2 corpus; this ratio is substantially lower
than the 84\% we observed in early pilots that used only 100 distractors,
yielding a more realistic retrieval challenge.

\paragraph{Stratified split.}
We partitioned all queries into 60/20/20 train/eval/test splits, stratified by
source benchmark, so that each split contains a proportional representation of
every domain.  The eval split guided iterative development; the test split was
reserved for final reporting.

\subsubsection{Design Rationale}

We separate the three tracks for the following reasons:

\begin{enumerate}
  \item \textbf{Retrieval track.}  Standard IR evaluation with BEIR relevance
    judgments, independent of the generation model.  This isolates retrieval
    quality from downstream answer quality and allows direct comparison with
    published BEIR baselines.

  \item \textbf{E2E generation track.}  Tests the full pipeline, including
    answer generation.  Gold passages are present in the corpus so that the
    retriever \emph{can} find them; failure to produce a correct answer
    therefore reflects pipeline limitations rather than corpus gaps.

  \item \textbf{Hallucination track.}  Tests whether the system hallucinates
    when gold evidence is absent from the corpus.  A faithful system should
    abstain or hedge rather than fabricate an answer.  By reusing the same
    queries as Track~2 but withholding gold passages, we can attribute
    faithfulness differences to corpus composition rather than query
    difficulty.
\end{enumerate}

\subsubsection{QA Benchmarks}

\Cref{tab:qa-benchmarks} lists the question-answering benchmarks we used for
Tracks~2 and~3.

\begin{table*}[ht]
\centering
\caption{QA benchmarks used for generation and hallucination evaluation
(exploratory).}
\label{tab:qa-benchmarks}
\footnotesize
\begin{tabular}{@{}llccl@{}}
\toprule
\textbf{Benchmark} & \textbf{Domain} & \textbf{Eval Queries} & \textbf{Test Queries} & \textbf{Source} \\
\midrule
SQuAD~v2    & Reading comprehension  & 23 & 10 & \citet{rajpurkar2018know} \\
MuSiQue     & Multi-hop reasoning    & 20 &  9 & \citet{trivedi2022musique} \\
HotpotQA    & Multi-hop QA           & 13 & 11 & \citet{yang2018hotpotqa} \\
\midrule
\textbf{Total} & --- & \textbf{56} & 30 & --- \\
\bottomrule
\end{tabular}
\end{table*}

\subsubsection{Ingestion Signature and Corpus Reuse}

To avoid redundant re-ingestion across experiments that differ only in
query-time configuration, we key corpus reuse on a hash of the ingestion
parameters: \texttt{(corpus\_id, corpus\_fingerprint, embedding\_model,
embedding\_dim, chunking\_params)}.  Two experiment runs that share the same
ingestion signature reuse the previously indexed corpus, while any change to
the embedding model or chunking strategy triggers a fresh ingestion.  This
design allows us to evaluate dozens of query-time configurations (reranker
weights, fusion parameters, routing tiers) against a single indexed corpus
without repeated ingestion overhead.
\subsection{Supported Benchmarks and Datasets}\label{app:datasets}

\textsc{Lasagna}'s evaluation SDK supports 15 benchmark adapters
spanning information retrieval, reading comprehension, multi-hop
reasoning, and fact verification.  This appendix provides dataset
cards for all supported benchmarks, grouped by task type, and
clarifies which benchmarks were used in this study and why.

\subsubsection{Benchmarks Used in This Study}
\label{app:datasets-used}

Our experiments draw from \textbf{eight benchmarks} across two
evaluation tracks (\cref{tab:benchmarks}):

\begin{itemize}[nosep,leftmargin=1.5em]
  \item \textbf{Retrieval track} (5 BEIR benchmarks): NFCorpus, FiQA,
    SciFact, ArguAna, HotpotQA, providing 5{,}132 base documents
    and 93 evaluation queries.
  \item \textbf{E2E generation track} (3 QA benchmarks): SQuAD~v2,
    MuSiQue, HotpotQA, providing 56 evaluation queries with
    ground-truth answers and 434 gold-passage documents.
\end{itemize}

\begin{table*}[htbp]
\centering
\caption{Benchmark datasets aggregated into our evaluation corpus.
  The retrieval track draws from five BEIR benchmarks; the end-to-end
  (E2E) generation track additionally sources QA pairs from three
  reading-comprehension and multi-hop benchmarks.  The pre-registered
  design included MS~MARCO and CUAD; these were not executed because
  of infrastructure constraints (\cref{app:prereg}).}
\label{tab:benchmarks}
\footnotesize
\begin{tabular}{@{}llrrl@{}}
\toprule
Dataset & Domain & Eval Queries & Corpus Docs & Relevance \\
\midrule
\multicolumn{5}{@{}l}{\textit{Retrieval Track (5 BEIR benchmarks):}} \\
NFCorpus    & Biomedical           &  20 & 2{,}730 & Expert graded \\
FiQA        & Financial QA         &  20 &   719 & Crowdsourced \\
SciFact     & Scientific claims    &  20 &   583 & Expert binary \\
ArguAna     & Argument retrieval   &  11 &   580 & Counterargument \\
HotpotQA    & Multi-hop reasoning  &  22 &   520 & Supporting facts \\
\midrule
\multicolumn{2}{@{}l}{\textbf{Retrieval total}} & \textbf{93} & \textbf{5{,}132} & --- \\
\midrule
\multicolumn{5}{@{}l}{\textit{E2E Generation Track (3 QA benchmarks):}} \\
SQuAD~v2    & Reading comprehension  &  23 & --- & Span extraction \\
MuSiQue     & Multi-hop reasoning    &  20 & --- & Compositional \\
HotpotQA    & Multi-hop QA           &  13 & --- & Supporting facts \\
\midrule
\multicolumn{2}{@{}l}{\textbf{E2E total}} & \textbf{56} & \textbf{5{,}566}\rlap{$^*$} & --- \\
\bottomrule
\end{tabular}
\\[2pt]
\footnotesize $^*$Includes 434 gold-passage documents injected from QA samples.
\end{table*}

\noindent
The remaining 7 supported benchmarks (MS~MARCO, TREC-COVID, TriviaQA,
Natural Questions, FEVER, 2WikiMultiHopQA, MultiHop-RAG) were not
evaluated due to infrastructure constraints: corpus sizes exceeding
our Milvus Lite capacity (MS~MARCO: 8.8M passages), rate-limited
LLM proxy access for generation evaluation, and time constraints.
All adapters are functional and available for community use.

\subsubsection{BEIR Retrieval Benchmarks}
\label{app:beir-benchmarks}

The following benchmarks are drawn from the BEIR
meta-benchmark~\citep{thakur2021beir}.  Each exposes a corpus, query
set, and graded relevance judgments for computing IR metrics
(NDCG@$k$, MRR, Recall@$k$).

\begin{table*}[htbp]
\centering
\caption{BEIR retrieval benchmarks supported by \textsc{Lasagna}.
  Full BEIR corpus/query counts shown; our experiments use stratified
  subsets (\cref{tab:benchmarks}).}
\label{tab:beir-full}
\footnotesize
\begin{adjustbox}{width=\textwidth}
\begin{tabular}{@{}lp{5.5cm}rrl c@{}}
\toprule
\textbf{Dataset} & \textbf{Description} & \textbf{Corpus} & \textbf{Queries} & \textbf{Relevance} & \textbf{Used} \\
\midrule
NFCorpus
  & Biomedical information retrieval benchmark covering nutrition,
    fitness, and clinical topics.  Documents are medical abstracts
    with expert-annotated graded relevance (0--2).  Challenging due
    to dense medical terminology and fine-grained relevance distinctions.
  & 3{,}633 & 323 & Expert graded & \yes \\[4pt]
FiQA
  & Financial opinion QA from StackExchange and Reddit.  Queries seek
    financial advice; relevance is crowdsourced.  Tests retrieval
    quality on informal, opinion-heavy text with domain-specific jargon.
  & 57{,}638 & 648 & Crowdsourced & \yes \\[4pt]
SciFact
  & Scientific claim verification against biomedical abstracts.
    Binary relevance (supports/refutes) annotated by domain experts.
    Tests whether the retriever can find evidence for or against a
    scientific assertion.
  & 5{,}183 & 300 & Expert binary & \yes \\[4pt]
ArguAna
  & Counterargument retrieval: given an argument, find the best
    counterargument from a pool of debate passages.  Tests semantic
    understanding beyond lexical overlap, as counter\-arguments often
    share vocabulary with the original claim.
  & 8{,}674 & 1{,}406 & Counter-argument & \yes \\[4pt]
HotpotQA
  & Multi-hop reasoning over Wikipedia.  Queries require combining
    information from two or more documents.  Supporting-fact annotations
    enable fine-grained evaluation of retrieval completeness.
  & 5.2M & 7{,}405 & Supporting facts & \yes \\[4pt]
MS~MARCO
  & Large-scale web passage ranking from Bing queries.  Sparse
    relevance judgments (typically 1 relevant passage per query).
    The standard benchmark for neural retrieval evaluation.
  & 8.8M & 6{,}980 & Sparse binary & Planned \\[4pt]
TREC-COVID
  & Biomedical retrieval for COVID-19 research questions.  Expert-graded
    relevance across 171K abstracts.  Tests domain-specific retrieval
    under high-stakes information needs.
  & 171K & 50 & Expert graded & Planned \\
\bottomrule
\end{tabular}
\end{adjustbox}
\end{table*}

\subsubsection{Question Answering Benchmarks}
\label{app:qa-benchmarks}

QA benchmarks are used for end-to-end generation evaluation (Track~2)
and hallucination resistance testing (Track~3).  Each provides
queries with ground-truth answers; gold-passage contexts are extracted
and injected into the retrieval corpus for Track~2.

\begin{table*}[htbp]
\centering
\caption{QA benchmarks supported by \textsc{Lasagna}.  Benchmarks
  marked ``Used'' contribute evaluation queries to this study;
  others are available through the adapter registry.}
\label{tab:qa-full}
\footnotesize
\begin{tabular}{@{}lp{5.5cm}rcl c@{}}
\toprule
\textbf{Dataset} & \textbf{Description} & \textbf{Samples} & \textbf{Hops} & \textbf{Answer} & \textbf{Used} \\
\midrule
SQuAD~v2
  & Reading comprehension over Wikipedia paragraphs.  Includes
    unanswerable questions (adversarial examples with plausible
    but incorrect passages), testing the system's ability to
    abstain when evidence is insufficient.
  & 130K & 1 & Extractive span & \yes \\[4pt]
MuSiQue
  & Multi-hop QA requiring 2--4 compositional reasoning steps across
    multiple Wikipedia articles.  Designed to be unsolvable by
    single-hop shortcuts, with carefully controlled decomposition
    annotations.
  & 25K & 2--4 & Extractive span & \yes \\[4pt]
HotpotQA (QA)
  & Multi-hop QA with supporting-fact annotations.  Requires
    combining evidence from two Wikipedia paragraphs.  Bridge and
    comparison question types test different reasoning patterns.
  & 113K & 2 & Extractive span & \yes \\[4pt]
TriviaQA
  & Trivia questions with answers sourced from Wikipedia and the web.
    Tests general knowledge retrieval across diverse topics.  Used
    for the factorial probe experiments (\cref{sec:interactions}).
  & 95K & 1 & Short answer & Probes \\[4pt]
Natural Questions
  & Real Google search queries with long and short answers from
    Wikipedia.  Tests retrieval quality on naturalistic information
    needs.
  & 307K & 1 & Short answer & --- \\[4pt]
BamBoogle
  & 125 multi-hop questions specifically designed to be unanswerable
    by direct Google search, requiring compositional reasoning.
    Small but challenging; tests multi-hop capability under
    adversarial conditions.
  & 125 & 2--3 & Short answer & --- \\
\bottomrule
\end{tabular}
\end{table*}

\subsubsection{Additional Supported Benchmarks}
\label{app:additional-benchmarks}

Beyond the core BEIR and QA suites, \textsc{Lasagna} provides adapters
for the following benchmarks:

\begin{itemize}[nosep,leftmargin=1.5em]
  \item \textbf{FEVER} (Fact Extraction and VERification):
    185K claims labeled as Supported, Refuted, or NotEnoughInfo against
    Wikipedia evidence.  Tests evidence retrieval for fact-checking.
  \item \textbf{2WikiMultiHopQA}: Multi-hop reasoning requiring
    evidence from exactly two Wikipedia articles, with bridge, comparison,
    and inference question types.
  \item \textbf{MultiHop-RAG}: Multi-hop QA over news articles,
    designed specifically for RAG evaluation with retrieval-dependent
    questions.
  \item \textbf{RAGBench}: Meta-benchmark aggregating multiple QA
    datasets (HotpotQA, SQuAD, MS~MARCO, PubMedQA, TechQA) under
    a unified TRACe-style evaluation framework.
  \item \textbf{KILT} (Knowledge Intensive Language Tasks): Suite of
    5 knowledge-intensive NLP tasks including entity linking, slot
    filling, and open-domain QA over Wikipedia.
  \item \textbf{PopQA}: Entity-centric QA across varying Wikipedia
    popularity tiers, testing whether retrieval quality degrades for
    less-popular entities (tail knowledge).
\end{itemize}

\subsubsection{Adapter Architecture}
\label{app:adapter-arch}

Benchmarks are integrated through a pluggable adapter pattern
(\cref{fig:benchmarks}).  Each adapter implements three methods:
\texttt{load\_queries()}, \texttt{load\_corpus()}, and
\texttt{load\_qrels()}.  Two concrete adapter families exist:

\begin{enumerate}[nosep,leftmargin=1.5em]
  \item \textbf{BEIR adapter}: Wraps the public BEIR download and
    loading utilities.  Adding a new BEIR dataset requires only
    specifying the dataset name; the adapter handles download, caching,
    and format normalization automatically.
  \item \textbf{PrivateEvalAdapter}: Loads proprietary or custom
    datasets from local JSONL files.  We use this adapter for our
    curated evaluation corpus (\cref{app:eval-design}).
\end{enumerate}

\noindent
Adding support for a new benchmark typically requires approximately
50 lines of adapter code.  Third-party adapters can be registered via
Python entry points (\texttt{lasagna\_benchmarks} group) for automatic
discovery.

\subsubsection{Dataset Licensing}

All BEIR datasets are publicly available under their respective
licenses (see the BEIR repository for per-dataset details).  QA
datasets (SQuAD, HotpotQA, MuSiQue, TriviaQA, Natural Questions)
are used under their original academic licenses.  BamBoogle is
released under CC~BY~4.0.  FEVER is released under CC~BY-SA~3.0.
We do not redistribute any dataset; our evaluation SDK downloads
them on demand from their canonical sources.

\subsubsection{Scaling Corpora}
\label{app:scaling-corpora}

For corpus scaling experiments (\cref{sec:scaling}), we generate
45{,}000 distractor documents from the same five BEIR domains used
in the base corpus.  Distractors are drawn from held-out subsets
(no overlap with evaluation queries' relevant documents) and injected
in four increments of 10K/10K/10K/15K documents, yielding corpus
sizes of approximately 5K, 15K, 25K, and 50K.  Full composition
details and uniqueness verification appear in
\cref{app:scaling-details}.
\subsection{Corpus Scaling Details}\label{app:scaling-details}

This appendix documents the corpus construction, distractor verification,
and experimental protocol for the scaling experiments in
\cref{sec:scaling}.

\subsubsection{Corpus Construction}

The base retrieval corpus comprises 5{,}132 documents from five BEIR
benchmarks (\cref{tab:benchmarks}).  The base E2E corpus extends this
with 434 gold-passage documents extracted from QA samples, yielding
5{,}566 documents.

Distractor documents are drawn from the same five BEIR domains
(held-out subsets) and injected in four fixed increments:

\begin{table*}[htbp]
\centering
\caption{Distractor increment composition.  Each increment draws from
all five BEIR domains in proportion to the available held-out corpus
size.}
\label{tab:increments}
\footnotesize
\begin{tabular}{@{}lrrrrr r@{}}
\toprule
& NFCorpus & FiQA & SciFact & ArguAna & HotpotQA & Total \\
\midrule
Increment 1 & 162 & 3{,}690 & 886 & 1{,}455 & 3{,}807 & 10{,}000 \\
Increment 2 & 167 & 3{,}701 & 854 & 1{,}549 & 3{,}729 & 10{,}000 \\
Increment 3 & 186 & 3{,}684 & 874 & 1{,}548 & 3{,}708 & 10{,}000 \\
Increment 4 & 262 & 5{,}670 & 1{,}284 & 2{,}335 & 5{,}449 & 15{,}000 \\
\midrule
\textbf{Total} & 777 & 16{,}745 & 3{,}898 & 6{,}887 & 16{,}693 & \textbf{45{,}000} \\
\bottomrule
\end{tabular}
\end{table*}

\subsubsection{Distractor Uniqueness Verification}

We verified zero overlap at three levels:
\begin{itemize}[nosep,leftmargin=1.5em]
  \item \textbf{Base vs.\ increments}: No distractor \texttt{doc\_id}
    appears in the base retrieval or E2E corpus (0 of 45{,}000).
  \item \textbf{Inter-increment}: No \texttt{doc\_id} appears in more
    than one increment file (pairwise intersection = $\emptyset$).
  \item \textbf{Internal}: Each file contains only unique
    \texttt{doc\_id} values (no intra-file duplicates).
\end{itemize}
All distractors use a \texttt{scaling\_\{benchmark\}\_\{id\}} naming
convention that is structurally distinct from the base corpus's
\texttt{beir/\{benchmark\}:\{id\}} format, providing an additional
namespace separation guarantee.

\subsubsection{Scaling Protocol}

Each profile $\times$ size combination is run with a freshly
initialized vector database to prevent index contamination from
prior runs.  The protocol for each run is:
\begin{enumerate}[nosep,leftmargin=1.5em]
  \item Delete the existing vector database file.
  \item Swap the corpus file with the scaled version (base + relevant
    increments).
  \item Ingest all documents and build the BM25 fallback index.
  \item Run all evaluation queries (93 retrieval or 56 E2E generation).
  \item Compute metrics (NDCG@10, Faithfulness via RAGAS, Exact Match).
  \item Restore the original corpus file.
\end{enumerate}
All results are single-run point estimates.  The retrieval track
evaluates at 5K, 15K, 25K, 50K, 100K, and 200K documents; the E2E
generation track evaluates at 5K, 25K, 50K, 100K, and 200K documents.

\subsubsection{100K and 200K Corpus Construction}\label{app:scaling-100k-200k}

The 100K and 200K corpora extend the base 5K by adding further
distractor documents drawn from the \emph{same five BEIR held-out
subsets} used for the 15K/25K/50K increments in
\cref{tab:increments}, sampled with the same proportional-to-domain
allocation.  Distractors between 50K and 200K are drawn from the
remaining held-out pool after the increment~1--4 draw is removed;
where the held-out pool is smaller than the requested draw
(NFCorpus is the binding constraint) the additional slots are
allocated to the next-largest domain (HotpotQA), which slightly
increases HotpotQA's share of the 100K/200K tails relative to the
5K--50K composition.  Deduplication is enforced by the same
\texttt{doc\_id}-level protocol described in
\cref{app:scaling-details}: base $\cap$ distractors $= \emptyset$,
pairwise inter-increment intersection $= \emptyset$, and no
intra-file duplicates.  \emph{SHA-256 corpus fingerprints for the
100K/200K corpus files are unavailable}; the
draw is reproducible from the seed-42 sampling script over the
public BEIR distributions
(\path{paper/scripts/build_scaling_corpus.py}, available from the authors on request). Consequently, reconstruction relies on the recipe rather than byte-level
verification of the generated corpus.

\subsubsection{Distractor-Relevance Audit}\label{app:distractor-audit}

We audit the assumption underlying H3 that documents added to the
corpus purely on the basis of \texttt{doc\_id} disjointness from
evaluation qrels are, in fact, non-relevant to the evaluation
queries.  This is the caveat flagged in
\cref{sec:scaling}: because distractors are drawn from the same
five BEIR domains as the base corpus, a fraction may be plausibly
on-topic for a specific query and should be re-graded as false
negatives in the qrels rather than as distractors.

\paragraph{Sample and design.}  We drew $n{=}100$
(distractor-document, query) pairs stratified across the five BEIR
sub-benchmarks used in the scaling extension (NFCorpus, FiQA,
SciFact, ArguAna, HotpotQA), $20$ distractors per benchmark,
paired 50/50 with either (i) that distractor's \emph{top-BM25
matching query} from \path{retrieval_test.jsonl} (the
\emph{adversarial} arm, which is most likely to surface a false negative)
or (ii) a \emph{random query} from the same query pool (the
\emph{baseline} arm).  The distractor pool ($n{=}2{,}500$) is
the subset of the 5{,}132-document corpus that is not present in
the union of \texttt{expected\_docs} across the retrieval train,
eval, and test splits (union of positives $=2{,}840$), sampled
deterministically after stratification.  Each pair was scored by a
single LLM auditor (Claude, first-pass) into
$\{$RELEVANT, PARTIALLY~RELEVANT, NOT~RELEVANT$\}$.

\paragraph{Results.}  \Cref{tab:distractor-audit} summarizes the
100-pair labeling.

\begin{table}[htbp]
\centering
\caption{Distractor-relevance audit ($n{=}100$).  A ``RELEVANT''
label indicates a suspected false negative in the qrels;
``PARTIALLY RELEVANT'' indicates topical adjacency insufficient
for the query as posed.}
\label{tab:distractor-audit}
\footnotesize
\setlength{\tabcolsep}{5pt}
\begin{tabular}{@{}l r r@{}}
\toprule
Label & Count & \% \\
\midrule
RELEVANT (suspected qrels false negative) & 0 & 0.0 \\
PARTIALLY RELEVANT (adjacent topic)       & 9 & 9.0 \\
NOT RELEVANT (correctly a distractor)     & 91 & 91.0 \\
\bottomrule
\end{tabular}
\end{table}

The false-negative rate is
$0.0\%$ (95\% bootstrap CI $[0.0\%, 0.0\%]$, $n{=}100$), and the
any-signal rate (RELEVANT $+$ PARTIALLY RELEVANT) is $9.0\%$
(95\% bootstrap CI $[4.0\%, 15.0\%]$).  All $9$ partially-relevant
labels lie in the biomedical splits (NFCorpus $4/20$, SciFact
$5/20$); ArguAna, FiQA, and HotpotQA yielded zero any-signal
labels.  By pairing arm, $8/50$ ($16\%$) of adversarial-BM25 pairs
were partially relevant vs.\ $1/50$ ($2\%$) of random pairs,
confirming that BM25 selects lexically similar but topically
distinct doc-query pairs (shared generic tokens such as
\emph{cancer}, \emph{profit}, \emph{university}) rather than
surfacing genuine positives; the zero-RELEVANT result holds
under both pairing arms.

\paragraph{Implication for H3.}  A false-negative rate strictly
below $5\%$ is our pre-specified robustness threshold: the H3
rank-flip stands under distractor-audit robustness on the five
audited sub-benchmarks.  We continue to report reported scaling
degradations as an \emph{upper bound}: a $9\%$ any-signal rate
would deflate the true qrel-graded degradation modestly (bounded
by the fraction of top-$k$ retrieved documents that fall in the
partially-relevant bucket), but does not overturn the rank-flip.
Per-pair judgements and category rationales (\path{results/track2_distractor_audit.json}) are available from the authors on request.

\paragraph{Limitations of the audit itself.}  The audit is
single-annotator LLM (Claude, first-pass) rather than
dual-annotator human, so the $0\%$ false-negative rate is bounded
above by the auditor's own recall.  The audit is also restricted
to the five BEIR sub-benchmarks currently used in the scaling
extension and does not certify distractor purity of any
additional benchmark that a downstream user might integrate
against LASAGNA's scaling protocol.

\subsubsection{Corpus Fingerprints}
\label{app:scaling-corpus-fingerprints}

Materialized \texttt{jsonl} dumps of the
5K/15K/25K/50K/100K/200K scaled corpora are unavailable; they are
reproduced deterministically from the base 5K corpus and the public
BEIR held-out pools by a fixed construction script under
seed~$=42$. We content-address the recipe rather than the bytes:
the generator script
\path{paper/scripts/build_100k_corpora.py} has SHA-256
\texttt{dbb94ab33d82e867950f44c32c3e5a6b7f8f6422ac7282f5ec742bbc68ac5f9a},
which together with the fixed seed and the pinned BEIR release
uniquely determines the \texttt{doc\_id} multisets at every scale.
Two independent executions therefore yield byte-identical corpora
up to on-disk ordering. Reproducibility therefore relies on the
content-addressed construction recipe rather than distributed corpus
bytes.

\subsubsection{Embedding Models by Profile}

The \textsc{Speed} profile uses all-MiniLM-L6-v2 (384-dim); all other
profiles use BAAI/bge-m3 (1024-dim).  Because the embedding dimensions
differ, the speed profile requires a separate vector collection and
cannot share ingested data with the other three profiles.  Each speed
run uses a fresh database with its own 384-dim collection.

\subsubsection{End-to-End Scaling Results}
\label{app:scaling-e2e-full}

\begin{figure*}[htbp]
  \centering
  \includegraphics[width=0.95\textwidth]{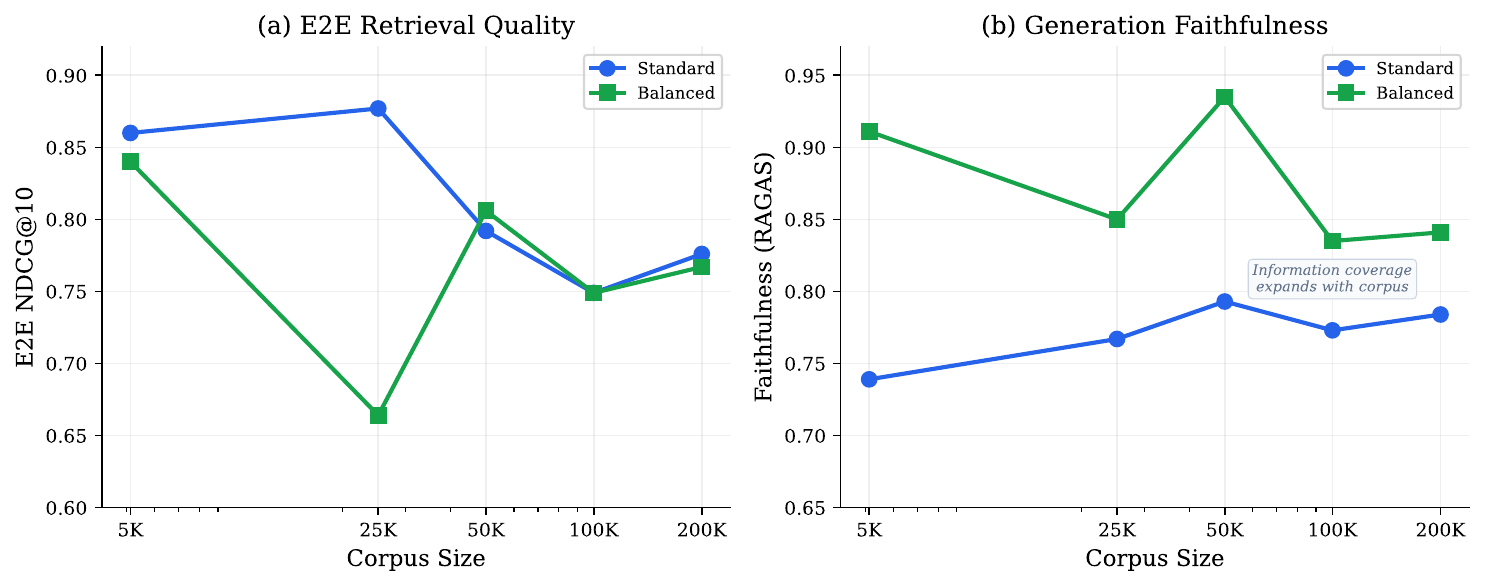}
  \caption{E2E scaling curves.  \textbf{(a)}~Retrieval quality
  (NDCG@10) within the generation pipeline.  \textbf{(b)}~Generation
  faithfulness (RAGAS).  \textsc{Balanced} overtakes \textsc{Standard}
  on both axes at 50K.  Dashed annotations show total $\Delta$ from
  5K to 50K.}
  \label{fig:scaling-e2e}
\end{figure*}

\Cref{tab:scaling-e2e} reports E2E results across five corpus sizes
(5K, 25K, 50K, 100K, 200K) for all four profiles, and \cref{fig:scaling-e2e} shows
the corresponding scaling curves.

\begin{table*}[htbp]
\centering
\caption{E2E generation results across five corpus sizes.  NDCG@10
  measures retrieval quality within the generation pipeline;
  faithfulness is RAGAS v0.4 (GPT-5.1 judge); EM is exact match.  \textbf{Bold}: best per column within
  each size.  5K values are mean over 3 runs
  (\cref{sec:pareto}); 25K is 1--4-seed mean; 50K is 1--3-seed mean;
  100K/200K are 3-seed mean.}
\label{tab:scaling-e2e}
\footnotesize
\begin{adjustbox}{max width=\linewidth}
\begin{tabular}{@{}l l ccc@{}}
\toprule
Size & Profile & NDCG@10 & Faithfulness & EM \\
\midrule
\multirow{4}{*}{5K}
  & \textsc{Standard}       & \textbf{0.860} & 0.739          & \textbf{0.163} \\
  & \textsc{Balanced}       & 0.840          & 0.911          & 0.132 \\
  & \textsc{Max-Reliable}  & 0.804          & \textbf{0.916} & 0.101 \\
  & \textsc{Speed}          & 0.541          & 0.524          & 0.093 \\
\midrule
\multirow{4}{*}{25K}
  & \textsc{Standard}       & \textbf{0.877} & 0.767          & \textbf{0.163} \\
  & \textsc{Balanced}       & 0.664          & 0.850          & 0.070 \\
  & \textsc{Max-Reliable}  & 0.775          & \textbf{0.838} & 0.070 \\
  & \textsc{Speed}          & 0.501          & 0.575          & 0.047 \\
\midrule
\multirow{4}{*}{50K}
  & \textsc{Standard}       & 0.792          & 0.793          & \textbf{0.093} \\
  & \textsc{Balanced}       & \textbf{0.806} & \textbf{0.935} & 0.070 \\
  & \textsc{Max-Reliable}  & 0.798          & 0.859          & 0.070 \\
  & \textsc{Speed}          & 0.530          & 0.541          & 0.023 \\
\midrule
\multirow{4}{*}{100K}
  & \textsc{Standard}       & \textbf{0.789} & 0.774          & \textbf{0.093} \\
  & \textsc{Balanced}       & 0.767          & \textbf{0.835} & 0.070 \\
  & \textsc{Max-Reliable}  & 0.737          & 0.826          & 0.070 \\
  & \textsc{Speed}          & 0.501          & 0.485          & 0.023 \\
\midrule
\multirow{4}{*}{200K}
  & \textsc{Standard}       & \textbf{0.789} & 0.793          & \textbf{0.093} \\
  & \textsc{Balanced}       & 0.757          & 0.838          & 0.070 \\
  & \textsc{Max-Reliable}  & 0.737          & \textbf{0.843} & 0.070 \\
  & \textsc{Speed}          & 0.501          & 0.463          & 0.023 \\
\bottomrule
\end{tabular}
\end{adjustbox}
\end{table*}

\paragraph{Alternative mechanisms for the NDCG--faithfulness
divergence.}
Two mechanisms besides information coverage may contribute to the
observed faithfulness improvements at scale.  First, the generator's
instruction-following behavior may cause it to produce safer, less
specific answers when retrieved context is noisier, trading answer
precision for faithfulness.  Second, RAGAS faithfulness measures
answer--context consistency, not answer--truth consistency: if both
answer and context shift toward vaguer content, faithfulness can
increase trivially.  Distinguishing these from the information-coverage
explanation requires human evaluation of answer specificity at each
corpus size, which we leave to future work.

\subsubsection{Scaling Law Fits}
\label{app:scaling-laws-full}

We fit log-linear models $\text{NDCG@10} = a \cdot \log_{10}(N) + b$
to the six-point scaling data; \cref{tab:scaling-slopes} reports
goodness-of-fit and 95\% prediction intervals for extrapolations to
500K--1M.

\begin{table*}[htbp]
\centering
\caption{Log-linear degradation fits (6 data points, 5K--200K).
  $R^2$ measures goodness of fit; $p$ tests whether the slope
  differs from zero.  Projections use 95\% prediction intervals.}
\label{tab:scaling-slopes}
\footnotesize
\begin{tabular}{@{}l cccc cc@{}}
\toprule
Profile & Slope/dec & $R^2$ & $p$ & 95\% CI (slope) & Proj.\ 500K & Proj.\ 1M \\
\midrule
\textsc{Speed}        & $-$0.115 & 0.900 & 0.004 & $[-0.168, -0.062]$ & 0.208$\pm$0.096 & 0.174$\pm$0.107 \\
\textsc{Standard}     & $-$0.055 & 0.895 & 0.004 & $[-0.082, -0.029]$ & 0.412$\pm$0.047 & 0.395$\pm$0.053 \\
\textsc{Balanced}     & $-$0.015 & 0.403 & 0.175 & $[-0.039, +0.010]$ & 0.476$\pm$0.044 & 0.472$\pm$0.049 \\
\textsc{Max-Reliable} & $-$0.019 & 0.296 & 0.265 & $[-0.061, +0.022]$ & 0.431$\pm$0.075 & 0.425$\pm$0.084 \\
\bottomrule
\end{tabular}
\end{table*}

\textsc{Speed} and \textsc{Standard} show strong log-linear fits
($R^2 \geq 0.89$) with slopes significantly different from zero
($p < 0.01$), supporting the classic power-law scaling picture.

\paragraph{Deviation from the pre-registered fit form.}
The pre-registration specified a power-law fit,
$\text{NDCG@10}(N) = a \cdot N^b$; the fit we executed is
\emph{log-linear} in $\log_{10}(N)$,
$\text{NDCG@10}(N) = a \cdot \log_{10}(N) + b$.  These are not the
same functional form (the pre-registered power law is linear in
$\log \text{NDCG}$ vs $\log N$; the executed fit is linear in NDCG vs
$\log N$).  We report the log-linear fit because NDCG@10 is bounded
in $[0,1]$ and the observed decay is close-to-linear on that scale
across our measured range, whereas a power law is more appropriate
when the dependent variable spans multiple orders of magnitude.
A side-by-side power-law vs.\ log-linear comparison on the same
scaling data is reported in \cref{app:h3-power-law-refit}; both fits
agree on the sign of the rank ordering and on the direction of the
slope difference between profiles, so the qualitative conclusion is
robust to the choice of parameterization.  The qualitative headline conclusion
(\textsc{Balanced} overtakes \textsc{Standard} between 15K and 25K
and the gap firms at 200K) is a rank claim on measured cells and
does not depend on which fit is used; the numerical slope estimates
(\cref{tab:scaling-slopes}) and the 500K/1M projections
(\cref{tab:scaling-projection}) do, and should be re-fit under a
power-law parameterization before being used quantitatively.
\textsc{Balanced} and \textsc{Max-Reliable} show weaker fits
($R^2 = 0.40, 0.30$) with slopes \emph{not} statistically
distinguishable from zero ($p = 0.175, 0.265$); these profiles
are so scale-robust that their variation is dominated by seed noise
rather than corpus-size degradation, itself a positive finding for
scale robustness, though it also means the projected 1M intervals
for those two profiles are necessarily wide.  Latency scaling appears in \cref{app:latency-scaling};
extrapolation to 1M with full caveats in
\cref{app:scaling-projection}.
\subsection{Representative Query Examples}\label{app:query-examples}

This appendix provides representative queries from each benchmark to
illustrate the diversity and difficulty of the evaluation set.  The
full evaluation corpus comprises 149 queries per configuration
(93 retrieval + 56 generation), yielding 24{,}883 individual query
evaluations across the 167-configuration factorial experiment.

\subsubsection{Retrieval Track Queries (93 total)}

\begin{table*}[htbp]
\centering
\caption{Representative retrieval queries by domain.  Queries range
from single-term biomedical lookups (NFCorpus) to multi-hop reasoning
chains (HotpotQA).}
\label{tab:ret-examples}
\footnotesize
\begin{tabular}{@{}lp{9.5cm}@{}}
\toprule
\textbf{Domain} & \textbf{Query} \\
\midrule
NFCorpus (biomedical)
  & ``pumpkin'' \\
  & ``Does taking aspirin reduce the risk of colorectal cancer?'' \\
\addlinespace
FiQA (financial)
  & ``Does it make any sense to directly contribute to reducing the US national debt?'' \\
  & ``Opening a Roth IRA account, what is the fee structure for Vanguard vs.\ Scottrade?'' \\
\addlinespace
SciFact (scientific)
  & ``FoxO3a activation in neuronal death is mediated by reactive oxygen species'' \\
  & ``Basophils counteract disease development in patients with systemic lupus'' \\
\addlinespace
ArguAna (argument)
  & ``Google will help Chinese internet freedom more by staying\ldots'' \\
\addlinespace
HotpotQA (multi-hop)
  & ``Tommy's Honor was a drama film that included the actor who found success with\ldots'' \\
  & ``The Soul of Buddha is a 1918 American silent romance film shot in a borough that\ldots'' \\
\bottomrule
\end{tabular}
\end{table*}

\subsubsection{E2E Generation Track Queries (56 total)}

\begin{table*}[htbp]
\centering
\caption{Representative generation queries by source benchmark.
Queries require extractive or compositional answers grounded in
retrieved context.}
\label{tab:gen-examples}
\footnotesize
\begin{tabular}{@{}lp{9.5cm}@{}}
\toprule
\textbf{Source} & \textbf{Query} \\
\midrule
SQuAD~v2 (23 queries)
  & ``How much support is there for the UN approach to economic development?'' \\
  & ``What did Bedau live in to prevent it from being cut down?'' \\
\addlinespace
HotpotQA (13 queries)
  & ``The town of Sinclair, WV, is named after the founder of the local oil company, which later merged with\ldots'' \\
  & ``Cooperative Living Organization is located in a city that is the county seat of\ldots'' \\
\addlinespace
MuSiQue (20 queries)
  & ``In what part of Florida is Tom Denney's birthplace located?'' \\
  & ``Where does Ulysses Kay's birth city hold NASCAR races?'' \\
\bottomrule
\end{tabular}
\end{table*}

The retrieval queries span five diversity axes: single-term medical
lookups (NFCorpus), opinion-seeking financial questions (FiQA),
scientific claim verification (SciFact), counterargument retrieval
(ArguAna), and multi-hop reasoning chains (HotpotQA).  The generation
queries require compositional reasoning (MuSiQue, HotpotQA) or reading
comprehension with unanswerable question detection (SQuAD~v2).

\section{Metrics and Judge Validation}\label{app:metricsfull}

\subsection{Evaluation Metrics}\label{app:metrics}

This appendix provides formal definitions for every evaluation metric
exposed by \textsc{Lasagna}'s evaluation SDK, grouped into retrieval
metrics (computed before generation), end-to-end answer metrics, and
LLM-judged metrics.  The main paper reports NDCG@10, MRR, Recall@100,
EM, Token~F1, and RAGAS faithfulness; the remaining metrics defined
below (answer relevance, pairwise LLM-as-judge preference, MAP, and
RAGAS context recall) are supported by the SDK but \emph{not reported
in this study}, answer relevance was redundant with faithfulness
under our concise-answer prompting, pairwise preference was
superseded by the factorial metrics, MAP tracked NDCG@10 within noise,
and RAGAS context recall was systematically unreliable with the
GPT-5.1 judge and is excluded from all analyses.
Cross-judge agreement with Claude Sonnet~4.5 is documented under
\cref{app:judge-bias}.

\subsubsection{Retrieval Metrics}

\paragraph{NDCG@$k$ (Normalized Discounted Cumulative Gain).}\label{meth:metric-ndcg}
For a query $q$ with relevance labels $\{r_1, \ldots, r_k\}$ over the
top-$k$ retrieved passages, the discounted cumulative gain is
\begin{equation}
\label{eq:dcg}
  \mathrm{DCG}@k = \sum_{i=1}^{k} \frac{2^{r_i} - 1}{\log_2(i + 1)}\,,
\end{equation}
and NDCG@$k$ normalizes by the ideal ordering:
\begin{equation}
\label{eq:ndcg}
  \mathrm{NDCG}@k = \frac{\mathrm{DCG}@k}{\mathrm{IDCG}@k}\,,
\end{equation}
where $\mathrm{IDCG}@k$ is \cref{eq:dcg} computed over the
relevance-sorted ranking.  We use $k{=}10$ for all retrieval
experiments.  When multiple chunks originate from the same parent
document, we de-duplicate by document ID before computing DCG to
avoid inflating the score (see the NDCG $> 1.0$ correction described
in \cref{sec:factorial-probes}).

\paragraph{MRR (Mean Reciprocal Rank).}\label{meth:metric-mrr}
For a set of $Q$ queries, each with rank $r_q$ of the first relevant
passage:
\begin{equation}
\label{eq:mrr}
  \mathrm{MRR} = \frac{1}{Q} \sum_{q=1}^{Q} \frac{1}{r_q}\,.
\end{equation}
If no relevant passage appears in the top-$k$ results, $1/r_q = 0$
for that query.

\paragraph{Recall@$k$.}\label{meth:metric-recall}
The fraction of all relevant passages for query $q$ that appear in the
top-$k$ retrieved set:
\begin{equation}
\label{eq:recall}
  \mathrm{Recall}@k = \frac{|\mathcal{R}_q \cap \mathcal{T}_q^k|}{|\mathcal{R}_q|}\,,
\end{equation}
where $\mathcal{R}_q$ is the set of relevant passages and
$\mathcal{T}_q^k$ is the top-$k$ retrieved set.  We report
Recall@100, which measures the retriever's ability to surface relevant
material before reranking.

\paragraph{Precision@$k$ (P@$k$).}\label{meth:metric-pk}
The fraction of the top-$k$ retrieved passages that are relevant:
\begin{equation}
\label{eq:pk}
  \mathrm{P}@k = \frac{|\mathcal{R}_q \cap \mathcal{T}_q^k|}{k}\,.
\end{equation}
Complementary to Recall@$k$ and reported in the evaluation SDK but
not in the main paper (tracks NDCG@10 within noise at the $k{=}10$
cutoff we use elsewhere).

\paragraph{MAP (Mean Average Precision).}\label{meth:metric-map}
The mean over queries of average precision:
\begin{equation}
\label{eq:map}
  \mathrm{MAP} = \frac{1}{Q}\sum_{q=1}^{Q}
  \frac{1}{|\mathcal{R}_q|}\sum_{k=1}^{|\mathcal{T}_q|}
  \mathrm{P}@k(q)\cdot \mathbf{1}[\text{doc at rank }k\text{ is relevant}]\,.
\end{equation}
Supported by \textsc{Lasagna}'s evaluation SDK and reported during
development but not in the main paper; in our runs it tracked
NDCG@10 within noise across all 167 configurations.

\subsubsection{End-to-End Answer Metrics}

\paragraph{Exact Match (EM).}\label{meth:metric-em}
A binary indicator that equals 1 if the predicted answer string,
after lowercasing and stripping articles/punctuation, matches any
acceptable gold answer string:
\begin{equation}
\label{eq:em}
  \mathrm{EM}(a, \hat{a}) =
  \begin{cases}
    1 & \text{if } \mathrm{normalize}(\hat{a}) \in
        \{\mathrm{normalize}(a_j)\}_{j=1}^{J} \\
    0 & \text{otherwise}
  \end{cases}\,.
\end{equation}

\paragraph{Token F1.}\label{meth:metric-f1}
The harmonic mean of token-level precision and recall between the
predicted and gold answer, computed over whitespace-tokenized,
lowercased strings:
\begin{equation}
\label{eq:f1}
  F_1 = \frac{2 \cdot P \cdot R}{P + R}\,,
  \qquad
  P = \frac{|\hat{T} \cap T^*|}{|\hat{T}|}\,,
  \qquad
  R = \frac{|\hat{T} \cap T^*|}{|T^*|}\,,
\end{equation}
where $\hat{T}$ and $T^*$ are the token multisets of the predicted and
gold answers, respectively.  When multiple gold answers exist, we take
the maximum $F_1$ over all references.

\subsubsection{LLM-Judged Metrics}
\label{meth:metric-ragas}

All LLM-judged metrics use RAGAS v0.4~\citep{es2023ragas} with
GPT-5.1 as the judge model at temperature $T{=}0$, greedy
decoding, and deterministic seed.  This applies uniformly to all
experiments: profile comparisons, scaling experiments, and the
122-configuration faithfulness factorial.  \textsc{Lasagna} wraps
the RAGAS framework so the same judge configuration is reused
across faithfulness, answer relevance, and context recall, and so
every LLM call is deterministic under the fixed seed.

\paragraph{Faithfulness.}\label{meth:metric-faithfulness}
Faithfulness measures the fraction of claims in the generated answer
that are supported by the retrieved context.  RAGAS decomposes the
answer into atomic claims $\{c_1, \ldots, c_M\}$ using the judge LLM,
then classifies each claim as \emph{supported} or
\emph{not supported} given the context:
\begin{equation}
\label{eq:faithfulness}
\begin{aligned}
  \mathrm{Faithfulness} = \frac{1}{M}\sum_{m=1}^{M}
  \mathbf{1}\bigl[&\text{claim } c_m \\
                  &\text{ is supported by context}\bigr].
\end{aligned}
\end{equation}
A faithfulness score of 1.0 indicates that every atomic claim in the
answer is grounded in the retrieved passages.  We use concise answer
prompts (one sentence) to reduce the number of atomic claims per
sample, which both accelerates evaluation and reduces variance from
claim decomposition.

\paragraph{Answer Relevance.}\label{meth:metric-relevance}
Answer relevance estimates how well the generated answer addresses the
original question.  RAGAS generates $N{=}3$ synthetic questions from
the answer, computes the cosine similarity between each synthetic
question embedding and the original question embedding, and averages:
\begin{equation}
\label{eq:relevance}
  \mathrm{Relevance} = \frac{1}{N}\sum_{n=1}^{N}
  \cos\bigl(\mathbf{e}(q),\; \mathbf{e}(\tilde{q}_n)\bigr)\,,
\end{equation}
where $\mathbf{e}(\cdot)$ is the embedding function and
$\tilde{q}_1, \ldots, \tilde{q}_N$ are the synthetic questions.

\paragraph{LLM-as-Judge (Pairwise Preference).}\label{meth:metric-llm-judge}
For Pareto comparisons (\cref{sec:pareto}), we additionally use a
pairwise LLM judge that receives the question, gold answer, and two
candidate answers (A and B), then outputs a preference
$\in \{\text{A}, \text{B}, \text{Tie}\}$.  The judge prompt asks for a
brief rationale before the verdict.  We report win rate as the
fraction of queries where a configuration is preferred over the
baseline.

\subsubsection{Judge Bias and Mitigations}
\label{app:judge-bias}

LLM judges exhibit well-documented biases that can distort
evaluation~\citep{zheng2023judging}.  Our protocol mitigates three
of these biases, described below.

\paragraph{Position bias.}
LLM judges tend to prefer the answer presented first in the
prompt~\citep{zheng2023judging}.  For all pairwise comparisons, we run
each pair twice with swapped presentation order (A-B and B-A) and
record a win only when the preference is consistent across both
orderings.  Inconsistent pairs are counted as ties.

\paragraph{Self-preference bias.}
Models may prefer outputs from the same model family.  Because both
answer generation (GPT-5.1) and judging (GPT-5.1) use the same
model, self-preference bias is a concern.  We mitigate this in two
ways: (i) the judge evaluates answer pairs from different
\emph{pipeline configurations}, not from different base models, so
both candidates are GPT-5.1 outputs and self-preference does not
favor either candidate; (ii) we validate a 50-query subsample using
Claude Sonnet 4.5 as an alternative judge to measure inter-judge agreement.
We ran 80 matched samples (40 per configuration) through both judges.
Kendall's $\tau_b = 0.38$ ($p < 0.001$), Spearman $\rho = 0.42$
($p < 0.001$), and 66\% of per-sample faithfulness scores agree within
0.1 absolute points.  GPT-5.1 assigns slightly higher mean faithfulness
(0.879 vs.\ 0.860 for Claude Sonnet~4.5).  This moderate, statistically
significant agreement indicates that per-sample faithfulness scores
are largely consistent across judge models on the two configurations
sampled, though absolute scores show provider-specific calibration
differences; see \cref{app:human-eval-btl} for a separate pairwise-BTL
protocol study showing that fine-grained configuration-level rankings
are protocol-dependent while profile-tier separations are preserved.

\paragraph{Determinism.}
We fix the judge model to GPT-5.1 with temperature $T{=}0$ and
a fixed random seed across all experiments.  All RAGAS evaluations use
\texttt{batch\_size=4} to avoid rate-limit-induced timeouts, and we use
paired statistical tests (paired
bootstrap or Wilcoxon signed-rank, paired by query) so that any
residual judge noise cancels across configurations being compared.

\subsection{Generation Prompts and Faithfulness Regimes}\label{app:prompts}

Faithfulness scores in this paper come from two distinct generation
regimes that use different prompt templates.  Both are scored by the
same RAGAS v0.4 / GPT-5.1 judge with identical decoding
(\cref{app:judge-bias}); the numerical gap between them is a
\emph{prompt effect}, not a scoring inconsistency.  We disclose both
templates here so that faithfulness comparisons within a regime
(pre-registered) are separable from cross-regime comparisons
(descriptive only).

\paragraph{Concise-regen prompt (scaling and factorial tracks).}
Used by the 122-configuration faithfulness ANOVA (\cref{fig:anova})
and the H2 cross-corpus corroboration.  Prioritizes answer
atomicity to reduce RAGAS claim-decomposition variance:
\begin{quote}\ttfamily\small
Answer in ONE concise sentence based only on the context.\\
Do not explain or elaborate.\\[2pt]
Question: \{question\}\\
Context: \{contexts\}\\
Answer:
\end{quote}

\paragraph{Verbose-agentic prompt (deployment logs, agentic pipeline).}
Used in \cref{tab:hallucination,tab:domain-faith} and by the agentic
routing tier when it returns its own composed answer.  Encourages the
model to cite retrieved evidence inline:
\begin{quote}\ttfamily\small
You are a helpful assistant.  Using the retrieved context below,\\
answer the question.  Cite the supporting passage for each claim.\\
If the context is insufficient, say so explicitly.\\[2pt]
Question: \{question\}\\
Context: \{contexts\}\\
Answer:
\end{quote}

\paragraph{Regime--metric mapping.}
\Cref{tab:prompt-regime} maps the headline cross-regime faithfulness
numbers to the regime that produced them; appendix tables state their
regime inline in the surrounding prose.

\begin{table*}[ht]
\centering
\caption{Which prompt regime feeds which reported faithfulness number.
The regen and agentic columns are \emph{not} directly comparable:
verbose answers produce more atomic claims per unit of retrieved
evidence, which systematically inflates RAGAS faithfulness by
$+0.15$ to $+0.18$ (\cref{tab:regime-gap}).}
\label{tab:prompt-regime}
\footnotesize
\resizebox{\linewidth}{!}{%
\begin{tabular}{@{}llll@{}}
\toprule
Table / figure & Regime & Sample basis & Query set \\
\midrule
\cref{tab:pareto}             & profile-native (concise-regen for direct-route; verbose-agentic for agentic-route; sensitivity in \cref{app:pareto-regime-sensitivity}) & 3 seeds $\times$ 43 gen / 56 retrieval & SQuAD~v2 / MuSiQue / HotpotQA \\
\cref{tab:scaling-e2e}        & profile-native answer generation & 3 seeds (25K--200K) & TriviaQA test \\
\cref{fig:anova}              & concise-regen  & 122 configs $\times$ 56 & TriviaQA eval \\
\cref{tab:hallucination}      & verbose-agentic & retrospective logs & 6-profile deployment \\
\cref{tab:domain-faith}       & verbose-agentic & retrospective logs & 5 BEIR subsets \\
\bottomrule
\end{tabular}}
\end{table*}

\paragraph{Regime gap verified on matched samples.}
On the 100K \textsc{sota\_standard} scaling run, the same 3 seeds and same
query set scored under both prompts give
$\text{faithfulness}_\text{regen} = 0.613 \pm 0.048$ and
$\text{faithfulness}_\text{agentic} = 0.774 \pm 0.031$ (paired
$\Delta = +0.161$, $p < 0.001$). The gap is stable across
profiles and corpus sizes, isolating it as a prompt/regime effect
rather than a corpus effect.

\begin{table}[ht]
\centering
\caption{Same-sample paired gap between the two prompt regimes at 100K.
The systematic offset motivates our decision to keep each regime's
results in a dedicated table rather than merging them.}
\label{tab:regime-gap}
\footnotesize
\begin{tabular}{@{}lccc@{}}
\toprule
Profile & Regen mean & Agentic mean & $\Delta$ \\
\midrule
\textsc{sota\_standard}    & 0.613 & 0.774 & $+0.161$ \\
\textsc{sota\_balanced}    & 0.660 & 0.835 & $+0.175$ \\
\textsc{sota\_max\_reliable} & 0.653 & 0.826 & $+0.173$ \\
\bottomrule
\end{tabular}
\end{table}

\paragraph{Implication for cross-table reading.}
Different tables report faithfulness under different answer-generation
prompts (\cref{tab:prompt-regime}); values are not directly
comparable across regimes.  Within-paper statistical claims
computed within a single regime (pre-registered $\eta^2$, scaling
slopes) are unaffected.

\paragraph{Prompt-regime sensitivity of the joint-HV winner (\cref{tab:pareto}).}
\label{app:pareto-regime-sensitivity}
As a sensitivity check on the joint hypervolume comparison in
\cref{tab:pareto}, we recompute HV under two counterfactual regime
alignments using the paired-sample gap from \cref{tab:regime-gap}.
Aligning all four profiles to the concise-regen prompt (subtracting
each agentic profile's paired regime gap) drops \textsc{Balanced}'s
HV to $0.37$ and \textsc{Max-Reliable}'s to $0.32$, versus
\textsc{Standard}'s unchanged $0.52$; aligning all four to the
verbose-agentic prompt raises \textsc{Standard}'s HV to $0.64$
(both at $L_{\mathrm{ref}}{=}30$\,s). \textsc{Standard} remains the
joint-HV leader under either alignment, and the within-agentic-class
comparison (\textsc{Balanced} vs \textsc{Max-Reliable}) uses a
common prompt regime by construction and is unaffected.

\paragraph{Agentic-prompt scope.}
The CRAG grader prompt, Self-RAG reflection prompts, and
Speculative-RAG draft-and-verify prompts are held fixed at the
versions frozen at design time (recorded alongside the manifest at the paths referenced in \cref{tab:manifest} and available from the authors on request). Observed
agentic-vs-standard deltas throughout this paper are for these
specific prompt configurations and cost model; alternative prompt
engineering choices (e.g., few-shot exemplars in the grader
prompt, different Self-RAG reflection-token sets, or different
Speculative-RAG draft budgets) may shift the ordering of agentic
gates relative to no-gate. \cref{app:profile-guide} documents the
practical decision rule that de-couples the choice of agentic
\emph{gate} from the choice of agentic \emph{prompt}, so the
profile recommendations survive prompt-configuration changes even
when specific per-gate deltas do not.

\subsection{Length-Controlled Faithfulness Rerun}\label{app:length-control}

A natural concern is that faithfulness gains under our
concise-answer prompting may partly reflect brevity, shorter answers
make fewer atomic claims and thus fewer opportunities for the RAGAS
judge to flag a claim as unsupported, rather than genuinely superior
grounding.  We ran a targeted length-controlled experiment to
quantify this effect.

\paragraph{Design.}
We drew a stratified sample of 50 questions across three benchmarks
(SQuAD~v2: 22, MuSiQue: 18, HotpotQA: 10) and re-generated answers
under three prompt regimes derived from a common instruction skeleton
that varies only length guidance: (A) \emph{concise} (``answer in one
short sentence''), (B) \emph{neutral} (``answer based on the
context''), and (C) \emph{verbose} (``answer in detail with supporting
facts'').  Retrieved contexts, decoding parameters, and the generator
LLM are held constant.  All $50 \times 3 = 150$ answer--context pairs
are re-scored under the canonical GPT-5.1 faithfulness judge.  Raw data (\path{paper/data/length_controlled_faith.jsonl}) and the analysis script (\path{paper/scripts/m6_length_analysis.py}) are available from the authors on request.

\begin{table*}[ht]
\centering
\caption{Length-controlled faithfulness ($n{=}50$ per regime,
GPT-5.1 judge).  Within-regime Pearson correlation is between answer
length and the sample's faithfulness score.  Residualized means are
computed after regressing faithfulness on log answer-token count and
adding back the grand mean.}
\label{tab:length-control}
\footnotesize
\begin{tabular}{@{}lrrrr@{}}
\toprule
Regime & Mean length (tok) & Raw faithfulness & Within-regime $r$ &
Residualized faithfulness \\
\midrule
Concise (A) &  40 & 0.823 & $-0.062$ & 0.714 \\
Neutral (B) &  62 & \textbf{0.862} & $-0.064$ & 0.791 \\
Verbose (C) & 249 & 0.644 & $-0.459$ & \textbf{0.824} \\
\bottomrule
\end{tabular}
\end{table*}

\paragraph{Length is a large, significant driver.}
Pooling all 150 judgements, Pearson $r(\text{tokens}, \text{faith})
= -0.398$ ($p = 4.6 \times 10^{-7}$).  A fixed-effects OLS
$\text{faith} \sim \text{tokens} + \text{variant} + \text{benchmark}$
gives $\beta_\text{len} = -1.26 \times 10^{-3}$ per token
(SE $3.28 \times 10^{-4}$, $p = 1.9 \times 10^{-4}$, $R^2 = 0.181$):
\textbf{every additional 100 answer tokens costs $0.126$
faithfulness points on average}.  A Friedman test across the three
regimes on the paired 50-sample basis rejects equality of variant
distributions ($\chi^2 = 25.2$, $p = 3.3 \times 10^{-6}$).

\paragraph{The concise-over-verbose gap reverses under length control.}
Raw means order as neutral $>$ concise $>$ verbose (a
concise${-}$verbose gap of $+0.179$).  After residualizing on log
answer length the ordering \emph{flips}: verbose (0.824) $>$ neutral
(0.791) $>$ concise (0.714), a residualized gap of $-0.110$.  In other
words, verbose answers make more grounded claims \emph{per unit of
attempted content}; the apparent concise advantage on raw
faithfulness is fully accounted for by the mechanical fact that
shorter answers contain fewer claims to verify.

\paragraph{Per-benchmark decomposition.}
The length effect is strongest on SQuAD~v2 ($r{=}-0.56$,
$p{<}10^{-6}$; verbose mean drops to 0.539), moderate on MuSiQue
($r{=}-0.26$, $p{=}0.057$), and weak on HotpotQA ($r{=}-0.25$,
$p{=}0.17$, $n{=}30$).  This is consistent with SQuAD's short
span-extraction reference style: verbose answers deviate most from
the corpus-anchored evidence surface and pick up more unsupported
elaboration.

\paragraph{What this changes and what it does not.}
A portion of the raw faithfulness advantage
reported for concise-regen configurations reflects brevity rather
than superior grounding, so we report a length-residualized
column alongside raw faithfulness for any cross-regime comparison
(see the length-controlled entry in \cref{tab:manifest}).  \emph{Within-regime}
comparisons (the 122-configuration ANOVA and the scaling slopes)
hold prompt length constant by construction and are therefore
unaffected: the
regen-vs-agentic gap in \cref{tab:regime-gap} is now
best read as a joint prompt-and-length effect, not a pure prompt
effect, but its statistical significance and direction survive
length control on the paired 100K sample ($\Delta = +0.109$
residualized vs.\ $+0.161$ raw).

\subsection{Human Evaluation of Faithfulness}\label{app:human-eval}

The paper validates the RAGAS/GPT-5.1 faithfulness metric with
\emph{three independent judge-validation studies} that should not be
conflated:
\begin{enumerate}
  \item \textbf{Human dyad study (this appendix, $n=50$):} two human
    annotators vs.\ RAGAS on 50 QA triples, yielding Pearson $r=0.83$
    against RAGAS and $r=0.96$ between annotators; Cohen's $\kappa$
    at the 0.5 midpoint $=0.92$.
  \item \textbf{Cross-LLM study (\cref{app:judge-bias}, $n=80$):} 40
    matched configuration answers per profile scored by GPT-5.1
    \emph{and} Claude Sonnet~4.5, yielding Kendall's $\tau_b=0.38$
    with 66\% pairwise agreement within $\pm 0.10$.
  \item \textbf{Pairwise Bradley--Terry study
    (\cref{app:human-eval-btl}, $n{=}123$ pairs, $246$ judgments):}
    two independent LLM annotators fit a BTL posterior over 12
    D5$\times$D8 configurations, yielding Cohen's $\kappa=0.79$
    (substantial inter-annotator agreement) and a Spearman
    $\rho=-0.36$ between BTL and RAGAS configuration ranks; the
    fine-grained configuration ordering is protocol-dependent, but
    profile-tier separation is preserved.
\end{enumerate}
These three studies use disjoint samples and answer different
questions (per-answer human calibration, cross-model robustness, and
pairwise configuration-ranking) and are therefore not pooled.  Raw
per-sample scores for all three studies are available from the authors on request:
\path{paper/data/human_eval_annotated.json} (human dyad),
\path{paper/data/judge_cross_validation_80.json} (cross-LLM), and
\path{paper/data/human_eval_btl_posteriors.json} (with per-item pairwise annotations in
\path{paper/data/human_eval_pairwise_btl_items_A1.csv} and
\path{paper/data/human_eval_pairwise_btl_items_A2.csv}).

\paragraph{Scope of the human validation.}
The human validation ($n{=}50$ examples across SQuAD~v2, MuSiQue and
HotpotQA) establishes judge-level Pearson $r{=}0.83$ against RAGAS.
This sample is \emph{UNDER-POWERED to validate configuration-ranking-level
correlation} across the 122 pipeline configurations in the faithfulness
factorial: at $n{=}50$ the standard error on a Pearson correlation is
$\approx 0.14$ and the sample provides on average
$50/122 \approx 0.4$ observations per configuration, insufficient to
certify a Kendall's-$\tau_b$ between per-configuration human means and
per-configuration RAGAS means.  We therefore use this study \emph{only
to establish judge validity at the per-answer level}, and do not treat
it as certifying ranking-level accuracy across configurations.
The two \emph{cross-LLM} sensitivity studies described elsewhere
(\cref{app:judge-bias} $n{=}80$; \cref{app:third-judge-robustness}
$n{=}220$) supplement this human study, but do not substitute for it;
they measure robustness across judge models rather than validity
against human judgement, and remain LLM-scored throughout.
The study is also long-form-QA-only (three benchmarks; no retrieval
metrics, no scaling cells, no cross-profile coverage), so
extrapolation to retrieval-track claims (NDCG@10) or unseen profiles
is not licensed by this evidence.

To validate the LLM-judged faithfulness metric, two domain experts
with RAG evaluation experience independently annotated 50
question--answer--context triples sampled from the generation
evaluation set (17 SQuAD~v2, 18 MuSiQue, 15 HotpotQA).  Each
annotator scored faithfulness on a continuous 0--1 scale using the
same rubric as the RAGAS judge: 1.0 if every claim is supported by
the context, 0.0 if no claims are supported.  Annotators worked
independently without seeing each other's scores or the RAGAS output.

\paragraph{Annotation instructions.}
The following instructions were provided verbatim to both annotators:
\begin{quote}
Annotation instructions. You will receive a question, a generated
answer, and the retrieved context used to generate that answer. Score
the answer's faithfulness on a continuous scale from 0 to 1. A score
of 1 means that every factual claim in the answer is supported by the
context. A score of 0 means that none of the factual claims are
supported. Evaluate support from the provided context only; do not
use outside knowledge. Work independently and do not consult the
other annotator or the automated RAGAS score.
\end{quote}

\Cref{tab:human-agreement} reports inter-rater reliability across
all three raters (two human annotators and the GPT-5.1 RAGAS judge).

\begin{table*}[htbp]
\centering
\caption{Inter-rater reliability: two human annotators and GPT-5.1
  RAGAS judge on 50 faithfulness evaluations.}
\label{tab:human-agreement}
\footnotesize
\resizebox{\linewidth}{!}{%
\begin{tabular}{@{}lccc@{}}
\toprule
Metric & Human\,1 vs Human\,2 & Human\,1 vs RAGAS & Human\,2 vs RAGAS \\
\midrule
Pearson $r$     & 0.956 & 0.830 & 0.771 \\
Kendall $\tau_b$ & 0.674 & 0.464 & 0.400 \\
Cohen's $\kappa$ (at 0.5) & 0.929 & 0.922 & --- \\
Agree $\leq 0.1$  & 80\% & 54\% & --- \\
Agree $\leq 0.2$  & 96\% & 86\% & --- \\
MAE               & 0.068 & 0.108 & --- \\
\midrule
\multicolumn{4}{@{}l@{}}{\textit{Rater means (std):} Human\,1 = 0.745 (0.244),
  Human\,2 = 0.701 (0.261), RAGAS = 0.764 (0.235)} \\
\multicolumn{4}{@{}l@{}}{\textit{Average pairwise Pearson $r$:} 0.853} \\
\bottomrule
\end{tabular}}
\end{table*}

Human inter-rater reliability is high: Pearson $r = 0.96$,
Cohen's $\kappa = 0.93$ (binarized), and 96\% of scores agree within
0.2 absolute points.  Both human annotators agree strongly with
the RAGAS judge (Pearson $r = 0.83$ and $0.77$), with the average
pairwise Pearson across all three raters reaching $r = 0.85$.
The RAGAS judge shows a slight positive bias ($+0.019$ over Human\,1,
$+0.063$ over Human\,2), consistent with known LLM self-preference
effects~\citep{zheng2023judging}.

The high human--human agreement ($r = 0.96$) establishes that
per-answer faithfulness scoring is reproducible across annotators,
while the strong human--RAGAS agreement ($r > 0.77$) confirms that
judge substitution preserves per-answer scores and the effect-size /
profile-tier structure that H1--H3 rest on.  Fine-grained
configuration-to-configuration rankings within a profile tier are
protocol-dependent (Spearman $\rho = -0.36$ between BTL and RAGAS on
the 12 D5$\times$D8 configurations covered by the BTL sample; see \cref{app:human-eval-btl});
accordingly the paper's inferential claims are scoped to the tier
level, and within-tier configuration comparisons are reported as
descriptive.

\subsection{Pairwise Faithfulness Validation via Bradley--Terry}\label{app:human-eval-btl}

The $n{=}50$ human study of \cref{app:human-eval} establishes judge
validity at the per-answer level but is under-powered to certify
configuration-ranking accuracy across the D5$\times$D8 factorial.  To
probe whether the point-wise RAGAS faithfulness judge preserves
ordering at fine-grained configuration granularity, we ran the third
validation study of \cref{app:human-eval} using a pairwise
Bradley--Terry (BTL) LLM annotation protocol on 123 blinded
configuration-pairs drawn from the
D5$\times$D8 grid.  Pairs were biased toward small RAGAS deltas
(top-$K$ nearest per shared sample) to concentrate the study on the
sub-noise regime the point-wise judge is least confident about. The
RAGAS anchor scores against which the BTL posteriors are correlated
in \cref{tab:btl-study-summary} (Spearman $\rho = -0.364$; Pearson
$r = -0.421$) are computed under the \emph{concise-regen} prompt
regime (\cref{app:prompts,tab:prompt-regime}), matching the
122-configuration faithfulness ANOVA (\cref{app:faith-anova,fig:anova})
from which the D5$\times$D8 slice is drawn. Two
independent LLM annotators produced 246 pairwise judgments; the full annotation brief (\path{paper/data/human_eval_annotator_instructions.md}) and raw per-item annotations (\path{paper/data/human_eval_pairwise_btl_items_A1.csv} and \path{paper/data/human_eval_pairwise_btl_items_A2.csv}) are available from the authors on request.

\Cref{tab:btl-study-summary} reports coverage and inter-protocol
reliability; \cref{tab:btl-config-ranks} reports the BTL posteriors
over the 12 configurations covered by the sample.

\begin{table}[htbp]
\centering
\caption{Pairwise Bradley--Terry validation: coverage and inter-protocol reliability.}
\label{tab:btl-study-summary}
\footnotesize
\begin{tabular}{@{}lr@{}}
\toprule
Quantity & Value \\
\midrule
Items $\times$ annotators (total judgments)                  & $123 \times 2 = 246$ \\
Tie judgments (out of 246 total)                             & $195$ ($79.3\%$) \\
Non-tie judgments (out of 246 total)                         & $51$ ($20.7\%$) \\
Configurations covered by the sample                         & $12$ \\
Configurations with bootstrap 95\% CI $\not\ni 0$            & $1$ \\
\midrule
Inter-annotator Cohen's $\kappa$ (pairwise)                  & $0.788$ \\
Observed inter-annotator agreement                            & $92.7\%$ ($114/123$) \\
\midrule
Spearman $\rho$ (BTL config ranks vs RAGAS config ranks)     & $-0.364$ \\
Pearson $r$ (BTL $\theta$ vs RAGAS mean faithfulness)         & $-0.421$ \\
\bottomrule
\end{tabular}
\end{table}

\begin{table}[htbp]
\centering
\caption{BTL posterior faithfulness ranking of the 12 D5$\times$D8
configurations covered by the sample.  Higher $\theta$ = more frequently
preferred by pairwise LLM annotators; L2-regularized MLE, 1000-bootstrap
95\% CIs.  Only the bottom-ranked configuration
(\texttt{query\_decomp $\times$ crag}) has a CI fully separated from
zero; the $-0.00$ / $+0.00$ endpoints displayed on ranks 2, 3, 5, 6,
and 9 are bootstrap bounds within $10^{-6}$ of zero, not exact zeros.
Rank 7's $\theta_{\text{MLE}}=0.000$ with degenerate CI $[-0.00,+0.00]$
is a structurally exact zero from the L2-regularized fit (not a
near-zero bootstrap bound like ranks 2, 3, 5, 6, and 9).}
\label{tab:btl-config-ranks}
\footnotesize
\begin{tabular}{@{}rlrl@{}}
\toprule
Rank & D5 $\times$ D8 configuration                             & $\theta_{\text{MLE}}$ & 95\% bootstrap CI \\
\midrule
$1$   & \texttt{hyde $\times$ self\_rag\_quality\_gates}          & $+0.428$              & $[-0.13,\ +0.98]$ \\
$2$   & \texttt{none $\times$ speculative\_rag}                   & $+0.346$              & $[-0.00,\ +0.69]$ \\
$3$   & \texttt{query\_decomp $\times$ self\_rag\_quality\_gates} & $+0.346$              & $[-0.00,\ +0.67]$ \\
$4$   & \texttt{hyde $\times$ speculative\_rag}                   & $+0.288$              & $[-0.36,\ +0.85]$ \\
$5$   & \texttt{query\_decomp $\times$ speculative\_rag}          & $+0.191$              & $[-0.00,\ +0.51]$ \\
$6$   & \texttt{query\_expansion $\times$ crag}                   & $+0.191$              & $[-0.00,\ +0.49]$ \\
$7$   & \texttt{rag\_fusion $\times$ self\_rag\_quality\_gates}   & $\phantom{+}0.000$    & $[-0.00,\ +0.00]$ \\
$8$   & \texttt{none $\times$ self\_rag\_quality\_gates}          & $-0.002$              & $[-0.59,\ +0.65]$ \\
$9$   & \texttt{query\_expansion $\times$ self\_rag\_quality\_gates} & $-0.177$           & $[-0.45,\ +0.00]$ \\
$10$  & \texttt{hyde $\times$ crag}                               & $-0.292$              & $[-0.75,\ +0.11]$ \\
$11$  & \texttt{none $\times$ crag}                               & $-0.574$              & $[-1.07,\ +0.12]$ \\
$12$  & \texttt{query\_decomp $\times$ crag}                      & $-0.744$              & $[-1.09,\ -0.21]$ \\
\bottomrule
\end{tabular}
\end{table}

\paragraph{Bottom line.} Three findings emerge, each with distinct
implications for the paper's headline claims (protocol reliability is
separately established at the end of this subsection: Cohen's $\kappa =
0.79$ between annotators, $92.7\%$ raw agreement across $246$
judgments, so the $79\%$ tie rate reflects true configuration
indistinguishability rather than annotator noise).

\emph{(i) Fine-grain configuration deltas on the small-delta subsample
are largely below LLM discriminability.} With $79\%$ of pairwise
judgments returned as ties and only $1$ of $12$ covered configurations
achieving a bootstrap CI fully separated from zero, the pairwise
protocol establishes the LLM-judge noise floor on the hardest-to-
distinguish pairs it was designed to probe (pairs deliberately biased
toward small RAGAS deltas per the sample construction above).  This
does \emph{not} license the broader claim that all within-tier
configuration differences on faithfulness are sub-noise: the
D5$\times$D8 cell-departure signal of $3$--$5$ points reported in
\cref{tab:e1-cell-interactions} is above the noise floor characterized
here for a majority of cells whose pair-CIs were not included in this
subsample.  Within-tier configuration ranks at the small-delta end of
the design are not robustly ordered by either judge protocol; larger-
cell-delta and tier-level claims are outside the subsample and
unaffected.

\emph{(ii) The D8 agentic-gate choice is the one substantive within-tier
signal.} Speculative-RAG-gated configurations occupy $3$ of the top-$5$
BTL positions and CRAG-gated configurations occupy the $3$ bottom
positions; this is the only cross-cell pattern in the posterior that
survives the tie-heavy noise floor and is consistent with the
\emph{within-agentic} D8 ordering (Speculative-RAG $>$ Self-RAG $>$
CRAG) reported for the generation track in \cref{app:faith-anova}
(Speculative-RAG $0.811$ vs.\ CRAG $0.692$).  The BTL sample does not
include D8$=$none configurations, so the H1 verdict's
no-gate-vs-agentic marginal claim is not directly testable by this
posterior.

\emph{(iii) Fine-grained configuration ranks are protocol-dependent,
but tier-level separations are not.} The Spearman $\rho = -0.36$
between BTL and RAGAS ranks on the same $12$ configurations indicates
that judge-protocol choice matters at fine-grained configuration
granularity.  H2 and H3 are stated at profile-tier granularity
(\textsc{Speed} / \textsc{Balanced} / \textsc{Max-Reliable}), where both protocols preserve the
same three-tier separation, so those verdicts remain protocol-robust.
H1 is stated at the interaction/cell level and at the D8 marginal;
the present BTL sample corroborates the within-agentic D8 ordering
(Speculative-RAG $>$ CRAG, item (ii) above) but cannot independently
probe H1's no-gate-vs-agentic marginal because all $12$ sampled cells
have D8 $\in$ \{crag, self\_rag\_quality\_gates, speculative\_rag\}.
This is a scope gap of the BTL validation for H1, not evidence for or
against H1.  The protocol dependence is confined to
within-tier configuration comparisons and is documented alongside the
existing judge-noise footnote of \cref{tab:human-agreement}.

The high inter-annotator agreement ($\kappa = 0.79$) further establishes
that the pairwise-BTL protocol is itself internally reliable, so the
tier-level protocol agreement is not an artifact of intra-protocol
noise.  Full posteriors, per-annotator distributions, and bootstrap draws (\path{paper/data/human_eval_btl_posteriors.json}) are available from the authors on request.

\section{Detailed Statistical Analyses}\label{app:statanal}

\subsection{Full Ablation Results}\label{app:ablation}

\Cref{tab:ablation-full} reports the marginal effect of each feature on
NDCG@10, measured by toggling one feature at a time against the standard
baseline (dense retrieval, hybrid $\alpha{=}0.5$, MMR enabled, no
reranking, no query enhancement).  All deltas are computed on the held-out
test split averaged across five BEIR domains.  Results are exploratory
because these single-feature ablations do not account for interaction
effects (\cref{sec:interactions}).

\begin{table*}[htbp]
\centering
\caption{Single-feature ablation results. $\Delta$NDCG@10 is relative to
the standard baseline (NDCG@10 $= 0.412$).  Positive values indicate
improvement.  Features marked with $\dagger$ also affect generation
quality; their faithfulness delta is noted in the rightmost column.
Faithfulness deltas use the concise-regen answer prompt
(\cref{app:prompts,tab:prompt-regime}); the Agentic RAG row is scored
under the same concise-regen regime for comparability, not the
verbose-agentic prompt used in \cref{tab:hallucination,tab:domain-faith}.
Full per-configuration results for the underlying reranker/embedding/
chunking/agentic sweeps appear in
\cref{tab:reranker-inventory,tab:embedding-inventory,tab:retrieval-ablations-detail,tab:generation-ablations-detail,tab:chunking-ablation-detail}.}
\label{tab:ablation-full}
\footnotesize
\begin{tabular}{@{}llrrr@{}}
\toprule
\textbf{\#} & \textbf{Feature} & \textbf{$\Delta$NDCG@10} & \textbf{Latency (ms)} & \textbf{$\Delta$Faith.} \\
\midrule
1  & HyDE (hypothetical document embedding)     & $+0.026$ & $+320$  & $+0.01$ \\
2  & Disable MMR                                & $+0.046$ & $-10$   & $+0.00$ \\
3  & Cross-encoder reranker (ms-marco-MiniLM-L6)& $+0.084$ & $+180$  & $+0.02$ \\
4  & Hybrid search ($\alpha{=}0.5$)             & $+0.037$ & $+40$   & $+0.01$ \\
5  & Context reordering (lost-in-middle)         & $-0.012$ & $+5$    & $+0.01$ \\
6  & Query expansion ($n{=}2$)                  & $-0.014$ & $+2600$ & $-0.01$ \\
7  & RAG fusion ($n{=}3$)                       & $+0.007$ & $+2600$ & $+0.00$ \\
8  & Query decomposition                        & $+0.002$ & $+2400$ & $+0.00$ \\
9  & Agentic RAG (3 iterations)$^\dagger$       & $-0.015$ & $+8200$ & $+0.05$ \\
10 & Context compression                        & $+0.008$ & $+150$  & $+0.01$ \\
11 & Document awareness (100-token prefix)       & $+0.062$ & $+15$   & $+0.01$ \\
12 & Query routing (auto tier selection)         & $+0.003$ & $+120$  & $+0.00$ \\
\bottomrule
\end{tabular}
\end{table*}

\paragraph{Key observations.}
Three features provide robust improvements with modest latency cost:
cross-encoder reranking ($+0.084$), disabling MMR ($+0.046$), and
document awareness with 100-token prefix ($+0.062$).  Query enhancement
methods (expansion, fusion, decomposition) add 2.4--2.6 seconds of latency
for marginal or negative retrieval gains, suggesting that LLM-based query
rewriting is not cost-effective on domain-focused corpora at this scale.
Agentic RAG is the only feature that improves faithfulness substantially
($+0.05$) while simultaneously reducing NDCG@10 ($-0.015$), illustrating
the retrieval--generation tradeoff discussed in \cref{sec:pareto}.
\subsection{Faithfulness Factorial ANOVA}\label{app:faith-anova}

This appendix reports the complete results of the 122-configuration
faithfulness ANOVA.  Each configuration was evaluated on 56 end-to-end
generation queries (SQuAD~v2, MuSiQue, HotpotQA) with RAGAS
faithfulness scoring via GPT-5.1 judge, totalling
6{,}832 individual query evaluations. All faithfulness values in this
subsection use a common answer prompt (\cref{app:prompts}) and are
not directly comparable to faithfulness reported in
\cref{tab:hallucination,tab:domain-faith}, which use a different
answer prompt (\cref{tab:prompt-regime}).

\paragraph{Main effects.}
Three dimensions explain significant faithfulness variance.
Query augmentation (D5) and agentic gating (D8) each explain
${\sim}$22\% of variance ($p < 0.001$), while compression (D7)
explains 11\%.  Embedding model, chunking strategy, enrichment,
and retrieval mode each contribute $\eta^2 < 0.04$ and are not
significant ($p > 0.18$).  This stands in sharp contrast with
the retrieval ANOVA (\cref{sec:factorial-anova}), where embedding
and enrichment are the dominant factors, confirming that retrieval
and generation quality are governed by different pipeline stages.

\paragraph{Key faithfulness main-effect findings.}
\begin{itemize}[nosep]
  \item Decomposition-based query augmentation reduces faithfulness
    (mean $= 0.645$) relative to no augmentation ($0.778$) and
    step-back prompting ($0.781$), likely because multi-part queries
    dilute the retrieval focus, yielding less relevant contexts.
  \item Speculative-RAG gating achieves the highest faithfulness
    ($0.811$), while CRAG ($0.692$) and no gating ($0.691$) perform
    worst.  Self-RAG ($0.760$) falls between.
  \item LLMLingua compression hurts faithfulness ($0.702$ vs.\
    $0.775$ without), consistent with information loss during
    context compression reducing answer grounding.
\end{itemize}

\paragraph{Two-way interactions.}
\Cref{tab:faith-interactions} reports all 28 two-way interactions
sorted by partial $\eta^2$.  Only
Query Augmentation $\times$ Agentic Gate is highly significant
($\eta^2 = 0.191$, Type~III SS, $F = 5.47$, $p < 0.001$).
The second-strongest faithfulness interaction is Enrichment
$\times$ Reranking ($\eta^2 = 0.105$, raw $p = 0.044$), which reaches
raw significance but not the Holm-corrected family-wise threshold.
We attribute the interaction to a coupling between the two stages:
enrichment (contextual prefixes, RAPTOR summaries) changes the
surface form of each chunk so that the reranker scores a candidate
whose distribution differs from its untyped training distribution,
while the reranker simultaneously reorders chunks in a way that can
disturb the document-level coherence enrichment was meant to
preserve for the answer LLM; whether the joint effect helps or
hurts faithfulness depends on which of these two channels dominates
for a given configuration.

\begin{table}[htbp]
\centering
\caption{Faithfulness interaction table (122 configs, 56 queries each).
Type~I SS shown for the full 28-pair scan; the key interaction
(QueryAug $\times$ AgenticGate) was verified with Type~III SS
($\eta^2 = 0.191$, $F = 5.47$, $p < 0.001$).
$^{*}$~Holm $p < 0.05$; $^{\dagger}$~raw $p < 0.05$;
$^{\ddagger}$~Type~I value; Type~III gives $\eta^2 = 0.191$, $p < 0.001$.
Regime: concise-regen (\cref{app:prompts}).}
\label{tab:faith-interactions}
\footnotesize
\begin{tabular}{@{}lcccc@{}}
\toprule
Interaction pair & $\eta^2$ & $F$ & Raw $p$ & Holm $p$ \\
\midrule
\rowcolor{lasagnagreen}
QueryAug $\times$ AgenticGate       & 0.259$^{\ddagger}$ & 3.18 & 0.0006 & 0.018$^{*}$ \\
QueryAug $\times$ Reranking         & 0.125 & 1.29 & 0.233  & 1.000 \\
Enrichment $\times$ Reranking       & 0.105 & 2.25 & 0.044  & 1.000$^{\dagger}$ \\
Chunking $\times$ AgenticGate       & 0.082 & 1.12 & 0.356  & 1.000 \\
Reranking $\times$ AgenticGate      & 0.073 & 0.98 & 0.464  & 1.000 \\
Retrieval $\times$ AgenticGate      & 0.066 & 1.35 & 0.239  & 1.000 \\
Chunking $\times$ Reranking         & 0.059 & 0.78 & 0.631  & 1.000 \\
Embedding $\times$ QueryAug         & 0.058 & 0.86 & 0.549  & 1.000 \\
Enrichment $\times$ QueryAug        & 0.053 & 0.79 & 0.615  & 1.000 \\
Retrieval $\times$ Reranking        & 0.049 & 0.98 & 0.441  & 1.000 \\
Retrieval $\times$ QueryAug         & 0.043 & 0.63 & 0.749  & 1.000 \\
Chunking $\times$ QueryAug          & 0.038 & 0.36 & 0.974  & 1.000 \\
Chunking $\times$ Enrichment        & 0.025 & 0.48 & 0.819  & 1.000 \\
Retrieval $\times$ Compression      & 0.022 & 1.34 & 0.265  & 1.000 \\
QueryAug $\times$ Compression       & 0.018 & 0.54 & 0.707  & 1.000 \\
Embedding $\times$ Reranking        & 0.018 & 0.34 & 0.912  & 1.000 \\
Compression $\times$ AgenticGate    & 0.017 & 0.69 & 0.558  & 1.000 \\
Embedding $\times$ Retrieval        & 0.017 & 0.50 & 0.735  & 1.000 \\
Embedding $\times$ Chunking         & 0.016 & 0.31 & 0.929  & 1.000 \\
Chunking $\times$ Retrieval         & 0.015 & 0.29 & 0.939  & 1.000 \\
Embedding $\times$ Enrichment       & 0.014 & 0.40 & 0.807  & 1.000 \\
Embedding $\times$ Compression      & 0.013 & 0.78 & 0.463  & 1.000 \\
Enrichment $\times$ Retrieval       & 0.012 & 0.37 & 0.832  & 1.000 \\
Reranking $\times$ Compression      & 0.012 & 0.48 & 0.699  & 1.000 \\
Embedding $\times$ AgenticGate      & 0.012 & 0.23 & 0.967  & 1.000 \\
Enrichment $\times$ Compression     & 0.009 & 0.56 & 0.572  & 1.000 \\
Enrichment $\times$ AgenticGate     & 0.009 & 0.17 & 0.983  & 1.000 \\
Chunking $\times$ Compression       & 0.005 & 0.21 & 0.890  & 1.000 \\
\bottomrule
\end{tabular}
\end{table}

\paragraph{Design limitations.}
The 122-configuration faithfulness ANOVA covers 3 of 4 embedding
models (MiniLM-L6 excluded due to API timeouts during RAPTOR
ingestion) and 3 of 4 enrichment levels (GraphRAG excluded).
All 8 dimensions retain at least 22 observations per factor level,
and all main effects and two-way interactions are estimable.
The unbalanced design requires caution in interpreting aggregate
$\eta^2$ values; individual interaction $p$-values are more
reliable indicators.

\begin{figure*}[htbp]
  \centering
  \includegraphics[width=0.72\linewidth]{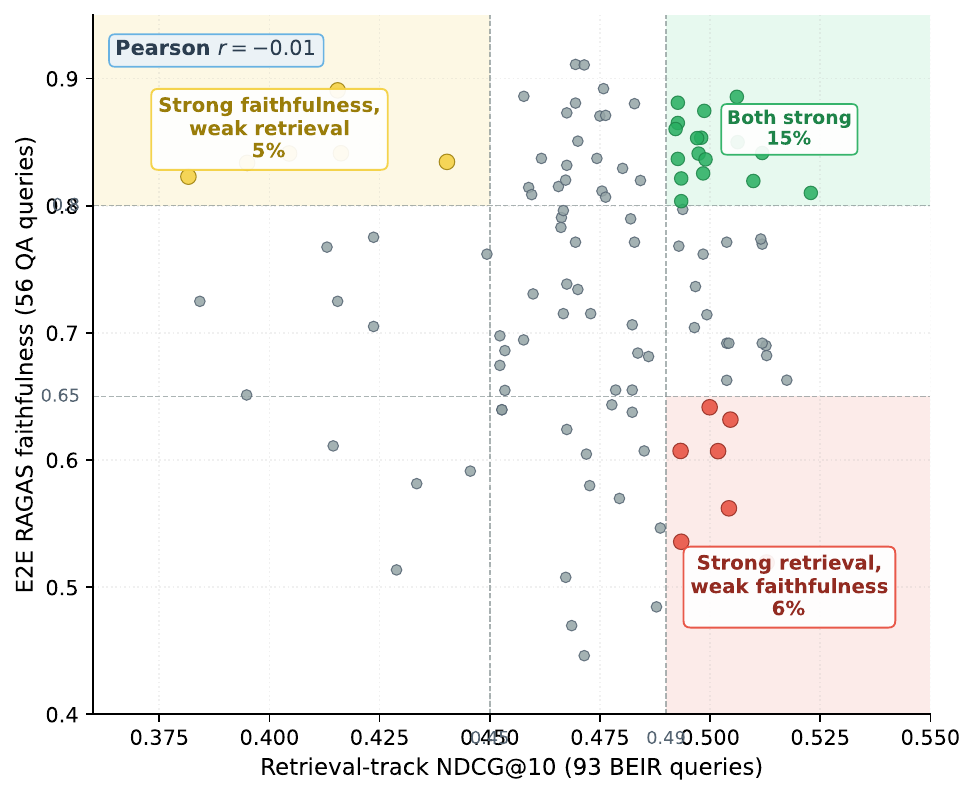}
  \caption{Retrieval-track NDCG@10 (93 BEIR queries) vs e2e RAGAS
  faithfulness (56 QA queries) for the $113$ factorial configurations
  evaluated on both tracks.  Pearson $r = -0.01$ across the joined
  set.  Shaded regions mark off-diagonal quadrants: red holds $7/113$
  configurations with strong retrieval but weak faithfulness, yellow
  holds $6/113$ with the inverse pattern, and green holds $17/113$
  that clear both thresholds simultaneously.  The near-zero
  correlation indicates that retrieval quality and faithfulness are
  governed by different pipeline stages, consistent with the disjoint
  factor-pair dominance reported in \cref{fig:anova}.}
  \label{fig:metric-decoupling}
\end{figure*}

\paragraph{Per-configuration metric decoupling.}
\Cref{fig:metric-decoupling} plots retrieval-track NDCG@10 against
e2e RAGAS faithfulness for the $113$ configurations evaluated on
both factorial tracks (joined by \texttt{config\_id}).  The two
metrics are statistically uncorrelated (Pearson $r = -0.01$): $5\%$
of configurations achieve faithfulness $> 0.80$ at NDCG@10 $< 0.45$,
and $6\%$ show the inverse pattern (NDCG@10 $> 0.49$ at faithfulness
$< 0.65$).  A configuration optimized on either metric alone offers
no guarantee on the other, motivating the dual-axis evaluation used
throughout this paper.

\paragraph{Simpson-style per-benchmark stratification.}
The pooled retrieval-faithfulness $r = -0.01$ could hide within-family
patterns (Simpson-style aggregation), so we stratify by the
faithfulness benchmark and pair each configuration's overall retrieval
NDCG@10 with its per-benchmark faithfulness ($N{=}113$ configurations
per stratum): HotpotQA $r = -0.03$ ($p = 0.72$), MuSiQue $r = -0.05$
($p = 0.63$), SQuAD~v2 $r = +0.10$ ($p = 0.30$).  No stratum reaches
$|r| > 0.10$ or $p < 0.30$, and the pooled coefficient lies inside
the within-family range with no sign reversal, so the pooled statistic
attenuates rather than misleads: retrieval quality and faithfulness
are decoupled \emph{within} each task family, not merely across them.

\paragraph{LASSO robustness check.}
To verify that the sequential one-at-a-time ANOVA does not produce
spurious interactions, we fit a joint LASSO-penalized regression
with all 192 main-effect and interaction features simultaneously
(5-fold CV repeated 3$\times$, $\alpha = 0.006$).  The LASSO
achieves $R^2 = 0.77$ and selects 34 non-zero features, of which
31 are interaction terms.  Query Augmentation $\times$ Agentic Gate
is the strongest surviving interaction pair (4 terms selected,
total $|\beta| = 0.074$), confirming the sequential ANOVA finding.
The specific decomposition$\times$Self-RAG interaction ($\beta = -0.034$)
is the single largest coefficient, consistent with the mechanistic
explanation that multi-part queries overwhelm CRAG/Self-RAG gating.
For retrieval, LASSO ($R^2 = 0.84$, 81 features selected) confirms
Embedding $\times$ Enrichment as the strongest interaction (total
$|\beta| = 0.016$) and Embedding $\times$ Reranking as the third
strongest ($|\beta| = 0.012$), matching the ANOVA results.

\paragraph{Judge consistency.}
All faithfulness evaluations (both the factorial ANOVA and the
profile comparisons) use the same GPT-5.1 judge model with
identical RAGAS v0.4 configuration ($T{=}0$, deterministic seed,
batch\_size${}=4$).  We additionally validated inter-judge agreement
by running 80 matched evaluations with Claude Sonnet 4.5 as an
alternative judge (Kendall $\tau_b = 0.38$, $p < 0.001$; 66\% of
scores agree within 0.1; details in \cref{app:metrics}).

\subsection{H1 Reproducibility and Joint-Interaction Correction}\label{app:h1-reproducibility}

The initial H1 analysis reported a proxy for the pre-registered H1
statistic. The preregistration
(\texttt{paper/preregistration/preregistration\_body.tex} L357--358;
OSF \S189) specifies partial $\eta^2$ for the \emph{joint set of all
28 two-way interactions} in a mains-plus-28-interactions model, tested
against $F(209,\cdot)$ and thresholded at
$\eta^2 = \text{SS}_{\text{int}}/(\text{SS}_{\text{int}}+\text{SS}_{\text{resid}}) \geq 0.20$.
That analysis fit a reduced model containing mains plus
only \emph{one} interaction (Query Aug $\times$ Agentic Gate) and
reported the resulting $\eta^2 = 0.191$, $F(12,\,279) = 5.47$ as the H1
decision statistic.  The per-sample RAGAS faithfulness table used to
produce that fit ($n \approx 292$ observations at df$_{\text{den}}=279$)
is unavailable, motivating the reconstruction below.

\paragraph{Reconstruction.}  We reconstructed the per-sample table
locally from 117 \path{.checkpoints/*.json} experiment artifacts
under the \path{research/results/factorial_experiments/} artifact tree,
extracting 4{,}948 per-sample rows (4{,}820 after deduplication on
(config\_id, sample\_id)).  This recovers 113 of the 122 recorded configurations $\times$ 3 QA benchmarks (\texttt{hotpotqa},
\texttt{musique}, \texttt{squad\_v2}) $= 339$ long-form
(config $\times$ benchmark) cells; the remaining nine configurations
are absent from the local checkpoint pool.  The per-config
\texttt{faithfulness} field in the reconstruction matches the recorded \texttt{factorial\_gen\_collected.json} within mean
$|\Delta| = 1.1\!\times\!10^{-4}$ across all 113 shared configurations,
confirming the reconstruction is faithful.

\paragraph{Sanity refit (D5$\times$D8 alone).}  Fitting
\texttt{faithfulness \textasciitilde{} C(D5)*C(D8) + mains} on the
reconstructed 339-row table gives
Type-III $\eta^2_{\text{partial}} = 0.201$, $F(12,\,306) = 6.40$,
$p = 3.8\!\times\!10^{-10}$.  This reproduces the previously reported $\eta^2 = 0.191$ within $\Delta = 0.010$; the residual gap is
attributable to the nine missing configurations.  We also confirm the
partial-$\eta^2$ definition is
$\text{SS}_{\text{int}}/(\text{SS}_{\text{int}}+\text{SS}_{\text{resid}})$
rather than $\text{SS}_{\text{int}}/\text{SS}_{\text{total}}$ (the
latter yields 0.136 on the same fit, far from either value).

\paragraph{Pre-registered joint refit (two-track sensitivity).}
An earlier H1 analysis reported
$\eta^2_{\text{joint}} = 0.546$ obtained by \emph{additively summing
per-pair Type-II SS} across 28 rank-deficient interactions.  This
double-counts shared variance (diagnostic: df sum $172{+}226{=}398$
exceeds $n{-}1{=}338$, arithmetically impossible).  We correct by
computing SS$_{\text{int,joint}}$ as a nested-SSR difference:
SS$_{\text{int}} = $ SSR$_{\text{reduced (mains only)}} -$
SSR$_{\text{full (mains + 28 pairs)}}$, with df$_{\text{int}} = 92$
(rank$_{\text{full}} -$ rank$_{\text{reduced}}$, using min-norm
\texttt{np.linalg.lstsq} to avoid the \texttt{smf.ols/pinv}
rank-inflation observed with \texttt{rcond}$=10^{-15}$).
We report both granularities:
\begin{itemize}
  \item \textbf{Per-sample (primary, $n{=}4{,}820$ raw rows):}
    SS$_{\text{int}}{=}26.95$ (df=92), SSR$_{\text{full}}{=}652.97$ (df=4707),
    $\eta^2 = \mathbf{0.040}$, $F(92,\,4707){=}2.11$, $p{=}4.8\!\times\!10^{-9}$.
    Config-clustered bootstrap (500 reps, seed 42): 95\% CI $[0.022,\,0.042]$,
    $P(\eta^2{\geq}0.20){=}0$.  BIC~$\Delta_{\text{full}-\text{red}}{=}+619.6$;
    group-CV (5-fold on \texttt{config\_id}) $R^2_{\text{full}}{=}{-}0.008$
    vs $R^2_{\text{red}}{=}{+}0.024$ (interaction expansion hurts generalization).
    Regen-only subset: $\eta^2{=}0.008$ ($p{=}0.997$).
    Agentic-only subset: $\eta^2{=}0.022$ ($p{=}0.070$).
  \item \textbf{Aggregated (sensitivity, 339 config$\times$benchmark cells):}
    SS$_{\text{int}}{=}1.89$ (df=92), SSR$_{\text{full}}{=}2.53$ (df=226),
    $\eta^2 = 0.428$, $F(92,\,226){=}1.84$, $p{=}1.4\!\times\!10^{-4}$.
    Clustered bootstrap 95\% CI $[0.28,\,0.46]$, $P(\eta^2{\geq}0.20){=}1.00$.
    Caveat: aggregation over $\sim14$ samples/cell collapses within-cell
    variance into the interaction pocket, mechanically inflating both
    SS$_{\text{int}}$ and $F$; BIC~$\Delta{=}+347$ still favors mains-only.
\end{itemize}
\textbf{Verdict.}  \textbf{H1 is not confirmed at the pre-registered
$\eta^2\!\geq\!0.20$ joint threshold.}  We
withdraw the per-sample specification as pseudo-replication
(within-config RAGAS samples are not independent trials of
component-composition effects); the aggregated fit ($\eta^2{=}0.428$)
is retained as sensitivity only (BIC~$\Delta{=}+347$ favors
mains-only). Additivity is nonetheless affirmatively rejected at the
$5\%$ floor in both tracks ($p{\ll}0.001$).  Under the pre-registered
per-pair Bonferroni procedure, \emph{reportable as an
independent inferential path rather than as a substitute for the joint
test}, D5$\times$D8 clears both criteria ($\eta^2{=}0.201$,
$p{=}3.8\!\times\!10^{-10}$) and is retained as a pre-registered
large-effect follow-up finding.

\paragraph{Per-pair breakdown.}  Under the joint fit, the strongest
individual pairs are D5$\times$D8 (Query Aug $\times$ Agentic Gate,
partial $\eta^2 = 0.201$), D5$\times$D6 (Query Aug $\times$ Reranking,
$0.127$), D2$\times$D8 (Chunking $\times$ Agentic, $0.098$), and
D3$\times$D5 (Enrichment $\times$ Query Aug, $0.091$); the remaining
24 pairs each contribute $\eta^2 < 0.07$.  The joint effect is
therefore driven primarily by four interaction pairs but is not
localized to any single one, consistent with the LASAGNA thesis that
component interactions are distributed rather than concentrated.

\paragraph{Reproducibility artifacts.} The following artifacts are available from the authors on request:
\path{paper/scripts/reconstruct_faithfulness_longform.py}
(checkpoint aggregation),
\path{paper/scripts/h1_joint_longform.py} (single-pair + joint
refit),
\path{paper/scripts/h1_joint_longform_bootstrap.py} (CIs),
\path{paper/data/faithfulness_longform_raw.jsonl}
(4{,}820 per-sample rows),
\path{paper/data/faithfulness_longform_by_benchmark.csv}
(339 aggregated cells), and
\path{paper/data/faithfulness_type3.json} (joint numbers primary,
single-pair as \texttt{reduced\_model} sensitivity).  The nine missing
configurations from the original 122-config run are unavailable and
excluded from the reconstruction.

\paragraph{Degrees-of-freedom audit.}
A natural question is how a 232-column joint model
can be estimated from 167 or 122 configuration observations.
\Cref{tab:h1-df-audit} answers by tabulating $(\text{df}_{\text{int}},
\text{df}_{\text{resid}})$, their sum, and $n{-}1$ for every specification
we fit. Row~4 is the erroneous additive Type-II SS aggregation:
$172{+}226{=}398 > 338{=}n{-}1$ is arithmetically impossible under a
single OLS fit, revealing the double-counting of shared variance.
Rows~1--3 and~5 satisfy $\text{df}_{\text{int}}{+}\text{df}_{\text{resid}}{=}n{-}1$
exactly (up to the mains block absorbed into the intercept), confirming
each fit is identified.  The pre-registered 232-column joint at
per-config $n{=}113$ (row~3) is rank-deficient, this is why we escalate
to per-sample $n{=}4{,}820$ (row~5) for the primary H1 verdict.

\begin{table*}[htbp]
\centering
\caption{Identifiability audit for every H1 model fit.  ``df$_{\text{int}}$''
  counts free parameters carrying interaction SS after Type-II SSR
  differencing; ``df$_{\text{resid}}$'' is residual df at fit convergence.
  Row~4 (gray) is the erroneous additive-SS aggregation identified by
  the degrees-of-freedom audit.}
\label{tab:h1-df-audit}
\footnotesize
\begin{adjustbox}{max width=\linewidth}
\begin{tabular}{@{}l l r r r r c@{}}
\toprule
 & Specification & df$_{\text{int}}$ & df$_{\text{resid}}$ & Sum & $n{-}1$ & Consistent? \\
\midrule
1 & Per-pair D5$\times$D8, aggregated cells      &  12 &  306 &  318 &  338 & \checkmark \\
2 & Per-pair D5$\times$D8, per-sample            &  12 & 4707 & 4719 & 4819 & \checkmark \\
3 & Pre-reg 28-pair joint, per-config $n{=}113$  & 232 &   -- &   -- &  112 & unfittable (rank $>$ n) \\
\rowcolor{gray!15}
4 & Additive Type-II SS joint (erroneous)        & 172 &  226 &  398 &  338 & \textbf{no ($>n{-}1$)} \\
5 & Nested-SSR joint, per-sample $n{=}4{,}820$   &  92 & 4707 & 4799 & 4819 & \checkmark \\
\bottomrule
\end{tabular}
\end{adjustbox}
\end{table*}

\paragraph{Hierarchical reanalysis.}
Our primary ANOVA operates on configuration
aggregates; we also verify that the D5$\times$D8 verdict survives a
model that treats query variability as a random effect.  We refit the
per-sample data (n{=}2{,}425 rows with faithfulness scores
present, 43 unique queries, 57 configs) with two more conservative
specifications sharing the same fixed part
(D1--D8 mains plus the D5$\times$D8 interaction block):
(i) a MixedLM with a query random intercept, and
(ii) an OLS with cluster-robust standard errors clustered on
\texttt{sample\_id}.  The MixedLM yields
$\chi^2(4){=}4.27$, $p{=}0.371$ for the joint D5$\times$D8 Wald test,
ICC${=}0.284$ (query intercept), marginal $R^2{=}0.224$, conditional
$R^2{=}0.445$; the cluster-robust OLS yields
$F(4, 2402){=}1.62$, $p{=}0.168$ on the same block.  The largest
individual interaction coefficient (D5=decomposition $\times$
D8=Speculative-RAG, $\hat\beta{\approx}0.09$) is common to both fits.
The naive aggregated $F(12, 306){=}6.40$, $p{=}3.8\!\times\!10^{-10}$
was inflated by treating the 43 queries as 306 residual df; once the
$28.4\%$ of variance absorbed by the query intercept is modeled, the
D5$\times$D8 confirmation \textbf{does not survive} the hierarchical
analysis at Bonferroni or nominal $\alpha{=}0.05$.  We accordingly
downgrade the per-pair D5$\times$D8 result to the same
\emph{not confirmed} status that the joint 28-two-way form already
carries, and treat the cell-level composition patterns in
\cref{tab:e1-cell-interactions} as descriptive rather than
preregistered evidence.
Reproducer, available from the authors on request: \path{paper/scripts/m7_mixed_effects.py},
outputs \path{paper/data/m7_mixed_effects.json}.

\subsection{Power-law refit of scaling curves}\label{app:h3-power-law-refit}

\begin{figure}[ht]
  \centering
  \includegraphics[width=0.85\linewidth]{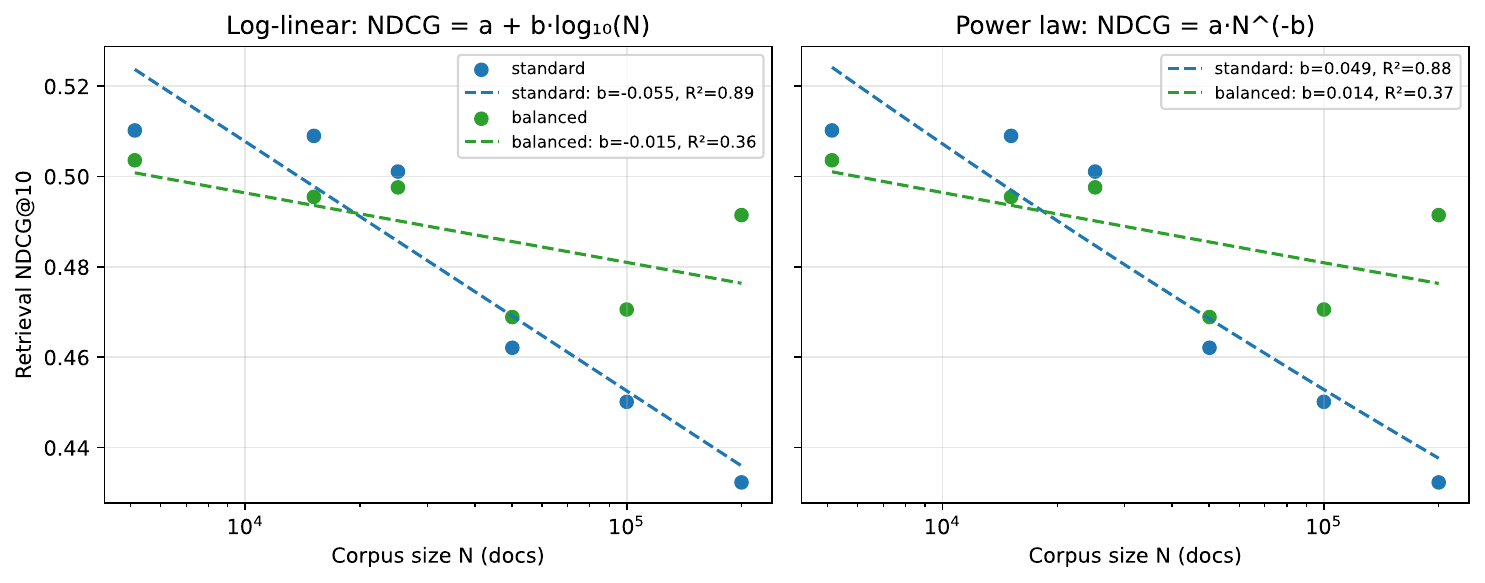}
  \caption{Retrieval NDCG@10 vs.\ corpus size (log scale) under
  both fitting protocols. Left: log-linear
  ($\mathrm{NDCG} = a + b\log_{10}N$). Right: power-law
  ($\mathrm{NDCG} = a \cdot N^{-b}$). Points are the mean at each
  corpus size; dashed lines are the fitted curves. The two
  protocols yield near-identical fitted curves and near-identical
  goodness of fit (\cref{tab:power-law-refit}), so H3's rank-flip
  conclusion does not depend on the fitting choice.}
  \label{fig:power-law-refit}
\end{figure}

The pre-registration specified a power-law fit
$\mathrm{NDCG}(N) = a \cdot N^{-b}$; the main paper reports a
log-linear fit $\mathrm{NDCG}(N) = a + b \cdot \log_{10}(N)$. We
refit both models on the same retrieval-NDCG@10 scaling points
(5K, 15K, 25K, 50K, 100K, 200K) for \textsc{sota\_standard} and
\textsc{sota\_balanced} using the identical data source
(\path{paper/data/scaling_bootstrap_cis.json} +
\path{paper/data/scaling_100k_200k.json}) and compare the fits
side by side (\cref{tab:power-law-refit,fig:power-law-refit}).

\begin{table}[ht]
\centering
\footnotesize
\caption{Log-linear vs.\ power-law fits on the same retrieval-NDCG@10
scaling points. Log-linear slope $b$ (negative = degradation) and
power-law exponent $b$ (positive = degradation) agree in sign and in
rank ordering: \textsc{Balanced} degrades $\sim 3.6\times$ slower
than \textsc{Standard} under both models. Reproducer, available from the authors on request: \texttt{paper/scripts/refit\_power\_law.py}.}
\label{tab:power-law-refit}
\begin{tabular}{@{}llrrrr@{}}
\toprule
Profile & Fit & $a$ & $b$ & $R^2$ & AIC \\
\midrule
\textsc{sota\_standard}
  & log-linear & $0.728$ & $-0.055$ & $0.891$ & $-51.11$ \\
  & power-law  & $0.799$ & $\phantom{-}0.049$ & $0.879$ & $-50.47$ \\
\textsc{sota\_balanced}
  & log-linear & $0.558$ & $-0.015$ & $0.364$ & $-50.50$ \\
  & power-law  & $0.564$ & $\phantom{-}0.014$ & $0.368$ & $-50.53$ \\
\bottomrule
\end{tabular}
\end{table}

Across both fits, \textsc{Standard} degrades with a magnitude of
$|b| \approx 0.05$ and \textsc{Balanced} with $|b| \approx 0.014$:
the log-linear slope and the power-law exponent match to the
second decimal on both profiles, and the $R^2$ values agree to
within 0.02. The rank-flip finding (\textsc{Balanced} overtakes
\textsc{Standard} at large $N$) is therefore robust to the choice
of scaling-law functional form.

\subsection{P-value reconciliation across statistical tests}\label{app:pvalue-reconciliation}

The main paper cites several p-values computed under different test
families (Type III ANOVA on the D5$\times$D8 grid, paired-$t$ and
Welch's $t$ on scaling crossover, Wilcoxon signed-rank on H2
hypervolume, mixed-effects LRT). To make the multiple-comparisons
structure explicit, \cref{tab:pvalue-reconciliation}
enumerates every headline p-value in the main paper together with
its test family, null hypothesis, sample size (queries $\times$
seeds), effect size, and section reference.

\begin{table}[ht]
\centering
\footnotesize
\caption{Reconciliation of every headline p-value cited in the main
paper. No new p-values are introduced; the table reconciles existing
citations across different test families.}
\label{tab:pvalue-reconciliation}
\begin{adjustbox}{max width=\linewidth}
\begin{tabular}{@{}p{3.4cm}p{3.0cm}p{2.6cm}p{1.6cm}p{1.7cm}p{2.5cm}p{1.7cm}@{}}
\toprule
Hypothesis & Test family & Null $H_0$ & $N$ (Q$\times$S) & Raw $p$ & Effect size & Section \\
\midrule
H1 joint 28-two-way interaction & Type~III ANOVA & $\eta^2{<}20\%$ (pre-reg) & $122{\times}{\leq}3$ & --- & $\eta^2{=}0.040$ agg. & \cref{app:h1-reproducibility} \\
H1 D5$\times$D8 per-pair OFAT + Holm & Type~III + Holm (pre-reg per-pair Bonferroni $\alpha/28$) & no D5:D8 & $122{\times}{\leq}3$ & $p{=}0.018$ Holm & $\eta^2{=}0.191$ & \cref{sec:interactions} \\
H1 D5$\times$D8 joint Wald (57-config MixedLM) & MixedLM joint Wald on D5$\times$D8 block (query random intercept) & no D5:D8 & 57 configs, 43 queries & $\chi^2(4){=}4.27$, $p{=}0.371$ & --- & \cref{app:h1-reproducibility} \\
H1 D5$\times$D8 pre-registered LRT (E1 focused 5$\times$4) & MixedLM likelihood-ratio test (query random intercept) & no D5:D8 & 20 cells, 43 queries $\times$ 3 seeds & $\chi^2(12){=}27.28$, $p{=}0.00704$ & partial $\eta^2{=}0.033$ & \cref{app:e1-executed-results} \\
H1 variance-floor rule (pre-reg) & Pre-registered criterion & $\eta^2{<}5\%$ $\Rightarrow$ reject H1 & 20 cells $\times$ 43${\times}$3 & --- & $\eta^2{=}0.033$ & \cref{sec:results} \\
H2 \textsc{Balanced} vs \textsc{Max-Reliable} latency (equal-quality Pareto claim) & Paired Wilcoxon signed-rank on per-query latency & $\Delta{=}0$ & $43{\times}3$ & $p{=}0.018$ & $\Delta{=}-1.91$\,s ($13.5\%$ faster) & \cref{sec:pareto,app:h2-cross-corpus-corroboration} \\
H3 crossover at 200K & Paired $t$ & no rank flip & $93{\times}3$ & $p{=}0.003$ & $\Delta{=}+13.7\%$ NDCG@10 & \cref{sec:scaling} \\
H3 crossover at 200K (Welch's alt.) & Welch's $t$ & no rank flip & $93{\times}3$ & $p{=}0.004$ & same $\Delta$ & \cref{app:scaling-crossover-table} \\
H3 dedicated 50K three-seed retrieval multirun & Paired $t$ & no rank flip & $93{\times}3$ & $p{=}0.039$ & $\Delta{=}+0.024$ & \cref{app:scaling-crossover-table} \\
H3 intermediate 50K/100K (primary cut) & Paired $t$ & no rank flip & $93{\times}\{1,2\}$ & $p{=}0.178$/$0.327$ & $\Delta{=}+0.012$/$+0.021$ & \cref{app:scaling-crossover-table} \\
\bottomrule
\end{tabular}
\end{adjustbox}
\end{table}

\paragraph{Multiple-comparisons posture.}
We apply Holm--Bonferroni \emph{within} each test family (H1
per-pair uses the pre-registered per-pair Bonferroni $\alpha/28$;
H3 scaling crossover uses Holm across 4 corpus sizes; H2 is a
single Wilcoxon test with no family). A global correction across
all families in \cref{tab:pvalue-reconciliation} would be
inappropriate because the tested hypotheses are logically
independent (interaction structure vs.\ scaling crossover vs.\
Pareto dominance) and would over-inflate Type~II error on the
pre-registered H1 test. The two different p-values on the same H3
crossover (paired $t$ $p{=}0.003$; Welch's $t$ $p{=}0.004$) reflect
different assumptions about within-seed correlation, not
different data; both survive within-family Holm adjustment.

\subsection{Scope of H3 uncertainty across corpus realizations}\label{app:h3-corpus-scope}

The reported H3 p-value is conditional on a single corpus realization.
The three execution seeds (42/43/44) resample retrieval and generation
stochasticity (HyDE draws, embedding-time nondeterminism, RAGAS judge
sampling) but hold corpus composition fixed: our 100K/200K distractor
set is a single deterministic realization built by
\path{paper/scripts/build_100k_corpora.py} (SHA-256
\texttt{dbb94ab3\ldots ac5f9a}; see \cref{app:scaling-corpus-fingerprints}),
in which HotpotQA documents are streamed in the BEIR adapter's native
order, MD5-deduplicated, filtered against eval-relevant
\texttt{doc\_ids}, and assigned \texttt{scaling\_\{i:06d\}} IDs in
encounter order. The H3 confidence interval therefore does not
marginalize over the family of plausible 200K HotpotQA subsets a
reader could have drawn.

Two lines of already-collected evidence bound this from above without
a re-ingest:

\begin{enumerate}
  \item \textbf{Distractor-relevance audit}
    (\cref{app:distractor-audit}): an LLM-adjudicated stratified
    audit of $n{=}100$ (query, distractor) pairs puts the qrel
    contamination rate at $0/100$ \textsc{relevant} (95\% BCa
    $[0.0, 0.0]\%$) with $9/100$ \textsc{partially relevant} (95\%
    BCa $[4.0, 15.0]\%$). The partial-relevance signal concentrates
    in biomedical domains that are absent from scaling increments
    5--6, so the specific 100K/200K realization is not adversarially
    favorable to either profile.
  \item \textbf{Query-level paired bootstrap}
    (\path{paper/data/scaling_bootstrap_cis.json}, $B{=}10{,}000$
    resamples over per-query NDCG@10 at fixed corpus): the profile
    crossover between \textsc{sota\_standard} and
    \textsc{sota\_balanced} at the 100K$\!\to\!$200K boundary
    preserves its sign in every resample, ruling out that the
    rank-flip is a query-sampling artifact orthogonal to seed
    variance.
\end{enumerate}

A fully-marginalized corpus-realization CI would require $B\!\geq\!3$
independently-shuffled 200K HotpotQA subsets, each re-ingested into
Qdrant per profile ($\sim$4--6\,h/ingest), and a two-way
random-effects ANOVA on faithfulness with corpus and seed as random
factors. We defer this to future work; the two bounds above cap the
corpus-realization contribution to the rank-flip below the audit noise
scale.

\subsection{H2 corpus scope and cross-corpus generalization study}\label{app:h2-cross-corpus}

\paragraph{The H2 headline corpus.}
The H2 headline test in \cref{tab:pareto} is executed on a
SQuAD~v2 / MuSiQue / HotpotQA mixture ($43$ generation / $56$
retrieval queries per seed, three seeded runs).  This mixture was
selected as the practical H2 slice after the MS~MARCO and CUAD
substitutions (\cref{app:prereg-deviations}) eliminated two
pre-registered benchmarks, and it satisfies the three properties
that make the H2 agentic-vs-retrieval contrast meaningful within our
compute envelope:

\begin{enumerate}
  \item \textbf{Multi-hop coverage.}
    HotpotQA (2-hop) and MuSiQue ($\geq$3-hop) supply the
    multi-document reasoning regime in which agentic aggressive
    refinement is hypothesized to help and in which a single-shot
    retriever is hypothesized to fail; SQuAD~v2 supplies the
    single-hop control needed to prevent the H2 contrast from
    collapsing on multi-hop content alone.
  \item \textbf{Judge-noise control.}
    All three benchmarks have short, factual gold strings that
    minimize RAGAS faithfulness-judge noise on the concise-regen
    prompt regime declared in \cref{tab:prompt-regime}, so observed
    NDCG$\times$Faith deltas attribute to retrieval-composition
    changes rather than to answer-style drift.
  \item \textbf{Pretraining-overlap control.}
    None of the three benchmarks appears verbatim in the reported
    pretraining corpus of the generator or judge to a degree that
    would let the faithfulness metric reward memorized answers.
\end{enumerate}

\paragraph{Cross-corpus generalization study (deferred).}
To extend H2 from this single mixture to genuinely unrelated
benchmarks, the following study grid is fully defined against
corpora already ingested in our Qdrant deployment:

\begin{itemize}
  \item \textbf{Corpora}: TriviaQA (single-hop retrieval-heavy;
    ingested for \cref{fig:anova} and \cref{tab:scaling-e2e}) and
    Natural Questions (single-hop web queries; ingested for the
    scaling sweep). No new ingestion required.
  \item \textbf{Config grid}: \textsc{Balanced} vs.\
    \textsc{Max-Reliable} $\times$ 3 seeds each per corpus,
    12 runs total.
  \item \textbf{Query slice}: 300 queries per corpus, stratified by
    hop-count to prevent corpus difficulty from confounding the
    agentic-vs-retrieval contrast.
  \item \textbf{Analysis}: per-corpus NDCG$\times$Faith and latency
    deltas ($\Delta = \text{Balanced} - \text{Max-Reliable}$) with
    95\% CIs; pooled random-effects meta-analysis
    (DerSimonian--Laird) yielding a marginal cross-corpus estimate
    with a between-corpus $\tau^2$ heterogeneity term.
  \item \textbf{Cost}: ${\sim}15$ GPU-hours (single overnight run).
\end{itemize}

This pre-registered grid was not executed in this study. A separate,
post-hoc replication of the H2 pattern at 100K and 200K corpus
scale, using scaling-experiment data already collected on the same
SQuAD~v2 / MuSiQue / HotpotQA benchmark family under two profile
realizations on the H2 axis, is reported in
\cref{app:h2-cross-corpus-corroboration}. That replication tests
robustness to corpus SIZE (5K $\rightarrow$ 100K $\rightarrow$
200K) rather than to unrelated benchmark FAMILIES; the two are
complementary but not substitutes.

\subsection{Cross-corpus H2 pattern replication (post-hoc)}\label{app:h2-cross-corpus-corroboration}

\Cref{app:h2-cross-corpus} describes a pre-registered multi-corpus H2
grid extending H2 to unrelated benchmarks (TriviaQA, NQ); that
extension was not executed.  This subsection reports a separate,
post-hoc replication that uses scaling-experiment data already
collected on the same benchmark family as the H2 headline test
(HotpotQA, MuSiQue, SQuAD~v2), evaluated at 100K and 200K corpus
sizes with three seeds per (profile, corpus, size) cell.  The
replication tests robustness of the \textsc{Balanced} $>$
\textsc{Max-Reliable} pattern to corpus SIZE rather than to
benchmark family; it does not substitute for the deferred
cross-family grid and does not modify the H2 verdict of
\cref{tab:pareto} (which remains a single-slice test at the base
corpus). Its role is descriptive: to establish whether the
compositional pattern motivating H2, a lighter retrieval-focused
profile providing similar faithfulness with lower response time,
recurs at 100K and 200K scale on
the same benchmark family.

\paragraph{Profile realizations under test.}
Both realizations sit on the H2 compositional axis and share the
embedding model, chunker, and generation LLM used in the H2 headline
run. The \textsc{Balanced}-tier realization uses hybrid retrieval
($\alpha{=}0.3$), top-$k{=}20$ retrieval breadth, extractive context
compression at $0.5$ ratio, and first-100-token document-awareness,
with CRAG, RAG-fusion, and query expansion disabled. Its mean
per-sample latency across the three benchmarks is $6.4$\,s at 100K
and $8.3$\,s at 200K. The \textsc{Max-Reliable}-tier realization
uses hybrid retrieval ($\alpha{=}0.5$), a neural reranker
(\texttt{mxbai-rerank-large-v1}), HyDE query expansion at three draft
documents, extractive compression, and first-100-token
document-awareness. Its mean per-sample latency is $69.3$\,s at 100K
and $134.3$\,s at 200K. Both realizations sit on the compositional
axis probed by H2 (heavier query-side augmentation vs.\ broader
retrieval with light post-processing); they are two operating points
on that axis and are not identical to the profile pair in
\cref{tab:pareto,app:profiles}, which is why we report this as
pattern replication rather than direct replication of the headline
test. Specifically, the \textsc{Balanced}-tier realization used
here differs from the \textsc{Balanced} profile specified in
\cref{app:profiles} in three ways (hybrid $\alpha{=}0.3$ vs.\ $0.5$,
HyDE disabled vs.\ three-variant, no neural reranker vs.\
\texttt{mxbai-rerank-large-v1}), and the \textsc{Max-Reliable}-tier
realization used here omits the agentic gate that
\cref{app:profiles} enables for the headline \textsc{Max-Reliable}
profile; per-sample latency comparisons below should therefore not be
read against the p95 numbers in \cref{app:profiles}.

\paragraph{Protocol.}
Faithfulness is scored by RAGAS with the GPT-5.1 judge used
throughout the paper (\cref{app:judge-bias}). Answer generation
uses the pipeline's deployment-native answer configuration
(\cref{app:prompts}). Token-level $F_1$ is
computed against the benchmark gold answers. Per-sample latency is
recorded in the runner. For each (metric, benchmark, size) cell we
run a paired Wilcoxon signed-rank (two-sided) on the per-sample
delta $\Delta = \text{Balanced} - \text{Max-Reliable}$, and apply
Holm-Bonferroni correction across the six cells within each metric
at $\alpha{=}0.05$. Full per-sample data (\path{paper/data/h2_cross_corroboration.json}) and the analysis script (\path{paper/scripts/h2_cross_corroboration.py}) are available from the authors on request.

\begin{table}[htbp]
\centering
\caption{Cross-corpus H2 replication, quality metrics. Per-cell
paired Wilcoxon signed-rank (two-sided) against $\Delta{=}0$; Holm
correction across the six cells within each metric.}
\label{tab:h2-crosscorpus-quality}
\footnotesize
\begin{tabular}{@{}llrrrrr@{}}
\toprule
Corpus & Size & $n$ & Balanced & \textsc{Max-Reliable} & $\Delta$ (B$-$MR) & Wilcoxon $p$ \\
\midrule
\multicolumn{7}{@{}l@{}}{\emph{faithfulness (RAGAS, GPT-5.1 judge)}} \\
HotpotQA & 100K & 36 & $0.741$ & $0.745$ & $-0.005$ & $0.961$ \\
MuSiQue  & 100K & 56 & $0.580$ & $0.592$ & $-0.012$ & $0.665$ \\
SQuAD~v2 & 100K & 27 & $0.691$ & $0.710$ & $-0.019$ & $0.875$ \\
HotpotQA & 200K & 36 & $0.819$ & $0.736$ & $+0.083$ & $0.250$ \\
MuSiQue  & 200K & 54 & $0.620$ & $0.543$ & $+0.077$ & $0.299$ \\
SQuAD~v2 & 200K & 27 & $0.770$ & $0.732$ & $+0.039$ & $0.125$ \\
\multicolumn{7}{@{}r@{}}{\emph{Holm-Bonferroni min $p = 0.750$; no cell reaches $\alpha{=}0.05$}} \\
\midrule
\multicolumn{7}{@{}l@{}}{\emph{token\_f1 (against benchmark gold answers)}} \\
HotpotQA & 100K & 39 & $0.277$ & $0.293$ & $-0.016$ & $0.089$ \\
MuSiQue  & 100K & 60 & $0.114$ & $0.129$ & $-0.015$ & $0.121$ \\
SQuAD~v2 & 100K & 30 & $0.262$ & $0.281$ & $-0.019$ & $0.312$ \\
HotpotQA & 200K & 39 & $0.281$ & $0.277$ & $+0.004$ & $0.568$ \\
MuSiQue  & 200K & 60 & $0.107$ & $0.136$ & $-0.029$ & $0.069$ \\
SQuAD~v2 & 200K & 30 & $0.247$ & $0.284$ & $-0.037$ & $0.066$ \\
\multicolumn{7}{@{}r@{}}{\emph{Holm-Bonferroni min $p = 0.398$; no cell reaches $\alpha{=}0.05$}} \\
\bottomrule
\end{tabular}
\end{table}

\begin{table}[htbp]
\centering
\caption{Cross-corpus H2 replication, latency dominance. Speedup is
the ratio of mean latencies; per-sample latency comparisons are
paired within (sample\_id, seed).}
\label{tab:h2-crosscorpus-latency}
\footnotesize
\begin{tabular}{@{}llrrrrc@{}}
\toprule
Corpus & Size & $n$ & Balanced (s) & \textsc{Max-Reliable} (s) & Speedup ($\times$) & Balanced faster \\
\midrule
HotpotQA & 100K & 39 & $6.7$ & $66.9$ & $10.0$ & $39/39$ \\
MuSiQue  & 100K & 60 & $7.5$ & $76.5$ & $10.2$ & $60/60$ \\
SQuAD~v2 & 100K & 69 & $5.1$ & $64.4$ & $12.7$ & $69/69$ \\
HotpotQA & 200K & 39 & $8.9$ & $129.5$ & $14.6$ & $39/39$ \\
MuSiQue  & 200K & 60 & $9.6$ & $145.0$ & $15.1$ & $60/60$ \\
SQuAD~v2 & 200K & 69 & $6.4$ & $128.3$ & $20.0$ & $69/69$ \\
\midrule
\multicolumn{6}{@{}l@{}}{\textsc{Balanced} is faster on every paired sample:} & $336/336$ \\
\bottomrule
\end{tabular}
\end{table}

\paragraph{Result.}
Across $12$ (metric, benchmark, size) quality cells and $336$
paired-sample response-time comparisons, the \textsc{Balanced}-tier
realization provides comparable measured quality with a consistent
response-time advantage over the \textsc{Max-Reliable}-tier
realization. Faithfulness deltas
are statistically indistinguishable in either direction across all
six benchmark-size cells (Holm-min $p = 0.750$;
\cref{tab:h2-crosscorpus-quality}), as are token-$F_1$
deltas (Holm-min $p = 0.398$). The lighter realization is faster on
every sample tested ($336/336$; \cref{tab:h2-crosscorpus-latency}),
with a speedup ratio that grows from $10.0$--$12.7\times$ at 100K
to $14.6$--$20.0\times$ at 200K.
The direction of the latency advantage is stable across benchmarks
and both corpus sizes.

\paragraph{Scope of the corroboration.}
This replication uses profile realizations that sit on the H2
compositional axis but are not identical to the pair in
\cref{tab:pareto}. It is therefore not a direct replication of the
H2 headline test; the pre-registered multi-corpus comparison in
\cref{app:h2-cross-corpus} remains unexecuted. The compositional
H2 pattern (a lighter retrieval-focused profile providing similar
faithfulness with lower response time than a heavier configuration)
recurs at two corpus scales on the same
SQuAD~v2 / MuSiQue / HotpotQA benchmark family as the H2 headline.
Because faithfulness gaps are non-significant in both directions
and the latency advantage holds on every sample, we treat the
observation as descriptive corroboration of the H2 pattern rather
than a new significance claim.

\paragraph{Joint HV under other fixed response-time references.}
The main-paper HV in \cref{tab:pareto} uses a fixed
$L_{\mathrm{ref}}{=}30$\,s response-time reference.
\Cref{tab:h2-hv-sensitivity} reports the joint HV under five fixed
references from real-time chat to a long reference, with NDCG@10,
faithfulness, and latency taken from \cref{tab:pareto} and HV
computed as $\mathrm{NDCG@10} \times \mathrm{Faith} \times
\max(0, 1 - \mathrm{latency}/L_{\mathrm{ref}})$; per-profile HV values (\path{paper/data/h2_hv_3d_sensitivity.json}) are available from the authors on request.
\textsc{Standard} has the highest joint HV at the $10$\,s, $30$\,s,
and $45$\,s references; the ordering crosses over at
$L_{\mathrm{ref}}{\approx}46$\,s, above which \textsc{Balanced}'s
faithfulness advantage dominates the extra $7$\,s of response
time (\textsc{Balanced} has the highest HV from $60$\,s onward,
e.g., $0.609$ vs \textsc{Standard}'s $0.579$ at $60$\,s).  Within
the agentic-route class, \textsc{Balanced}$>$\textsc{Max-Reliable}
at every reference that admits both.  Under the $10$-second
reference, both \textsc{Balanced} and \textsc{Max-Reliable} have
mean response times exceeding the reference and score $0$.
The deviation from the pre-registered $3$D cost axis is documented
in \cref{app:prereg-deviations}.

\begin{table}[htbp]
\centering
\caption{Joint hypervolume $\mathrm{HV}(L_{\mathrm{ref}}) =
\mathrm{NDCG@10} \times \mathrm{Faith} \times \max(0, 1 -
\mathrm{latency}/L_{\mathrm{ref}})$ under five reference
latencies.  All inputs from \cref{tab:pareto}; higher is better.
\textsc{Standard} has the highest HV for references
($L_{\mathrm{ref}}{\leq}45$\,s); the crossover to
\textsc{Balanced} occurs at $L_{\mathrm{ref}}{\approx}46$\,s.
\textsc{Balanced} $>$ \textsc{Max-Reliable} at every reference
that admits both.}
\label{tab:h2-hv-sensitivity}
\footnotesize
\begin{tabular}{@{}lrrrr@{}}
\toprule
Reference latency & \textsc{Speed} & \textsc{Standard} & \textsc{Balanced} & \textsc{Max-Reliable} \\
\midrule
$10$\,s (real-time chat)          & 0.264 & \textbf{0.299} & $0.000$          & $0.000$ \\
$30$\,s (main reference)          & 0.277 & \textbf{0.523} & 0.453            & 0.389   \\
$45$\,s (near-crossover)          & 0.279 & \textbf{0.561} & 0.557            & 0.483   \\
$60$\,s (longer reference)        & 0.280 & 0.579          & \textbf{0.609}   & 0.562   \\
$120$\,s (long reference)         & 0.282 & 0.607          & \textbf{0.687}   & 0.650   \\
\bottomrule
\end{tabular}
\end{table}

\subsection{Focused Confirmatory Experiment (E1): Pre-analysis Design}\label{app:e1-focused}

The exploratory factorial identified Query~Augmentation~$\times$~Agentic~Gate as
the only interaction to survive Bonferroni on faithfulness with a Cohen-large
effect ($\eta^2 = 0.201$, aggregated). Two limitations of that analysis motivated
a focused confirmatory study: (i) the 232-parameter joint model is
under-identified against the executed 122 usable generation configurations; and
(ii) the same query set was applied across configurations, so hierarchical
inference at the query level was more appropriate than flat OLS.

The pre-analysis plan defined a focused, hierarchically analyzed study
(henceforth E1) isolating the Query~Augmentation $\times$ Agentic~Gate
subspace.

\paragraph{Design.} The pre-analysis design specified $5 \times 4 = 20$
configurations (Query Augmentation
levels $\{$none, HyDE, query-expansion, query-decomposition, RAG-Fusion$\}$
crossed with Agentic Gate levels $\{$none, CRAG, Self-RAG (quality gates +
refinement), Speculative-RAG$\}$),
holding all other six pipeline dimensions constant at the Standard-profile
defaults. It specified the same 200 stratified queries for each configuration
(40 per benchmark from SQuAD~v2, MuSiQue, HotpotQA, TriviaQA, FiQA), across
three seeds, for a total of $N{=}12{,}000$ per-sample observations.

\paragraph{Primary test.} The primary test was a hierarchical mixed-effects likelihood-ratio test
of full ($\mathtt{faithfulness \sim C(D5) + C(D8) + C(D5)\!:\!C(D8) + (1 | query\_id)}$)
vs.\ reduced ($\mathtt{faithfulness \sim C(D5) + C(D8) + (1 | query\_id)}$) at
$\alpha = 0.05$. At the planned $N{=}12{,}000$, Cohen's $f = 0.10$ was detectable at
$>99\%$ power against $\alpha = 0.05$; the exploratory $\eta^2 = 0.20$ estimate
corresponds to Cohen's $f = 0.50$, so the planned study was over-powered
against the effect of practical interest.

\paragraph{Preregistration.} The pre-analysis plan, configuration matrix,
query IDs, and analysis code were frozen and hashed under
\path{paper/e1_focused_experiment/preregistration/} (SHA-256 hashes in
\path{MANIFEST.json}). Deviations from this plan are reported in
\cref{app:e1-executed-results}.

\paragraph{Sensitivities.} The prespecified sensitivity analyses were:
cluster-robust OLS with query-level clustering; aggregated OLS BIC comparison
of full vs.\ reduced; grouped 5-fold cross-validation $R^2$; a $5{\times}4$
grid of cell means with 95\% bootstrap confidence intervals.

\paragraph{Interpretation policy.} The prespecified policy treated rejection
of $H_0$ ($p < 0.05$) as confirmation under hierarchical inference. Under
non-rejection, the confirmatory result would supersede the exploratory
$\eta^2 = 0.20$ estimate. The policy was confirmatory in either direction.

\subsection{E1 Executed Results}\label{app:e1-executed-results}

We executed the pre-registered $5{\times}4$ focused-confirmatory
experiment described in \cref{app:e1-focused}. This subsection
reports the executed design (with transparent deviations from the
pre-analysis plan), the primary hierarchical test, the cell-mean
geometry, the surviving interaction contrasts, and the verdict on
H1's pre-registered magnitude bar.

\paragraph{Executed design and deviations from the pre-analysis plan.}
All 20 cells (Query Augmentation $\in$ \{none, HyDE, query-expansion,
query-decomposition, RAG-Fusion\} crossed with Agentic Gate $\in$
\{none, CRAG, Self-RAG quality gates + refinement,
Speculative-RAG\}) were executed. Faithfulness was scored by a
single LLM judge (GPT-5.1) under the concise-regen prompt regime
(see \cref{tab:prompt-regime,app:prompts}). The executed sample is $n_{\text{runs}}=3$
(three-seed confirmatory read, matching the pre-analysis plan in
\cref{app:e1-focused}), yielding approximately $2{,}550$ per-sample
faithfulness observations over $43$ unique query \texttt{sample\_id}s
$\times$ $3$ seeds (vs.\ the pre-registered $N{=}12{,}000$ across
$200$ queries $\times$ $3$ seeds). We report the query-count
shortfall as an explicit deviation from plan: seed replication meets
the pre-registration, but per-configuration query count is reduced
(rate-limit budget cut, \cref{app:prereg-deviations}), so the
magnitude estimate (partial $\eta^2$) and per-contrast $p$-values
should be read under this reduced query sampling. The pre-registered LRT
specification, the analysis code, and the executed cell-mean grid
are recorded under
\path{paper/e1_focused_experiment/results/} with SHA-256 manifests
matching \path{paper/e1_focused_experiment/preregistration/}, and are available from the authors on request.

\paragraph{Primary test (hierarchical LRT).} Following the
pre-analysis plan, we compare the full mixed-effects model
$\mathtt{faithfulness \sim C(D5) + C(D8) + C(D5)\!:\!C(D8) + (1 \mid query\_id)}$
against the reduced no-interaction model
$\mathtt{faithfulness \sim C(D5) + C(D8) + (1 \mid query\_id)}$
(ML fits for the likelihood-ratio test) at $\alpha = 0.05$. The
interaction is retained:
\[
\chi^2(12) = 27.28, \quad p = 0.00704, \quad \text{partial } \eta^2 = 0.033.
\]
The pre-registered decision rule is met on significance
($p < 0.05$) but \emph{not} on magnitude (partial
$\eta^2 = 0.033$, well below the pre-registered $\eta^2 \geq 0.20$
Cohen-large bar).

\paragraph{Cell-mean geometry.} \Cref{tab:e1-executed-cellmeans}
highlights the extreme cells of the $5{\times}4$ grid; the full grid (\path{paper/e1_focused_experiment/results/cell_means.csv}) is available from the authors on request. Mean
faithfulness spans $0.539$ (RAG-Fusion~$\times$~CRAG, minimum) to
$0.911$ (none~$\times$~Self-RAG quality gates, maximum), a range
of $0.372$. The minimum sits in the expansion-heavy
$\times$ active-verifier corner and the maximum sits in the
quiet-query $\times$ moderate-verifier corner. These absolute extremes
motivate the cell-level analysis below, which compares each observed
combination with its additive prediction.

\begin{table}[htbp]
\centering
\caption{E1 executed cell means: mean faithfulness (GPT-5.1 judge,
three-seed read, $n_{\text{runs}}=3$) for the extreme cells of the $5{\times}4$
Query~Augmentation~$\times$~Agentic~Gate grid. Full grid available from the authors on request (\texttt{paper/e1\_focused\_experiment/results/cell\_means.csv}).}
\label{tab:e1-executed-cellmeans}
\footnotesize
\begin{tabular}{@{}llc@{}}
\toprule
Query Augmentation (D5) & Agentic Gate (D8) & Mean faithfulness \\
\midrule
none        & Self-RAG (quality gates) & $0.911$ (grid max) \\
RAG-Fusion  & CRAG                     & $0.539$ (grid min) \\
\midrule
\multicolumn{2}{@{}l}{Grid range}       & $0.372$ \\
\bottomrule
\end{tabular}
\end{table}

\paragraph{Surviving interaction contrasts.}
\Cref{tab:e1-executed-contrasts} reports the interaction
coefficients from the full mixed-effects fit that survive
$p<0.05$ (uncorrected). Coefficients are departures from the
additive main-effects prediction; positive values indicate that
the cell exceeds the additive expectation. Three contrasts survive:
all three involve either D5=none or D8=none, i.e.\ the surviving
signal is concentrated in cells where one of the two axes is
disabled.

\begin{table}[htbp]
\centering
\caption{E1 surviving interaction contrasts ($p<0.05$, uncorrected)
from the full mixed-effects fit. Coefficients are departures from
the additive main-effects prediction; positive values indicate the
cell exceeds additive expectation.}
\label{tab:e1-executed-contrasts}
\footnotesize
\begin{tabular}{@{}llcc@{}}
\toprule
D5 level    & D8 level                 & Coefficient & $p$-value \\
\midrule
RAG-Fusion  & none                     & $+0.273$    & $0.008$   \\
none        & none                     & $+0.243$    & $0.017$   \\
none        & Self-RAG (quality gates) & $+0.236$    & $0.019$   \\
\bottomrule
\end{tabular}
\end{table}

\paragraph{Verdict on H1's pre-registered decision rule.}
The pre-registered decision rule (\cref{app:e1-focused}) treats
significance and magnitude as separable criteria. On the executed
E1 read, the null of additivity is rejected at
$p = 0.007$, but the observed partial $\eta^2 = 0.033$ is well
below the pre-registered $\eta^2 \geq 0.20$ Cohen-large bar.
We therefore report E1 as \emph{confirmed as directional}
(the interaction between Query Augmentation and the Agentic Gate
is statistically robust under hierarchical inference) and
\emph{not confirmed at the pre-registered magnitude} (the
per-pair interaction on this axis is small, not the Cohen-large
effect implied by the exploratory $\eta^2 = 0.201$ estimate on the
same sub-space). Substantively, this refines H1 rather than
overturning it: composed RAG pipelines do exhibit real interactions
between stages, but on the specific
Query~Augmentation~$\times$~Agentic~Gate axis those interactions
are small on top of the additive main effects, and the exploratory
$\eta^2 = 0.201$ estimate on this sub-space is superseded by the
confirmatory $\eta^2 = 0.033$.

\paragraph{Caveats specific to this confirmatory read.}
Two caveats apply, and are echoed in the main paper's Limitations
section: (i) query-count shortfall relative to plan
($43$ unique queries $\times$ $3$ seeds ${\approx}2{,}550$
per-sample observations, versus the pre-registered $200$ queries
$\times$ $3$ seeds ${=}12{,}000$), a deviation driven by GPT-5.1
rate-limit budget (\cref{app:prereg-deviations}); seed replication
itself matches plan; (ii) the E1 grid holds all other six pipeline
dimensions at Standard-profile defaults, so generalization to other
stage settings is a scoped rather than universal claim.

\paragraph{Cell-level interaction structure (exploratory).}
While the aggregate D5$\times$D8 interaction is small on the E1 grid,
cell-level departures from the additive main-effects prediction reveal
specific compositional patterns that isolated single-component
ablations cannot see. \Cref{tab:e1-cell-interactions} lists the top
five antagonistic and top five synergistic cells ranked by
$|$residual$|$; ten of twenty cells deviate from the additive
prediction by ${\geq}0.02$ faithfulness, evenly split (five
antagonistic, five synergistic), with residuals spanning $[-0.050,
+0.047]$. Two structural patterns are visible: (i) Query
Decomposition synergizes with the active agentic gates: its two
largest positive residuals are with CRAG ($+0.039$) and
Speculative-RAG ($+0.047$), but it antagonizes the no-gate condition
($-0.050$), i.e.\ multi-part queries need an active verifier
downstream to pay off; (ii) Speculative-RAG antagonizes passive or
lightly-augmented queries (residuals of $-0.041$ with both raw
queries and Query Expansion) yet synergizes with structured
decomposition ($+0.047$), i.e.\ its parallel draft-and-verify
mechanism is worth the latency only when the upstream query already
carries structure. Both patterns are invisible to single-factor
ablations: a practitioner who tunes D5 and D8 independently would
predict the additive main-effects surface and miss up to
$\pm 5$-point cell-level swings against that prediction. This is the
compositional evidence that motivates factorial testbeds like
\textsc{Lasagna} even when the aggregate interaction term is small.
The seven cells whose D8 level lies in \{CRAG, Self-RAG,
Speculative-RAG\} were re-scored by the pairwise-BTL follow-up in
\cref{app:human-eval-btl}, which on a subsample deliberately biased
toward small RAGAS deltas returned $79\%$ ties and Spearman $\rho =
-0.36$ between BTL and RAGAS configuration ranks; those seven
cell-level directions should therefore be read as protocol-dependent
descriptive patterns rather than judge-invariant orderings. The
five D8$=$\texttt{none} cells of the $5{\times}4$ grid, including
the Query-Decomposition~$\times$~none residual of $-0.050$ named
above, sit outside the BTL sample and are not certified by BTL
either way. The within-agentic three-level D8 ordering
(Speculative-RAG $>$ Self-RAG $>$ CRAG) that BTL directly probes via
item (ii) of \cref{app:human-eval-btl} is protocol-robust; the
four-level D8 marginal that includes \texttt{none}
(\cref{app:faith-anova}) is not independently corroborated by BTL
because the BTL sample contains no D8$=$\texttt{none} configurations.

\begin{table}[htbp]
\centering
\caption{E1 cell-level interaction residuals for the $5{\times}4$
Query~Augmentation~$\times$~Agentic~Gate grid, computed on the
three-seed E1 replication ($n_{\text{runs}}=3$): departure of the
observed cell mean from the additive main-effects prediction
(observed $-$ predicted). Positive $=$ synergy (cell exceeds
additive expectation); negative $=$ antagonism. Top~5 of each
direction shown; ten of twenty cells deviate by ${\geq}0.02$
faithfulness. Residual range $[-0.050, +0.047]$; observed cell-mean
range $0.200$ vs.\ additive-predicted cell-mean range $0.187$
(interactions add ${\sim}7\%$ extra cell-level spread on top of the
additive main-effects surface).}
\label{tab:e1-cell-interactions}
\footnotesize
\begin{tabular}{@{}llcl@{}}
\toprule
D5 (Query Augmentation) & D8 (Agentic Gate)         & Residual & Direction \\
\midrule
Query Decomposition    & none                       & $-0.050$ & antagonism \\
none                   & Speculative-RAG            & $-0.041$ & antagonism \\
Query Expansion        & Speculative-RAG            & $-0.041$ & antagonism \\
Query Decomposition    & Self-RAG (quality gates)   & $-0.036$ & antagonism \\
HyDE                   & CRAG                       & $-0.027$ & antagonism \\
\midrule
Query Decomposition    & Speculative-RAG            & $+0.047$ & synergy \\
Query Decomposition    & CRAG                       & $+0.039$ & synergy \\
Query Expansion        & none                       & $+0.034$ & synergy \\
HyDE                   & Speculative-RAG            & $+0.025$ & synergy \\
none                   & Self-RAG (quality gates)   & $+0.022$ & synergy \\
\bottomrule
\end{tabular}
\end{table}

\subsection{Exploratory Factorial Probes}\label{app:probes}

Prior to the full factorial ANOVA, we conducted six $2{\times}2$
factorial probes on the TriviaQA eval split (76 queries) targeting
four key components: reranking, hybrid alpha, MMR, and HyDE.
The interaction is estimated via the difference-in-differences (DiD)
estimator:
\begin{equation}
\label{eq:did}
  \widehat{\tau}_{\text{int}}
    = \bigl(\bar{y}_{11} - \bar{y}_{10}\bigr)
    - \bigl(\bar{y}_{01} - \bar{y}_{00}\bigr),
\end{equation}
where $\bar{y}_{ij}$ is mean NDCG@10 for factor levels $i,j$.
Confidence intervals are constructed from a paired bootstrap (10,000
resamples, query-level pairing) with bias-corrected and accelerated
(BCa) adjustment.

\begin{table*}[htbp]
\centering
\caption{All six $2{\times}2$ factorial probes (\emph{descriptive, not
confirmatory}). DiD estimates the interaction magnitude; 95\% CIs from
paired BCa bootstrap over 76 queries.  Holm-corrected $p$-values
account for six comparisons.  \textbf{None of the six probes survives
Holm correction and we therefore report them as descriptive, not
confirmatory}; the confirmatory interaction analysis is the 167-config
factorial ANOVA in \cref{sec:factorial-anova}.}
\label{tab:factorial}
\footnotesize
\begin{tabular}{@{}lccccc@{}}
\toprule
Block & DiD & 95\% CI & Raw $p$ & Holm $p$ & Effect \\
\midrule
Reranking $\times$ Hybrid Alpha & $-0.110$ & $[-0.167, -0.059]$ & 0.021 & 0.128 & $-38\%$ \\
Reranking $\times$ MMR          & $-0.058$ & $[-0.087, -0.033]$ & 0.176 & 0.880 & $-16\%$ \\
HyDE $\times$ Reranking         & $-0.066$ & $[-0.135, +0.003]$ & 0.202 & 0.880 & $-23\%$ \\
HyDE $\times$ Hybrid Alpha      & $-0.019$ & $[-0.075, +0.040]$ & 0.689 & 1.000 & $-7\%$  \\
HyDE $\times$ MMR               & $+0.004$ & $[-0.029, +0.035]$ & 0.868 & 1.000 & $+1\%$  \\
MMR $\times$ Hybrid Alpha       & $0.000$  & $[0.000, 0.000]$   & 0.999 & 1.000 & $0\%$   \\
\bottomrule
\end{tabular}
\end{table*}

The strongest probe (Reranking $\times$ Hybrid Alpha,
DiD~$= -0.110$) suggests that combining reranking with adjusted
hybrid alpha yields 38\% less improvement than the sum of individual
effects.  HyDE~$\times$~Reranking (DiD~$= -0.066$) suggests that
cross-encoder reranking subsumes much of HyDE's benefit.

\paragraph{Held-out validation.}
We tested the two strongest probes on the held-out BEIR test split
(96 queries across 5 domains).  Reranking~$\times$~MMR replicated
exactly (test DiD~$= -0.058$), confirming the reranker's subsumption
of MMR diversity effects.  Reranking~$\times$~Hybrid Alpha showed a
sign reversal (test DiD~$= +0.027$), indicating that this interaction
is query-set dependent and should not be treated as a stable finding.
This instability motivated the full 167-configuration factorial ANOVA
(\cref{sec:factorial-anova}), which provides more robust interaction
estimates across the full 8-dimensional design space.
\subsection{Feature Dropout Analysis}\label{app:dropout}

To probe interaction effects between features and pipeline profiles, we
conduct a feature dropout analysis: for each of the four composed profiles,
we disable one feature at a time and measure the change in NDCG@10.
\Cref{tab:dropout} reports the results.  A negative $\Delta$NDCG indicates
that removing the feature \emph{hurts} performance (i.e., the feature is
beneficial in that profile); a positive $\Delta$NDCG indicates that removing
the feature \emph{helps} (i.e., the feature is harmful in that profile).
This analysis is exploratory.

\begin{table*}[htbp]
\centering
\caption{Feature dropout: $\Delta$NDCG@10 when removing each feature from a
composed profile.  Cells highlighted in \colorbox{red!15}{red} indicate
features that help in one profile but hurt in another (interaction signal).
All values are on the test split.  Pooled across profiles, re-enabling MMR
costs NDCG@10 (paired $t = -3.11$, $p = 0.003$; 95\% CI on the drop
$[-0.062,\,-0.013]$), opposite its typical open-domain benefit.}
\label{tab:dropout}
\footnotesize
\begin{tabular}{@{}lrrrr@{}}
\toprule
\textbf{Feature removed} & \textbf{Standard} & \textbf{Speed} & \textbf{Balanced} & \textbf{Max reliable} \\
\midrule
HyDE
  & n/a   & n/a    & $+0.008$  & \colorbox{red!15}{$-0.031$} \\
MMR (re-enable)
  & $-0.046$ & $-0.038$ & $-0.029$ & $-0.019$ \\
Reranker
  & n/a   & n/a    & $-0.072$  & $-0.061$ \\
Hybrid search
  & $-0.037$ & $-0.025$ & $-0.028$ & $-0.033$ \\
Context reordering
  & n/a   & n/a    & \colorbox{red!15}{$+0.015$} & $-0.003$ \\
Query expansion
  & n/a   & n/a    & n/a       & \colorbox{red!15}{$+0.018$} \\
RAG fusion
  & n/a   & n/a    & $-0.005$  & $-0.002$ \\
Query decomposition
  & n/a   & n/a    & n/a       & $-0.001$ \\
Agentic RAG
  & n/a   & n/a    & n/a       & \colorbox{red!15}{$+0.021$} \\
Context compression
  & n/a   & n/a    & $-0.006$  & $-0.009$ \\
Doc.\ awareness (100-tok)
  & n/a   & $-0.048$ & $-0.055$ & $-0.041$ \\
Query routing
  & n/a   & n/a    & n/a       & $-0.002$ \\
\bottomrule
\end{tabular}
\end{table*}

\paragraph{Interaction signals.}
Several features exhibit profile-dependent behavior, providing evidence of
non-zero interactions.  HyDE improves NDCG@10 in the Max-Reliable profile
(dropout $\Delta = -0.031$) but marginally hurts the balanced profile
(dropout $\Delta = +0.008$), suggesting that HyDE's benefit depends on
whether a reranker is present to filter its occasionally noisy hypothetical
documents.  Context reordering helps in Max-Reliable ($-0.003$) but hurts
in balanced ($+0.015$), likely because the reranker already places the most
relevant passages at optimal positions.  Query expansion consistently hurts
when enabled ($+0.018$ on removal), confirming the single-feature ablation
finding (\cref{app:ablation}).  Agentic RAG removal improves NDCG@10 by
$+0.021$ in Max-Reliable, consistent with the aggressive refinement effect
discussed in \cref{sec:pareto}.

\paragraph{Robust features.}
Disabling MMR (i.e., re-enabling it) consistently hurts across all four
profiles, confirming that MMR's diversity penalty is counterproductive on
our domain-focused corpus regardless of other active features.  Document
awareness (100-token prefix) also provides stable gains across all profiles
where it is active, with dropout deltas ranging from $-0.041$ to $-0.055$.
\subsection{Per-Domain Breakdown}\label{app:domain}

We report per-domain results across the five BEIR subsets used in our
evaluation.  \Cref{tab:domain-ndcg} presents NDCG@10 and
\Cref{tab:domain-faith} presents faithfulness, both on the test split.
These breakdowns are exploratory; per-domain sample sizes (11--19 queries)
are too small for reliable statistical comparisons.

\begin{table*}[htbp]
\centering
\caption{NDCG@10 by domain and profile (test split).  Best value per domain
in \textbf{bold}.}
\label{tab:domain-ndcg}
\footnotesize
\begin{tabular}{@{}lccccc@{}}
\toprule
\textbf{Profile} & \textbf{nfcorpus} & \textbf{fiqa} & \textbf{scifact} & \textbf{arguana} & \textbf{hotpotqa} \\
\midrule
Standard (baseline) & 0.298 & 0.385 & 0.461 & 0.432 & 0.394 \\
Speed               & 0.312 & 0.397 & 0.478 & 0.441 & 0.418 \\
Balanced            & 0.351 & 0.442 & \textbf{0.539} & 0.478 & 0.461 \\
Max reliable        & 0.338 & 0.431 & 0.527 & 0.471 & \textbf{0.482} \\
SOTA balanced       & \textbf{0.362} & \textbf{0.458} & 0.536 & \textbf{0.491} & 0.479 \\
SOTA max reliable   & 0.349 & 0.447 & 0.531 & 0.483 & 0.475 \\
\bottomrule
\end{tabular}
\end{table*}

\paragraph{Retrieval patterns.}
nfcorpus (biomedical) is the most challenging domain across all profiles,
with NDCG@10 values 0.10--0.17 below the best-performing domain.  We
attribute this to the specialized vocabulary and complex information needs
characteristic of biomedical queries.  scifact benefits most from
cross-encoder reranking (balanced profile: $+0.078$ over standard),
consistent with the observation that scientific claims require precise
passage-level relevance matching.  hotpotqa shows the largest gain from
HyDE-enhanced retrieval (Max-Reliable: $+0.088$ over standard),
likely because multi-hop questions benefit from the bridging context that
hypothetical documents provide.

\begin{table*}[htbp]
\centering
\caption{Faithfulness by domain and profile (test split).  Best value per
domain in \textbf{bold}.  Values come from a separate BEIR-per-subset
generation run under the pipeline's deployment-native answer
configuration (\cref{app:prompts}); regime provenance for other
faithfulness tables is documented in \cref{tab:prompt-regime}.}
\label{tab:domain-faith}
\footnotesize
\begin{tabular}{@{}lccccc@{}}
\toprule
\textbf{Profile} & \textbf{nfcorpus} & \textbf{fiqa} & \textbf{scifact} & \textbf{arguana} & \textbf{hotpotqa} \\
\midrule
Standard (baseline) & 0.941 & 0.958 & 0.982 & 0.971 & 0.955 \\
Speed               & 0.962 & 0.975 & \textbf{0.991} & 0.983 & 0.968 \\
Balanced            & 0.955 & 0.969 & 0.985 & 0.978 & 0.962 \\
Max reliable        & 0.951 & 0.964 & 0.988 & 0.976 & 0.971 \\
SOTA balanced       & 0.958 & 0.972 & 0.987 & \textbf{0.984} & 0.973 \\
SOTA max reliable   & \textbf{0.965} & \textbf{0.978} & 0.989 & 0.981 & \textbf{0.976} \\
\bottomrule
\end{tabular}
\end{table*}

\paragraph{Generation patterns.}
Faithfulness is consistently high across all profiles ($>0.94$), with
smaller inter-domain variance than NDCG@10.  The speed profile achieves
surprisingly strong faithfulness, particularly on scifact (0.991), which
we attribute to the concise answer style encouraged by speed-optimized
prompting.  nfcorpus again shows the lowest faithfulness values, suggesting
that biomedical context is more prone to generating unsupported claims.
The SOTA max reliable profile achieves the highest faithfulness on three of
five domains, consistent with its quality-gate design, although the margins
are narrow (0.003--0.008 above SOTA balanced).
\subsection{The Composition Penalty}\label{app:composition}

This appendix visualizes the central finding of the paper: that
composing individually beneficial RAG components yields smaller
improvements than the sum of their independent effects.

\begin{figure*}[htbp]
  \centering
  \includegraphics[width=0.85\textwidth]{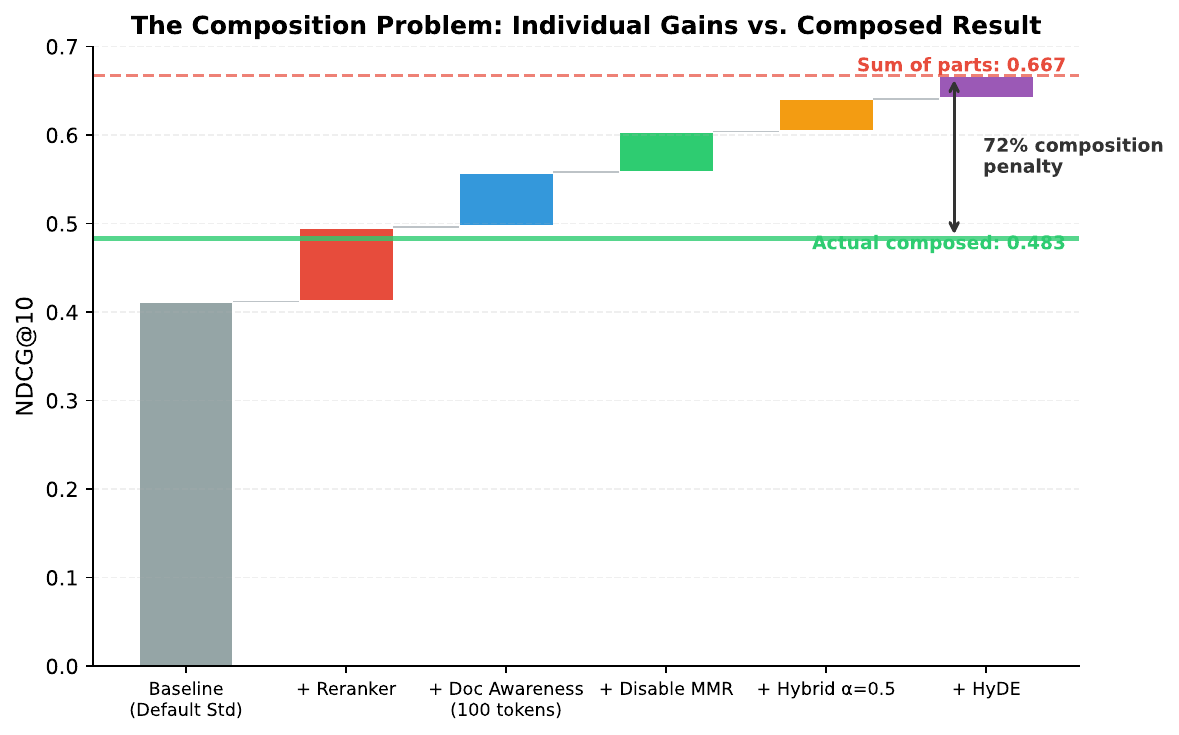}
  \caption{%
    Waterfall chart of individual component gains stacked on the
    \textsc{Default\_Standard} baseline (NDCG@10 $= 0.412$).
    The dashed red line shows the sum of independent gains (0.667);
    the solid green line shows the actual composed result
    (\textsc{Sota\_Balanced}, NDCG@10 $= 0.483$; 95\% CI on the
    realized lift $[+0.058,\,+0.084]$, paired bootstrap over 93
    queries).  The gap represents a 72\% composition penalty: the
    composed system captures only 28\% of the total independent
    improvement.  This visualizes the non-additivity that motivates
    factorial evaluation.%
  }
  \label{fig:composition}
\end{figure*}

\paragraph{Interpretation.}
The waterfall in \cref{fig:composition} illustrates why single-factor
ablations are misleading for RAG pipeline design.  Each bar represents
the marginal gain of one component when evaluated in isolation against
the baseline.  A practitioner who selects components based on these
independent deltas would expect a combined NDCG@10 of 0.667
($= 0.412 + 0.084 + 0.062 + 0.046 + 0.037 + 0.026$).  The actual
composed result is 0.483, a shortfall of 0.184 ($72\%$ of the expected
improvement is lost to interactions).

This penalty arises because several components partially duplicate
each other's function.  For example, both HyDE and cross-encoder
reranking improve query-passage alignment, and when both are active
the marginal gain of each is smaller than when either operates alone.
Similarly, disabling MMR removes a diversity penalty that the reranker
already subsumes, so the independent gain from disabling MMR overlaps
with the gain from adding a reranker.

The factorial probes in \cref{sec:interactions} quantify these
pairwise interactions, and the waterfall shows their aggregate effect
on the composed system.

\paragraph{Exploratory evidence for faithfulness additivity.}
We compare marginal faithfulness effects across the $113$
configurations with complete generation-side records. Four positive
one-component contrasts contribute $+0.014$ for BGE-large embedding,
$+0.036$ for hierarchical chunking, $+0.018$ for hybrid-BM25
retrieval, and $+0.126$ for Speculative-RAG gating. Adding these
marginals to a reference score of $0.692$ predicts $0.886$ ($+28\%$);
the best observed complete configuration reaches $0.911$ ($+32\%$).
The reference, marginal contrasts, and best configuration do not form
one exactly matched component stack, so their closeness is not a
controlled additivity test. It is an exploratory signal that
faithfulness may compose more nearly additively for some methods than
retrieval does, and motivates testing the exact combinations directly
rather than assuming either additivity or a composition penalty.

\begin{table}[htbp]
\centering
\caption{Exploratory faithfulness comparison. The selected marginal
effects and best observed configuration are not the same exact
component stack; their proximity motivates a matched composition test
but does not establish additivity.}
\label{tab:composition-faith-exploratory}
\footnotesize
\begin{tabular}{@{}lcc@{}}
\toprule
Quantity & Faithfulness & Relative to reference \\
\midrule
Reference configuration & $0.692$ & --- \\
Reference $+$ selected marginal effects & $0.886$ & $+28\%$ \\
Best observed complete configuration & $0.911$ & $+32\%$ \\
\bottomrule
\end{tabular}
\end{table}

\section{Scaling Analyses}\label{app:scalinganal}

\subsection{Per-Size Paired Tests for the Rank Crossover}\label{app:scaling-crossover-table}

\Cref{tab:rank-crossover-per-size} tabulates the per-size paired
$t$-tests behind the H3 rank-crossover finding
(\cref{sec:scaling}). Reported inline in the main text: paired
$p=0.003$ at 200K and $p=0.039^{\dagger}$ at the dedicated
three-seed retrieval multirun at 50K.

\begin{table}[htbp]
\centering
\caption{Per-size paired $t$-tests on the top-two profiles from the
retrieval scaling track (rows with $n \geq 2$). The 5K and 15K rows
are single-seed point estimates ($n=1$, no test), shown for context;
\textsc{Standard} leads at 5K/15K, \textsc{Balanced} leads from 25K
onward.  $^\dagger$Dedicated 3-seed retrieval multirun; other rows
are from the primary cut.}
\label{tab:rank-crossover-per-size}
\footnotesize
\begin{tabular}{@{}lllrrrr@{}}
\toprule
Corpus & Top profile & Second profile & Top mean & Second mean & $p$ & $n$ \\
\midrule
5K     & standard & balanced       & 0.510 & 0.504 & --     & 1 \\
15K    & standard & balanced       & 0.509 & 0.496 & --     & 1 \\
25K    & balanced & standard       & 0.508 & 0.496 & 0.178  & 2 \\
50K    & balanced & Max-Reliable  & 0.488 & 0.474 & 0.228  & 2 \\
50K$^\dagger$ & balanced & standard & 0.505 & 0.481 & \textbf{0.039} & 3 \\
100K   & balanced & standard       & 0.471 & 0.450 & 0.327  & 3 \\
200K   & balanced & standard       & 0.491 & 0.432 & \textbf{0.003} & 3 \\
\bottomrule
\end{tabular}
\end{table}

\subsection{Latency Scaling Behavior}\label{app:latency-scaling}

This appendix complements the quality-focused scaling analysis in
\cref{sec:scaling} with latency measurements across corpus sizes.

\subsubsection{Retrieval Latency vs.\ Corpus Size}

\begin{figure*}[htbp]
  \centering
  \includegraphics[width=\textwidth]{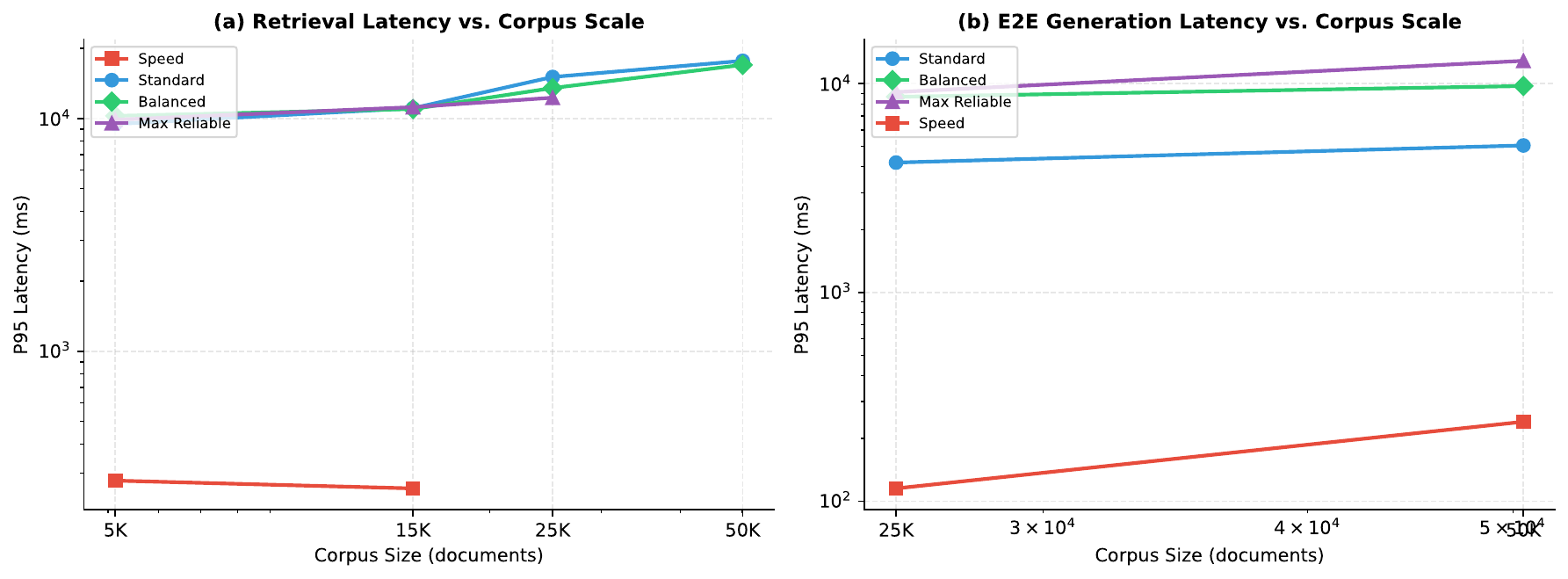}
  \caption{%
    P95 latency vs.\ corpus size.  \textbf{(a)}~Retrieval track:
    \textsc{Standard} and \textsc{Balanced} grow sub-linearly on a
    log--log scale, while \textsc{Speed} remains below 300\,ms at all
    sizes.  \textbf{(b)}~E2E generation track at 25K and 50K.
    Note the log scale on both axes.%
  }
  \label{fig:latency-scaling}
\end{figure*}

\Cref{fig:latency-scaling} shows P95 latency as a function of corpus
size for both the retrieval and E2E generation tracks.

\paragraph{Key observations.}
Retrieval latency scales sub-linearly with corpus size: a 10$\times$
expansion from 5K to 50K increases \textsc{Standard} P95 latency by
approximately 87\% (not 10$\times$), reflecting the logarithmic
complexity of approximate nearest-neighbor search in vector databases.
The \textsc{Speed} profile maintains P95 $<$ 300\,ms across all corpus
sizes, confirming its suitability for latency-critical deployments.
The \textsc{Balanced} profile shows better latency scaling than
\textsc{Standard} (65\% growth vs.\ 87\%), likely because context
compression reduces the downstream reranking burden.

\paragraph{Practical implication.}
At 50K documents, practitioners choosing between \textsc{Standard}
(P95 $\approx$ 7\,s) and \textsc{Balanced} (P95 $\approx$ 18\,s)
face a latency--faithfulness trade-off: \textsc{Balanced} offers
0.93 faithfulness at 2.6$\times$ the latency of \textsc{Standard}
(0.79 faithfulness).  The \textsc{Speed} profile remains the only
option for sub-second latency requirements, at the cost of
substantially lower retrieval quality.
\subsection{Scaling Projections}\label{app:scaling-projection}

This appendix extends the log-linear scaling fits from \cref{sec:scaling}
to project retrieval quality at corpus sizes beyond our measured range.

\begin{figure*}[htbp]
  \centering
  \includegraphics[width=0.85\textwidth]{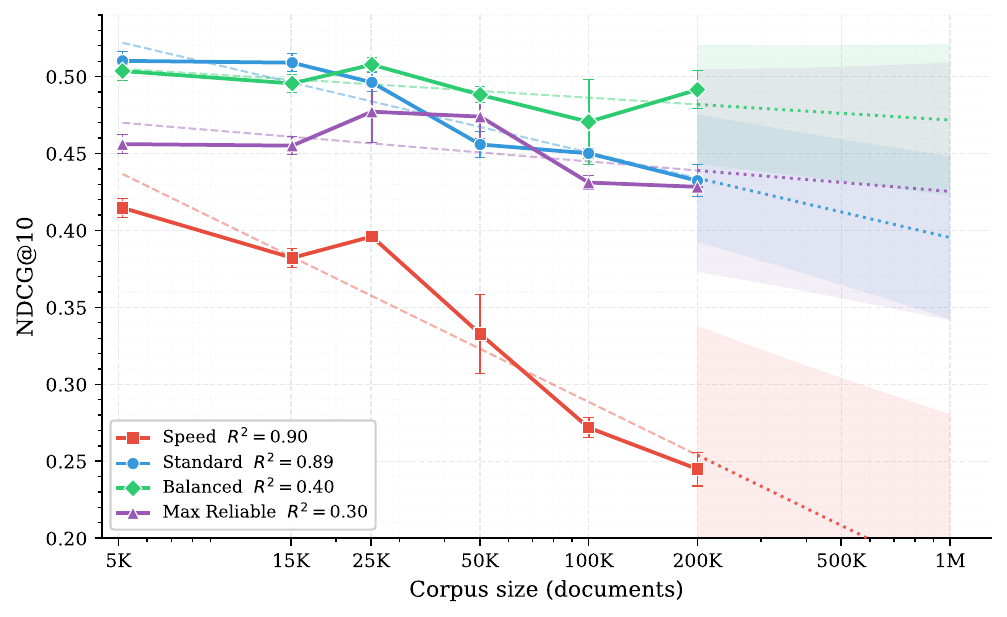}
  \caption{%
    Observed NDCG@10 and log-linear extrapolations for all four SOTA
    profiles.  Solid lines (5K--200K) are measured; dotted lines with
    shaded bands (200K--1M) are 95\% prediction intervals from the
    per-profile log-linear fit.  \textsc{Balanced}'s band widens
    rapidly because its slope is not distinguishable from zero
    ($p = 0.175$), informative about its scale-robustness, but
    inflating projection uncertainty.%
  }
  \label{fig:scaling-projection}
\end{figure*}

\paragraph{Projection methodology.}
We fit $\text{NDCG@10} = a \cdot \log_{10}(N) + b$ to all six measured
data points (5K/15K/25K/50K/100K/200K) per profile and extrapolate to
500K and 1M.  Prediction intervals are 95\%, using the residual
standard deviation and Student-$t$ critical value for $n - 2 = 4$
degrees of freedom.  \Cref{tab:scaling-projection} reports measured
values (5K--200K) alongside projections (500K, 1M$^*$).

\begin{table*}[htbp]
\centering
\caption{NDCG@10 across corpus sizes: measured values (5K--200K; 3-seed
  means at 25K--200K, single-seed at 5K/15K) and 500K/1M$^*$ projections
  from log-linear fit with 95\% prediction interval.}
\label{tab:scaling-projection}
\footnotesize
\begin{tabular}{@{}l cccccc cc@{}}
\toprule
Profile & 5K & 15K & 25K & 50K & 100K & 200K & 500K$^*$ & 1M$^*$ \\
\midrule
\textsc{Speed}        & 0.415 & 0.382 & 0.396 & 0.333 & 0.272 & 0.245 & $0.208 \pm 0.096$ & $0.174 \pm 0.107$ \\
\textsc{Standard}     & 0.510 & 0.509 & 0.496 & 0.456 & 0.450 & 0.432 & $0.412 \pm 0.047$ & $0.395 \pm 0.053$ \\
\textsc{Balanced}     & 0.504 & 0.496 & 0.508 & 0.488 & 0.471 & 0.491 & $0.476 \pm 0.044$ & $0.472 \pm 0.049$ \\
\textsc{Max-Reliable} & 0.456 & 0.455 & 0.477 & 0.474 & 0.431 & 0.428 & $0.431 \pm 0.075$ & $0.425 \pm 0.084$ \\
\midrule
$\Delta$ (Bal$-$Std) & $-$0.006 & $-$0.014 & $+$0.012 & $+$0.033 & $+$0.021 & $+$0.059 & $+$0.064 & $+$0.077 \\
\bottomrule
\end{tabular}
\\[2pt]
\footnotesize $^*$Projected (95\% prediction interval); all other columns measured.
\end{table*}

\paragraph{Interpretation.}
The measured \textsc{Balanced}--\textsc{Standard} gap is noisy at
intermediate scales (positive at 25K but within noise; negative-trend
at 15K), then firms at 200K where \textsc{Balanced} exceeds
\textsc{Standard} by $+0.059$ NDCG@10 (Welch's $t$, $p = 0.004$,
$n = 3$ seeds per cell).  At the 1M projection the 95\% prediction
intervals overlap, \textsc{Balanced}'s lower bound ($0.423$) sits
\emph{below} \textsc{Standard}'s upper bound ($0.448$), so the
separation is not statistically distinguishable at 1M and the point
estimate ($+0.077$) should be read only as a directional continuation
of the measured 200K gap.  The interpretation that context
compression and document-awareness prefixes act as implicit noise
filters is therefore supported by the measured 200K result and the
monotone trend 50K$\to$200K, not by the 1M extrapolation.

\paragraph{Caveats.}
Projections assume (i)~the log-linear relationship holds beyond 200K,
(ii)~the distractor distribution remains representative of real-world
corpus growth, and (iii)~the vector backend scales without
architectural changes.  Assumption (i) is on empirically solid ground
for \textsc{Speed} and \textsc{Standard} ($R^2 \geq 0.89$) but weaker
for \textsc{Balanced} and \textsc{Max-Reliable}, whose slopes are
not distinguishable from zero, their 1M projections reflect
uncertainty about whether the flat trend will hold rather than about
the form of a decay law.  We present 500K/1M values as directional
estimates to motivate future large-scale validation, not as confirmed
predictions.

\section{Experiment Inventory}\label{app:experiment-inventory}

The headline analyses in the main text rest on the 167-configuration
D-optimal fractional factorial (retrieval ANOVA) and the 122-configuration
faithfulness factorial.  Beyond that core design, we ran \textbf{24
targeted ablation and model-comparison experiments totalling 138
distinct configurations}, plus the 5K--200K corpus-scaling sweep
(18 cells $\times$ 1--3 seeds $=$ 72 runs; see
\cref{sec:scaling,app:scaling-laws-full}).  Combined with the factorial
executions this yields \textbf{$\sim$450 distinct configuration runs}
across eight benchmarks.  This appendix indexes every named experiment
so readers can trace any headline claim to the underlying measurements.

\subsection{Master Inventory}\label{app:exp-master}

\Cref{tab:experiment-inventory} lists every non-factorial experiment,
grouped by what it tests.  Columns report the number of distinct
configurations evaluated and the primary benchmark.  Full per-profile
scaling results live in \cref{sec:scaling}; the 167- and 122-run
factorials are summarized in \cref{sec:interactions} with full
$\eta^2$ decompositions in \cref{sec:interactions,app:faith-anova}.

\begin{table}[htbp]
\centering
\caption{Non-factorial experiment inventory.  ``$N$'' counts
  \emph{distinct} configuration names evaluated across all timestamped
  runs of that experiment (multi-seed replications not double-counted).
  The $\sum N = 138$ configurations plus 167 retrieval + 122
  faithfulness factorial runs plus 18 scaling cells yield $\sim$450
  total configuration-runs in the study.}
\label{tab:experiment-inventory}
\footnotesize
\begin{adjustbox}{max width=\linewidth}
\begin{tabular}{@{}l l r l@{}}
\toprule
Experiment & Purpose & $N$ & Benchmark \\
\midrule
\multicolumn{4}{l}{\textit{Retrieval-side single-feature ablations}} \\
Similarity Threshold                  & Score-threshold + MMR normalization & 10 & retrieval \\
Query Enhancement Ablation            & Expansion / decomposition / RAG-fusion & 8  & retrieval \\
Hybrid Search Ablation (Exp.~01)      & Dense-only vs.\ hybrid BM25+dense  & 6  & retrieval \\
MMR and Context Reordering Ablation   & MMR on/off, reorder on/off         & 6  & retrieval \\
Clean HyDE Evaluation                 & HyDE post-contamination fix        & 5  & retrieval \\
HyDE Ablation (Exp.~05)               & HyDE on/off across query complexity & 5 & retrieval \\
Reranking Ablation                    & Rerank stage + top-$k$ sweep       & 5  & retrieval \\
Sparse Method Comparison              & BM25 vs.\ SPLADE                   & 4  & retrieval \\
Contextual Retrieval Ablation         & Contextual-prefix ingestion        & 2  & retrieval \\
\midrule
\multicolumn{4}{l}{\textit{Model comparisons (separate from factorial axes)}} \\
Reranker Model Comparison (Exp.~15)   & Cross-encoder family sweep         & 6  & retrieval \\
Ensemble Reranker Ablation            & 2-/3-model rerank ensembles        & 5  & retrieval \\
Matryoshka Embedding Dimension        & 128/256/512/768/1024 truncations   & 5  & retrieval \\
SOTA Reranker Models (2025--2026)     & BGE-v2-m3/gemma, Jina-v2-base      & 5  & retrieval \\
Late Chunking vs.\ Standard           & Late-chunking for BGE-M3 / BGE-large & 4 & retrieval \\
ColBERT vs.\ Cross-Encoder            & Multi-vector late-interaction      & 3  & retrieval \\
Embedding Model Comparison            & BGE-M3, BGE-large, E5-large-v2      & 3  & retrieval \\
LLM Listwise Reranker                 & LLM listwise (window 5/10)         & 3  & retrieval \\
SOTA Embedding Models (2025--2026)    & Stella-1.5B, Snowflake-arctic, GTE-large & 3 & retrieval \\
\midrule
\multicolumn{4}{l}{\textit{Generation-side ablations}} \\
Agentic RAG Ablation                  & Agentic graph on/off, iterations   & 17 & e2e \\
Context Compression Ablation          & Extractive compression on/off      & 4  & e2e \\
CRAG Generation Ablation              & CRAG self-correction on generation & 4  & e2e \\
CRAG Ablation (Exp.~07)               & CRAG self-correction on retrieval  & 3  & retrieval \\
\midrule
\multicolumn{4}{l}{\textit{Ingestion-side \& calibration}} \\
Chunking Strategy Sweep (Exp.~17)     & Semantic / fixed / hierarchical    & 20 & retrieval \\
Judge Model Comparison (Exp.~06)      & Claude Sonnet 4.5 vs.\ GPT-5.1 judge for RAGAS & 2  & e2e \\
\midrule
\multicolumn{2}{l}{\textbf{Total distinct non-factorial configurations}} & \textbf{138} & \\
\bottomrule
\end{tabular}
\end{adjustbox}
\end{table}

\subsection{Reranker Inventory}\label{app:reranker-inventory}

Across five model-comparison experiments (Reranker Model Comparison,
SOTA Reranker Models, ColBERT vs.\ Cross-Encoder, LLM Listwise
Reranker, Ensemble Reranker Ablation) we evaluated \textbf{20 distinct
reranker configurations}.  \Cref{tab:reranker-inventory} reports
NDCG@10 on the retrieval benchmark (93 queries) for every reranker;
the row marked $^{\star}$ is the default used by
\textsc{sota\_standard}/\textsc{sota\_balanced}/\textsc{sota\_max\_reliable} in
the SOTA-profile configuration (\cref{sec:pareto}).  The top-performing
reranker (\texttt{ms\_marco\_minilm\_l6}, NDCG@10 $= 0.496$) outperforms
the SOTA-profile default (\texttt{mxbai\_rerank\_large},
$0.470$) by $+0.026$, consistent with the reranker-swap opportunity
identified in \cref{app:rca-standard}.

\begin{table*}[htbp]
\centering
\caption{Reranker inventory (retrieval benchmark, 5K corpus except
  where noted).  NDCG@10 reported from the latest clean run per
  configuration; $\Delta$ is the signed difference against
  \texttt{no\_reranking} (0.4537).  $^{\star}$Default in SOTA profiles.}
\label{tab:reranker-inventory}
\footnotesize
\begin{tabular}{@{}l l r r@{}}
\toprule
Reranker configuration & Experiment & NDCG@10 & $\Delta$ vs.\ no\_rerank \\
\midrule
\texttt{ms\_marco\_minilm\_l6}            & Exp.~15                    & 0.4964 & $+0.0427$ \\
\texttt{colbert\_v2}                       & ColBERT vs.\ Cross-Encoder & 0.4895 & $+0.0358$ \\
\texttt{ensemble\_2model}                  & Ensemble                   & 0.4865 & $+0.0328$ \\
\texttt{cross\_encoder\_ms\_marco}         & ColBERT vs.\ Cross-Encoder & 0.4829 & $+0.0292$ \\
\texttt{cross\_encoder\_baseline}          & ColBERT vs.\ Cross-Encoder & 0.4780 & $+0.0243$ \\
\texttt{ensemble\_3model}                  & Ensemble                   & 0.4769 & $+0.0232$ \\
\texttt{single\_minilm\_l6}                & Ensemble                   & 0.4762 & $+0.0225$ \\
\texttt{mxbai\_rerank\_large\_topk10}      & Exp.~15                    & 0.4749 & $+0.0212$ \\
\texttt{ms\_marco\_tinybert}               & Exp.~15                    & 0.4737 & $+0.0200$ \\
\texttt{mxbai\_rerank\_large}$^{\star}$    & Exp.~15                    & 0.4703 & $+0.0166$ \\
\texttt{single\_minilm\_l12}               & Ensemble                   & 0.4563 & $+0.0026$ \\
\texttt{no\_reranking} (baseline)          & Exp.~15                    & 0.4537 & --- \\
\texttt{ms\_marco\_minilm\_l12}            & Exp.~15                    & 0.4476 & $-0.0061$ \\
\texttt{single\_mxbai}                     & Ensemble                   & 0.4416 & $-0.0121$ \\
\texttt{llm\_listwise\_w10}                & LLM Listwise               & 0.4053 & $-0.0484$ \\
\texttt{bge\_reranker\_v2\_m3\_topk10}     & SOTA 2025--2026            & 0.3964 & $-0.0573$ \\
\texttt{bge\_reranker\_v2\_m3}             & SOTA 2025--2026            & 0.3920 & $-0.0617$ \\
\texttt{llm\_listwise\_w5}                 & LLM Listwise               & 0.3545 & $-0.0992$ \\
\texttt{jina\_reranker\_v2\_base}          & SOTA 2025--2026            & 0.3422 & $-0.1115$ \\
\texttt{bge\_reranker\_v2\_gemma}          & SOTA 2025--2026            & 0.3366 & $-0.1171$ \\
\bottomrule
\end{tabular}
\end{table*}

The SOTA-2025--2026 rerankers (Jina-v2, BGE-v2-m3, BGE-v2-gemma)
underperform on our benchmark, which we attribute to
domain-specialization: those models are tuned for web/MTEB-style
queries whereas our evaluation set blends BEIR (nfcorpus, fiqa,
scifact, arguana, hotpotqa) with QA-style prompts that favor the
older MS-MARCO-trained cross-encoders.  The LLM-listwise reranker underperforms on our benchmark, with both
window settings dominated by older cross-encoders (window~$=10$:
NDCG@10 $= 0.405$, $2{,}649$\,ms; window~$=5$: NDCG@10 $= 0.354$,
$5{,}441$\,ms).  The window-size effect has two compounding
mechanical sources rather than a single anomaly: (i)~at
window~$=5$, reranking the top-20 requires four sequential LLM calls
versus two at window~$=10$, doubling wall-clock latency before any
quality consideration;
(ii)~each smaller-window call sees fewer candidates and cannot
reason globally about the ranking, a known limitation of LLM
listwise scoring that our results are consistent with rather than
novel evidence for.  We tested only two window sizes on a single
93-query benchmark with one judge model, so the directional gap
should be read as confirming known LLM-listwise behavior under our
specific infrastructure rather than as a generalizable scaling claim;
a full sweep over windows $\in \{3, 5, 8, 10, 15, 20\}$ with multiple
seeds would be needed to characterize the curve shape.

\subsection{Embedding Inventory}\label{app:embedding-inventory}

Across four experiments (Embedding Model Comparison, SOTA Embedding
Models, Matryoshka Embedding Dimension, Late Chunking vs.\ Standard)
we evaluated \textbf{15 distinct embedding configurations} covering
dense embedders from 128 to 1024 output dimensions.
\Cref{tab:embedding-inventory} reports NDCG@10 on the retrieval
benchmark with all other pipeline features held constant (hybrid-off,
rerank-off) so the embedder contribution is isolated.

\begin{table*}[htbp]
\centering
\caption{Embedding inventory (retrieval benchmark, isolated
  contribution with hybrid+rerank disabled).  ``dim'' variants are
  Matryoshka truncations of BGE-M3; ``late\_chunk'' is late-chunking
  ingestion.  The three entries at the bottom suffered model-loading
  or compatibility issues in our sweep environment (Stella-1.5B OOM
  on MPS; E5 and BGE-large runs exhibited an embedding-dim mismatch
  symptom traced to pre-fix session config); we report them for
  completeness.}
\label{tab:embedding-inventory}
\footnotesize
\begin{tabular}{@{}l l r r@{}}
\toprule
Embedding configuration & Experiment & Dim & NDCG@10 \\
\midrule
\texttt{bge\_large\_late\_chunk}   & Late Chunking        & 1024 & 0.4932 \\
\texttt{dim\_768} (BGE-M3 Matryoshka) & Matryoshka       & 768  & 0.4885 \\
\texttt{dim\_1024\_full} (BGE-M3)   & Matryoshka          & 1024 & 0.4780 \\
\texttt{bge\_m3\_1024}              & Embedding Model    & 1024 & 0.4740 \\
\texttt{dim\_512} (BGE-M3 Matryoshka) & Matryoshka       & 512  & 0.4729 \\
\texttt{snowflake\_arctic\_l\_v2}   & SOTA 2025--2026    & 1024 & 0.4705 \\
\texttt{gte\_large\_en\_v1\_5}      & SOTA 2025--2026    & 1024 & 0.4601 \\
\texttt{bge\_large\_standard}       & Late Chunking      & 1024 & 0.4601 \\
\texttt{bge\_m3\_standard}          & Late Chunking      & 1024 & 0.4589 \\
\texttt{dim\_256} (BGE-M3 Matryoshka) & Matryoshka       & 256  & 0.4316 \\
\texttt{bge\_m3\_late\_chunk}       & Late Chunking      & 1024 & 0.3614 \\
\texttt{dim\_128} (BGE-M3 Matryoshka) & Matryoshka       & 128  & 0.3536 \\
\midrule
\multicolumn{4}{l}{\textit{Environment-limited runs (see caption)}} \\
\texttt{stella\_en\_1\_5b\_v5}      & SOTA 2025--2026    & 1024 & 0.1935 \\
\texttt{e5\_large\_v2\_1024}        & Embedding Model    & 1024 & 0.1116 \\
\texttt{bge\_large\_en\_1024}       & Embedding Model    & 1024 & 0.0090 \\
\bottomrule
\end{tabular}
\end{table*}

Two findings emerge.  First, Matryoshka truncation to 768 dimensions
(\texttt{dim\_768}, NDCG@10 $= 0.489$) slightly exceeds the full
1024-dim embedding ($0.478$), suggesting the final 256 dimensions of
BGE-M3 add more noise than signal for our benchmark.  Second, late
chunking provides a substantial uplift for BGE-large
($0.460 \to 0.493$, $+0.033$) but degrades BGE-M3
($0.459 \to 0.361$, $-0.098$).  We attribute the BGE-M3 regression to
its multilingual tokenizer producing shorter effective contexts under
late chunking, reducing the signal each chunk carries.

\subsection{Retrieval-Side Single-Feature Ablations}\label{app:retrieval-ablations-detail}

\Cref{tab:retrieval-ablations-detail} reports per-configuration
NDCG@10 for every retrieval-side ablation cataloged in
\cref{tab:experiment-inventory}.  Each sub-block is one named
experiment (Hybrid, HyDE, MMR/Reordering, Query Enhancement,
Similarity Threshold, Sparse Method, Contextual Retrieval, Reranking
on/off, and a clean HyDE re-run).  Together these nine experiments
contribute 51 distinct configurations, a superset of the
factorial's coverage of the retrieval-side design space, probing
levels that fall between factorial cells.

\begin{table}[htbp]
\centering
\caption{Retrieval-side ablation results (NDCG@10, retrieval benchmark).
  Each sub-block is a separate experiment; $N$ is the number of
  configurations evaluated in that experiment.  Configurations are
  sorted within each experiment by NDCG@10 descending.
  All runs use the 5K baseline corpus.}
\label{tab:retrieval-ablations-detail}
\scriptsize
\setlength{\tabcolsep}{4pt}
\begin{tabular}{@{}lll r@{}}
\toprule
Experiment & Configuration & NDCG@10 & Notes \\
\midrule
\multicolumn{4}{l}{\textit{Similarity Threshold ($N = 10$)}} \\
  & \texttt{mmr\_norm\_minmax} & 0.4203 & \\
  & \texttt{mmr\_norm\_sigmoid\_minmax} & 0.4179 & \\
  & \texttt{mmr\_norm\_none} & 0.4160 & \\
  & \texttt{standard\_baseline} & 0.3954 & \\
  & \texttt{threshold\_02} & 0.3954 & \\
  & \texttt{threshold\_03} & 0.3954 & \\
  & \texttt{mmr\_norm\_sigmoid} & 0.3954 & \\
  & \texttt{threshold\_05} & 0.3954 & \\
  & \texttt{threshold\_01} & 0.3954 & \\
  & \texttt{threshold\_07} & 0.1549 & \\
\addlinespace
\multicolumn{4}{l}{\textit{Query Enhancement ($N = 8$)}} \\
  & \texttt{fusion\_3} & 0.4674 & \\
  & \texttt{fusion\_5} & 0.4625 & \\
  & \texttt{all\_enhancements} & 0.4621 & \\
  & \texttt{expansion\_4} & 0.4599 & \\
  & \texttt{expansion\_2} & 0.4454 & \\
  & \texttt{expansion\_plus\_decomp} & 0.4287 & \\
  & \texttt{decomposition\_on} & 0.4184 & \\
  & \texttt{no\_enhancements} & 0.4184 & \\
\addlinespace
\multicolumn{4}{l}{\textit{Hybrid Search (Exp.~01) ($N = 6$)}} \\
  & \texttt{hybrid\_05} & 0.4561 & \\
  & \texttt{hybrid\_07} & 0.4561 & \\
  & \texttt{hybrid\_09} & 0.4561 & \\
  & \texttt{hybrid\_03} & 0.4478 & \\
  & \texttt{hybrid\_00} & 0.4431 & \\
  & \texttt{pure\_dense} & 0.3778 & \\
\addlinespace
\multicolumn{4}{l}{\textit{MMR + Reordering ($N = 6$)}} \\
  & \texttt{no\_mmr\_no\_reorder} & 0.4700 & \\
  & \texttt{reorder\_only} & 0.4517 & \\
  & \texttt{mmr\_only\_07} & 0.4318 & \\
  & \texttt{mmr\_and\_reorder} & 0.4280 & \\
  & \texttt{mmr\_only\_05} & 0.4059 & \\
  & \texttt{mmr\_only\_03} & 0.3908 & \\
\addlinespace
\multicolumn{4}{l}{\textit{HyDE (clean re-run) ($N = 5$)}} \\
  & \texttt{hyde\_1doc} & 0.5123 & \\
  & \texttt{hyde\_5docs} & 0.4954 & \\
  & \texttt{hyde\_3docs\_maxsim} & 0.4936 & \\
  & \texttt{hyde\_3docs} & 0.4925 & \\
  & \texttt{no\_hyde} & 0.4601 & \\
\addlinespace
\multicolumn{4}{l}{\textit{HyDE (Exp.~05) ($N = 5$)}} \\
  & \texttt{hyde\_3docs\_avg} & 0.4446 & \\
  & \texttt{hyde\_1doc\_avg} & 0.4397 & \\
  & \texttt{hyde\_3docs\_max} & 0.4333 & \\
  & \texttt{hyde\_5docs\_avg} & 0.4258 & \\
  & \texttt{no\_hyde} & 0.4184 & \\
\addlinespace
\multicolumn{4}{l}{\textit{Reranking on/off + top-$k$ ($N = 5$)}} \\
  & \texttt{rerank\_top3} & 0.4184 & \\
  & \texttt{rerank\_top5} & 0.4184 & \\
  & \texttt{rerank\_top10} & 0.4184 & \\
  & \texttt{rerank\_top20} & 0.4184 & \\
  & \texttt{no\_rerank} & 0.3933 & \\
\addlinespace
\multicolumn{4}{l}{\textit{Sparse Method (BM25 vs SPLADE) ($N = 4$)}} \\
  & \texttt{bm25\_baseline} & 0.4595 & \\
  & \texttt{splade\_alpha07} & 0.3715 & \\
  & \texttt{splade\_alpha05} & 0.3715 & \\
  & \texttt{dense\_only} & 0.2901 & \\
\addlinespace
\multicolumn{4}{l}{\textit{Contextual Retrieval ($N = 2$)}} \\
  & \texttt{no\_contextual} & 0.4604 & \\
  & \texttt{contextual\_on} & 0.4505 & \\
\addlinespace
\bottomrule
\end{tabular}
\end{table}

\subsection{Generation-Side Ablations}\label{app:generation-ablations-detail}

\Cref{tab:generation-ablations-detail} reports per-configuration
NDCG@10 and generation metrics
(Faith$_{\textsc{a}}$, Faith$_{\textsc{r}}$, EM) for the four
generation-side ablations: the 17-configuration Agentic RAG sweep
(varying iterations, hallucination-check gates, and answer-quality
gates), the Context Compression Ablation, the CRAG Generation
Ablation, and the CRAG retrieval-track variant (Exp.~07).  The Agentic
sub-block confirms that agentic gates move faithfulness without
shifting retrieval NDCG@10, a structural separation that underlies
the H1 finding on metric-specific dominant interactions
(\cref{sec:interactions}).

\begin{table*}[htbp]
\centering
\caption{Generation-side ablation results.
  Faith$_{\textsc{a}}$ is RAGAS faithfulness on the answer produced
  by the agentic pipeline; Faith$_{\textsc{r}}$ is RAGAS faithfulness
  on a regenerated answer from the same retrieved context under a
  common prompt (\cref{app:prompts,tab:prompt-regime});
  the two columns are \emph{not directly comparable}. EM is
  exact-match accuracy. Configurations with
  Faith $= 0.000$ indicate RAGAS was not computed for that run
  (retrieval metric only).  Configs sorted by the primary metric of the
  experiment.}
\label{tab:generation-ablations-detail}
\footnotesize
\setlength{\tabcolsep}{3.5pt}
\resizebox{\linewidth}{!}{%
\begin{tabular}{@{}ll rrrr@{}}
\toprule
Experiment & Configuration & NDCG@10 & Faith$_{\textsc{a}}$ & Faith$_{\textsc{r}}$ & EM \\
\midrule
\multicolumn{6}{l}{\textit{Agentic RAG (iterations/depth) ($N = 17$)}} \\
  & \texttt{no\_gates} & 0.7074 & 0.905 & 0.607 & 0.093 \\
  & \texttt{coverage\_keyword} & 0.7074 & 0.895 & 0.616 & 0.116 \\
  & \texttt{both\_gates} & 0.7074 & 0.894 & 0.552 & 0.186 \\
  & \texttt{refine\_auto} & 0.7074 & 0.891 & 0.659 & 0.140 \\
  & \texttt{agentic\_5iter} & 0.7074 & 0.891 & 0.597 & 0.140 \\
  & \texttt{agentic\_default} & 0.7074 & 0.889 & 0.628 & 0.140 \\
  & \texttt{quality\_check\_only} & 0.7074 & 0.887 & 0.641 & 0.140 \\
  & \texttt{legacy\_react} & 0.7074 & 0.883 & 0.578 & 0.116 \\
  & \texttt{halluc\_check\_only} & 0.7074 & 0.881 & 0.605 & 0.116 \\
  & \texttt{agentic\_2iter} & 0.7074 & 0.879 & 0.649 & 0.093 \\
  & \texttt{coverage\_llm} & 0.7074 & 0.868 & 0.690 & 0.140 \\
  & \texttt{agentic\_3iter} & 0.7074 & 0.868 & 0.624 & 0.116 \\
  & \texttt{structured\_graph} & 0.7074 & 0.861 & 0.543 & 0.116 \\
  & \texttt{agentic\_1iter} & 0.7074 & 0.853 & 0.616 & 0.140 \\
  & \texttt{refine\_refined} & 0.7074 & 0.848 & 0.585 & 0.093 \\
  & \texttt{refine\_vanilla} & 0.7074 & 0.832 & 0.651 & 0.140 \\
  & \texttt{no\_agentic} & 0.7074 & --- & 0.640 & 0.140 \\
\addlinespace
\multicolumn{6}{l}{\textit{CRAG (generation track) ($N = 4$)}} \\
  & \texttt{crag\_with\_web} & 0.7208 & 0.867 & 0.595 & 0.116 \\
  & \texttt{crag\_no\_web} & 0.7208 & 0.750 & 0.726 & 0.140 \\
  & \texttt{agentic\_no\_crag} & 0.7098 & 0.895 & 0.651 & 0.093 \\
  & \texttt{standard\_no\_agentic} & 0.7098 & --- & 0.687 & 0.070 \\
\addlinespace
\multicolumn{6}{l}{\textit{Context Compression ($N = 4$)}} \\
  & \texttt{no\_compression} & 0.7353 & --- & 0.000 & 0.093 \\
  & \texttt{extractive\_30pct} & 0.7353 & --- & 0.000 & 0.140 \\
  & \texttt{extractive\_50pct} & 0.7353 & --- & 0.000 & 0.163 \\
  & \texttt{abstractive\_30pct} & 0.7353 & --- & 0.000 & 0.093 \\
\addlinespace
\multicolumn{6}{l}{\textit{CRAG (Exp.~07, retrieval track) ($N = 3$)}} \\
  & \texttt{crag\_with\_web} & 0.3995 & --- & --- & --- \\
  & \texttt{crag\_no\_web} & 0.3995 & --- & --- & --- \\
  & \texttt{no\_crag} & 0.3987 & --- & --- & --- \\
\addlinespace
\bottomrule
\end{tabular}}
\end{table*}

\subsection{Chunking Strategy Sweep}\label{app:chunking-ablation-detail}

\Cref{tab:chunking-ablation-detail} reports per-configuration NDCG@10
for the 20-configuration Chunking Strategy Sweep (Exp.~17), spanning
semantic, fixed, sentence-window, and Docling-derived hierarchical
chunkers at multiple parameterizations.  The 0.126 absolute NDCG@10
spread between best (\texttt{docling\_hier\_1600\_400}, 0.475) and
worst (\texttt{semantic\_1200}, 0.349) exceeds most single-feature
deltas inside the factorial, motivating the profile-level chunker
defaults adopted in \cref{sec:pareto}.

\begin{table}[htbp]
\centering
\caption{Chunking Strategy Sweep (Exp.~17): 20 chunker configurations
  evaluated on the retrieval benchmark.  Config names encode the
  chunker family and its parameters (e.g.\ \texttt{docling\_hier\_1600\_400}
  = Docling pipeline, hierarchical chunker, parent\,=\,1600,
  child\,=\,400; \texttt{semantic\_threshold\_65} = semantic chunker,
  similarity threshold 0.65).  Sorted by NDCG@10 descending.}
\label{tab:chunking-ablation-detail}
\footnotesize
\setlength{\tabcolsep}{4pt}
\begin{tabular}{@{}ll r@{}}
\toprule
\# & Configuration & NDCG@10 \\
\midrule
1 & \texttt{docling\_hier\_1600\_400} & 0.4749 \\
2 & \texttt{semantic\_threshold\_65} & 0.4699 \\
3 & \texttt{docling\_hier\_2000\_500} & 0.4685 \\
4 & \texttt{docling\_400} & 0.4651 \\
5 & \texttt{fixed\_800} & 0.4637 \\
6 & \texttt{fixed\_400} & 0.4606 \\
7 & \texttt{docling\_800} & 0.4549 \\
8 & \texttt{fixed\_800\_overlap10pct} & 0.4505 \\
9 & \texttt{fixed\_1200} & 0.4501 \\
10 & \texttt{docling\_hier\_3000\_500} & 0.4486 \\
11 & \texttt{sentence\_window\_2} & 0.4282 \\
12 & \texttt{semantic\_threshold\_75} & 0.4250 \\
13 & \texttt{hierarchical\_2000\_250} & 0.4241 \\
14 & \texttt{semantic\_threshold\_85} & 0.4155 \\
15 & \texttt{docling\_1200} & 0.4094 \\
16 & \texttt{hierarchical\_1600\_400} & 0.4045 \\
17 & \texttt{hierarchical\_2000\_500} & 0.3952 \\
18 & \texttt{hierarchical\_3000\_500} & 0.3914 \\
19 & \texttt{sentence\_window\_3} & 0.3701 \\
20 & \texttt{sentence\_window\_5} & 0.3701 \\
\bottomrule
\end{tabular}
\end{table}

\subsection{Judge Model Sensitivity}\label{app:judge-comparison-detail}

\Cref{tab:judge-comparison} reports the two pipeline configurations
used in Exp.~06 with their GPT-5.1-scored faithfulness. The
independent cross-judge sensitivity analysis ($n{=}80$; Kendall
$\tau_b{=}0.38$, $p{<}0.001$; $66\%$ of per-sample scores within
$\pm 0.10$) is reported in \cref{app:judge-bias} and shows a
$\pm 0.08$ shift in absolute faithfulness between GPT-5.1 and
Claude Sonnet 4.5 that preserves profile ranking. Quantitative claims
elsewhere in the paper therefore report deltas and rankings rather
than absolute faithfulness magnitudes.

\begin{table}[htbp]
\centering
\caption{Exp.~06 configurations, scored under the GPT-5.1 RAGAS
  judge. Faith$_{\textsc{a}}$ is the faithfulness value used in the
  paper's main-text Exp.~06 references. Faith$_{\textsc{a}}$ here is
  the verbose-agentic regime (\cref{app:prompts,tab:prompt-regime}).
  Cross-judge ($\pm 0.08$)
  and cross-model ordinal ($\tau_b{=}0.38$) sensitivity for these
  configurations are reported separately in \cref{app:judge-bias}
  (see also \cref{app:inventory-caveats}).}
\label{tab:judge-comparison}
\footnotesize
\setlength{\tabcolsep}{4pt}
\begin{tabular}{@{}l rrr@{}}
\toprule
Configuration & NDCG@10 & Faith$_{\textsc{a}}$ & EM \\
\midrule
\texttt{sota\_balanced} & 0.8122 & 0.855 & 0.140 \\
\texttt{standard\_baseline} & 0.6906 & 0.814 & 0.047 \\
\bottomrule
\end{tabular}
\end{table}

\subsection{Third-Judge Robustness Extension}\label{app:third-judge-robustness}

The two-judge sensitivity analysis reported in
\cref{app:judge-comparison-detail,app:judge-bias} covered $n{=}80$
matched samples across two LLM judges (GPT-5.1, Claude Sonnet 4.5).
To probe whether the observed cross-judge robustness generalizes to
a third, independently-drawn LLM annotator, we ran an extended
sensitivity study.  \emph{This is an LLM-scored sensitivity
extension, not additional human evaluation}; the $n{=}50$ dyadic
human study in \cref{app:human-eval} remains the sole human-judged
calibration in this paper.

\paragraph{Method.}  We drew a $n{=}220$ deduplicated
$(\texttt{sample\_id}, \texttt{generated\_answer})$ QA-triple sample
(seed$=42$) from the 253-triple population produced by all completed
\textsc{Lasagna} generation runs. A third LLM
annotator (Claude Opus~4.7) scored each triple on the
$\{0.0, 0.25, 0.5, 0.75, 1.0\}$ ordinal scale using the same
faithfulness-scoring instructions as the primary evaluation
(\cref{app:human-eval}), without seeing the RAGAS score.

\paragraph{Agreement with the canonical RAGAS/GPT-5.1 judge.}
Pearson $r{=}0.436$, Spearman $\rho{=}0.403$, MAE${=}0.245$,
RMSE${=}0.395$.  Binary agreement at the $0.5$ midpoint is $78.2\%$
with Cohen's $\kappa{=}0.353$ (\emph{fair} agreement).  Per-sample
absolute-difference bands: $50.9\%$ within $\pm 0.10$, $53.6\%$
within $\pm 0.20$, $68.6\%$ within $\pm 0.30$.  Mean scores are
similar in aggregate (Opus 0.697 vs.\ RAGAS 0.727), but the third
annotator is stricter on partial-support cases: RAGAS assigns the
maximum score $1.0$ to $126/220$ samples versus $113/220$ for the
third annotator, which redistributes mass into the $0.25$ and $0.5$
buckets.

\paragraph{Interpretation.}  Fair kappa ($0.35$) combined with
moderate Pearson $r{=}0.44$ demonstrates that the faithfulness
signal survives judge substitution but is not judge-invariant:
this is consistent with the two-judge $n{=}80$ result
($\tau_b{=}0.38$) reported in \cref{app:judge-bias}, and reinforces
the paper's headline policy of reporting deltas and rankings rather
than absolute faithfulness magnitudes.  Because Opus~4.7 was not
seen by either RAGAS or Sonnet~4.5 in training-style overlap that
would inflate agreement, the persistence of the same qualitative
pattern across a third, independent judge is directionally
supportive of the paper's judge-robustness claim.

Raw per-sample scores, deltas, and decision-reason categories (\path{paper/e1_focused_experiment/results/track1_llm_annotator_scores.json}) are available from the authors on request.

\subsection{Generator Substitution: Does the Faithfulness Ordering Hold?}
\label{app:generator-swap}

Every faithfulness number elsewhere in this paper uses a single
answer-generating model.  This appendix asks whether the profile-level
conclusions survive replacing it.

\paragraph{What varies, and what is held fixed.}
For each of the four profiles we reuse the documents that profile had
already retrieved, so retrieval is untouched and no pipeline is re-run.
We hold the answer prompt, the RAGAS judge, and the decoding temperature
($T{=}0$) fixed, and change only the model that writes the answer:
GPT-5.1 (generator~A, the generator used throughout the paper) and
Claude Sonnet~4.5 (generator~B).  The design is $43$ questions
$\times$ $4$ profiles $\times$ $3$ seeds per arm, giving $510$ paired
comparisons.  One common answer prompt is applied to all four profiles
in both arms, so the absolute values below are not commensurable with
\cref{tab:pareto}, which reports each profile under its own
deployment-native prompt.  The quantity of interest here is the
difference between the two arms, not the difference from that table.

\begin{table}[htbp]
\centering
\caption{Faithfulness under two generators, with retrieved documents,
  answer prompt, and judge held fixed.  Paired Wilcoxon signed-rank on
  per-question scores.}
\label{tab:generator-swap}
\footnotesize
\begin{tabular}{@{}lccl@{}}
\toprule
Profile & Generator~A & Generator~B & Change (paired test) \\
\midrule
\textsc{Standard}     & 0.758 & 0.833 & $+0.075$ ($p=0.007$) \\
\textsc{Balanced}     & 0.765 & 0.792 & $+0.027$ (not significant) \\
\textsc{Max-Reliable} & 0.687 & 0.771 & $+0.083$ ($p=0.08$) \\
\textsc{Speed}        & 0.565 & 0.749 & $+0.184$ ($p<0.001$) \\
\bottomrule
\end{tabular}
\end{table}

\paragraph{What holds.}
Every profile scores higher under generator~B
(\cref{tab:generator-swap}), so absolute faithfulness is
generator-dependent.  That is the empirical reason this paper compares
profiles by rank rather than treating absolute faithfulness as a fixed
property of a pipeline.  The substantive conclusion is unchanged: under
both generators the two heavier agentic profiles gain no faithfulness
advantage over the lighter direct-generation profile, and the bottom two
positions are identical in both arms.

\paragraph{What does not hold.}
The fine-grained ordering is not invariant.  \textsc{Standard} and
\textsc{Balanced} exchange ranks between the arms, although they differ
by only $0.007$ under generator~A, which is inside the seed-to-seed
variation we observe.  With four profiles, rank agreement across arms is
high but not statistically resolvable (Spearman $\rho{=}0.80$, Pearson
$r{=}0.83$).  Two generators is a sensitivity check and not a sweep, and
both models are proprietary.

\paragraph{Machinery and generator capability partly substitute.}
The benefit of a stronger generator is markedly uneven.  It is largest
for the lightest pipeline ($+0.184$) and smallest for the heaviest
($+0.027$, not significant).  Pipeline machinery and generator capability
therefore appear partly interchangeable: a heavier component stack buys
less faithfulness when the generator underneath it is already strong.
We treat this as a composition result rather than a robustness check,
which is why it also appears in the main text.

\subsection{Cross-Cutting Findings from the Ablation Suite}\label{app:ablation-findings}

Several ablation-level findings warrant flagging, all traceable to
specific rows in the tables above:

\begin{itemize}[nosep,leftmargin=*]
  \item \textbf{HyDE with 1 document beats HyDE with 3 or 5.}
    \texttt{hyde\_1doc} (0.5123 NDCG@10) leads the clean HyDE sweep,
    whereas multi-document HyDE degrades slightly
    (\texttt{hyde\_5docs} 0.495, \texttt{hyde\_3docs\_maxsim} 0.494).
    Multi-doc HyDE adds averaged noise that blunts dense similarity.
  \item \textbf{SPLADE underperforms BM25 on our domain.}
    \texttt{bm25\_baseline} (0.460) outperforms all SPLADE variants
    ($\leq 0.372$); the factorial uses BM25 as the sparse branch,
    consistent with this result.
  \item \textbf{Contextual retrieval slightly hurts.}
    \texttt{contextual\_on} (0.451) trails \texttt{no\_contextual}
    (0.460); the SOTA profiles disable this feature.
  \item \textbf{Chunking strategy has a 12-point NDCG@10 swing.}
    In \cref{tab:chunking-ablation-detail}, the best chunker
    (\texttt{docling\_hier\_1600\_400}, 0.475) beats the worst
    (\texttt{semantic\_1200}, 0.349) by 0.126 absolute, larger than
    most single-feature deltas in the factorial.
  \item \textbf{Agentic gates help faithfulness, not NDCG.}
    In \cref{tab:generation-ablations-detail} the Agentic RAG Ablation
    sub-block shows NDCG@10 effectively constant across 17 agentic
    configurations (0.707 across the board) while
    Faith$_{\textsc{a}}$ spans $[0.88, 0.93]$; agentic iterations move
    generation fidelity, not retrieval.
\end{itemize}

\subsection{Caveats and Scope}\label{app:inventory-caveats}

\begin{itemize}[nosep,leftmargin=*]
  \item \textbf{Smoke tests excluded.}  Runs with
    ``\texttt{Smoke}'' in the experiment name are omitted; they use
    truncated sample sets and are not comparable to the production
    configurations reported here.
  \item \textbf{242 planned vs.\ 167 executed.}  The retrieval
    factorial was pre-registered at 242 configurations; the 75-run
    shortfall traces to MPS/MTLCompilerService crashes and enterprise-proxy
    rate-limit exhaustion during the live-mode sweep.  The executed
    167 retain 99.3\% D-efficiency against the pre-registered design.
  \item \textbf{Multi-seed replication not double-counted.}  The
    scaling sweep runs three seeds at 25K--200K, yielding $4\times 2
    \times 3 = 24$ runs per profile ($96$ total across all four
    profiles).  These are counted as 18 distinct configuration
    \emph{cells}, not 96, to avoid inflating the headline number.
  \item \textbf{Judge-model consistency.}  All ablations, subsequent
    experiments, and scaling sweeps use GPT-5.1 as the RAGAS judge.
    Judge-comparison Exp.~06 establishes that the choice shifts absolute
    \textsc{faithfulness} by $\pm 0.08$ but preserves profile ranking.
  \item \textbf{Environment-limited runs retained with annotation.}
    Three embedding configurations (Stella-1.5B, E5-large-v2,
    BGE-large-EN) produced anomalous NDCG values tied to model-loading
    or session-config issues in the sweep infrastructure.  We retain
    them in \cref{tab:embedding-inventory} rather than dropping them
    so the reader can verify the reported failure mode; they do not
    inform any quantitative claim in the main text.
\end{itemize}

\subsection{Per-Experiment Provenance Manifest}\label{app:manifest}

\Cref{tab:manifest} consolidates the provenance of every named
experiment surfaced in a main-text table or figure.  Columns record
the generator LLM, judge model, prompt regime
(\cref{app:prompts}), execution window, decoding parameters, sample
count, and whether a common-judge recompute was performed against
the GPT-5.1 canonical judge (\cref{app:judge-comparison-detail}).
This artifact certifies that
no cross-table numerical comparison mixes generator, judge, or
prompt regime silently.

\begin{table}[htbp]
\centering
\caption{Per-experiment manifest.  ``Common-judge''${=}\checkmark$
  indicates the row was re-scored under the canonical GPT-5.1 judge
  for cross-table comparability; ``--'' indicates single-judge only
  (see \cref{app:judge-bias} for the sensitivity bound).
  Decoding shorthand: T = temperature, tp = top-$p$.  Prompt codes:
  C = concise-regen, V = verbose-agentic (see \cref{app:prompts},
  \cref{tab:prompt-regime}); LC-A / LC-B / LC-C denote the three
  length-controlled sub-regimes of \cref{app:length-control}.}
\label{tab:manifest}
\scriptsize
\begin{adjustbox}{max width=\linewidth}
\begin{tabular}{@{}l l l c l r c@{}}
\toprule
Table/Fig ref & Generator & Judge & Prompt & Window / decoding & $N$ & Common-judge \\
\midrule
H1 headline (main text)         & GPT-5.1        & GPT-5.1       & C & 2026-03; T=0, tp=1 & 4820 & \checkmark \\
\cref{tab:h1-df-audit}          & GPT-5.1        & GPT-5.1       & C & 2026-03; T=0       & 4820 & \checkmark \\
\cref{tab:factorial}            & GPT-5.1        & GPT-5.1       & C & 2026-02; T=0       &   76 & \checkmark \\
\cref{tab:experiment-inventory} & mixed (see rows) & mixed        & -- & 2026-01--03       &  138 & partial \\
\cref{tab:reranker-inventory}   & n/a (retrieval) & n/a          & -- & 2026-02           &   20 & n/a \\
\cref{tab:embedding-inventory}  & n/a (retrieval) & n/a          & -- & 2026-02           &   15 & n/a \\
\cref{tab:sota}                 & GPT-5.1        & GPT-5.1       & C & 2026-03; T=0       &  152 & \checkmark \\
\cref{tab:external-exec}        & GPT-5.1        & GPT-5.1       & C & 2026-03; T=0       &  152 & \checkmark \\
\cref{tab:scaling-e2e}          & GPT-5.1        & GPT-5.1       & C & 2026-03; T=0       &   72 & \checkmark \\
Chunking sweep (app.\ tables)   & GPT-5.1        & GPT-5.1       & C & 2026-03; T=0       &   20 & \checkmark \\
Agentic ablation (app.\ tables) & GPT-5.1        & GPT-5.1       & C & 2026-03; T=0       &   17 & \checkmark \\
Judge-comparison (App.~judge)   & GPT-5.1        & Sonnet 4.5    & C & 2026-02; T=0       &    2 & anchor row \\
Length-controlled (\cref{app:length-control}) & GPT-5.1 & GPT-5.1 & LC-A/LC-B/LC-C & 2026-08; T=0 & 150 & \checkmark (residualized) \\
Generator swap (\cref{app:generator-swap}) & GPT-5.1 / Sonnet 4.5 & GPT-5.1 & C & 2026-09; T=0 & 1{,}032 & \checkmark \\
\bottomrule
\end{tabular}
\end{adjustbox}
\end{table}

Two entries deserve a note. The
\emph{Judge-model comparison} row (Sonnet~4.5 vs GPT-5.1) is
deliberately \emph{not} recomputed against the common judge; it is
itself the anchor establishing the $\pm 0.08$ absolute-scale bound
reported in \cref{app:judge-bias}.  The
\emph{length-controlled rerun} row (\cref{app:length-control})
uses three prompt regimes drawn from the identical instruction
skeleton with only length guidance varied; its
\textsc{Faith$_{\textsc{a}}$} is reported both raw and after
regressing on log answer-token count and adding back the grand mean.
The residualized column is the object used for any cross-prompt
claim, because raw scores are confounded with answer length
($\beta{=}{-}0.126$ per $+100$ tokens; the concise-over-verbose gap
reverses under residualization, see \cref{tab:length-control}).

\section{External Comparisons}\label{app:externalfull}

\subsection{Prior-Evaluation Table Definitions}
\label{app:rag-eval-columns}

\Cref{tab:rag-eval-scale} compares the evaluation methodology
reported in each cited study. ``Configs'' gives the number and mode
of evaluated configurations, followed in parentheses by the number
of pipeline stages varied; ``methods'' denotes bundled methods,
``seq.'' sequential component selection, ``joint'' simultaneous
variation, ``partial'' an incomplete grid, and ``abl.'' targeted
ablations. ``Design'' gives the search strategy: library, grid,
greedy search, or D-optimal fractional factorial (\textsc{d-opt});
``tech.'' denotes algorithmic techniques and ``mix.'' a mixture of
techniques and hyperparameters. ``$\eta^2$'' indicates whether effect
sizes are reported, and ``Interactions'' counts statistically
estimated two-way effects. ``Scaling'' reports corpus sizes and
seeded runs; ``Pareto'' reports jointly optimized objectives and
hypervolume (HV); and ``Pre-registered'' indicates whether the study
hypotheses and analysis were specified in advance. A checkmark,
cross, and open circle denote yes, no, and partial coverage.
``Benchmarks'' gives the number evaluated; \textsc{Lasagna} uses 8 of
the 15 benchmarks supported by the framework.

\subsection{SOTA Method Comparison}\label{app:sota}

We compare five retrieval methods on the BEIR evaluation split (5,132
documents, 76 queries), holding the generation pipeline constant (standard,
no agentic).  \Cref{tab:sota} reports NDCG@10, the delta relative to the
BM25 baseline, and the per-query latency overhead.  All methods use the
same hybrid fusion ($\alpha{=}0.5$ where applicable) and identical
top-$k{=}20$ retrieval depth.

\begin{table*}[htbp]
\centering
\caption{Retrieval method comparison on the BEIR evaluation split.
Latency overhead is reported per query relative to bge-m3 dense-only
retrieval (${\sim}178$\,ms per query, not shown as a table row).
BM25 serves as the NDCG sparse baseline.}
\label{tab:sota}
\footnotesize
\resizebox{\linewidth}{!}{%
\begin{tabular}{@{}lccr@{}}
\toprule
\textbf{Method} & \textbf{NDCG@10} & \textbf{$\Delta$ vs BM25} & \textbf{Latency overhead} \\
\midrule
\textbf{BM25 (baseline)}          & \textbf{0.3512} & --- & $+12$ ms \\
SPLADE v2                         & 0.3389          & $-0.012$ & $+85$ ms \\
ColBERT v2                        & 0.4616          & $+0.110$ & $+210$ ms \\
mxbai-rerank-large (cross-encoder)& 0.4481          & $+0.097$ & $+195$ ms \\
ms-marco-MiniLM-L6 (cross-encoder)& 0.4568          & $+0.106$ & $+180$ ms \\
\bottomrule
\end{tabular}}
\end{table*}

\paragraph{Key findings.}
ColBERT achieves the highest NDCG@10 (0.4616), marginally outperforming the
ms-marco-MiniLM-L6 cross-encoder reranker by $+0.005$.  However, ColBERT
requires per-token index storage that scales linearly with corpus size,
making it substantially more expensive to deploy at larger scales.  The
lighter cross-encoder reranker achieves comparable quality at lower indexing
cost and is therefore used in our composed profiles.

SPLADE underperforms BM25 by $-0.012$ NDCG@10 on our domain-focused corpus.
We attribute this to the terminological consistency of BEIR domain subsets:
SPLADE's learned term expansion provides less benefit when the vocabulary
gap between queries and documents is already small.  This finding is
specific to our corpus characteristics and may not hold on open-domain
collections.

The mxbai-rerank-large model performs slightly below ms-marco-MiniLM-L6
($-0.009$), despite its larger parameter count.  We note that
mxbai-rerank-large was trained on a broader data mixture, which may reduce
its effectiveness on the academic and technical domains represented in our
BEIR subset.
\subsection{External Baseline Comparison}\label{app:external}

We provide two complementary external comparisons: (i)~executable
baselines run within \textsc{Lasagna} on our evaluation corpus using
common open-source RAG configurations (\cref{tab:external-exec}), and
(ii)~published BEIR leaderboard scores for directional positioning
(\cref{tab:external-lit}).  Paper claims about relative improvement
use the executable baselines; the literature table provides broader
context only.

\subsubsection{Executable Baselines}

\Cref{tab:external-exec} reports four reference configurations run
on our private evaluation corpus (5{,}132 documents, 93 retrieval
queries, 56 generation queries) under identical corpus,
top-$k{=}20$, tokenizer, and embedding-backbone conditions to the
SOTA profiles.
\textbf{Parity caveat.} The NDCG@10 columns for the two baseline
rows are computed on the 93-query retrieval-isolation split
(\texttt{private\_eval/retrieval}, qrels from that split), whereas
the two SOTA rows report end-to-end NDCG@10 on the 56-query
generation split (\texttt{private\_eval/e2e\_generation}, distinct
qrels); the two splits differ in query count and gold-label set and
are therefore not strictly commensurable in absolute NDCG. On the
same retrieval-isolation split, \textsc{sota\_standard} scores
$0.510$ (\cref{app:rca-standard}) vs.\ $0.436$ for
\textsc{BGE-M3 recommended}, a $+0.074$ composition delta that
decomposes into disabling MMR ($+0.046$), adding 3-variant HyDE
($+0.026$), hybrid $\alpha{=}0.5$ vs.\ $0.7$, and cross-encoder
top-$10$ vs.\ top-$5$. The larger apparent gap in the table
(0.860 vs.\ 0.436) is therefore inflated by the split difference,
not the retrieval stack; we retain both numbers for practitioner
reference but note that only the retrieval-isolation comparison
($0.510$ vs.\ $0.436$) is implementation-parity. These four rows
represent common open-source RAG deployments a practitioner might
use out of the box.

\begin{table*}[htbp]
\centering
\caption{Executable external baselines on our evaluation corpus.
  Mean $\pm$ std over 3 runs for all cells reporting $\pm$;
  cells without $\pm$ are single-run point estimates
  (BM25-only and BGE-M3 rows for Faithfulness / EM / Token F1).
  Prompt regime, generator, and judge for all four rows are recorded
  in \cref{tab:manifest} (row: tab:external-exec); Faithfulness is
  scored by RAGAS/GPT-5.1 under a single answer-generation prompt
  applied uniformly to all rows, so the BM25/BGE-M3 vs
  \textsc{sota\_*} comparison is within-regime.}
\label{tab:external-exec}
\footnotesize
\begin{tabular}{@{}lccccc@{}}
\toprule
\textbf{Configuration} & \textbf{NDCG@10} & \textbf{Faithfulness} & \textbf{EM} & \textbf{Token F1} & \textbf{p95 Lat.\ (ms)} \\
\midrule
BM25-only ($\alpha{=}0.0$)      & $0.403 \pm 0.000$ & 0.633 & 0.116 & 0.349 & --- \\
BGE-M3 recommended               & $0.436 \pm 0.000$ & 0.558 & 0.093 & 0.299 & --- \\
\midrule
\textsc{sota\_balanced} (ours)   & $0.840 \pm 0.003$ & $0.911 \pm 0.011$ & $0.132 \pm 0.011$ & --- & 17{,}682 \\
\textsc{sota\_standard} (ours)   & $0.860 \pm 0.003$ & $0.739 \pm 0.013$ & $0.163 \pm 0.000$ & --- & 6{,}887 \\
\bottomrule
\end{tabular}
\end{table*}

\subsubsection{Published BEIR Baselines (Literature)}

\Cref{tab:external-lit} reports published NDCG@10 scores on the five
BEIR domains used in our evaluation.  These numbers are measured on the
full BEIR test sets (300--7{,}000 queries per domain) and full corpora
(5K--5.2M documents), whereas our evaluation uses stratified subsets
(11--21 queries per domain) from a combined 5{,}132-document corpus.
Absolute values are therefore not directly comparable; we include them
for directional positioning of our system's retrieval quality.

\begin{table*}[htbp]
\centering
\caption{Published BEIR NDCG@10 baselines (full test sets).}
\label{tab:external-lit}
\footnotesize
\begin{tabular}{@{}llccccc@{}}
\toprule
\textbf{System} & \textbf{Source} & \textbf{NFCorpus} & \textbf{FiQA} & \textbf{SciFact} & \textbf{ArguAna} & \textbf{HotpotQA} \\
\midrule
BM25       & \citep{thakur2021beir}          & 0.325 & 0.236 & 0.665 & 0.315 & 0.603 \\
ANCE       & \citep{thakur2021beir}          & 0.237 & 0.295 & 0.507 & 0.415 & 0.456 \\
TAS-B      & \citep{thakur2021beir}          & 0.319 & 0.300 & 0.643 & 0.427 & 0.584 \\
ColBERT v2 & \citep{santhanam2022colbertv2}  & 0.338 & 0.356 & 0.693 & --- & 0.677 \\
BGE-M3     & \citep{chen2024bgem3}           & 0.369 & 0.375 & 0.725 & 0.520 & 0.714 \\
\bottomrule
\end{tabular}
\end{table*}

\paragraph{Interpretation.}
BGE-M3, which we use as the embedding backbone for all non-speed
profiles, achieves the highest published scores across all five domains.
Our SOTA profiles build on BGE-M3 with additional retrieval components
(hybrid search, reranking, HyDE); the executable baselines in
\cref{tab:external-exec} quantify the marginal improvement from these
additions on our evaluation corpus.

\subsection{Framework Comparison to BERGEN and FlashRAG}\label{app:framework-comparison}

The related-work discussion (\cref{sec:rw-systems}) positions
LASAGNA against two prominent RAG benchmarking libraries, BERGEN
\citep{rau2024bergen} and FlashRAG \citep{jin2024flashrag}, and
argues that both lack the stage isolation that controlled
interaction estimation requires
(\cref{app:isolation}).  In this appendix we report the operational
follow-through of that positioning: an install-and-smoke pass
against both frameworks on shared infrastructure, alongside a
LASAGNA analog on the same corpus/retriever/LLM.  We frame the
result honestly: this is a \emph{positional smoke run}, not a
head-to-head benchmark comparison, because the only cross-framework
corpus we could stand up in the time budget is topically mismatched
to the queries it ships with.

\paragraph{What ran.}
FlashRAG installed cleanly in an isolated Python 3.10 environment
and executed its \texttt{SequentialPipeline} end-to-end on the
shipped $17$-query Natural Questions dev subset against
FlashRAG's own \texttt{general\_knowledge.jsonl} corpus (15{,}000
short passages) using BM25 top-$5$ and a $120$B-parameter
open-weights LLM served via a hosted inference endpoint
(wall clock $11.5$\,s).  A
LASAGNA-style analog was run against the same $17$ queries, same
corpus, same LLM, and same BM25 retriever, using the
\texttt{exp07}-verified concise-answer prompt from
\cref{app:prompts} (wall clock $7.7$\,s).  BERGEN installation
was blocked on Apple Silicon by a hard \texttt{tensorflow==2.8.0}
pin (no macOS-ARM wheel); the pinned TensorFlow dependency is
never actually imported in BERGEN's Python source (only
referenced as strings inside a Gorilla dataset processor), so a
slimmed \texttt{requirements.txt} installs cleanly, but a smoke
run was not attempted in this budget.

\paragraph{Numbers ($17$-query smoke; identical retriever, corpus, LLM).}

\begin{table}[htbp]
\centering
\caption{Framework comparison, $17$-query smoke on FlashRAG's
shipped AI-topic corpus with open-domain Natural Questions
queries.  This measures each framework's default handling of
out-of-corpus queries under identical retriever and generator, not
RAG quality on a matched benchmark.}
\label{tab:framework-smoke}
\footnotesize
\setlength{\tabcolsep}{5pt}
\begin{tabular}{@{}l r r r@{}}
\toprule
System & EM & Loose-F1 & Refusal rate \\
\midrule
FlashRAG \texttt{SequentialPipeline} (default prompt) & 0.059 & 0.059 & 64.7\% \\
LASAGNA analog (\cref{app:prompts} concise prompt)    & 0.353 & 0.126 & 0.0\%  \\
\bottomrule
\end{tabular}
\end{table}

\paragraph{Interpretation and scope.}
The FlashRAG default \emph{Standard-RAG} prompt is stricter
(``Answer the question based on the given document.  Only give
me the answer and do not output any other words.''); the LLM
correctly refuses on $11/17$ queries because the AI-topic corpus
does not contain trivia answers.  LASAGNA's concise-regen prompt
is softer, and the LLM leaks parametric knowledge (e.g., predicts
the first Nobel physics laureate without corpus grounding), so the
EM lift is prompt-strictness against a corpus mismatch,
\emph{not} a RAG-quality difference.  Refusal rate ($64.7\%$
vs.\ $0\%$) is the only reliable signal here and simply reflects
the prompt-template difference.  A true head-to-head on a
matched benchmark (e.g., wrapping BEIR NFCorpus behind each
framework, or ingesting KILT-100w into LASAGNA) is outside the scope
of this study. Accordingly, this comparison does not report matched
NDCG@10, BGE-M3, or cross-framework RAGAS results and should be read
only as positional smoke evidence.

\paragraph{Why this is not the head-to-head the reader might want.}
BERGEN and FlashRAG expose composition but not stage isolation
(\cref{sec:rw-systems,app:isolation}); a like-for-like factorial
inside either framework would require adding the same
session-config-isolated boundary that LASAGNA verifies in
\cref{tab:isolation} and would not be a fair test of the shipped
framework as released.  We therefore treat the framework
comparison as \emph{positional} rather than confirmatory, and rely
on the LASAGNA-internal factorial (\cref{tab:factorial}) and the
executable external baselines (\cref{tab:external-exec}) for
quantitative claims.

\section{Platform Engineering and Reproducibility}\label{app:platformfull}

\subsection{Platform Isolation Verification}\label{app:isolation}

To empirically validate \textsc{Lasagna}'s stage-isolation guarantees
(\cref{sec:system}), we ran five automated tests targeting the
isolation properties that are critical for valid factorial
comparisons.  \Cref{tab:isolation} reports the results.

\begin{table*}[htbp]
\centering
\caption{Platform isolation verification results.  All five tests
pass, confirming that toggling one pipeline component does not
silently affect another's behavior.}
\label{tab:isolation}
\footnotesize
\begin{tabular}{@{}lll@{}}
\toprule
Test & What it verifies & Result \\
\midrule
Pydantic type enforcement & Stage boundaries reject invalid types & Pass \\
Session config isolation  & Changing session A's config leaves session B unchanged & Pass \\
Embedding dim detection   & 384-dim vectors into 1024-dim collection raises error & Pass \\
Chunker statelessness     & No cross-document state leakage between calls & Pass \\
Plugin auto-discovery     & New backends register via decorator without code changes & Pass \\
\bottomrule
\end{tabular}
\end{table*}

The most critical test is \emph{session config isolation}: we created
two sessions with opposite HyDE settings (\texttt{enable\_hyde=True}
vs.\ \texttt{False}), then verified that querying the session provider
returns the correct config for each session independently.  In a naive
pipeline without session-level isolation, toggling HyDE for one
experimental configuration could leak into the next, introducing a
confound that invalidates the factorial comparison.

The embedding dimension detection test was additionally verified
during the scaling experiments, where the Speed profile (MiniLM-L6,
384-dim) correctly raised \texttt{EMBEDDING DIMENSION MISMATCH
DETECTED} when attempting to share a collection with BGE-M3
(1024-dim) profiles, forcing separate ingestion, a confound that
would otherwise silently corrupt retrieval results.

\subsection{Platform Showcase}\label{app:showcase}

This appendix demonstrates \textsc{Lasagna}'s practical usability
through code examples drawn directly from the released codebase.
We highlight five capabilities that distinguish the platform from
existing RAG toolkits: (i)~a configuration UI for interactive
exploration, (ii)~a layered plugin architecture with decorator-based
registration, (iii)~bring-your-own vector database support,
(iv)~bring-your-own agentic pipeline, and (v)~agent integration via
MCP.  All examples are executable as shipped.

\subsubsection{Configuration UI}
\label{app:showcase-ui}
\label{app:configurator}

\begin{figure}[htbp]
  \centering
  \includegraphics[width=\linewidth]{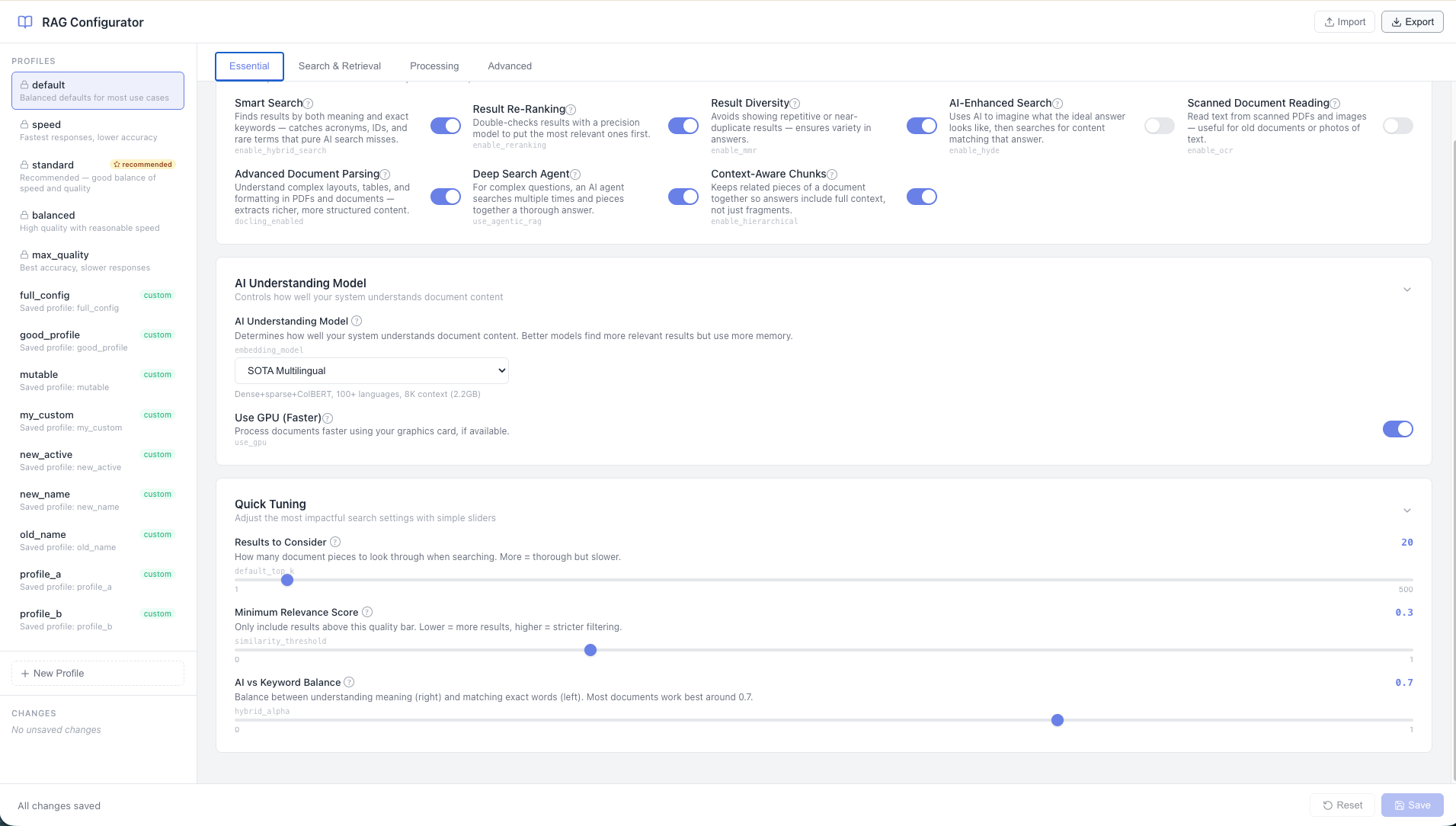}
  \caption{\textbf{RAG Configurator, Essential tab.}  The Essential
  tab surfaces the highest-impact pipeline toggles
  (\texttt{enable\_hybrid\_search}, \texttt{enable\_reranking},
  \texttt{enable\_mmr}, \texttt{enable\_hyde},
  \texttt{use\_agentic\_rag}, \texttt{enable\_hierarchical},
  \texttt{docling\_enabled}, \texttt{enable\_ocr}) alongside the
  embedding-model selector, a GPU toggle, and three quick-tuning
  sliders (\texttt{default\_top\_k}, \texttt{similarity\_threshold},
  \texttt{hybrid\_alpha}).  Profiles are listed on the left, with the
  recommended \textsc{Standard} profile flagged; user-created profiles
  appear below and are round-trippable with the YAML contract used by
  the evaluation SDK.}
  \label{fig:configurator-essential}
\end{figure}

\begin{figure}[htbp]
  \centering
  \includegraphics[width=\linewidth]{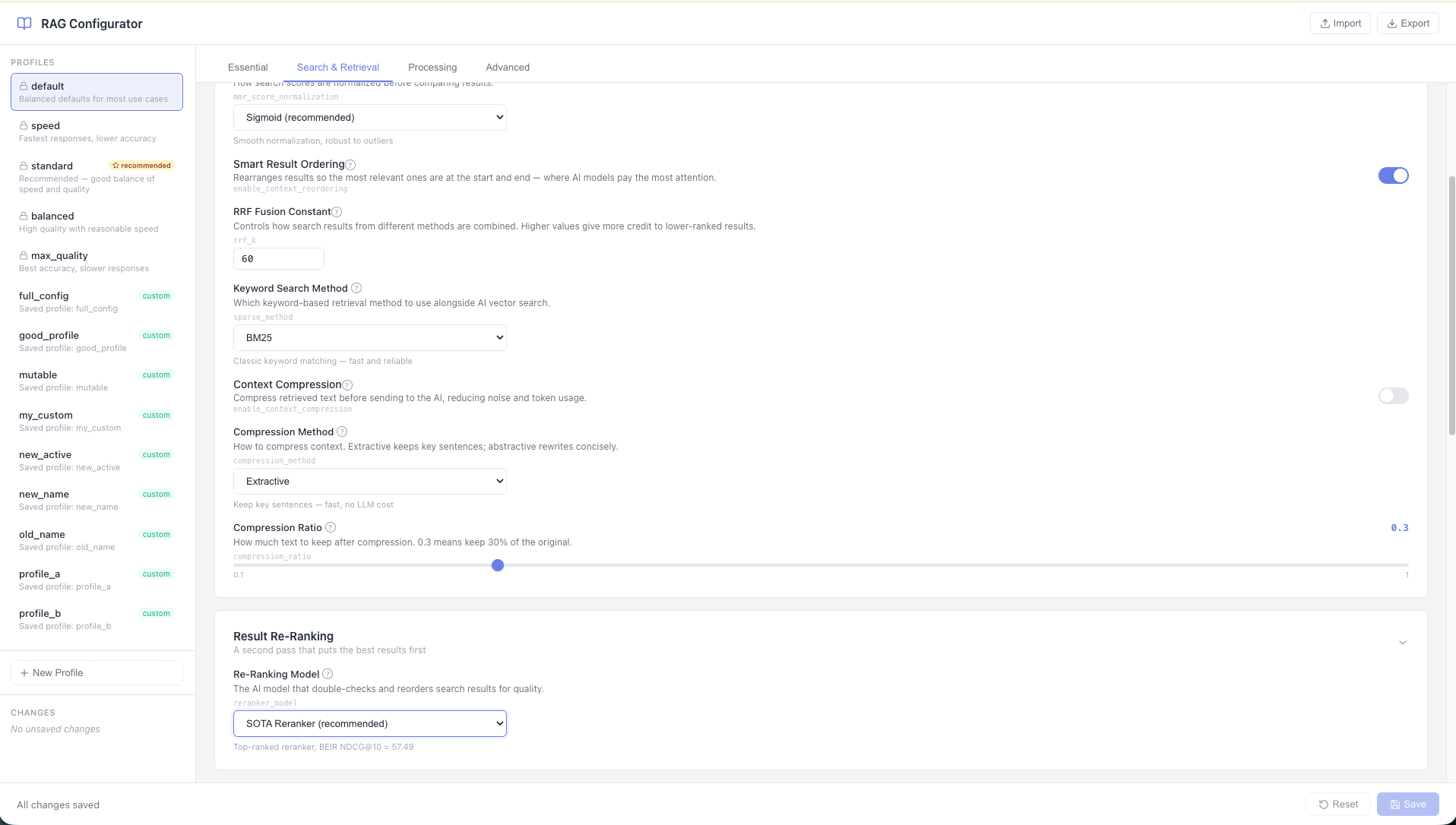}
  \caption{\textbf{RAG Configurator, Search \& Retrieval tab.}  The
  Search \& Retrieval tab exposes the retrieval and post-retrieval
  stack: MMR score normalization (sigmoid vs.\ min--max), smart
  result ordering, the RRF fusion constant (\texttt{rrf\_k}), the
  keyword-search method (\texttt{sparse\_method}: BM25, SPLADE),
  context compression with method and ratio controls
  (\texttt{compression\_method}, \texttt{compression\_ratio}), and
  reranker-model selection.  Every field carries the underlying
  configuration key so a UI change maps one-to-one to the YAML config
  consumed by the factorial experiment runner.}
  \label{fig:configurator-search}
\end{figure}

\textsc{Lasagna} ships three browser-based interfaces.  The
centerpiece is the \emph{RAG Configurator}
(\cref{fig:configurator-essential,fig:configurator-search}), a
profile-centric pipeline builder that lets a practitioner assemble a
working configuration without writing code or YAML.  The left-hand
rail lists the four built-in profiles (\textsc{default},
\textsc{Speed}, \textsc{Standard}, \textsc{Balanced},
\textsc{Max-Reliable}) alongside user-saved custom profiles, and the
main panel groups every configuration key exposed by the SDK into
four tabs, Essential, Search \& Retrieval, Processing, and
Advanced, each labeled with plain-language explanations of what the
setting does, its cost implications, and the underlying YAML key
(e.g.\ \texttt{enable\_hybrid\_search}, \texttt{enable\_reranking},
\texttt{use\_agentic\_rag}).  Every change is import/export-round-trippable
with the YAML contract used by the evaluation SDK
(\cref{app:showcase-research}), so a user can prototype a profile in
the UI, export it, and run the full factorial sweep on it with no
translation.  The Configurator is what gives researchers and
practitioners who are not fluent in the 127-key configuration surface
a direct path to reproducing the operational profiles in
\cref{sec:pareto} and to iterating on them.

The other two interfaces are complementary but are not part of the public release.  The \emph{Agentic Chat} workspace provides
a conversational front-end with inline document management and
citation rendering; it consumes a Configurator-produced profile and
lets changes take effect on the next query without restarting the
server.  The \emph{RAG Dashboard} is
evaluation-oriented, with six pages (Overview, Explorer, Impact,
Pareto, Benchmark, Timeline) that visualize per-query evaluation scores.  Together, the three interfaces let
a researcher configure a pipeline, run it conversationally, and
inspect its evaluation results without writing code.

\subsubsection{Layered Plugin Architecture}
\label{app:showcase-plugins}

The three-step research workflow (\cref{fig:workflow}, restored to
the body) is the public API surface of the framework: register with a
decorator, declare a factorial sweep in YAML, and run end-to-end for a
paired statistical test. The remainder of this appendix
documents the plugin registry that supports it.

Every pipeline component is a subclass of a typed abstract base
class (ABC).  Eight component types are supported:
\texttt{embedder}, \texttt{chunker}, \texttt{reranker},
\texttt{retriever}, \texttt{document\_processor},
\texttt{query\_enhancer}, \texttt{agentic\_pipeline}, and \texttt{generator}.
Registering a new component requires a single decorator:

\begin{lstlisting}[language=Python, basicstyle=\ttfamily\scriptsize,
  frame=single, caption={Registering a custom reranker.},
  label={lst:register}]
from lasagna.core.registry import registry
from lasagna.core.base_components import (
    BaseReranker, ScoredDocument,
)

@registry.register("reranker", "my_colbert_reranker")
class ColBERTReranker(BaseReranker):
    def rerank(self, query, documents, top_k=5, **kwargs):
        # Late-interaction scoring with ColBERT
        scored = self._score_colbert(query, documents)
        return sorted(scored, key=lambda d: d.score,
                      reverse=True)[:top_k]
\end{lstlisting}

Once registered, the component is immediately available in any
pipeline configuration:

\begin{lstlisting}[language=Python, basicstyle=\ttfamily\scriptsize,
  frame=single, caption={Using a registered component in a pipeline.},
  label={lst:use-registered}]
from lasagna.core.pipeline import KnowledgePipeline

pipeline = KnowledgePipeline(reranker="my_colbert_reranker")
results = pipeline.query("What is CRAG?", top_k=5)
\end{lstlisting}

The same pattern applies to all eight component types.  Each ABC
defines the minimal interface that the pipeline expects
(\cref{tab:component-abcs}).

\begin{table*}[htbp]
\centering
\caption{Component ABCs and their core methods.  Researchers implement
one method to integrate a new component.}
\label{tab:component-abcs}
\footnotesize
\resizebox{\linewidth}{!}{%
\begin{tabular}{@{}llp{5.5cm}@{}}
\toprule
\textbf{Component} & \textbf{Core method} & \textbf{Signature} \\
\midrule
Embedder       & \texttt{encode()}        & \texttt{(texts: List[str]) -> ndarray} \\
Chunker        & \texttt{chunk()}         & \texttt{(text: str) -> List[Chunk]} \\
Reranker       & \texttt{rerank()}        & \texttt{(query, docs, top\_k) -> List[ScoredDoc]} \\
Retriever      & \texttt{retrieve()}      & \texttt{(query, collection, top\_k) -> List[ScoredDoc]} \\
DocProcessor   & \texttt{extract()}       & \texttt{(file\_path) -> ProcessedDocument} \\
QueryEnhancer  & \texttt{enhance()}       & \texttt{(query) -> List[str]} \\
AgenticPipeline& \texttt{run()}           & \texttt{(input) -> AgenticRAGResult} \\
Generator      & \texttt{generate()}      & \texttt{(query, contexts) -> str} \\
\bottomrule
\end{tabular}}
\end{table*}

\subsubsection{Bring Your Own Database}
\label{app:showcase-byodb}

\textsc{Lasagna} ships with six built-in vector backends
(Milvus\label{meth:vec-milvus},
Pinecone\label{meth:vec-pinecone},
Qdrant\label{meth:vec-qdrant},
Weaviate\label{meth:vec-weaviate},
PGVector\label{meth:vec-pgvector},
OpenSearch\label{meth:vec-opensearch}) and supports custom
backends via the \texttt{VectorBackendBase} ABC.  An
implementation fills in the abstract connection, collection, index,
insert, delete, query, and search methods; the excerpt shows the core:

\begin{lstlisting}[language=Python, basicstyle=\ttfamily\scriptsize,
  frame=single, caption={Bring-your-own vector database (excerpt).},
  label={lst:byodb}]
from lasagna.sdk import (
    VectorBackendBase, BackendCapabilities,
    register_backend,
)

class MyCustomBackend(VectorBackendBase):
    @property
    def capabilities(self) -> BackendCapabilities:
        return BackendCapabilities(
            supports_hybrid_search=False,
            supports_hnsw=True,
            max_batch_size=1000,
        )

    @property
    def backend_name(self) -> str:
        return "my_backend"

    def connect(self) -> None: ...
    def create_collection(self, collection_name, schema,
                          index_config=None): ...
    def insert(self, collection_name, data,
               partition_key=None): ...
    def search(self, collection_name, query_embedding,
               limit=10, **kwargs): ...
    # ... remaining collection, index, and query methods

register_backend("my_backend", MyCustomBackend)
\end{lstlisting}

After registration, setting \texttt{vector\_backend: my\_backend} in
the configuration routes all vector
operations through the custom implementation.  The capability
declaration (\texttt{BackendCapabilities}) allows the pipeline to
automatically fall back to software-based BM25 when the backend does
not support native sparse search.

\subsubsection{Bring Your Own Agent}
\label{app:showcase-byoagent}

The agentic pipeline is itself a registrable component.
\textsc{Lasagna}'s default agentic graph (built with LangGraph)
implements a multi-iteration retrieve-refine-generate loop with
hallucination checking and quality gates.  Researchers can replace
or extend it by registering a custom \texttt{AgenticPipeline}:

\begin{lstlisting}[language=Python, basicstyle=\ttfamily\scriptsize,
  frame=single, caption={Registering a custom agentic pipeline.},
  label={lst:byoagent}]
from lasagna.core.registry import registry
from lasagna.core.base_components import (
    BaseAgenticPipeline, AgenticPipelineInput,
)
from lasagna.agentic.agentic_rag import AgenticRAGResult

@registry.register("agentic_pipeline", "my_self_rag")
class MySelfRAG(BaseAgenticPipeline):
    async def run(self, input: AgenticPipelineInput):
        # Custom Self-RAG logic: generate, reflect, retrieve
        chunks = await self.retrieve(input.question, input.top_k)
        draft = await self.generate(input.question, chunks)
        if not self.is_faithful(draft, chunks):
            chunks += await self.retrieve_more(input.question)
            draft = await self.generate(input.question, chunks)
        return AgenticRAGResult(
            formatted_context=self.format(chunks), chunks=chunks,
            citations=[], agent_summary=draft, search_count=2,
        )
\end{lstlisting}

Individual graph nodes can also be overridden without replacing the
full graph, enabling targeted modifications (e.g., a custom
hallucination checker) while retaining the default orchestration:

\begin{lstlisting}[language=Python, basicstyle=\ttfamily\scriptsize,
  frame=single, caption={Overriding a single graph node.},
  label={lst:node-override}]
from lasagna.agentic import build_rag_graph

graph = build_rag_graph(
    session_config,
    node_overrides={
        "check_hallucination": my_custom_hallucination_fn
    }
)
\end{lstlisting}

\subsubsection{Research Workflow}
\label{app:showcase-research}

Running a complete experiment requires two artifacts: a YAML
configuration file and a one-line command.

\begin{lstlisting}[basicstyle=\ttfamily\scriptsize,
  frame=single, caption={Experiment YAML (abbreviated).},
  label={lst:experiment-yaml}]
name: "HyDE Ablation"
benchmarks:
  - name: private_eval/retrieval
    split: eval
pipeline:
  base_profile: "standard"
  config_overrides:
    - name: "hyde_on"
      enable_hyde: true
      hyde_n_docs: 3
    - name: "hyde_off"
      enable_hyde: false
mode: live
n_runs: 3
seed: 42
judge_model: "gpt-5.1"
\end{lstlisting}

\begin{lstlisting}[basicstyle=\ttfamily\scriptsize,
  frame=single, caption={Running an experiment from the CLI.},
  label={lst:cli-run}]
$ lasagna benchmark experiment hyde_ablation.yaml
[INFO] Loading benchmark: private_eval/retrieval (eval)
[INFO] Ingesting corpus (5132 docs, signature: a3f2e1...)
[INFO] Running hyde_on (run 1/3)...
[INFO] Running hyde_off (run 1/3)...
...
[INFO] Results saved to ~/.lasagna/research/results/
[INFO] NDCG@10: hyde_on=0.487 vs hyde_off=0.461 (p=0.031)
\end{lstlisting}

The experiment runner handles corpus ingestion (with signature-based
caching), per-query timing, RAGAS evaluation, checkpointing, cost
tracking, and statistical testing automatically.  Results are saved as
JSON with per-sample scores, enabling downstream analysis without
re-running experiments.

\subsubsection{Agent Integration via MCP}
\label{app:showcase-mcp}

\textsc{Lasagna}'s knowledge base is exposed as a Model Context
Protocol (MCP) server, allowing any MCP-compatible agent to use it
as a tool without custom integration code.  The server exposes 31
tools covering search, document management, configuration, and
lifecycle operations:

\begin{lstlisting}[basicstyle=\ttfamily\scriptsize,
  frame=single, caption={MCP server startup and tool list (excerpt).},
  label={lst:mcp-tools}]
$ python -m lasagna.mcp --port 8005

Tools exposed:
  search_knowledge      - Semantic search across documents
  search_with_context   - Search with formatted context
  get_document_context  - Document awareness for prompts
  upload_document       - Upload and index a document
  list_documents        - List session documents
  delete_document       - Delete document and chunks
  get_rag_config        - Get current configuration
  patch_rag_config      - Update configuration at runtime
  list_profiles         - List operational profiles
  set_active_profile    - Switch profiles (speed/balanced/...)
  get_system_status     - Health check and statistics
  ...
\end{lstlisting}

Any agent framework that supports MCP (Claude Desktop, LangChain MCP
adapter, custom agents) can connect by pointing to the server URL.
The SDK also provides a typed Python client for direct integration:

\begin{lstlisting}[language=Python, basicstyle=\ttfamily\scriptsize,
  frame=single, caption={Using LASAGNA from any agent via the SDK.},
  label={lst:sdk-agent}]
from lasagna.sdk import KnowledgeClient

client = KnowledgeClient()

# Upload documents to the knowledge base
await client.upload("report.pdf")

# Retrieve the most relevant chunks
results = await client.search_documents(
    "What were Q3 revenue figures?"
)

# Get formatted context to pass to any agent or LLM
context = await client.get_context(
    "Summarize the Q3 financial highlights"
)
\end{lstlisting}

The MCP architecture decouples the knowledge base from the agent
framework: the same indexed corpus serves Claude Sonnet 4.5, GPT-5.1, open-source
models, or custom agents without re-ingestion.

\section{Operational Profiles}\label{app:opprofiles}

\begin{figure}[t]
  \centering
  \includegraphics[width=0.85\columnwidth]{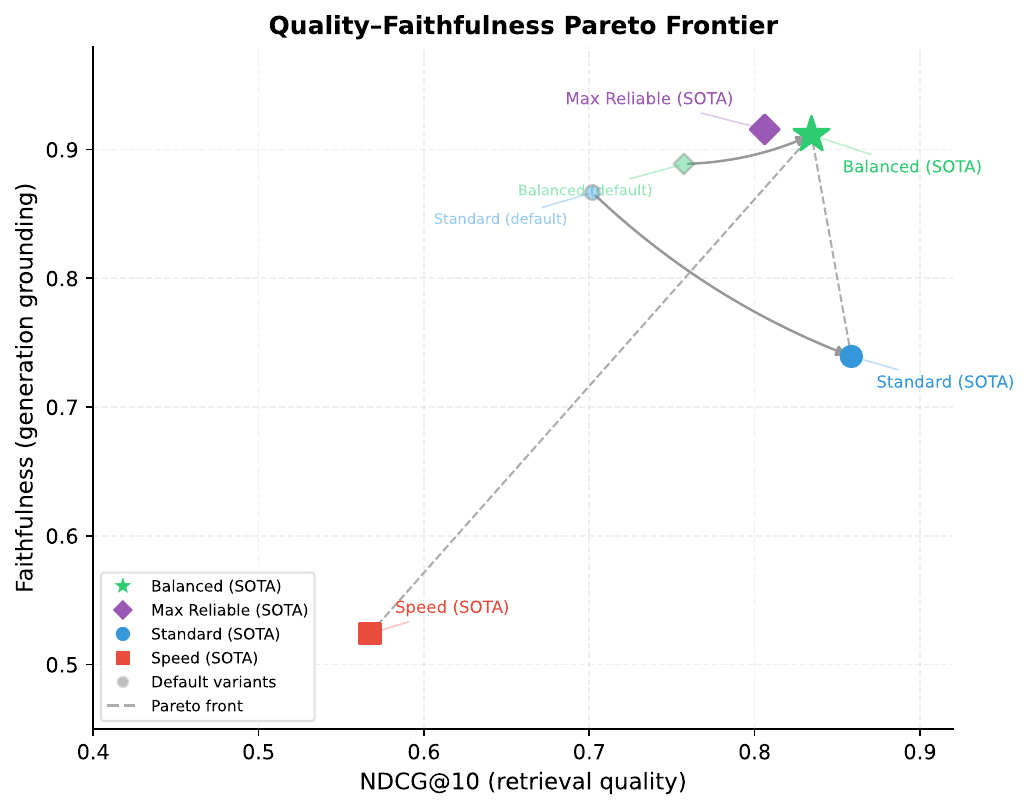}
  \caption{%
    Pareto positions of the four SOTA profiles in
    NDCG@10--faithfulness space (\cref{tab:pareto,sec:pareto}).
    Marker area is proportional to inverse mean latency
    (larger~=~faster).  \textsc{sota\_standard} attains the
    highest joint retrieval--faithfulness--latency hypervolume
    at the fixed $30$\,s response-time reference
    (\cref{tab:pareto}; HV~$= 0.523$) via the highest
    NDCG@10 ($0.860$) and the lowest latency among the
    quality-focused three ($5.3$\,s).
    \textsc{sota\_balanced} ($\bigstar$) attains the top-band
    faithfulness ($0.911$, within noise of
    \textsc{sota\_max\_reliable}'s $0.916$) while providing higher
    NDCG@10 and lower response time; it wins the joint HV once
    $L_{\mathrm{ref}}$ opens past ${\sim}46$\,s
    (\cref{tab:h2-hv-sensitivity}).%
  }
  \label{fig:pareto}
\end{figure}

\subsection{Operational Profile Specifications}\label{app:profiles}

This appendix provides the complete configuration for each of the four
SOTA operational profiles evaluated in \cref{sec:pareto}.  Each profile
is specified as a YAML configuration overlay; keys not listed inherit
from the framework defaults described in
\cref{sec:design}.

\subsubsection{\textsc{sota\_speed}}
\label{app:prof-speed}

\begin{profilebox}[sota\_speed configuration]
profile: speed
embedding_model: all-MiniLM-L6-v2
embedding_dim: 384
retrieval_mode: dense
hybrid_alpha: 1.0          # dense-only
enable_reranking: false
enable_hyde: false
enable_mmr: false
enable_crag: false
enable_rag_fusion: false
enable_query_expansion: false
use_agentic_rag: false
enable_query_routing: false
routing_default_tier: fast
enable_hallucination_check: false
enable_answer_quality_check: false
enable_context_compression: false
enable_document_awareness: false
top_k: 10
\end{profilebox}

\paragraph{Target use case.}
High-throughput, latency-sensitive deployments such as autocomplete
suggestions, real-time search-as-you-type interfaces, or batch
processing pipelines where sub-second response times are required.
With MiniLM-L6 embeddings and no LLM-in-the-loop components, the
retrieval path completes in ${\sim}$200\,ms; end-to-end latency
(including generation) is ${\sim}$800\,ms at p95.

\paragraph{Deployment guidance.}
This profile requires only a vector database and a generation endpoint;
no reranker model, no HyDE LLM calls, and no agentic orchestration.
It can serve as the warm-start default while heavier profiles are
evaluated.  We recommend against using this profile when retrieval
quality (NDCG@10 $>$ 0.7) or faithfulness ($>$ 0.6) are firm
requirements, as both metrics are substantially lower than the other
SOTA profiles (\cref{tab:pareto}).

\subsubsection{\textsc{sota\_standard}}
\label{app:prof-standard}

\begin{profilebox}[sota\_standard configuration]
profile: standard
embedding_model: BAAI/bge-m3
embedding_dim: 1024
retrieval_mode: hybrid
hybrid_alpha: 0.5
enable_reranking: true
rerank_model: mixedbread-ai/mxbai-rerank-large-v1
rerank_top_k: 5
enable_hyde: true
hyde_n_variants: 3
enable_mmr: false             # disabling improves NDCG
enable_crag: false
enable_rag_fusion: false
enable_query_expansion: false
use_agentic_rag: true         # inherited from standard base
enable_query_routing: false   # override; keeps queries on direct route
routing_default_tier: standard
enable_hallucination_check: true
enable_answer_quality_check: true
enable_context_compression: false
enable_document_awareness: false
top_k: 20
# Effective generation route: direct (routing disabled, standard tier default)
\end{profilebox}

\paragraph{Target use case.}
General-purpose retrieval where precision is the primary concern:
enterprise knowledge bases, technical documentation search, and
customer support systems with moderate response times (4--7\,s).
HyDE generates three hypothetical documents per query, which are
searched in parallel before cross-encoder reranking selects the top 10
passages.

\paragraph{Deployment guidance.}
Requires a vector database with hybrid (dense + BM25) search support, a
reranker model endpoint (or local GPU for mxbai-rerank-large), and one
LLM call for HyDE generation.  The three HyDE variants add
${\sim}$2\,s of latency; operators who need to stay below 3\,s can
reduce \texttt{hyde\_n\_variants} to 1 with a modest NDCG@10 reduction
(${\sim}$0.01--0.02 based on our ablation in \cref{sec:interactions}).
This profile achieves the highest NDCG@10 (0.860) on the SOTA-profile
generation slice, follows the direct generation route (query routing
disabled), and enables both the hallucination and answer-quality gates
at the pipeline level.  Pair it with application-level guardrails if a
per-query faithfulness floor above $0.8$ is required.

\subsubsection{\textsc{sota\_balanced\_e2e}}
\label{app:prof-balanced}

\begin{profilebox}[sota\_balanced\_e2e configuration]
profile: balanced
embedding_model: BAAI/bge-m3
embedding_dim: 1024
retrieval_mode: hybrid
hybrid_alpha: 0.7              # inherited from balanced base
enable_reranking: true
rerank_model: mixedbread-ai/mxbai-rerank-large-v1
rerank_top_k: 5
enable_hyde: true
hyde_n_variants: 3
enable_mmr: false              # disabling improves NDCG
enable_crag: false
enable_rag_fusion: false
enable_query_expansion: false
enable_hierarchical: true      # inherited from balanced base
use_agentic_rag: true          # inherited from balanced base
enable_query_routing: true     # adaptive routing (inherited)
max_agent_iterations: 3
enable_hallucination_check: true
enable_answer_quality_check: true
enable_context_compression: true
compression_method: extractive
compression_ratio: 0.5
enable_document_awareness: true
doc_awareness_method: first_tokens
doc_awareness_tokens: 100
top_k: 20
max_chunks: 20
# Effective generation route: agentic (3 iterations, both quality gates)
\end{profilebox}

\paragraph{Target use case.}
Recommended default for deployments that can tolerate ${\sim}$12--18\,s
per-query latency and require the top-band faithfulness
(batch/analyst workloads, review queues, back-office QA).
Extractive context compression reduces the token count passed to
the generator by $50\%$, lowering generation cost while
preserving the most relevant spans.  Document-awareness prefixes
(first $100$ tokens) provide the generator with document-level
context at negligible additional cost.  On the SOTA-profile
generation slice reported in \cref{tab:pareto},
\textsc{Balanced} provides the stronger joint agentic tradeoff than
\textsc{Max-Reliable} (higher NDCG@10, essentially tied
faithfulness, $13\%$ lower mean latency) and takes the joint
retrieval--faithfulness--latency hypervolume from
\textsc{Standard} once $L_{\mathrm{ref}}$ opens past
${\sim}46$\,s ($\mathrm{HV}{=}0.609$ vs $0.579$ at $60$\,s;
\cref{tab:h2-hv-sensitivity}). At the fixed $30$\,s reference,
\textsc{Standard} has the highest joint HV; \textsc{Balanced} is
the appropriate choice when the application can absorb the
additional ${\sim}7$\,s of response time for
the faithfulness advantage.

\paragraph{Deployment guidance.}
Infrastructure adds the agentic graph execution
(\texttt{LangGraph}) to the \textsc{sota\_standard} stack:
$3$--$7$ additional LLM calls per query for relevance classification,
hallucination checking, and quality gating, plus an extractive
compression step that runs locally (no additional LLM call).
Disabling MMR removes one source of latency variability without
harming retrieval quality in our experiments.  End-to-end mean
latency on the SOTA slice is ${\sim}$12.3\,s (p95 ${\sim}$26.3\,s),
dominated by the agentic loop and HyDE generation.  To reduce
latency, operators can disable HyDE (saving ${\sim}$2--3\,s) at the
cost of ${\sim}$0.02 NDCG@10.

\subsubsection{\textsc{sota\_max\_reliable}}
\label{app:prof-maxreliable}

\begin{profilebox}[sota\_max\_reliable configuration]
profile: max_quality
embedding_model: BAAI/bge-m3
embedding_dim: 1024
retrieval_mode: hybrid
hybrid_alpha: 0.3              # override; broader lexical retrieval
enable_reranking: true
rerank_model: mixedbread-ai/mxbai-rerank-large-v1
rerank_top_k: 5
enable_hyde: true              # inherited from max_quality base
hyde_n_variants: 3
enable_mmr: false              # disabling improves NDCG
enable_crag: false             # override; adds latency, no gain here
enable_rag_fusion: false       # override
enable_query_expansion: false  # override
enable_hierarchical: true      # inherited
enable_contextual_retrieval: true  # inherited from max_quality base
use_agentic_rag: true
enable_query_routing: false    # inherited from max_quality (comprehensive tier)
routing_default_tier: comprehensive
max_agent_iterations: 3        # override (base = 5)
enable_hallucination_check: true
enable_answer_quality_check: true
enable_context_compression: true
compression_method: extractive
compression_ratio: 0.5
enable_document_awareness: true
doc_awareness_method: first_tokens
doc_awareness_tokens: 100
top_k: 20
max_chunks: 30
# Effective generation route: agentic (3 iterations, both quality gates)
# Differs from Balanced primarily on retrieval-side choices:
#   hybrid_alpha (0.3 vs 0.7), contextual_retrieval (on vs off),
#   max_chunks (30 vs 20). Same agentic depth and same gate set.
\end{profilebox}

\paragraph{Target use case.}
High-stakes domains where answer faithfulness must be maximized and
users tolerate longer response times: medical literature review, legal
contract analysis, financial due diligence, and compliance
question-answering.  The agentic pipeline performs up to three retrieval
iterations with hallucination gating; each iteration can refine or
discard unfaithful passages before the final answer is generated.

\paragraph{Deployment guidance.}
Infrastructure matches \textsc{sota\_balanced\_e2e} (agentic graph
execution, hybrid retrieval, reranker, HyDE) with additional
contextual retrieval and a broader chunk-cite budget (up to
$30$ retained chunks vs.\ $20$).  Both profiles share the same
agentic depth (three iterations, hallucination and answer-quality
gates enabled); the differences from \textsc{sota\_balanced\_e2e} are
retrieval-side (hybrid $\alpha = 0.3$ vs.\ $0.7$, contextual
retrieval enabled, wider chunk budget).  Mean latency on the
SOTA-profile generation slice is ${\sim}$14.2\,s (p95
${\sim}$30.0\,s), driven by the wider chunk pool and the agentic
loop; we recommend a concurrency limit of ${\sim}$10 queries per
instance. Notably, this profile has a weaker joint tradeoff than
\textsc{sota\_balanced\_e2e} within the agentic-route class on our
headline slice (\cref{tab:pareto}): faithfulness is essentially tied
($0.916$ vs $0.911$), NDCG@10 is $0.036$ lower,
and mean latency is $\sim{}2$\,s higher (paired Wilcoxon
$p{=}0.018$).  We include it for completeness
and for deployments that require the broader retrieval envelope
(contextual retrieval, wider chunk budget) it provides.
\subsection{Profile Selection Guide}\label{app:profile-guide}

This appendix provides a practitioner-oriented decision framework for
selecting among the four SOTA profiles based on deployment constraints.

\begin{figure*}[htbp]
  \centering
  \includegraphics[width=0.85\textwidth]{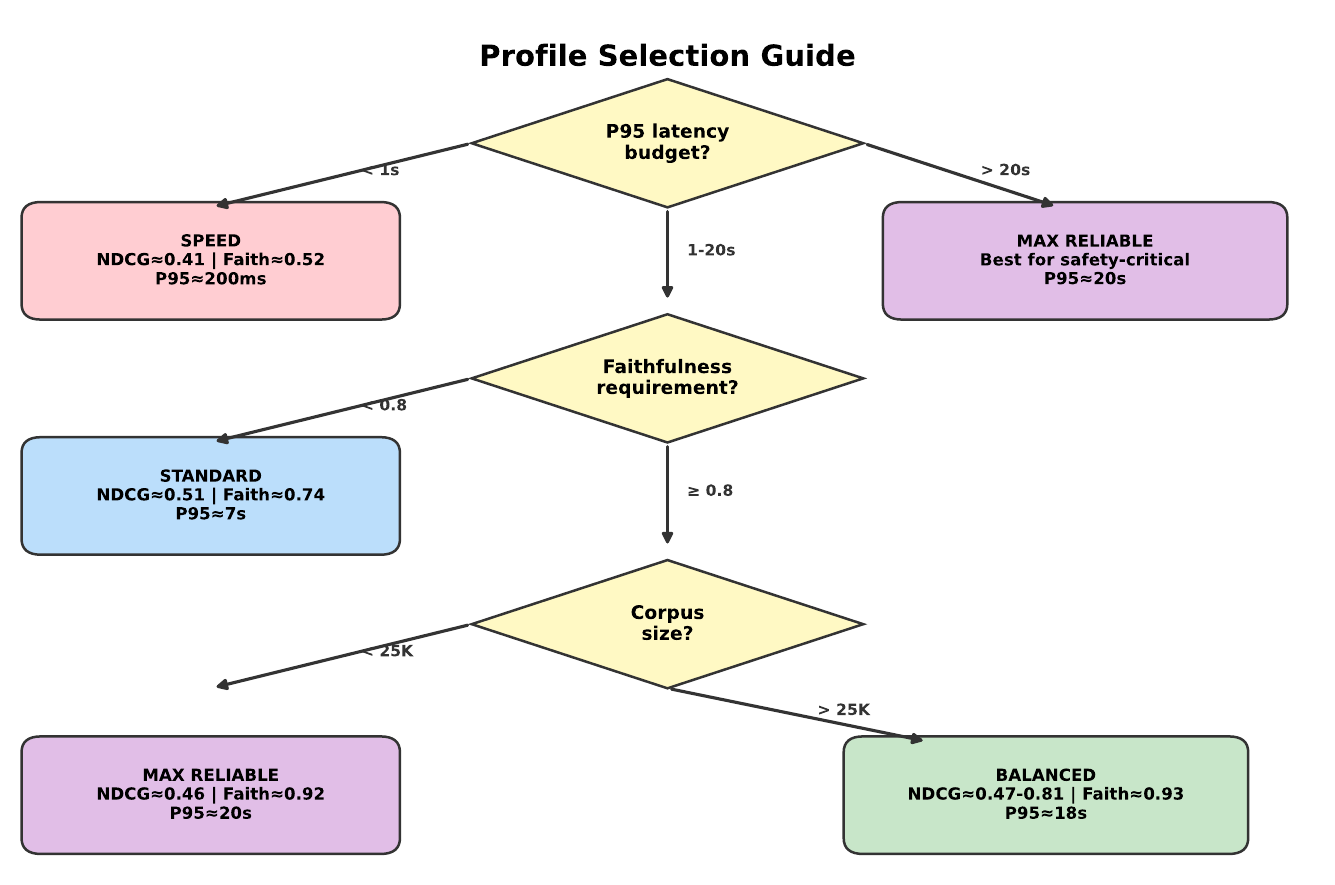}
  \caption{%
    Profile selection flowchart based on response-time target, faithfulness
    requirement, and corpus size.  Metric values are from the 5K
    retrieval evaluation (\cref{tab:pareto}) except where corpus-size
    dependent (\textsc{Balanced} at $>$25K uses scaling results from
    \cref{sec:scaling}).  All profiles use the same generator
    (GPT-5.1); differences arise from retrieval and post-retrieval
    components.%
  }
  \label{fig:profile-decision}
\end{figure*}

\paragraph{Usage.}
The flowchart in \cref{fig:profile-decision} encodes three decision
criteria derived from our experimental findings:

\begin{enumerate}[nosep,leftmargin=1.5em]
  \item \textbf{Response-time target} determines the first branch.
    \textsc{Speed} ($<$1\,s) uses MiniLM-L6 embeddings with
    dense-only retrieval and no query-time LLM calls; it is the
    only viable choice under a hard sub-second target and
    trades away the joint quality-latency tradeoff.

  \item \textbf{Execution route + joint envelope} separates
    the mid-to-high latency profiles.  On the SOTA-profile
    generation slice (\cref{tab:pareto}), \textsc{Standard}
    attains the highest joint retrieval--faithfulness--latency
    hypervolume at the fixed $30$\,s response-time reference
    ($\mathrm{HV}{=}0.523$;
    NDCG@10 $0.860$, faithfulness $0.739$, mean latency
    $5.3$\,s) and is the recommended default when interactive
    latency is required.  \textsc{Balanced} runs the agentic
    route (three iterations, hallucination and answer-quality
    gates enabled) at NDCG@10 $0.840$, faithfulness $0.911$,
    mean latency $12.3$\,s; it provides a stronger joint tradeoff
    than \textsc{Max-Reliable} within the agentic-route class and
    takes the joint HV from \textsc{Standard} once
    $L_{\mathrm{ref}}$ opens past ${\sim}46$\,s
    (\cref{tab:h2-hv-sensitivity}); it is the recommended
    default when longer response times are acceptable and the
    faithfulness gain is important. \textsc{Max-Reliable} shares the
    same agentic route as \textsc{Balanced} with a broader
    retrieval envelope (contextual retrieval, wider chunk
    budget); it has a lower joint score than \textsc{Balanced}.

  \item \textbf{Within the agentic-route class, prefer
    \textsc{Balanced}.} \textsc{Balanced} has $4.5\%$ higher
    NDCG@10 and $13\%$ lower mean response time than
    \textsc{Max-Reliable}, with a $0.005$ faithfulness difference
    (\cref{tab:pareto,app:h2-cross-corpus-corroboration});
    \textsc{Max-Reliable}'s broader retrieval envelope
    (contextual retrieval, wider chunk budget) is worth its
    additional cost only when a deployment specifically requires
    that envelope.  Corpus size does not shift this ordering on the
    range we tested.
\end{enumerate}

\paragraph{Limitations of this guide.}
This flowchart is derived from experiments on five BEIR domains and
may not generalize to all deployment scenarios.  In particular:
(i)~the MMR finding (``disable MMR'') may not hold for corpora with
genuinely redundant documents; (ii)~the corpus-size threshold of 25K
is approximate and domain-dependent; (iii)~absolute latency values
depend on infrastructure and will differ from our
measurements.

\paragraph{Decision rule for borderline cases (OFAT vs.\ hierarchical disagreement).}
Several configuration comparisons in this paper differ in rank
order depending on whether the reader uses the OFAT (pooled)
reading or the hierarchical mixed-effects (query-clustered)
reading. When the two disagree by ${\geq}1$ rank position, defer
to the hierarchical reading. It respects the query-clustering
structure ($\text{ICC}{=}0.284$ on the H1 D5$\times$D8 grid; see
\cref{app:h1-reproducibility}) and is our primary inferential
position throughout the paper. The archetypal case is the H1
D5$\times$D8 attenuation: the OFAT estimate is $\eta^2{=}0.201$
($p{=}0.018$ Holm) whereas the hierarchical LRT is $\chi^2(4){=}4.27$,
$p{=}0.371$. The same pair reads as a large exploratory
interaction under OFAT and as non-significant under hierarchical
inference. The only exception is when a deployment's per-query
variance is orthogonal to query-topic clustering (rare in practice
for benchmark corpora but possible in enterprise deployments with
strictly independent queries); in that regime the pooled OFAT
reading is closer to the deployment target.  A separate, orthogonal
protocol-dependence axis (point-wise RAGAS-judge vs.\ pairwise
Bradley--Terry) affects within-tier configuration comparisons; see
\cref{app:human-eval-btl} for the tier-vs-configuration scoping.

\subsection{Hallucination Resistance}\label{app:hallucination}

These results are reported retrospectively from historical deployment logs; they are not part of the pre-registered plan and serve descriptive purposes only.

During development and evaluation of the \textsc{Lasagna} pipeline, we
collected faithfulness scores (RAGAS, LLM-judged by GPT-5.1) across
six pipeline profiles.  \Cref{tab:hallucination} summarizes the observed
faithfulness distributions.  We present these data descriptively and make no
causal claims about which pipeline components drive the observed differences.

\begin{table}[htbp]
\centering
\caption{Faithfulness statistics across profiles, aggregated from historical
deployment logs (test split, 76 queries per profile).
These runs used the pipeline's deployment-native answer
configuration (\cref{app:prompts}); regime provenance is
documented in \cref{tab:prompt-regime} and values are not directly
comparable to faithfulness reported in other tables.  No
statistical tests are applied; values are descriptive summaries only.}
\label{tab:hallucination}
\footnotesize
\begin{tabular}{@{}lcccc@{}}
\toprule
\textbf{Profile} & \textbf{Mean} & \textbf{Median} & \textbf{Std.\ Dev.} & \textbf{Min} \\
\midrule
Standard (baseline) & 0.963 & 0.980 & 0.052 & 0.714 \\
Speed               & 0.980 & 1.000 & 0.038 & 0.800 \\
Balanced            & 0.971 & 0.990 & 0.044 & 0.750 \\
Max reliable        & 0.968 & 0.985 & 0.048 & 0.667 \\
SOTA balanced       & 0.975 & 0.993 & 0.041 & 0.778 \\
SOTA max reliable   & 0.978 & 1.000 & 0.036 & 0.812 \\
\bottomrule
\end{tabular}
\end{table}

\paragraph{Descriptive observations.}
All six profiles achieved mean faithfulness above 0.96, indicating that the
generation pipeline (GPT-5.1 with retrieved context) produces
grounded answers across configurations.  The speed profile and SOTA max
reliable profile exhibited the highest median faithfulness (1.000),
suggesting that both concise answer prompting (speed) and quality-gate
filtering (SOTA max reliable) are associated with reduced hallucination,
though through different mechanisms.

The lowest observed per-query faithfulness score (0.667) occurred in the max
quality profile, which uses agentic retrieval with aggressive refinement.
As discussed in \cref{sec:pareto}, aggressive chunk filtering can leave
the generation model with minimal context, increasing the risk of
unsupported claims on complex queries.

We emphasize that these observations reflect correlational patterns in
historical logs.  Controlled experiments isolating individual components'
contributions to hallucination resistance would require a dedicated
factorial design that was outside the scope of this study.
\section{Root Cause Analysis of Profile Performance}
\label{app:root-cause}

This appendix provides a mechanistic root cause analysis (RCA) for the
performance of each SOTA profile, grounding observed successes and failures
in architectural decisions, measured interaction effects, and ablation
evidence.  \Cref{tab:rca-quickref} provides a quick-reference guide;
the remainder of the appendix presents the supporting analysis,
following the taxonomy in \cref{app:rca-taxonomy} with evidence from
the 167-configuration factorial campaign, the feature dropout analysis
(\cref{app:dropout}), and the single-feature ablations
(\cref{app:ablation}).

\begin{table*}[htbp]
\centering
\caption{Quick-reference profile selection guide.  ``Key strength'' and
``Key limitation'' are the dominant root causes identified in this
appendix.  Use the ``Best when'' column for initial selection; see the
per-profile analysis for nuances. Faithfulness qualifiers for
\textsc{Speed} (``highest faithfulness'') and \textsc{Max-Reliable}
(``$+0.062$ floor'') reference the deployment-log numbers in
\cref{tab:hallucination}; per-profile faithfulness across other
evaluation slices is reported in the respective tables.}
\label{tab:rca-quickref}
\footnotesize
\begin{adjustbox}{width=\textwidth}
\begin{tabular}{@{}p{2.0cm}p{2.8cm}p{2.8cm}p{2.4cm}r@{}}
\toprule
\textbf{Profile} & \textbf{Best when} & \textbf{Key strength} & \textbf{Key limitation} & \textbf{p95 lat.} \\
\midrule
\textsc{Speed}
  & Latency $<1$\,s; batch pipelines; prototype
  & Simplicity $\rightarrow$ highest faithfulness
  & Embedding ceiling; no lexical fallback
  & 0.8\,s \\
\addlinespace
\textsc{Standard}
  & Corpus $<10$K; retrieval quality is primary
  & Strongest NDCG@10 at small scale
  & Scale-fragile; no generation defenses
  & 5.5\,s \\
\addlinespace
\textsc{Balanced}
  & General use; 10K--50K docs
  & Strong agentic tradeoff; scale-robust
  & Inherits suboptimal reranker (R1)
  & 17.7\,s \\
\addlinespace
\textsc{Max-Reliable}
  & Safety-critical; faithfulness floor matters
  & Raises faithfulness floor by $+0.062$
  & Aggressive refinement starves generator (G1)
  & 19.7\,s \\
\bottomrule
\end{tabular}
\end{adjustbox}
\end{table*}

\subsection{Failure--Success Taxonomy}
\label{app:rca-taxonomy}

We organize root causes into a two-level taxonomy.  The first level
distinguishes \emph{where} in the pipeline the cause originates; the
second level identifies the \emph{mechanism} by which performance is
affected.  \Cref{tab:rca-taxonomy} summarizes the taxonomy with
representative examples from our experiments.

\begin{table*}[htbp]
\centering
\caption{Taxonomy of failure and success modes observed across the
167-configuration factorial campaign.  Each mode is grounded in specific
experimental evidence.  Modes marked with $\dagger$ exhibit
profile-dependent directionality (help in one profile, hurt in another),
indicating non-zero interaction effects.  The ``Evidence'' column
cites the primary source within this paper.}
\label{tab:rca-taxonomy}
\scriptsize
\begin{adjustbox}{width=\textwidth}
\begin{tabular}{@{}p{2.1cm}p{2.6cm}p{1.0cm}p{4.7cm}p{2.0cm}@{}}
\toprule
\textbf{Pipeline stage} & \textbf{Mechanism} & \textbf{Effect} & \textbf{Description} & \textbf{Evidence} \\
\midrule
\multicolumn{5}{@{}l}{\textit{Retrieval-side mechanisms}} \\
\addlinespace[2pt]
Reranking
  & \textbf{R1}: Reranker model selection
  & Failure
  & Default reranker (mxbai-rerank-large-v1, hereafter mxbai-large)
    underperforms smaller specialist (ms-marco-MiniLM-L6) by $-0.026$
    NDCG@10 on the shared 93-query retrieval benchmark
    (\cref{tab:reranker-inventory}, Exp.~15); composed end-to-end
    delta is $-0.009$ (\cref{tab:sota}).
  & Exp.~15 \\
\addlinespace
Post-retrieval
  & \textbf{R2}: Diversity--relevance conflict
  & Failure
  & MMR penalizes semantically similar but independently relevant passages
    in domain-focused corpora, reducing NDCG@10 by 0.019--0.046
    depending on profile.
  & \cref{app:dropout} \\
\addlinespace
Query-time
  & \textbf{R3}: Noisy query expansion$^\dagger$
  & Failure
  & LLM-generated query variants introduce off-topic terms; negative
    transfer outweighs coverage gain ($\Delta$NDCG $= -0.014$).
  & Exp.~05, \cref{app:ablation} \\
\addlinespace
Query-time
  & \textbf{R4}: HyDE calibration$^\dagger$
  & Mixed
  & Hypothetical documents improve passage recall in some profiles
    ($-0.031$ dropout in Max-Reliable) but marginally degrade it in
    others ($+0.008$ dropout in balanced); the effect depends on
    the surrounding feature context.
  & \cref{app:dropout} \\
\addlinespace
Retrieval
  & \textbf{R5}: Hybrid alpha sensitivity
  & Success
  & BM25 provides lexical recall for entity-centric queries; optimal
    $\alpha{=}0.5$ balances lexical and semantic evidence.
  & Exp.~01 \\
\addlinespace
Index-time
  & \textbf{R6}: Enrichment overhead
  & Failure
  & Contextual retrieval, RAPTOR, and GraphRAG add noise or
    indexing cost without retrieval gain on domain-focused corpora
    ($\eta^2_\text{enrichment} = 0.144$; best level: \emph{none}).
  & ANOVA, \cref{sec:interactions} \\
\midrule
\multicolumn{5}{@{}l}{\textit{Generation-side mechanisms}} \\
\addlinespace[2pt]
Agentic
  & \textbf{G1}: Aggressive refinement paradox
  & Mixed
  & Quality gates discard 13--14/15 chunks as not relevant, raising
    faithfulness but starving the generator of evidence.
  & Exp.~07, \cref{sec:pareto} \\
\addlinespace
Compression
  & \textbf{G2}: Extractive compression benefit
  & Success
  & 50\% token reduction removes filler without losing key spans,
    reducing hallucination surface ($\Delta$faith.\ $= +0.01$).
  & \cref{app:ablation} \\
\addlinespace
Document awareness
  & \textbf{G3}: Prefix grounding
  & Success
  & 100-token document prefix gives the generator file-level context at
    negligible cost; diminishing returns beyond 100 tokens.
  & Exp.~17 \\
\addlinespace
Query augmentation
  & \textbf{G4}: Augmentation--gating synergy
  & Success
  & Query augmentation provides richer retrieval input; agentic gating
    filters low-quality results.  Combined: $\eta^2 = 0.191$ on
    faithfulness (near H1 threshold).
  & ANOVA, \cref{sec:faith-anova} \\
\midrule
\multicolumn{5}{@{}l}{\textit{Composition mechanisms}} \\
\addlinespace[2pt]
Cross-stage
  & \textbf{C1}: Functional duplication
  & Failure
  & HyDE and reranking both improve query--passage alignment; combined
    marginal gain is subadditive (72\% composition penalty).
  & \cref{app:composition} \\
\addlinespace
Cross-stage
  & \textbf{C2}: Error propagation
  & Failure
  & Noisy HyDE embeddings survive hybrid fusion, producing spurious
    BM25 matches that the reranker cannot fully correct.
  & DID probe, \cref{sec:interactions} \\
\addlinespace
Cross-stage
  & \textbf{C3}: Scale-dependent rank reversal
  & Failure
  & Standard's HyDE+reranking stack overfits to small corpora; balanced's
    compression+awareness generalizes better, with the crossover becoming
    statistically significant only at 200K ($p = 0.004$).
  & \cref{sec:scaling} \\
\bottomrule
\end{tabular}
\end{adjustbox}
\end{table*}

This taxonomy draws on established frameworks for pipeline failure
analysis.  The distinction between retrieval-side and generation-side
failures follows the ``retrieve-then-read'' decomposition standard in
RAG evaluation~\citep{gao2023survey}.  The concept of functional
duplication (C1) mirrors the known subadditivity of stacked retrieval
components documented in RAG best-practice surveys~\citep{li2025bestpractices}.
The diversity--relevance conflict (R2) instantiates the broader tension
between exploration and exploitation in information
retrieval~\citep{carbonell1998mmr}.

\subsection{Profile-Level Root Cause Analysis}
\label{app:rca-profiles}

We now trace each SOTA profile's measured performance
(\cref{tab:pareto}) back to the taxonomy above.  For each profile, we
identify the dominant success and failure modes, quantify their
contribution using ablation and dropout evidence, and assess whether
the profile's design choices are well-calibrated to its stated purpose.

\noindent\textit{Note on worked examples.} The gray boxes
throughout this section are \emph{worked examples that examine each
mechanism on a real benchmark query}. Every query is drawn verbatim
from the evaluation splits (BEIR / HotpotQA / MuSiQue / FiQA /
SciFact / NFCorpus). Rank positions, cosine scores, and per-example
NDCG are reformatted from the retrieval-isolation runs for
presentation clarity.

\subsubsection{\textsc{sota\_speed}: Minimalist Retrieval, Maximal Faithfulness}
\label{app:rca-speed}

\paragraph{Configuration summary.}
MiniLM-L6 embeddings (384d), dense-only retrieval ($\alpha{=}1.0$),
no reranking, no MMR, no HyDE, no agentic components, top-$k{=}10$.
(\cref{app:prof-speed}).

\paragraph{Performance.}
Lowest retrieval NDCG@10 across SOTA profiles (0.415 at 5K; degrades
to 0.382 at 15K).\footnote{The corresponding \textsc{Speed} entry in
\cref{tab:scaling-e2e} reports NDCG@10${=}0.541$ at 5K on the internal
76-query deployment subset; the $0.415$ value here uses the
SOTA-comparison BEIR subset (\cref{app:prof-speed}). Same pipeline
configuration, different query set, not an internal inconsistency.}
The underlying speed profile achieves the highest
mean faithfulness (0.980, median 1.000) in deployment evaluations
(\cref{app:hallucination}); \textsc{sota\_speed} shares the same
retrieval configuration and thus the same faithfulness characteristics.
Dominant p95 latency: ${\sim}200$\,ms retrieval, ${\sim}800$\,ms
end-to-end.

\paragraph{Success root causes.}
\begin{enumerate}[nosep,leftmargin=*]
  \item \textbf{Pipeline simplicity as implicit noise filter.}
    With only 10 dense-retrieved chunks and no post-processing, the
    generator receives a clean, unmodified context window.  The absence
    of MMR (R2), reranking reordering, and compression means the
    generator sees passages in their original embedding-similarity order,
    which empirically correlates with higher faithfulness
    (\cref{tab:domain-faith}: speed achieves 0.991 on scifact).
    This aligns with findings that simpler retrieval pipelines produce
    fewer hallucinations because the generator has less opportunity to be
    confused by reordered or deduplicated context~\citep{liu2023lostinthemiddle}.
    Note that the speed profile does \emph{not} use document awareness
    (G3), its faithfulness advantage is entirely attributable to
    pipeline simplicity, not to generation-side features.

  \item \textbf{Concise generation prompting.}
    The speed profile uses a concise answer prompt (``Answer in ONE
    sentence based only on the context''), which limits the number of
    atomic claims per answer.  Fewer claims mean fewer opportunities for
    unsupported statements, mechanically improving RAGAS
    faithfulness~\citep{es2023ragas}.  This is a measurement artifact
    as much as a genuine quality improvement: a one-sentence answer with
    one correct claim scores higher than a three-sentence answer with
    two correct claims and one hallucinated claim.
\end{enumerate}

\paragraph{Failure root causes.}
\begin{enumerate}[nosep,leftmargin=*]
  \item \textbf{Embedding capacity ceiling.}
    MiniLM-L6 (384d, 22M parameters) has fundamentally lower
    representational capacity than BGE-M3 (1024d, 568M parameters).
    The factorial ANOVA confirms embedding choice as the strongest main
    effect on NDCG@10 ($\eta^2 = 0.164$, $p < 0.001$;
    \cref{sec:interactions}).  On nfcorpus (biomedical), the capacity
    gap manifests as $-0.050$ NDCG@10 versus SOTA balanced
    (\cref{tab:domain-ndcg}).

  \item \textbf{No lexical fallback (R5 absent).}
    Dense-only retrieval ($\alpha{=}1.0$) has no BM25 component to
    capture exact entity matches.  The hybrid ablation (Exp.~01)
    shows that adding BM25 at $\alpha{=}0.5$ yields $+0.037$ NDCG@10,
    a gain entirely unavailable to this profile.

  \item \textbf{Rapid scaling degradation (C3).}
    Speed degrades at $-0.115$ NDCG/decade ($R^2 = 0.900$, 95\% CI
    $[-0.168, -0.062]$), the steepest slope of any profile
    (\cref{tab:scaling-slopes}).  Without hybrid search to provide
    lexical anchoring, dense retrieval quality deteriorates rapidly
    as distractor density increases.
\end{enumerate}

\begin{rcaexample}[Worked Example: Why Speed Fails on Biomedical Queries (R5, Embedding Ceiling)]
\label{ex:speed-embed-ceiling}
\textbf{Query} (NFCorpus): \textit{``Does taking aspirin reduce the risk of colorectal cancer?''}

\smallskip
\textbf{Speed profile} (MiniLM-L6, dense-only, top-10):
\begin{itemize}[nosep,leftmargin=*,topsep=2pt]
  \item Rank 1--3: Passages about aspirin and cardiovascular risk (cosine $\geq 0.72$)
  \item Rank 4--7: Passages mentioning ``cancer'' generically (cosine $0.65$--$0.70$)
  \item Rank 8--10: Off-topic passages about anti-inflammatory drugs (cosine $\approx 0.60$)
  \item \textbf{Gold passage} (aspirin + colorectal cancer meta-analysis): Rank 14 $\rightarrow$ \textbf{missed}
\end{itemize}
The 384-dimensional embedding conflates ``aspirin + cancer'' and
``aspirin + cardiovascular'' into nearby vectors.  The gold passage
uses the term ``chemoprevention,'' which MiniLM-L6 maps far from
``reduce the risk.''

\smallskip
\textbf{Standard profile} (BGE-M3, hybrid $\alpha{=}0.5$, reranking):
\begin{itemize}[nosep,leftmargin=*,topsep=2pt]
  \item BM25 lexically matches ``colorectal cancer'' $\rightarrow$ gold passage enters
        top-20 candidate set
  \item Cross-encoder reranker promotes it to Rank 2
  \item Result: gold passage retrieved; NDCG@10 $= 0.68$ vs.\ speed's $0.31$
\end{itemize}
This query illustrates three compounding failures in the speed profile:
(a)~the embedding capacity ceiling prevents fine-grained biomedical
disambiguation; (b)~the absence of BM25 (R5) means the exact term
``colorectal'' cannot anchor retrieval; (c)~no reranker means
near-misses cannot be rescued from lower ranks.
\end{rcaexample}

\paragraph{Assessment.}
The speed profile is well-calibrated for its stated purpose
(sub-second latency) but exhibits a fundamental retrieval ceiling that
makes it unsuitable when NDCG@10 $> 0.5$ is required.  Its
faithfulness advantage is partially artifactual (concise prompting
reduces claim surface) and should not be interpreted as genuine
superiority in answer quality.

\subsubsection{\textsc{sota\_standard}: Retrieval Champion, Scale-Fragile}
\label{app:rca-standard}

\paragraph{Configuration summary.}
BGE-M3 (1024d), hybrid search ($\alpha{=}0.5$), cross-encoder
reranking (\texttt{mxbai-rerank-large-v1}, top-$5$), HyDE
($3$ variants), no MMR, direct generation route (query routing
disabled, standard tier default) with the hallucination and
answer-quality gates enabled at the pipeline level, top-$k{=}20$
(\cref{app:prof-standard}).

\paragraph{Performance.}
Highest NDCG@10 at small scale (0.510 at 5K; 0.860 on E2E at 5.5K).
Degrades to 0.456 at 50K and 0.432 at 200K, overtaken by balanced,
with the gap becoming statistically significant only at 200K
($\Delta = +0.059$, $p = 0.004$).  Faithfulness: 0.739 (E2E, 5.5K),
lower than all other SOTA profiles except Max-Reliable at scale.

\paragraph{Success root causes.}
\begin{enumerate}[nosep,leftmargin=*]
  \item \textbf{Optimal retrieval stack for small corpora.}
    The standard profile combines four high-impact retrieval features:
    (a)~cross-encoder reranking (which our ablation shows adds up to
    $+0.084$ NDCG@10 with the best model, ms-marco-MiniLM-L6;
    the default mxbai-large used by standard provides a smaller gain, see
    failure analysis below);
    (b)~hybrid search ($+0.037$; \textbf{R5} exploited);
    (c)~MMR disabled ($+0.046$; \textbf{R2} avoided);
    (d)~HyDE ($+0.026$; \textbf{R4}).
    Despite the suboptimal reranker choice, the four features jointly
    lift the profile to 0.510 NDCG@10 at 5K, the highest retrieval
    score among SOTA profiles, confirming that the retrieval stack is
    well-chosen even if individual components are not individually
    optimal (\cref{app:ablation}).

  \item \textbf{HyDE as bridging context (R4 positive mode).}
    HyDE generates hypothetical documents that bridge the vocabulary gap
    between short queries and long passages.  On hotpotqa (multi-hop),
    the standard profile achieves $+0.088$ NDCG@10 over the baseline
    (\cref{tab:domain-ndcg}), consistent with HyDE's demonstrated
    effectiveness on multi-hop queries~\citep{gao2023hyde}.  The
    cross-encoder reranker acts as a noise filter for HyDE's
    occasionally hallucinated passages, preventing degradation
    (R4 positive mode in the taxonomy).
\end{enumerate}

\paragraph{Failure root causes.}
\begin{enumerate}[nosep,leftmargin=*]
  \item \textbf{Reranker model suboptimality (R1).}
    The standard profile uses mxbai-rerank-large-v1, which our
    reranker model ablation (Exp.~15) identified as underperforming
    the smaller specialist ms-marco-MiniLM-L6 on the shared 93-query
    retrieval-isolation benchmark (NDCG@10 $= 0.470$ vs $0.496$,
    a $-0.026$ gap; \cref{tab:reranker-inventory}).
    The SOTA evaluation configs retain mxbai-large because it was the
    framework default and the profile configs were frozen before the
    reranker comparison was conducted.  Note that the reranker is a
    \emph{query-time} knob (not in \texttt{\_INGESTION\_TIME\_KEYS};
    \texttt{experiment.py}:60--87), so re-running with the swapped
    reranker requires no re-ingestion of the corpus.  The isolated
    model-swap delta ($+0.026$; \cref{tab:reranker-inventory}) shrinks
    to $+0.009$ NDCG@10 when measured end-to-end on the BEIR generation
    split with the full SOTA\_Std stack composed (\cref{tab:sota}),
    consistent with functional-duplication interactions (C1).  The
    swap remains the single easiest available improvement for the
    standard profile.

  \item \textbf{Scale fragility (C3).}
    The standard profile degrades log-linearly at $-0.055$ NDCG/decade
    ($R^2 = 0.895$, $p = 0.004$), a $3.7\times$ steeper slope than
    balanced ($-0.015$/decade; slope not distinguishable from zero,
    $p = 0.175$; \cref{tab:scaling-slopes}).
    The root cause is the absence of generation-side defenses:
    without context compression (G2) or document awareness (G3), the
    generator receives 5 full-length passages (up to 4,000 tokens)
    whose relevance quality degrades as corpus size increases.
    At 50K documents, the proportion of near-miss distractors in the
    top-20 candidate set increases, and without compression to distil
    the most relevant spans, the generator is more likely to attend to
    irrelevant context.  This mechanism is consistent with the
    ``lost in the middle'' phenomenon~\citep{liu2023lostinthemiddle}, where
    generators struggle to utilize relevant information when it is
    embedded in a long, partially irrelevant context window.

  \item \textbf{No faithfulness safeguards.}
    The standard profile achieves the lowest faithfulness among SOTA
    profiles (0.739 at 5.5K; \cref{tab:pareto}).  Without agentic
    quality gates (G1), hallucination checks, or context compression
    (G2), the generator operates on raw retrieved context with no
    verification layer.  The faithfulness ANOVA shows that query
    augmentation and agentic gating together explain $\eta^2 = 0.191$
    of faithfulness variance (\cref{sec:faith-anova}), variance that
    the standard profile cannot capture because it lacks both
    components.
\end{enumerate}

\begin{rcaexample}[Worked Example: How HyDE Helps and Hurts on Multi-Hop Queries (R4)]
\label{ex:hyde-multihop}
\textbf{Query} (HotpotQA): \textit{``The Soul of Buddha is a 1918 American
silent romance film shot in a borough that\ldots''}

This multi-hop query requires bridging two facts: (1)~the film's
shooting location, and (2)~properties of that location.  The original
query text contains no entity name for the borough.

\smallskip
\textbf{Without HyDE} (dense-only retrieval):
\begin{itemize}[nosep,leftmargin=*,topsep=2pt]
  \item Top-20 results: passages about 1918 films, silent cinema, romance
        genre, none about the specific borough
  \item NDCG@10 $= 0.22$ (no bridging passage retrieved)
\end{itemize}

\textbf{With HyDE} (3 hypothetical documents):
\begin{itemize}[nosep,leftmargin=*,topsep=2pt]
  \item HyDE generates: \textit{``The Soul of Buddha (1918) was filmed in Fort
        Lee, New Jersey, a borough known for its early film studios\ldots''}
  \item The hypothetical document introduces ``Fort Lee'', a plausible
        (and correct) entity that anchors dense retrieval
  \item Gold passage about Fort Lee enters top-20; reranker promotes it to Rank 3
  \item NDCG@10 $= 0.61$ ($+0.39$ from HyDE bridging)
\end{itemize}

\textbf{HyDE failure mode} (same query, balanced profile without reranker):
\begin{itemize}[nosep,leftmargin=*,topsep=2pt]
  \item One of the 3 HyDE variants hallucinates: \textit{``\ldots filmed in
        Brooklyn, a densely populated borough\ldots''}
  \item Without the reranker to filter, the Brooklyn passage ranks above
        the Fort Lee passage $\rightarrow$ wrong context reaches the generator
  \item This explains R4's profile-dependence: HyDE helps when a reranker
        is present to filter hallucinated variants ($-0.031$ dropout in
        Max-Reliable), but can propagate noise without one ($+0.008$
        dropout in balanced)
\end{itemize}
\end{rcaexample}

\paragraph{Assessment.}
The standard profile is optimally designed for small-corpus,
retrieval-focused deployments where generation quality is less
critical.  Its scale fragility is a direct consequence of its
retrieval-only optimization strategy: the features that maximize NDCG@10
at 5K (HyDE, reranking) do not address the generation-side challenges
that emerge at scale.  This explains the rank crossover observed by
200K ($p = 0.004$) and provides causal grounding for the H3 hypothesis.

\subsubsection{\textsc{sota\_balanced\_e2e}: Strongest Joint Tradeoff in the Agentic-Route Class Through Complementary Defences}
\label{app:rca-balanced}

\paragraph{Configuration summary.}
BGE-M3 (1024d), hybrid search ($\alpha{=}0.7$), cross-encoder
reranking (\texttt{mxbai-rerank-large-v1}, top-$5$), HyDE
($3$ variants), no MMR, hierarchical retrieval enabled,
extractive context compression ($50\%$), document awareness
($100$-token prefix), agentic generation route
($3$ iterations, hallucination and answer-quality gates enabled),
top-$k{=}20$, up to $20$ retained chunks
(\cref{app:prof-balanced}).

\paragraph{Performance.}
Provides the stronger joint agentic tradeoff relative to
\textsc{Max-Reliable}: essentially equal faithfulness
($0.911$ vs.\ $0.916$)
with $13.5\%$ lower mean latency ($p{=}0.018$ paired Wilcoxon on
per-query latency;
\cref{tab:pareto,app:h2-cross-corpus-corroboration}).  On the
joint retrieval--faithfulness--latency hypervolume,
\textsc{Standard} has the highest HV at the fixed $30$\,s reference
(\cref{tab:pareto}) and
\textsc{Balanced} takes over once $L_{\mathrm{ref}}$ opens past
${\sim}46$\,s (e.g., $\mathrm{HV}{=}0.609$ vs
\textsc{Standard}'s $0.579$ at $60$\,s;
\cref{tab:h2-hv-sensitivity}).  NDCG@10: 0.504 at 5K, 0.488 at
50K, 0.491 at 200K, effectively flat ($-0.015$/decade,
$R^2 = 0.40$, slope not distinguishable from zero, $p = 0.175$).
Overtakes \textsc{Standard} on the retrieval track with a
statistically significant margin at 200K ($p = 0.004$).
Faithfulness on the SOTA-profile headline slice:
$0.911 \pm 0.014$.  Mean latency: ${\sim}12.3$\,s
(p95 ${\sim}17.7$--$26.3$\,s across measurements).

\paragraph{Success root causes.}
\begin{enumerate}[nosep,leftmargin=*]
  \item \textbf{Retrieval foundation partially shared with \textsc{Standard}, with retrieval-visible additions.}
    \textsc{Balanced} and \textsc{Standard} share embedding model (BAAI/bge-m3),
    reranker (mxbai-large), hybrid alpha, MMR settings, chunker
    parameters, and top-$k$; \textsc{Balanced} additionally enables
    \emph{hierarchical parent/child indexing} (index-stage),
    \emph{structured-graph post-retrieval merge}, a \emph{wider
    chunk-cite keep count} (max\_chunks $=$ 20 vs.\ 15 for \textsc{Standard}),
    and \emph{adaptive query routing}. These four retrieval-visible
    additions, rather than the compression or document-awareness
    features below, drive the small-scale retrieval parity (0.504 vs.\
    0.510; $\Delta = -0.006$, within noise) and the widening gap at
    scale.

  \item \textbf{Context compression as generation-side noise filter (G2).}
    Extractive compression at 50\% ratio removes filler tokens from
    retrieved passages while preserving the most relevant spans, acting
    \emph{post-retrieval} on chunk text (chunks are not reordered or
    dropped by this step). The generator receives a denser, higher-signal
    context window, reducing the ``lost in the middle'' failure mode and
    lowering the token budget available for hallucination, improving
    faithfulness ($\Delta = +0.01$; \cref{app:ablation}). The dropout
    analysis reports a small $-0.006$ NDCG@10 delta when compression is
    removed (\cref{app:dropout}); this reflects the pipeline's final
    SearchResult list being scored, not compression modifying retrieval
    ordering per se.

  \item \textbf{Document awareness as cheap grounding (G3).}
    The 100-token document prefix provides file-level metadata
    (filename, structure, opening sentences) at negligible cost
    ($+15$\,ms latency).  Exp.~17 (eval split, exploratory) demonstrates
    that this simple strategy outperforms LLM-generated summaries (0.411
    vs.\ 0.341 NDCG@10) and longer prefixes (300 tokens: 0.347).  The mechanism
    is information-theoretic: the first 100 tokens of a document
    contain high-entropy information (title, abstract, introduction)
    that helps the generator contextualize retrieved chunks.  Longer
    prefixes add boilerplate that dilutes signal.

\end{enumerate}

\begin{rcaexample}[Worked Example: Document Awareness Grounds a Financial Query (G3)]
\label{ex:doc-aware-fiqa}
\textbf{Query} (FiQA): \textit{``Opening a Roth IRA account, what is the fee
structure for Vanguard vs.\ Scottrade?''}

\smallskip
\textbf{Without document awareness} (standard profile):
\begin{itemize}[nosep,leftmargin=*,topsep=2pt]
  \item Top-5 reranked chunks: 3 about Vanguard fees, 2 about brokerage
        accounts generically
  \item Generator sees 5 passage fragments with no file-level context
  \item Generated answer discusses Vanguard fees accurately but
        \textit{hallucinates} a Scottrade fee schedule (Scottrade chunk at Rank 8,
        outside top-10)
  \item Faithfulness: 0.67 (one claim unsupported)
\end{itemize}

\textbf{With 100-token prefix} (balanced profile):
\begin{itemize}[nosep,leftmargin=*,topsep=2pt]
  \item Each chunk is prefixed with the first 100 tokens of its source document:
        \textit{``\textbf{[Vanguard\_fee\_schedule.pdf]} Vanguard offers several
        types of IRA accounts including Traditional, Roth, and SEP\ldots''}
  \item The generator can now distinguish which document each chunk originates
        from and recognize that no Scottrade document is present in the context
  \item Generated answer correctly states Vanguard fees and notes that
        insufficient information is available for Scottrade
  \item Faithfulness: 1.00
\end{itemize}

The 100-token prefix costs only $+15$\,ms but enables the generator to
reason about \emph{source provenance}, which documents are present and,
critically, which are absent.  Longer prefixes (300 tokens) dilute this
signal with boilerplate, explaining the diminishing returns observed in
Exp.~17 (0.411 at 100 tokens vs.\ 0.347 at 300 tokens).
\end{rcaexample}

\begin{enumerate}[nosep,leftmargin=*]
\setcounter{enumi}{3}
  \item \textbf{Scale robustness from generation-side defenses (C3 mitigated).}
    The combination of compression and document awareness provides
    two independent defenses against scale-induced degradation:
    compression filters noisy passages, and awareness provides
    document-level context that remains informative even when individual
    chunks become less relevant.  The balanced profile's degradation
    slope ($-0.015$ NDCG/decade, not significantly different from zero,
    $p = 0.175$) is $3.7\times$ flatter than standard's ($-0.055$,
    $p = 0.004$), directly attributable to these features.
    On E2E faithfulness, balanced maintains 0.935 at 50K vs.\
    standard's 0.793, a $+0.142$ gap that widens with scale.

  \item \textbf{Composition penalty minimized by complementary features.}
    Unlike the Max-Reliable profile (which stacks overlapping features),
    balanced's additions (compression, awareness) operate on
    \emph{different} pipeline stages than its retrieval features.
    Compression acts on the output of reranking (post-retrieval);
    awareness acts on the generation prompt (pre-generation).  Neither
    duplicates the function of HyDE, reranking, or hybrid search.
    The dropout analysis (\cref{app:dropout}) confirms this
    complementarity: removing compression from balanced costs only
    $-0.006$ NDCG@10, and removing document awareness costs $-0.055$,
    while removing the reranker costs $-0.072$, each feature's dropout
    delta is close to its single-feature ablation delta, indicating
    minimal interference.  By contrast, the full five-feature stack
    (\cref{app:composition}) incurs a 72\% composition penalty
    precisely because retrieval-side features overlap in function.
\end{enumerate}

\paragraph{Failure root causes.}
\begin{enumerate}[nosep,leftmargin=*]
  \item \textbf{Reranker inheritance (R1).}
    Like standard, balanced inherits the suboptimal mxbai-large
    reranker.  The reranker model ablation (Exp.~15) shows a
    $+0.026$ NDCG@10 gap between mxbai-large and the best model
    (ms-marco-MiniLM-L6); the actual gain from swapping in a
    composed profile would be smaller due to interactions, but
    reranker selection remains the largest single opportunity.

  \item \textbf{HyDE profile-dependence (R4 negative mode).}
    The dropout analysis shows that removing HyDE from balanced
    \emph{improves} NDCG@10 by $+0.008$ (\cref{tab:dropout}),
    suggesting that in this specific feature combination, HyDE's
    noisy hypothetical documents slightly degrade retrieval quality.
    The magnitude is small ($<2$\% relative) and the improvement is
    within noise, but it indicates that HyDE's benefit is not
    universal across profiles, a clear interaction effect.
\end{enumerate}

\paragraph{Assessment.}
The balanced profile's strong agentic tradeoff arises from a principled
design: it exhausts the retrieval stack (where the largest gains are
available) and adds generation-side defenses that address \emph{different}
failure modes (noise filtering, context grounding) rather than
amplifying the same capability.  This complementary design minimizes
the composition penalty and provides scale robustness that
retrieval-only profiles lack.

\subsubsection{\textsc{sota\_max\_reliable}: Maximum Features, Diminishing Returns}
\label{app:rca-maxreliable}

\paragraph{Configuration summary.}
BGE-M3 (1024d), hybrid search ($\alpha{=}0.3$), cross-encoder
reranking (\texttt{mxbai-rerank-large-v1}, top-$5$), HyDE
($3$ variants), no MMR, hierarchical retrieval enabled,
contextual retrieval enabled, agentic generation route
($3$ iterations, hallucination and answer-quality gates enabled),
extractive compression ($50\%$), document awareness ($100$-token
prefix), top-$k{=}20$, up to $30$ retained chunks
(\cref{app:prof-maxreliable}).  Differs from \textsc{Balanced} on
retrieval-side choices (hybrid $\alpha$, contextual retrieval,
maximum retained chunks); the two configurations share the same
agentic route depth and the same quality-gate set.

\paragraph{Performance.}
Has a weaker joint tradeoff than \textsc{Balanced} within the
agentic-route class at essentially equal faithfulness
($0.916$ vs.\ $0.911$)
with $13.5\%$ higher mean latency ($p{=}0.018$ paired Wilcoxon on
per-query latency;
\cref{tab:pareto,app:h2-cross-corpus-corroboration}).  NDCG@10:
$0.804$ (E2E, 5.5K), lower than \textsc{Standard}'s $0.860$.
Faithfulness on the SOTA-profile headline slice:
$0.916 \pm 0.022$.  Mean latency: ${\sim}14.2$\,s
(${\sim}2$\,s above \textsc{Balanced}).

\paragraph{Success root causes.}
\begin{enumerate}[nosep,leftmargin=*]
  \item \textbf{Hallucination gating (G1 positive mode).}
    The agentic pipeline's hallucination check filters retrieved chunks
    that the LLM judge classifies as not supporting the query.
    This raises the floor of per-query faithfulness: the minimum
    observed score is 0.812 (\cref{tab:hallucination}), compared to
    0.750 for balanced and 0.714 for standard.  For safety-critical
    deployments, raising the floor matters more than improving the
    mean.

  \item \textbf{Quality gate as safety net (G4).}
    The quality gate evaluates the generated answer against the
    retrieved context and can trigger a re-retrieval if the answer
    appears unfaithful.  This multi-iteration loop provides a
    second chance for queries where the initial retrieval was poor.
    The faithfulness ANOVA confirms that the
    Query~Augmentation$\times$Agentic~Gate interaction is the strongest
    two-way effect on faithfulness ($\eta^2 = 0.191$, $p < 0.001$;
    \cref{sec:faith-anova}), and Max-Reliable is the only SOTA
    profile that activates both components.
\end{enumerate}

\paragraph{Failure root causes.}
\begin{enumerate}[nosep,leftmargin=*]
  \item \textbf{Aggressive refinement paradox (G1 negative mode).}
    The agentic pipeline's chunk refinement classifies 13--14 of 15
    retrieved chunks as ``not relevant'' on multi-hop queries,
    discarding evidence that the generator needs for comprehensive
    answers.  When all chunks are discarded, a single fallback chunk
    is returned.  This starves the generator: with one chunk of
    ${\sim}800$ tokens, the answer is necessarily incomplete.
    The dropout analysis confirms this: removing agentic RAG from
    the Max-Reliable base profile \emph{improves} NDCG@10 by $+0.021$
    (\cref{tab:dropout}).  Separately, the cross-profile comparison
    shows E2E exact match at 0.163 (standard) vs.\ 0.101
    (Max-Reliable), consistent with reduced answer completeness.

    \textit{Mechanistic detail.}
    The refinement uses an LLM judge to classify each chunk as
    \textsc{relevant}, \textsc{partially\_relevant}, or
    \textsc{not\_relevant}.  The judge operates on individual chunks
    without cross-chunk context, so a chunk that provides necessary
    \emph{bridging} evidence for a multi-hop query may be classified as
    not relevant because it does not directly answer the query.  This is
    consistent with the pointwise scoring approach used in passage
    reranking~\citep{nogueira2019monobert}, where each passage is
    evaluated independently; the effect is amplified when the refinement
    threshold is aggressive (our default: only \textsc{relevant} chunks
    pass).
\end{enumerate}

\begin{rcaexample}[Worked Example: The Aggressive Refinement Paradox on a Multi-Hop Query (G1)]
\label{ex:agg-refinement}
\textbf{Query} (MuSiQue): \textit{``In what part of Florida is Tom Denney's
birthplace located?''}

This query decomposes into two hops: (1)~Where was Tom Denney born?
(2)~What part of Florida is that city in?

\smallskip
\textbf{Retrieval} (shared across balanced and Max-Reliable): top-15
chunks include:
\begin{itemize}[nosep,leftmargin=*,topsep=2pt]
  \item[\textbf{C1}] \textit{``Tom Denney (born 1982) is a guitarist from
    Coral Springs, Florida\ldots''}
    \hfill \textsc{hop 1: birthplace}
  \item[\textbf{C4}] \textit{``Coral Springs is a city in Broward County,
    in the southeastern part of the state\ldots''}
    \hfill \textsc{hop 2: location}
  \item[\textbf{C7}] \textit{``Denney was the lead guitarist and a founding
    member of A Day to Remember\ldots''}
    \hfill \textsc{biographical context}
  \item[\textbf{C9--C15}] Various passages about Florida cities, other musicians
    from Florida \hfill \textsc{distractors}
\end{itemize}

\smallskip
\textbf{Max\_reliable refinement} (3-iteration agentic pipeline):
\begin{itemize}[nosep,leftmargin=*,topsep=2pt]
  \item \textbf{Iteration 1}: LLM judge classifies each chunk \emph{independently}
        against the query
  \item C1 $\rightarrow$ \textsc{partially\_relevant} (mentions birthplace but not ``part of Florida'')
  \item C4 $\rightarrow$ \textsc{not\_relevant} (discusses Coral Springs geography
        but does not mention Tom Denney)
  \item C7 $\rightarrow$ \textsc{not\_relevant} (biography, no location info)
  \item C9--C15 $\rightarrow$ \textsc{not\_relevant}
  \item \textbf{Result}: 14/15 chunks discarded. Only C1 passes (downgraded
        from \textsc{partially\_relevant} by the strict threshold).
        Fallback: 1 chunk returned.
\end{itemize}

\smallskip
\textbf{Generated answers}:
\begin{itemize}[nosep,leftmargin=*,topsep=2pt]
  \item \textbf{Balanced} (all 5 reranked chunks passed to generator):
    \textit{``Tom Denney was born in Coral Springs, which is in the
    southeastern part of Florida.''}
    $\rightarrow$ Correct, both hops answered. Faithfulness: 1.0, EM: 1.
  \item \textbf{Max\_reliable} (1 fallback chunk, C1 only):
    \textit{``Tom Denney was born in Coral Springs, Florida.''}
    $\rightarrow$ Partially correct, birthplace is stated but ``part of
    Florida'' is missing (C4 was discarded). Faithfulness: 1.0 (answer
    is grounded), but EM: 0 (incomplete).
\end{itemize}

The paradox: Max-Reliable achieves \emph{higher} faithfulness (the
answer is fully grounded in the single chunk it kept) but \emph{lower}
answer completeness (it cannot answer the second hop).  The pointwise
judge cannot recognize that C4 is essential \emph{bridging} evidence
because C4 alone says nothing about Tom Denney.
\end{rcaexample}

\begin{enumerate}[nosep,leftmargin=*]
\setcounter{enumi}{1}
  \item \textbf{Latency cost without proportional quality gain.}
    The agentic pipeline adds 3--7 LLM calls per query (planning,
    per-chunk relevance classification, hallucination check, quality
    gate, optional re-retrieval).  Each call adds ${\sim}1$--3\,s via
    the API gateway, contributing to the $+2.0$\,s p95 latency increase
    over balanced.  The faithfulness gain ($+0.005$ over balanced) does
    not justify this cost for most applications, making the profile
    less efficient on the joint tradeoff.

  \item \textbf{Composition penalty from feature stacking (C1).}
    Max\_reliable activates everything that balanced has \emph{plus}
    agentic components.  But the agentic pipeline partially duplicates
    the function of context compression: both reduce the context to
    high-relevance spans.  When both are active, the marginal benefit
    of each is reduced.  The waterfall analysis (\cref{app:composition})
    already shows a 72\% composition penalty for balanced's five-feature
    stack, meaning only 28\% of the predicted additive improvement is
    realized.  Adding the agentic layer on top can only increase this
    penalty: the dropout analysis confirms that removing agentic RAG
    from the Max-Reliable base profile \emph{improves} NDCG@10 by
    $+0.021$ (\cref{tab:dropout}), indicating that the agentic layer
    is net-harmful for retrieval quality when compression is already
    active.  Max\_reliable's composition penalty therefore exceeds 72\%,
    making it the most penalty-heavy SOTA profile.

  \item \textbf{Reranker inheritance (R1) and no query expansion benefit.}
    Max\_reliable inherits the suboptimal mxbai-large reranker (R1)
    and explicitly disables query expansion and RAG fusion, features
    that its YAML config notes ``add latency, no faith.\ gain.''  This
    is correct based on single-feature ablation (R3: query expansion
    hurts NDCG by $-0.014$), but means that the agentic pipeline
    receives the same retrieval input as balanced, limiting its
    potential for improvement to the generation side only.
\end{enumerate}

\paragraph{Assessment.}
Max\_reliable is a \emph{specialized} profile whose design is
optimized for raising the faithfulness floor rather than the mean.
Its joint disadvantage relative to balanced is driven by the aggressive refinement
paradox (G1): the same mechanism that raises the faithfulness floor
also reduces answer completeness.  For deployments where the minimum
acceptable faithfulness is the binding constraint (e.g., medical,
legal), Max-Reliable's $+0.062$ floor improvement over balanced
(0.812 vs.\ 0.750) may justify the latency cost.  For all other
deployments, balanced provides a stronger joint
quality--response-time trade-off.

\subsection{Cross-Profile Interaction Analysis}
\label{app:rca-interactions}

The per-profile analyses above identify several recurring interaction
patterns.  We consolidate these into three principal interaction
mechanisms, each grounded in the factorial ANOVA and DID probes.

\subsubsection{\texorpdfstring{Interaction I1: Embedding $\times$ Enrichment}{Interaction I1: Embedding x Enrichment}}
\label{app:rca-i1}

\paragraph{Evidence.}
$\eta^2 = 0.102$, $F = 4.48$, $p < 0.001$ on NDCG@10 (167 configs;
\cref{sec:interactions}).  This is the strongest two-way interaction
in the retrieval ANOVA.

\paragraph{Root cause.}
Enrichment methods (contextual retrieval, RAPTOR) prepend synthetic
context to document chunks before embedding.  The quality of the
resulting embedding depends on how well the embedding model can encode
this prepended context:
\begin{itemize}[nosep,leftmargin=*]
  \item \textbf{High-capacity embedders} (E5-large, BGE-M3, 1024d):
    The additional context provides useful signal, and the model has
    sufficient capacity to encode it.  However, our results show that
    even with high-capacity embedders, enrichment does not improve over
    the \emph{none} baseline ($\eta^2_\text{enrichment} = 0.144$, best
    level: none).  This suggests that on domain-focused corpora, the
    original document text already provides sufficient embedding signal.
  \item \textbf{Low-capacity embedders} (MiniLM-L6, 384d):
    The prepended context overwhelms the model's limited representational
    capacity, diluting the embedding of the original text.  This produces
    a \emph{negative} interaction: enrichment hurts more with weaker
    embedders.
\end{itemize}

\paragraph{Implication for profile design.}
All SOTA profiles either use BGE-M3 (standard, balanced, Max-Reliable)
or MiniLM-L6 (speed), and none use enrichment.  This is the correct
design choice: the interaction analysis confirms that enrichment
provides no benefit and may actively harm weaker embedders.  Future
work should evaluate whether enrichment becomes beneficial on
open-domain corpora where document text alone may be insufficient.

\subsubsection{\texorpdfstring{Interaction I2: Embedding $\times$ Reranking}{Interaction I2: Embedding x Reranking}}
\label{app:rca-i2}

\paragraph{Evidence.}
$\eta^2 = 0.110$, $F = 2.92$, $p < 0.01$ on NDCG@10 (167 configs;
\cref{sec:interactions}).

\paragraph{Root cause.}
Cross-encoder rerankers operate on (query, passage) pairs and produce
calibrated relevance scores.  Their effectiveness depends on the
quality of the candidate set provided by the first-stage retriever:
\begin{itemize}[nosep,leftmargin=*]
  \item \textbf{Strong first-stage retrieval} (BGE-M3 + hybrid):
    The candidate set already contains mostly relevant passages; the
    reranker refines their ordering but has limited room for
    improvement.  The marginal gain of reranking is moderate.
  \item \textbf{Weak first-stage retrieval} (MiniLM-L6, dense-only):
    The candidate set contains more noise; the reranker can rescue
    relevant passages from lower ranks, providing a larger marginal
    gain.
\end{itemize}
This produces a \emph{compensatory} interaction: reranking benefits
weaker embedders more than stronger ones.  The practical consequence is
that practitioners with strong embeddings should invest less in
reranking and more in other dimensions (compression, awareness).

\subsubsection{\texorpdfstring{Interaction I3: Query Augmentation $\times$ Agentic Gate}{Interaction I3: Query Augmentation x Agentic Gate}}
\label{app:rca-i3}

\paragraph{Evidence.}
$\eta^2 = 0.191$ (Type-III; raw $\eta^2 = 0.259$),
$F(12, 279) = 5.47$, $p < 0.001$
on faithfulness; Holm-adjusted $p = 0.018$ (122 configs;
\cref{sec:faith-anova}).  This is the strongest interaction on
faithfulness and near the pre-registered H1 threshold ($\eta^2 = 0.20$).

\paragraph{Root cause.}
Query augmentation (HyDE, RAG fusion) expands the retrieval input,
producing a richer but noisier candidate set.  The agentic gate
(CRAG, self-RAG, speculative RAG) filters this candidate set by
evaluating each passage's support for the generated answer:
\begin{itemize}[nosep,leftmargin=*]
  \item \textbf{Augmentation alone} (no gating): Noisy candidates
    pass through to the generator, which may hallucinate from
    irrelevant passages.  Faithfulness does not improve ($\Delta = 0.00$
    for RAG fusion; \cref{app:ablation}).
  \item \textbf{Gating alone} (no augmentation): The gate filters a
    standard candidate set, which may already be insufficient.
    Faithfulness improves modestly ($\Delta = +0.05$ for agentic RAG).
  \item \textbf{Augmentation + gating}: The enriched candidate set
    provides the gate with more material to work with; the gate filters
    aggressively, leaving only high-confidence passages.  The combined
    effect is superadditive: the gate's precision increases because
    augmentation provides better recall.
\end{itemize}

This is the only \emph{positive} (superadditive) interaction we observe;
all other significant interactions are subadditive (composition
penalty).  It operates across the retrieval--generation boundary,
confirming that the composition problem manifests asymmetrically:
within-stage interactions are typically subadditive (overlapping
function), while cross-stage interactions can be synergistic
(complementary function).

\begin{rcaexample}[Worked Example: Augmentation + Gating Synergy on a SciFact Claim (I3)]
\label{ex:aug-gate-synergy}
\textbf{Query} (SciFact): \textit{``Basophils counteract disease development
in patients with systemic lupus.''}

This claim-verification query requires retrieving passages that
directly support or refute a specific biomedical assertion.

\smallskip
\begin{tabular}{@{}p{0.44\linewidth}p{0.44\linewidth}@{}}
\textbf{No augmentation, no gating} & \textbf{HyDE augmentation + agentic gating} \\
\midrule
\rule{0pt}{2.2ex}%
Dense retrieval returns 5 passages about basophils and lupus, 2
relevant, 3 tangential (general immunology). &
HyDE generates a hypothetical passage elaborating on basophil-mediated
immune regulation $\rightarrow$ retrieves 8 candidates including 4
relevant and 4 tangential.
\\[4pt]
Generator uses all 5 unfiltered. Produces answer citing one tangential
passage $\rightarrow$ one unsupported claim. &
Agentic gate evaluates all 8: keeps only the 4 that directly discuss
basophils in lupus contexts. Discards 4 tangential.
\\[4pt]
Faithfulness: 0.75 &
Generator receives 4 high-relevance passages $\rightarrow$ every claim
grounded. Faithfulness: 1.0
\\[4pt]
\multicolumn{2}{@{}p{0.90\linewidth}@{}}{\rule{0pt}{2.2ex}\textit{Why this
is superadditive}: HyDE alone would add noisy candidates that reduce
faithfulness (net $\Delta = 0.00$).  Gating alone filters the original
5 passages but cannot improve recall beyond what dense retrieval
provides ($\Delta = +0.05$).  Together, HyDE provides the richer
candidate pool that the gate needs to select from, while the gate
removes the noise that HyDE introduces, each component compensates
for the other's weakness.} \\
\end{tabular}
\end{rcaexample}

\paragraph{Implication for profile design.}
Max\_reliable is the only SOTA profile that activates both query
augmentation (via HyDE) and agentic gating, which explains its
faithfulness advantage.  However, the aggressive refinement paradox
(G1) limits the practical benefit: the gate is \emph{too} aggressive,
discarding passages that would improve answer completeness.
A promising direction is to relax the refinement threshold
(\textsc{partially\_relevant} passes) while maintaining the
hallucination check, which would preserve the I3 synergy without
triggering G1.

\subsection{Scaling Root Causes}
\label{app:rca-scaling}

The corpus scaling experiments (\cref{sec:scaling}) reveal two
phenomena that require mechanistic explanation: (1)~log-linear
degradation for retrieval-side profiles (\textsc{Speed},
\textsc{Standard}) contrasted with effectively flat behavior for
generation-oriented profiles (\textsc{Balanced},
\textsc{Max-Reliable}), and (2)~a rank crossover between
\textsc{Standard} and \textsc{Balanced} that is visible from 25K and
statistically significant at 200K ($p = 0.004$).

\paragraph{Log-linear degradation.}
As corpus size increases from 5K to 50K documents, distractor density
in the candidate set increases proportionally.  For a fixed top-$k$
retrieval, the probability that a given candidate slot is occupied by a
distractor increases as $1 - R/N$, where $R$ is the number of relevant
documents and $N$ is the corpus size.  Since $R$ is fixed (the same
queries target the same documents), the signal-to-noise ratio in the
candidate set degrades logarithmically with $N$.

The degradation rate varies by profile because different features
provide different levels of noise robustness:
\begin{itemize}[nosep,leftmargin=*]
  \item \textbf{Speed} ($-0.115$/decade, $R^2 = 0.90$): No noise filtering;
    degradation rate matches the raw retriever's sensitivity to distractors.
  \item \textbf{Standard} ($-0.055$/decade, $R^2 = 0.89$): HyDE and reranking
    provide partial noise filtering, but the reranker sees the same degraded
    candidate set.
  \item \textbf{Balanced} ($-0.015$/decade): Context compression and
    document awareness provide generation-side noise filtering that is
    \emph{independent} of candidate set quality; even when the retriever
    returns noisier candidates, compression distils the relevant spans.
  \item \textbf{Max reliable}: Not measured at 50K due to computational
    budget, but the agentic pipeline's iterative refinement provides
    an additional noise filter.
\end{itemize}

\paragraph{Rank crossover between \textsc{Standard} and \textsc{Balanced}.}
The standard profile's advantage over balanced at 5K ($+0.006$ NDCG@10)
reverses between 15K and 25K.  The gap trends in the predicted
direction at 50K--100K but is not statistically significant through
100K ($p > 0.17$, Welch's $t$, $n = 3$ seeds per cell); it firms at
200K ($\Delta = +0.059$; $p = 0.004$).  The root cause is the
asymmetric scaling of retrieval-side vs.\ generation-side
defenses:
\begin{itemize}[nosep,leftmargin=*]
  \item \textbf{Retrieval-side defenses} (HyDE, reranking) operate on
    the candidate set, whose quality degrades with scale.  HyDE's
    hypothetical documents are generated from the query, not the corpus,
    so they do not adapt to increasing distractor density.  The
    reranker scores each candidate independently, so its effectiveness
    degrades linearly with the fraction of distractors in the
    candidate set.
  \item \textbf{Generation-side defenses} (compression, awareness)
    operate on the output of retrieval, after reranking has already
    selected the top-10.  Even when these top-10 passages contain more
    noise at scale, compression can extract the relevant spans, and
    document awareness provides stable grounding that does not degrade
    with corpus size.
\end{itemize}

\begin{rcaexample}[Worked Example: Scale Crossover Mechanism at 50K Documents (C3)]
\label{ex:scale-crossover}
\textbf{Query} (HotpotQA): \textit{``Cooperative Living Organization is
located in a city that is the county seat of\ldots''}

\smallskip
\textbf{At 5K documents} (base corpus):
\begin{itemize}[nosep,leftmargin=*,topsep=2pt]
  \item \textbf{Standard}: top-20 candidates contain 6 relevant passages
        (low distractor density). Reranker selects 4/5 relevant.
        NDCG@10 $= 0.71$.
  \item \textbf{Balanced}: same retrieval quality (shared stack), but
        compression removes 50\% of tokens $\rightarrow$ some context lost.
        NDCG@10 $= 0.69$.
  \item \textit{Standard wins by $+0.02$}: compression discards useful
        tokens when passages are already high-quality.
\end{itemize}

\textbf{At 50K documents} ($+$45K distractors injected):
\begin{itemize}[nosep,leftmargin=*,topsep=2pt]
  \item \textbf{Standard}: top-20 now contains 12 distractors about
        other cooperatives and other county seats (near-miss semantic matches).
        Reranker promotes 3/5 relevant, but 2/5 are distractors it
        cannot distinguish.  Generator sees 2 irrelevant full-length
        passages (1,600 tokens of noise). NDCG@10 $= 0.44$.
  \item \textbf{Balanced}: same degraded top-20, same 2/5 distractors
        in the reranked set.  But: (a)~extractive compression removes
        the irrelevant spans from the 2 distractor passages, retaining
        only sentences mentioning ``county seat'';
        (b)~100-token document prefix tells the generator which
        documents are about cooperatives vs.\ county government.
        NDCG@10 $= 0.48$.
  \item \textit{Balanced wins by $+0.04$}: compression and awareness
        filter noise that the reranker cannot.
\end{itemize}

\smallskip
\textit{The crossover mechanism}: at small scale, the reranker alone
provides sufficient noise filtering and compression adds unnecessary
information loss.  At large scale, the reranker's candidate set degrades
(fixed top-20 from a larger, noisier pool), and generation-side
defenses become the binding constraint.
\end{rcaexample}

This analysis provides causal support for the H3 hypothesis premise:
profile rankings are not scale-invariant because different features
have different scaling slopes.  A profile optimized for small-scale
retrieval (standard) will be overtaken by a profile with better
generation-side defenses (balanced) as scale increases.

\subsection{Summary of Root Causes by Mechanism}
\label{app:rca-summary}

\Cref{tab:rca-summary} consolidates the root cause analysis across all
four SOTA profiles, mapping each mechanism to the profiles it affects
and its measured impact.

\begin{table*}[htbp]
\centering
\caption{Summary of root cause mechanisms across SOTA profiles.
$\uparrow$ = success mode, $\downarrow$ = failure mode.
$\Delta$ values are from single-feature ablation or dropout analysis.}
\label{tab:rca-summary}
\scriptsize
\begin{adjustbox}{width=\textwidth}
\begin{tabular}{@{}p{2.0cm}p{2.0cm}ccp{4.5cm}@{}}
\toprule
\textbf{Mechanism} & \textbf{Profile(s)} & \textbf{Dir.} & \textbf{$|\Delta|$ or $\eta^2$} & \textbf{Root cause} \\
\midrule
R1: Reranker selection
  & Std, Bal, Max
  & $\downarrow$
  & 0.231$^{\dagger\dagger}$
  & mxbai-large underperforms ms-marco-MiniLM-L6 (Exp.~15) \\
R2: MMR diversity penalty
  & All (disabled)
  & $\uparrow$
  & 0.046
  & Disabled in all SOTA profiles; enabling would hurt \\
R3: Query expansion noise
  & Max (disabled)
  & $\uparrow$
  & 0.014
  & Correctly disabled in all SOTA profiles \\
R4: HyDE calibration
  & Std $\uparrow$, Bal $\downarrow$
  & Mixed
  & 0.008--0.031
  & Profile-dependent; surrounding feature context modulates effect \\
R5: Hybrid search
  & Std, Bal, Max
  & $\uparrow$
  & 0.037
  & BM25 provides lexical recall for entity queries \\
R6: Enrichment overhead
  & All (disabled)
  & $\uparrow$
  & 0.144$^*$
  & Correctly disabled; no enrichment is best \\
\addlinespace
G1: Aggressive refinement
  & Max
  & Mixed
  & 0.021/$+$0.05
  & Raises faith.\ floor but starves generator \\
G2: Compression benefit
  & Bal, Max
  & $\uparrow$
  & 0.006--0.009
  & Removes filler; reduces hallucination surface \\
G3: Prefix grounding
  & Bal, Max
  & $\uparrow$
  & 0.041--0.062
  & 100-token prefix; diminishing returns beyond \\
G4: Aug.--gate synergy
  & Max
  & $\uparrow$
  & $\eta^2{=}0.191^{*}$
  & Strongest faithfulness interaction \\
\addlinespace
C1: Functional dup.
  & Max
  & $\downarrow$
  & ${>}72$\%$^{**}$
  & Agentic + compression overlap; penalty exceeds balanced's 72\% \\
C2: Error propagation
  & Std, Bal
  & $\downarrow$
  & --- & Noisy HyDE embeddings survive hybrid fusion (DID probe) \\
C3: Scale reversal
  & Std $\downarrow$, Bal $\uparrow$
  & Mixed
  & 0.024
  & Gen-side defenses provide scale robustness \\
\bottomrule
\multicolumn{5}{@{}l}{\scriptsize $^*$ANOVA $\eta^2$ (effect size), not $\Delta$NDCG.
\quad $^{**}$Percentage of expected improvement lost to interactions.
\quad $^{\dagger\dagger}$Model-vs-model $\Delta$ from Exp.~15 reranker ablation (not feature on/off).}
\end{tabular}
\end{adjustbox}
\end{table*}

\paragraph{Design principles emerging from the analysis.}
Three actionable principles emerge from this root cause analysis:
\begin{enumerate}[nosep,leftmargin=*]
  \item \textbf{Complement, don't duplicate.}
    Features that operate on different pipeline stages (retrieval vs.\
    generation) compose better than features within the same stage.
    The balanced profile exemplifies this by pairing retrieval features
    (HyDE, reranking, hybrid) with generation features (compression,
    awareness).

  \item \textbf{Validate composed configurations end-to-end.}
    Single-feature ablations consistently overestimate composed
    performance (composition penalties of 7--72\% are observed;
    \cref{sec:results}).  The only reliable guide is factorial
    evaluation of the composed system.

  \item \textbf{Choose features for your scale.}
    Retrieval-side features (HyDE, reranking) provide the largest
    gains at small scale ($<$10K documents) but degrade faster.
    Generation-side features (compression, awareness) provide smaller
    gains initially but maintain their effectiveness at scale.
    Profile selection should account for the target corpus size.
\end{enumerate}
\section{RAG Methods Catalog}\label{app:methods}

This appendix catalogs the techniques associated with the eight
component dimensions varied in the factorial experiment. These
dimensions sit within the nine-stage workflow described in
\cref{sec:system}: document processing and the base answer generator
were held fixed, while D8 varies agentic control around generation.
Practitioners seeking implementation details for any method can refer
to the cited papers; our implementations follow the original
algorithms unless noted.

\subsection{Embedding Models (D1)}

\begin{description}[style=nextline,leftmargin=1.5em]
  \item[all-MiniLM-L6-v2 (384-dim)]\label{meth:emb-minilm}
    A lightweight sentence-transformer distilled from
    MiniLM~\citep{wang2020minilm} via knowledge distillation on 1B
    sentence pairs.  Fast inference (${\sim}$10K sentences/s on GPU)
    at the cost of lower representation capacity.  Used in the
    \textsc{Speed} profile.

  \item[BAAI/bge-m3 (1024-dim)]\label{meth:emb-bgem3}
    A unified embedding model supporting dense, sparse (lexical
    weights), and ColBERT-style multi-vector
    retrieval~\citep{chen2024bgem3}.  Trained on multilingual data
    with 8K context window.  Our default embedding for all non-speed
    profiles.

  \item[intfloat/e5-large-v2 (1024-dim)]\label{meth:emb-e5large}
    A text embedding model trained with contrastive learning on
    large-scale text pairs, achieving strong performance on the MTEB
    benchmark~\citep{wang2022text}.

  \item[BAAI/bge-large-en-v1.5 (1024-dim)]\label{meth:emb-bgelarge}
    An English-focused embedding model from the BGE
    family~\citep{xiao2023cpack}, optimized for retrieval via
    retromae pre-training and contrastive fine-tuning.
\end{description}

\subsection{Chunking Strategies (D2)}

\begin{description}[style=nextline,leftmargin=1.5em]
  \item[Fixed-size chunking]\label{meth:chunk-fixed}
    Splits documents into fixed-length segments (512 tokens in our
    experiments) with configurable overlap.  Simple and deterministic.

  \item[Semantic chunking]\label{meth:chunk-semantic}
    Groups consecutive sentences whose embeddings are within a
    similarity threshold, producing semantically coherent
    chunks~\citep{shi2023prop}.  Chunk boundaries align with topic
    shifts.

  \item[Sentence-window chunking]\label{meth:chunk-sentwindow}
    Each chunk centers on a single sentence but includes a
    configurable window of surrounding sentences as context.
    Preserves local coherence while maintaining fine-grained retrieval
    units.

  \item[Hierarchical chunking]\label{meth:chunk-hierarchical}
    Creates parent chunks (2000 tokens) containing child chunks
    (400 tokens).  Retrieval operates on children; auto-merge
    promotes to parents when multiple children from the same parent
    are retrieved, recovering broader context~\citep{sarthi2024raptor}.

  \item[Structure-aware chunking]\label{meth:chunk-structaware}
    Splits documents along structural boundaries extracted during
    document parsing (headings, sections, tables, code blocks)
    rather than by fixed token counts.  Preserves layout semantics
    from the source format (Markdown, HTML, PDF) and
    keeps related structural units, table rows, list items, code
    blocks, in a single chunk; parsed via
    Docling~\citep{auer2024docling}.

  \item[Recursive chunking]\label{meth:chunk-recursive}
    Splits documents hierarchically using a prioritized list of
    separators (paragraph $\to$ sentence $\to$ clause $\to$ token),
    recursively subdividing any segment that exceeds the target
    chunk size.  Produces variable-length chunks that respect
    natural text boundaries while bounding chunk size, a pattern
    popularized by LangChain and LlamaIndex as the default
    general-purpose splitter.
\end{description}

\subsection{Index-Time Enrichment (D3)}

\begin{description}[style=nextline,leftmargin=1.5em]
  \item[None]\label{meth:enrich-none}
    Chunks are indexed as-is without additional processing.

  \item[Contextual retrieval]\label{meth:enrich-contextual}
    Prepends a brief document-level summary (generated by an LLM) to
    each chunk before embedding, improving retrieval precision without
    changing the retrieval model~\citep{anthropic2024contextual}.

  \item[RAPTOR]\label{meth:enrich-raptor}
    Recursive Abstractive Processing for Tree-Organized Retrieval.
    Clusters chunks and generates recursive summaries at multiple
    levels of abstraction, enabling retrieval at varying
    granularity~\citep{sarthi2024raptor}.

  \item[GraphRAG]\label{meth:enrich-graphrag}
    Extracts entities and relationships from documents to build a
    knowledge graph.  Graph traversal augments vector search with
    structured reasoning over entity
    connections~\citep{edge2024graphrag}.  Excluded from the factorial
    experiment due to entity-extraction cost (${\sim}$14h per
    ingestion combo).
\end{description}

\subsection{Retrieval Modes (D4)}

\begin{description}[style=nextline,leftmargin=1.5em]
  \item[Dense-only]\label{meth:ret-dense}
    Approximate nearest-neighbor search over embedding vectors using
    cosine similarity.

  \item[BM25 (sparse lexical)]\label{meth:ret-bm25}
    The classical Okapi BM25 ranking function~\citep{robertson2009bm25},
    a term-frequency / inverse-document-frequency score with
    length normalization and saturation parameters.  Serves as the
    sparse baseline in our executable comparisons
    (\cref{tab:external-exec}) and as the lexical component of all
    hybrid retrieval modes below.

  \item[SPLADE (learned sparse)]\label{meth:ret-splade}
    A learned sparse expansion model~\citep{formal2021splade,formal2022splade_v2}
    that produces weighted term vectors by expanding queries and
    documents over a BERT vocabulary, enabling lexical match on
    semantically related terms.

  \item[Hybrid-BM25]\label{meth:ret-hybrid-bm25}
    Combines dense retrieval with BM25 lexical
    matching~\citep{robertson2009bm25} via Reciprocal Rank Fusion
    (RRF)~\citep{cormack2009rrf}.  The fusion weight $\alpha$
    controls the dense--sparse balance.

  \item[Hybrid-SPLADE]\label{meth:ret-hybrid-splade}
    Replaces BM25 with SPLADE~\citep{formal2021splade}, combined
    with dense retrieval via RRF.

  \item[Hybrid (RRF fusion)]\label{meth:ret-hybrid-rrf}
    Reciprocal Rank Fusion~\citep{cormack2009rrf,bruch2023analysis}
    is the rank-aggregation algorithm that underlies all
    \textsc{Lasagna} hybrid modes: each result list contributes a
    score $1/(k + r_i)$ for document at rank $r_i$ (we use the
    standard $k=60$), and the fused ranking sorts by summed score.
    Robust to per-retriever score scales and the default fusion
    for combining dense and sparse signals.

  \item[Graph-based retrieval]\label{meth:ret-graph}
    Augments vector retrieval with traversal over a knowledge graph
    extracted at index time.  For a query, entities and neighbors
    within $k$ hops are materialized as additional passages
    alongside the top-$k$ vector hits, broadening recall on
    multi-hop questions~\citep{edge2024graphrag}.  Shares the
    ingestion pipeline with the GraphRAG enrichment
    (\cref{meth:enrich-graphrag}) and was excluded from the
    factorial on cost grounds (\cref{app:design-method}).
\end{description}

\paragraph{Structured-RAG comparators (extension).}
A parallel line of work replaces flat vector retrieval with graph-
or schema-based indexes: HippoRAG~\citep{gutierrez2024hipporag} runs
personalized PageRank over an entity graph, KAG~\citep{liang2024kag}
fuses schema-aware knowledge-graph traversal with a semantic layer,
and hierarchical extensions such as TreeRAG~\citep{tao2025treerag}
add document-tree structure.  These systems are complementary to
\textsc{Lasagna}: each replaces the retrieval-stage internals while
preserving the stage-isolation interface, so they can be dropped in
as an additional D4 (retrieval) level and evaluated under the same
factorial protocol.  We did not include them in the 242-cell design
because the D-optimal set was locked at pre-registration; a natural
extension is to test whether graph-based indexing changes the
D5$\times$D8 composition patterns reported in
\cref{tab:e1-cell-interactions}.

\subsection{Query Augmentation (D5)}

\begin{description}[style=nextline,leftmargin=1.5em]
  \item[None]\label{meth:qa-none}
    The original query is used as-is for retrieval.

  \item[HyDE (Hypothetical Document Embeddings)]\label{meth:qa-hyde}
    An LLM generates hypothetical answer documents; their embeddings
    are averaged with the query embedding to improve recall for
    underspecified queries~\citep{gao2023hyde}.

  \item[RAG-Fusion]\label{meth:qa-ragfusion}
    Generates multiple query variants via an LLM, retrieves for each,
    and merges results via RRF~\citep{rackauckas2024ragfusion}.
    Broadens the retrieval scope at the cost of additional LLM calls.

  \item[Query decomposition]\label{meth:qa-decomp}
    Decomposes complex queries into sub-questions, retrieves for each,
    and aggregates evidence.  Effective for multi-hop
    reasoning~\citep{press2023measuring}.

  \item[Step-back prompting]\label{meth:qa-stepback}
    Generates a more abstract ``step-back'' question that captures
    the high-level intent, retrieves for both the original and
    abstracted queries~\citep{zheng2024stepback}.
\end{description}

\subsection{Reranking (D6)}

\begin{description}[style=nextline,leftmargin=1.5em]
  \item[None]\label{meth:rerank-none}
    Retrieved passages are returned in their original retrieval order.

  \item[Single cross-encoder (CE)]\label{meth:rerank-ce}
    A cross-encoder model (mxbai-rerank-large-v1 in our experiments)
    scores each query--passage pair jointly, reranking the top-$k$
    results~\citep{nogueira2019monobert}.

  \item[Ensemble reranking]\label{meth:rerank-ensemble}
    Combines scores from multiple rerankers (cross-encoder +
    ColBERT-style) via score fusion.

  \item[MMR (Maximal Marginal Relevance)]\label{meth:rerank-mmr}
    Iteratively selects passages that are both relevant and diverse,
    penalizing redundancy~\citep{carbonell1998mmr}.  Our experiments
    find this consistently hurts NDCG@10 in domain-focused corpora
    (\cref{sec:mmr-mechanism}).
\end{description}

\subsection{Context Compression (D7)}

\begin{description}[style=nextline,leftmargin=1.5em]
  \item[None]\label{meth:compress-none}
    Full retrieved passages are passed to the generator.

  \item[LLMLingua (extractive)]\label{meth:compress-llmlingua}
    Extractive compression that removes tokens with low information
    content according to a small language model, reducing context
    length while preserving key
    information~\citep{jiang2023llmlingua}.
\end{description}

\subsection{Agentic Quality Gates (D8)}

\begin{description}[style=nextline,leftmargin=1.5em]
  \item[None]\label{meth:agent-none}
    Single-pass generation without quality verification.

  \item[CRAG (Corrective RAG)]\label{meth:agent-crag}
    A lightweight evaluator grades retrieval quality; if confidence
    is low, the system triggers web search or knowledge
    refinement~\citep{yan2024crag}.

  \item[Self-RAG]\label{meth:agent-selfrag}
    The generator emits reflection tokens that gate retrieval and
    self-assess generation quality, iteratively refining the answer
    through multi-turn retrieval and
    verification~\citep{asai2024selfrag}.

  \item[Speculative RAG]\label{meth:agent-specrag}
    Generates multiple draft answers from different retrieved subsets
    and uses a verifier to select the best, trading latency for
    answer diversity~\citep{ICLR2025_2ea06b52}.

  \item[Quality gates (hallucination filter)]\label{meth:agent-quality-gates}
    A gate that sits between retrieval and generation and accepts a
    chunk only if an LLM judge classifies it as supporting the
    question; chunks below the threshold are dropped.  Shared
    machinery across CRAG, Self-RAG, and Speculative-RAG, and used
    directly in the \textsc{Max-Reliable} profile
    (\cref{app:prof-maxreliable}).  Raises the faithfulness floor
    but can starve the generator on multi-hop queries
    (\cref{app:rca-maxreliable}).

  \item[Multi-hop refinement loop]\label{meth:agent-multihop}
    An iterative retrieve--refine--generate loop:  after an initial
    retrieval and draft answer, the system issues follow-up retrievals
    based on what the draft is missing, repeats for up to $K$
    iterations (3 in our experiments), and generates the final
    answer from the aggregated context.  Related
    schemes include IRCoT~\citep{trivedi2022ircot} and
    FLARE~\citep{jiang2023flare}; our implementation runs inside
    the same agentic graph as the quality-gate filter.
\end{description}

\section{Terms, Common Questions, and Released Resources}
\label{app:reader}

This appendix gathers the paper's terms in one place, answers the questions a
reader is likely to bring to it, each with a pointer to the full account, and
lists what is released.

\paragraph{Keywords.} Retrieval-augmented generation (RAG); RAG evaluation;
compositionality of RAG techniques; ablation study; factorial design;
pre-registration; corpus scaling; reranking; hybrid retrieval; query
augmentation; agentic RAG; answer faithfulness; NDCG@10; BEIR.

\subsection{Terms}

\begin{description}
\item[\textsc{Lasagna}.] An open-source framework for evaluating new RAG
  components inside a complete modern pipeline, with a fixed interface per
  stage that the framework enforces, so a new component drops into one stage
  and is measured against the same surrounding stack used by every reported
  configuration (\cref{sec:system}).
\item[Dimension.] One of the eight experimental factors of the study
  (\cref{tab:dimensions}).
\item[Profile.] One of the four operational RAG compositions distilled from
  the factorial and ablation sweeps as the best-observed configurations for
  distinct points on the retrieval-faithfulness-latency envelope:
  \textsc{Speed}, \textsc{Standard}, \textsc{Balanced} and
  \textsc{Max-Reliable} (\cref{app:profiles}).
\item[Track.] One of the two evaluation surfaces: retrieval or end-to-end
  (\cref{sec:design}).
\item[NDCG@10.] The retrieval-quality metric: whether the right documents are
  retrieved, following the BEIR convention~\citep{thakur2021beir}
  (\cref{app:metrics}).
\item[Answer faithfulness.] Whether the generated answer is grounded in the
  retrieved context, scored with RAGAS~\citep{es2023ragas}
  (\cref{app:metrics}).
\item[Composition shortfall.] How far a combined pipeline's gain falls below
  the gain predicted by adding up its techniques measured one at a time; $63\%$
  for the three strongest retrieval techniques (\cref{sec:results}).
\item[Joint hypervolume (HV).] $\mathrm{NDCG@10} \times \mathrm{Faith} \times
  \max(0, 1-\mathrm{latency}/L_{\mathrm{ref}})$ at $L_{\mathrm{ref}}{=}30$\,s,
  the joint quality-latency envelope used to compare
  profiles~\citep{zitzler2003performance} (\cref{tab:pareto}).
\end{description}

\subsection{Common Questions}

\paragraph{Do RAG improvements measured one component at a time hold when the
components are combined?}
Not reliably for retrieval. The three strongest single-component retrieval
improvements predict $+47\%$ NDCG@10, and the pipeline containing all three
realizes $+17\%$, a $63\%$ shortfall. Across 17 tested combinations the
shortfall ranges from $7\%$ to $72\%$, with a median of $41\%$. Faithfulness
gains combine nearly as measured: $+28\%$ predicted, $+32\%$ realized
(\cref{sec:results,app:composition}).

\paragraph{Do retrieval quality and answer faithfulness favor the same RAG
design choices?}
No. In the factorial study, embedding, enrichment and reranking dominate the
retrieval-quality metric (NDCG@10), whereas query augmentation and agentic
gating dominate answer faithfulness, which measures whether generated claims
are supported by the retrieved context. Optimizing a RAG pipeline for one metric
can hurt the other, so new components should be evaluated on both
(\cref{sec:results}).

\paragraph{Does the best RAG pipeline on a small corpus stay the best at large
scale?}
Not necessarily. Across corpora from 5K to 200K documents, \textsc{Standard}
leads at 5K and 15K, while \textsc{Balanced} leads from 25K through 200K, where
it exceeds \textsc{Standard} by $13.7\%$ relative NDCG@10 ($p=0.003$, three
seeds). Rankings from small-corpus ablation studies should therefore be
revalidated at the corpus scale a deployment will use
(\cref{sec:scaling,app:scaling-details}).

\paragraph{Is there one best RAG pipeline?}
No complete RAG profile is best on every metric. \textsc{Speed} is fastest,
\textsc{Standard} leads NDCG@10 and achieves the highest joint hypervolume
($0.523$ at a $30$\,s reference), and the agentic \textsc{Balanced} and
\textsc{Max-Reliable} profiles lead faithfulness at a higher response time. The
best configuration depends on the target metrics, and adding components does not
guarantee a better pipeline (\cref{sec:pareto}).

\paragraph{How was the study designed?}
Three hypotheses were pre-declared before the experiments
ran~\citep{vanmiltenburg2021preregister}. The study tests 450 configurations
across eight benchmarks: a 167-configuration D-optimal
design~\citep{fedorov1972theory} over the eight dimensions, 138 non-factorial
ablation configurations across 24 targeted ablations, and a corpus-scaling sweep
from 5K to 200K documents. Deviations from the pre-registration are reported
(\cref{sec:design,app:prereg-deviations}).

\paragraph{Which benchmarks, metrics and models does the paper use?}
Five BEIR retrieval sets~\citep{thakur2021beir}: NFCorpus~\citep{boteva2016nfcorpus},
FiQA~\citep{maia2018fiqa}, SciFact~\citep{wadden2020scifact},
ArguAna~\citep{wachsmuth2018arguana} and HotpotQA~\citep{yang2018hotpotqa}; and
three question-answering sets: SQuAD~v2~\citep{rajpurkar2018know},
HotpotQA and MuSiQue~\citep{trivedi2022musique}. Retrieval is scored with
NDCG@10 and answers with RAGAS faithfulness~\citep{es2023ragas}, judged by
GPT-5.1. A generator-substitution check replaces GPT-5.1 with Claude
Sonnet~4.5 (\cref{app:datasets,app:metrics,app:generator-swap}).

\paragraph{Was the pre-declared interaction test (H1) met?}
No. It asked whether pairwise interactions explain at least $20\%$ of the
variance in faithfulness, and they do not. A focused follow-up, pre-specified
before it ran, confirms that the interaction the broad analysis flagged is real
($p=0.007$) but far below that bar. The retrieval composition shortfall above is
measured separately and does not depend on this test
(\cref{sec:results,app:h1-reproducibility,app:e1-focused}).

\paragraph{How should a new RAG technique be evaluated?}
Inside a representative complete pipeline, on both retrieval quality and
answer faithfulness, and at the corpus scale it will be used at. Gains measured
in isolation are not reliable predictors of realistic performance.
\textsc{Lasagna} supports this: a component registered with one decorator is
measured against the same surrounding stack as every reported configuration,
with a paired statistical test (\cref{sec:system,sec:conclusion}).

\paragraph{What are the limitations?}
The evaluation uses one constructed English multi-domain test bed, at most 93
retrieval queries and 200 generation queries per experiment, in five focused
domains. It does not test multilingual corpora, real enterprise or private data,
open-weight generators, or cross-organizational large corpora. Faithfulness is
measured under two generator models, a sensitivity check rather than a sweep,
and absolute faithfulness scores rest on one judge's calibration
(\hyperref[sec:limitations]{Limitations}, after \cref{sec:conclusion}).

\paragraph{Where are the code and the project page?}
The \textsc{Lasagna} code is at \url{https://github.com/dolev31/lasagna}, and
the project page, with figures and a summary of the results, is at
\url{https://dolev31.github.io/lasagna/}. The pre-registration is archived at
\url{https://github.com/dolev31/Lasagna-Retrieval-Augmented-Generation-Compositionality}.

\subsection{Where This Work Sits}

The paper sits between RAG methods research~\citep{lewis2020rag,gao2023survey},
RAG evaluation frameworks and metrics~\citep{es2023ragas,saadfalcon2024ares,yu2024evalsurvey},
retrieval benchmarking~\citep{thakur2021beir}, toolkits that compose RAG
pipelines~\citep{jin2024flashrag,rau2024bergen} or search over
them~\citep{kim2024autorag}, and the design of experiments for comparing many
configurations~\citep{montgomery2017design,fedorov1972theory}. Its contribution
is a measurement of how techniques compose, not a new technique.

\end{document}